%% file: main.tex
\documentclass[pdflatex,sn-mathphys-num,iicol]{sn-jnl}%

\usepackage{natbib}
\setcitestyle{authoryear,open={(},close={)}} %

\usepackage[dvipsnames]{xcolor}
\usepackage{array}
\usepackage{minitoc}
\usepackage{tabularx}
\usepackage{comment}
\usepackage{afterpage}

\usepackage{graphicx}%
\usepackage{multirow}%
\usepackage{amsmath,amssymb,amsfonts}%
\usepackage{amsthm}%
\usepackage{mathrsfs}%
\usepackage[title]{appendix}%
\usepackage{textcomp}%
\usepackage{manyfoot}%
\usepackage{booktabs}%
\usepackage{algorithm}%
\usepackage{algorithmicx}%
\usepackage{algpseudocode}%
\usepackage{listings}%
\usepackage{makecell}
\usepackage{colortbl}

\input{preamble}

\theoremstyle{thmstyleone}%

\theoremstyle{thmstyletwo}%

\theoremstyle{thmstylethree}%

\begin{document}

\title{Ego-Exo4D: Understanding Skilled Human Activity\\from First- and Third-Person Perspectives}

\makeatletter
\makeatother
\author[1,2]{Kristen Grauman}
\author[1]{Andrew Westbury}
\author[1]{Lorenzo Torresani}
\author[1,3]{Kris Kitani}
\author[1,4]{Jitendra Malik}
\author[1]{Triantafyllos Afouras$^\ast$}
\author[1,2]{Kumar Ashutosh$^\ast$}
\author[5]{Vijay Baiyya$^\ast$}
\author[6,7]{Siddhant Bansal$^\ast$}
\author[8]{Bikram Boote$^\ast$}
% ...
\affil[1]{FAIR, Meta}
\affil[2]{University of Texas at Austin}
\affil[3]{Carnegie Mellon University}
\affil[4]{University of California, Berkeley}
\affil[5]{Project Aria, Meta}
\affil[6]{University of Bristol}
\affil[7]{IIIT, Hyderabad}
\affil[8]{University of Illinois, Urbana Champaign}
% ...

\input{sec/0_abstract.tex}

\maketitle 
\input{sec/1_intro.tex}

\input{sec/2_related}
\input{sec/3_dataset}

\input{sec/4_language}

\input{sec/5_benchmarks}
\input{sec/6_conclusion}

\input{sec/7_contribution}

\begin{appendices}

\input{sec/appendices/A_Aria}

\input{sec/appendices/A_collection}

\input{sec/appendices/A_people} %
\input{sec/appendices/A_language} %
\input{sec/appendices/A_benchmarks}
\input{sec/contributions}
\end{appendices}

\clearpage
\bibliography{main}

\end{document}

%% file: preamble.tex
\newcommand{\valrng}[3]{#1 \in [#2, #3)}
\newcolumntype{P}[1]{>{\centering\arraybackslash}p{#1}}
\newcolumntype{s}{>{\hsize=.5\hsize}X}

\usepackage{pifont}
\newcommand{\cmark}{\ding{51}}%
\newcommand{\xmark}{\ding{55}}%

\definecolor{LightCyan}{rgb}{0.88,1,1}
\newcolumntype{a}{>{\columncolor{pink}}c}
\newcolumntype{e}{>{\columncolor{pink}}r}
\newcolumntype{d}{>{\columncolor{LightCyan}}r}

\newcommand{\KGCR}[1]{{\color{black}{}#1}}
\newcommand{\arxivb}[1]{{\color{black}{}#1}}

\newcommand{\NEW}[1]{{\color{black}{}#1}}
\newcommand{\NEWRESULT}[1]{{\color{black}{}#1}}

\usepackage{subcaption}
\usepackage{bbm}

\newcommand{\MRnote}[1]{{\color{violet}{\bf MR: }#1}} %
\newcommand{\SMnote}[1]{{\color{teal}{\bf SM: }#1}} %
\newcommand{\SM}[1]{{\color{black}#1}} %
\newcommand{\KGnew}[1]{{\color{black}#1}} %
\newcommand{\note}[1]{{\color{blue}#1}} %

\renewcommand{\MRnote}[1]{{\color{black}}} %
\renewcommand{\SMnote}[1]{{\color{black}}} %
\renewcommand{\SM}[1]{{\color{black}#1}} %
\newcommand{\KGupdate}[1]{{\color{black}#1}} %
\newcommand{\TNIJCV}[1]{{\color{black}#1}} %

\definecolor{lavenderpurple}{rgb}{0.59, 0.48, 0.71}
\newcommand*{\efi}[1]{}
\newcommand{\EM}[1]{#1}

\newcommand{\numcamerawearers}{\KGCR{740} }

\newcommand{\numhourstotal}{\KGCR{1,286} }
\newcommand{\numtakes}{\KGCR{5,035} }
\newcommand{\numscenes}{\KGCR{123} }

\newcommand{\numlocations}{13 }
\newcommand{\numrecordings}{\KGCR{783} }
\newcommand{\numgoprorecordings}{\KGCR{3,724} }
\newcommand{\numlabscollecting}{12 }
\newcommand{\numexperts}{52 }
\newcommand{\numexpertcommentaries}{\KGCR{117,812 }}

%% file: sec/0_abstract.tex
\abstract{We present Ego-Exo4D, a diverse, large-scale multimodal multiview video dataset and benchmark challenge.  Ego-Exo4D centers around simultaneously-captured egocentric and exocentric video of skilled human activities (e.g., sports, music, dance, bike repair). %
\KGCR{740} participants from \numlocations cities worldwide performed these activities %
in \numscenes different natural scene contexts, yielding long-form captures from 1 to 42 minutes each and \numhourstotal hours of video combined.  The multimodal nature of the dataset is unprecedented: the video is accompanied by multichannel audio, eye gaze, 3D point clouds, camera poses, IMU, and multiple paired language descriptions---\KGupdate{including a novel ``expert commentary" done by coaches and teachers and tailored to the skilled-activity domain.} 
To push the frontier of first-person video understanding of skilled human activity, we also present a suite of benchmark tasks and their annotations, including fine-grained activity understanding, proficiency estimation, cross-view translation, and 3D hand/body pose.  All resources are open sourced to fuel new research in the community. 
 \url{https://ego-exo4d-data.org/}
}

\keywords{video understanding, first-person video, egocentric, video-language, 3D, body pose}

%% file: sec/1_intro.tex
\section{Introduction}

\begin{figure*}[t]
    \centering
    \includegraphics[width=1\linewidth]{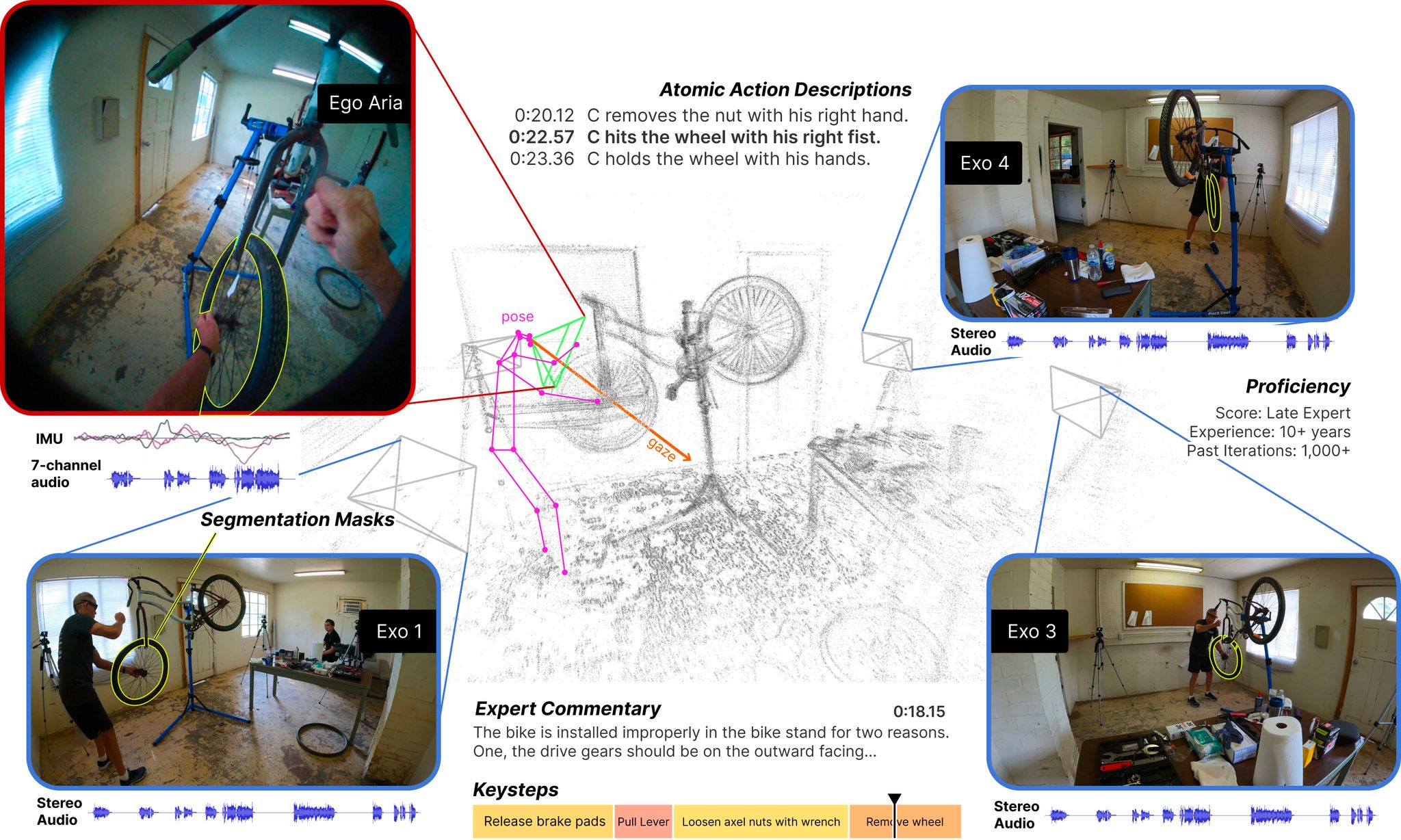} 
    \caption{Ego-Exo4D offers egocentric video alongside  multiple time-synchronized exocentric video streams for an array of skilled human activities---\numhourstotal hours of ego and exo video in total.  The data is both multiview and multimodal, and it is extensively annotated with language, 3D body and hand pose, keysteps, procedural dependencies, and proficiency ratings in support of our proposed benchmark tasks.}
    \label{fig:concept}
\end{figure*}

A dancer leaps across a stage; Lionel Messi delivers a precise pass; your grandmother prepares her famous dumplings.  We observe and seek  human skills in a myriad of settings, from the practical (fixing a bike) to the aspirational (dancing beautifully).   What would it mean for AI to understand human skills?  And what would it take to get there?  

Advances in AI understanding of human skill %
could facilitate many  applications.  In augmented reality (AR), a person wearing smart glasses could quickly pick up new skills with a virtual AI coach that %
\KGupdate{provides real-time guidance.}
In robot learning, a robot watching people in its environment could acquire new dexterous manipulation skills with less physical experience.  In social networks, new communities could form based on how people share their expertise and complementary skills in video.

We contend that both the \emph{egocentric} and \emph{exocentric} viewpoints are critical for capturing human skill.   Firstly, the two viewpoints are synergistic.  The first-person (ego) perspective captures the details of close-by hand-object interactions and the camera wearer's attention, whereas the third-person (exo) perspective captures the full body pose and surrounding environment context. %
See Figure~\ref{fig:concept}.  Not coincidentally, instructional or ``how-to" videos often alternate between a third-person view of the demonstrator and a close-up view of their near-field demonstration. For example, a chef may describe their approach and the equipment from an exo view, then cut to clips showing their hands manipulating the ingredients and tools from an ego-like view.

Secondly, not only are the ego and exo viewpoints synergistic, but there is a need to \emph{translate} fluently from one to the other when acquiring skill.  For example, imagine watching an expert repair a bike tire, juggle a soccer ball, or fold an origami swan---then mapping their steps to your own body.  Cognitive science tells us that even from a very young age we can observe others’ behavior (exo) and map it onto our own (ego)~\citep{flavell,newcombe}, and this actor-observer translation remains the foundation of visual learning.

Realizing this potential, however, is not possible using today's datasets and learning paradigms.  Existing datasets comprised of both ego and exo views (i.e., ego-exo) are  few~\citep{sigurdsson2018charades,sener2022assembly101,Kwon_2021_ICCV,cmu-kitchens,homage}, small in scale, lack synchronization across cameras, and/or are too staged or curated to be resilient to the diversity of the real world.  
Thus the current literature for activity understanding primarily attends to \emph{either} the ego~\citep{Damen2020RESCALING,grauman2022ego4d} or exo~\citep{kinetics,ava,moments,ucf} view,
leaving the ability to move fluidly between the first- and third-person perspectives out of reach.  Instructional video datasets~\citep{miech19howto100m,tang2020comprehensive,zhukov2019cross,youcook2} offer a compelling window into skilled human activity, but (like the above) are limited to single-viewpoint video, \KGupdate{whether purely exocentric or mixed with ``ego-like" views at certain time points}.

We introduce Ego-Exo4D, a foundational dataset to support research on ego-exo video learning and multimodal perception.  The result of a two-year effort by a consortium of 15 research institutions, Ego-Exo4D is a first-of-its-kind large-scale multimodal multiview dataset and benchmark suite.  It constitutes the largest public dataset of time-synchronized first- and third- person video, captured by \numcamerawearers diverse camera wearers in \numscenes distinct scenes and \numlocations cities worldwide. 
For every sequence, Ego-Exo4D provides both the camera wearer's egocentric video, %
as well as \emph{multiple} (4-5) exocentric videos from tripods placed around the camera wearer. %
All views are time-synchronized %
\KGupdate{and precisely localized in a metric, gravity-aligned frame of reference.}
The total collection has \numhourstotal hours of video and \numtakes instances, each spanning   1 to 42 minutes of continuous capture.  

Ego-Exo4D focuses on skilled single-person activities.  The \numcamerawearers participants perform skilled physical and/or procedural activities---dance, soccer, basketball, bouldering, music, cooking, bike repair, health care---in an unscripted manner and in natural settings (e.g., gym, soccer field, kitchens, bike shops, etc.), exhibiting a variety of skill levels from novice to expert.  All video is recorded with rigorous privacy and ethics policies and
 formal consent of participants. 

Ego-Exo4D is not only multiview, it is also multimodal.  Captured with the unique open-source Aria glasses~\citep{aria_2023}, all ego video is accompanied by 7-channel audio, IMU, eye gaze, both RGB and two grayscale SLAM cameras, \KGupdate{and 3D environment point clouds}.  
Additionally, Ego-Exo4D provides multiple new video-language resources, all time indexed: first-person narrations by the camera wearers describing their own actions; third-person play-by-play descriptions of every camera wearer action; and third-person spoken expert commentary critiquing their performance. 
The latter is particularly novel: performed by domain-specific experienced coaches and teachers, it focuses on \emph{how} an activity is executed rather than merely \emph{what} is being done, surfacing subtleties in skilled execution not perceivable by the untrained eye.
All three language corpora are time-stamped against the video.
To our knowledge, there is no prior video resource with such extensive and high quality multimodal data.

Alongside this data, we introduce benchmarks for  foundational tasks for ego-exo video and we formalize them with annotations and evaluation protocols
to spur the community's efforts.  
We propose four families of tasks:
\begin{enumerate}
    \item \emph{ego-exo relation}, for relating the actions of a teacher (exo) to a learner (ego) by estimating semantic correspondences and translating viewpoints; 
    \item \emph{ego(-exo) recognition}, for recognizing fine-grained keysteps %
    and task structure; 
    \item \emph{ego(-exo) proficiency estimation}, for inferring how well a person is executing a skill; 
    \item \emph{ego pose}, for recovering %
skilled 3D body and hand movements %
from ego-video. 
\end{enumerate} We provide annotations for %
each task---the result of more than 200,000 hours of annotator effort.  To kickstart work in these new challenges, we also develop baseline models and report their results.  We are hosting the first public benchmark challenges in 2024.

\begin{figure*}[t]
    \centering
    \includegraphics[width=1\linewidth]{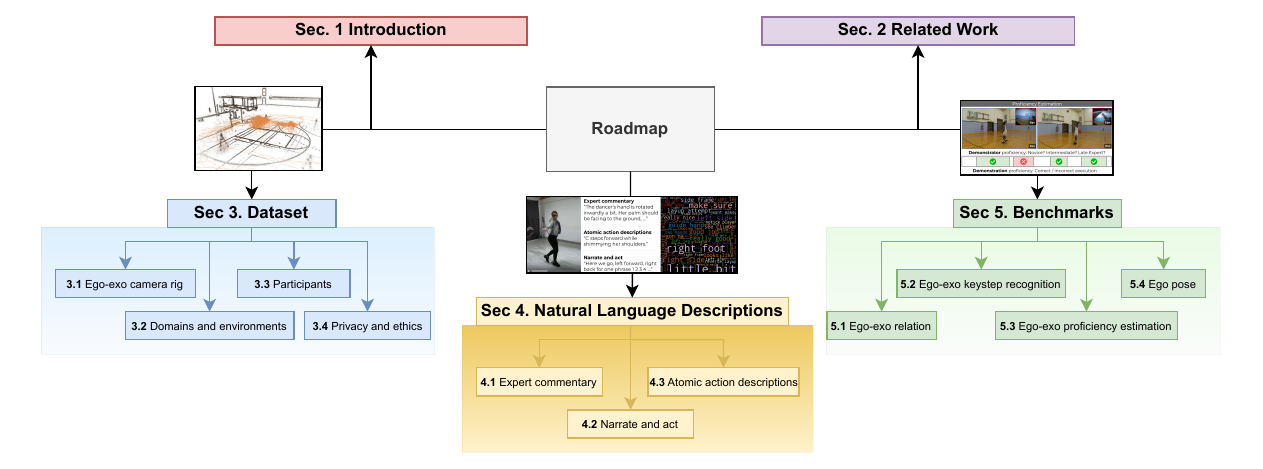}
    \caption{Overview of the paper and its sections, including the Ego-Exo4D dataset (Sec.~\ref{sec:rig},~\ref{sec:domains},~\ref{sec:participants}), the natural language descriptions collected alongside the dataset (Sec.~\ref{sec:language}), and the benchmark tasks (Sec.~\ref{sec:benchmarks}).}
    \label{fig:roadmap_v1}
\end{figure*}

Though we are motivated by skill learning, %
Ego-Exo4D is %
poised for even broader influence, beyond the proposed benchmarks.  %
Whereas existing datasets lack activity modeling in real-world 3D contexts  (e.g., restricted to mocap suits and/or lab settings) %
and existing 3D datasets typically
focus on static scenes and objects. 
Ego-Exo4D is a resource for \textbf{general 3D vision}---such as environment reconstruction, camera relocalization, audio-visual mapping, and many others.  Similarly, our novel \textbf{video-language} resources will offer many opportunities for grounding of actions and objects, multimodal representation learning, and language generation.  Finally, though our tasks prioritize perception from the ``ego-only" perspective, the exo component of our data ensures its utility for the more \textbf{traditional exo viewpoint} too, e.g., for activity recognition and body pose estimation.

In summary, Ego-Exo4D is the community's first diverse, large-scale multimodal multiview video resource.   We have open sourced all the data, annotations, camera rig protocol, and benchmarks. With this release, we aim to fuel new research in ego-exo, multimodal activity, and beyond.  

Figure~\ref{fig:roadmap_v1} provides a roadmap for this paper.  After reviewing related work (Sec.~\ref{sec:related}), we describe the dataset---its contents, camera setup, participants, and our approach to collection (Sec.~\ref{sec:dataset})---followed by an overview of its three forms of natural language annotations (Sec.~\ref{sec:language}).  Finally, we introduce the benchmark tasks organized into the four families described above, outlining the motivation, task definitions, metrics, and baseline results for each (Sec.~\ref{sec:benchmarks}).

%% file: sec/2_related.tex
\KGCR{

\begin{table*}[t]
\setlength{\tabcolsep}{3pt} %
  \centering
  \resizebox{\linewidth}{!}{
  \begin{tabular}{lrddddeeeeaa}
    \toprule
    \textbf{Dataset} & \textbf{Year} & \textbf{Modalities} & \textbf{\#Subj.} & \textbf{\#Scenes} & \textbf{\#Tasks} & \textbf{\#Actions} & \textbf{\#Masks}  & \textbf{\#BP}  & \textbf{\#HP} & \textbf{Nar.} & \textbf{EC}     \\
    \midrule
       \multicolumn{9}{c}{\emph{Multimodal Egocentric Datasets}} \\
    EGTEA-Gaze~\citep{li2018eye}  & 2018                 & V,A,G & 32    & 1   & 7  & 106            & 15k & - & - & \xmark  & \xmark \\
    MECCANO~\citep{ragusa2021meccano} & 2021 & V,D,G  & 20 & 2 & 1 & 61 & - & - & - & \xmark & \xmark  \\
    EK100~\citep{epic-kitchens-100}  & 2022      & V,A  & 37   & 45 & N/A   & (97;300)$*$   & - & - & - & \cmark & \xmark        \\
    Ego4D~\citep{grauman2022ego4d} & 2022 & V,A,3D,S,G,I & 931   & 74 & N/A  & 110$\dagger$  & - & - & - & \cmark & \xmark         \\
    HoloAssist~\citep{wang2023holoassist} & 2024 & V,A,D,G,3D,I & 222 & ? & 20 & (49;165)$*$ & - & ? & ? & \cmark & \xmark \\
    \midrule
    \multicolumn{9}{c}{\emph{Multiview Datasets}} \\
    IXMAS~\citep{ixmas}  & 2006 & V & 10 & 1 & N/A & 11 & - & - & - & \xmark & \xmark\\
    MEVA~\citep{meva}   & 2021                                &  V,T,GPS & 100 & 28 & N/A & 37 & - & - & - & \xmark & \xmark   \\
    \midrule
    \multicolumn{9}{c}{\emph{Ego-Exo Datasets}} \\
        CMU-MMAC~\citep{cmu-kitchens}    & 2009                  & V,A,M,I  & 43    & 1    & 5 & -  & - & - & - & \xmark & \xmark          \\
    Charades-Ego~\citep{sigurdsson2018actor}  & 2018  & TODO & 71 & N/A & N/A & 157 & - & - & - & \xmark & \xmark \\
    LEMMA~\citep{jia2020lemma} & 2020 & V,D & 8 & 14 & 15 & (24;64)$*$ & - & - & - & \xmark & \xmark \\
    HOMAGE~\citep{homage} & 2020 & V,A,T,B,Ma & 27 & 10 & 70 & 453 & - & - & - & \xmark & \xmark \\ %
     H2O~\citep{Kwon_2021_ICCV}     & 2021                 & V,D  & 4    & 3   & N/A  & 36   & - & - & 0.5M & \xmark & \xmark         \\
    
    Assembly101~\citep{sener2022assembly101,ohkawa2023assemblyhands}  & 2022                    & TODO & 53    & 1    & \textbf{101}  & (24;90)$^*$            & - & - & 0.2M & \xmark & \xmark  \\

    EgoExoLearn~\citep{huang2024egoexolearn} & 2024 & V,A,G,I & 136 & 7 & 8 & (95;254)$^*$ & - & - & - & \cmark & \xmark \\

    \textbf{EgoExo4D}     & 2024               & V,A,I,G,3D,6D,B,Ma  & \textbf{\numcamerawearers}   & \textbf{\numscenes}     & \KGCR{43}$^\ddag$   & \textbf{689} & \KGCR{\textbf{2.2M}} & \KGCR{\textbf{9.6M}} & \KGCR{\textbf{4.4M}} & \cmark & \cmark \\
    \midrule
    \bottomrule
  \end{tabular}
  }
  \caption{\KGCR{Comparison between Ego-Exo4D and relevant datasets. Compared to existing datasets capturing both egocentric and exocentric views, Ego-Exo4D features more modalities, more subjects, and significantly larger scene diversity, as well as rich annotations including key-step segments, object masks, and three meticulously synchronized natural language descriptions paired with the videos (narrations, narrate-and-act, and expert commentary). 
  To our knowledge, Ego-Exo4D also offers the largest available manual ground truth egocentric body pose annotations to date (in the above datasets or any others), and it has $\sim$14M total frames of 3D pose annotations and pseudo-annotations.
  \emph{\#Tasks} denotes the number of tasks that subjects were asked to execute in each dataset, \emph{Subj.} denotes recorded subjects, \emph{\#BP} refers to number of 3D body poses, \emph{\#HP} refers to number of 3D hand poses, \emph{Nar.} denotes narrations, and \emph{EC} refers to expert commentary annotations. Modality abbreviations: \textbf{V}ideo, \textbf{A}udio,  \textbf{D}epth, \textbf{G}aze, \textbf{S}tereo, \textbf{I}MU, \textbf{3D} Environments, \textbf{T}hermal IR, \textbf{GPS}, \textbf{M}otion Capture, \textbf{6D}OF, \textbf{B}arometer, \textbf{Ma}gnetometer. $^*$ denotes action taxonomies defined in terms of verbs and nouns, statistics reported as (number of verbs; number of nouns). $\dagger$ The number has been taken from the Moment Query benchmark. $\ddag$ Number of tasks for Ego-Exo4D includes $21$ procedural activities and $22$ physical activities (listed in Table~\ref{tab:scenarioslist}).}}
\label{tab:comparison_with_datasets}
\end{table*}
}

\section{Related work}\label{sec:related}

Next we review prior work in datasets, %
human skill, and cross-view analysis.  
Section~\ref{sec:benchmarks} will discuss additional related work for each benchmark task. %

\vspace*{-0.1in}
\paragraph{Egocentric datasets}
There has been a surge of interest in egocentric video understanding, facilitated by recent ego-video datasets showing unscripted daily-life activity as in Ego4D~\citep{grauman2022ego4d}, EPIC-Kitchens~\citep{Damen2018EPICKITCHENS,Damen2020RESCALING,EPICFields2023}, UT Ego~\citep{lee-cvpr2012}, 
ADL~\citep{pirsiavash2012detecting}, and KrishnaCam~\citep{krishnacam}, or  procedural activities as in  EGTea~\citep{li2018eye}, AssistQ~\citep{wong2022assistq}, Meccano~\citep{ragusa2021meccano}, CMU-MMAC~\citep{cmu-kitchens}, EgoProcel~\citep{EgoProceLECCV2022}, and HoloAssist~\citep{wang2023holoassist}.  %
Unlike any of the above, Ego-Exo4D focuses on multimodal ego \emph{and} exo capture, and it is focused on the domain of skilled activities.  %

Many members of our Ego-Exo4D team worked together to create Ego4D~\citep{grauman2022ego4d}.  The two datasets share some properties: both emphasize unscripted data, with long continuous captures in authentic environments with diverse participants, and both offer novel benchmark tasks and video-language annotations.  
However, whereas Ego4D focuses on daily-life activity from the egocentric view alone,
Ego-Exo4D focuses on skilled activity in specific domains, captures both egocentric and exocentric viewpoints, and is significantly more multimodal.  Ego-Exo4D's language annotations are also broader in scope compared to Ego4D, going beyond play-by-play action narrations to also include first-person how-to descriptions and third-person expert commentary about the skilled activities.

\vspace*{-0.1in}

\paragraph{Multiview and ego-exo datasets}

Most existing multiview datasets focus on static scenes~\citep{matterport3d,gibson,replica19arxiv,ramakrishnan2021habitat,sun3d} and objects~\citep{co3d,shapenet}, with limited (exo only) multiview human activity~\citep{ixmas,meva}.
CMU-MMAC~\citep{cmu-kitchens} and CharadesEgo~\citep{sigurdsson2018charades} are early efforts to capture both ego and exo video.  
CMU-MMAC~\citep{cmu-kitchens} features %
43 participants in mocap suits who cook 5 recipes in a lab kitchen.  
In CharadesEgo~\citep{sigurdsson2018charades}, 71 Mechanical Turkers record 34 hours of scripted scenarios (e.g., ``type on laptop, then pick up a pillow") from the ego and exo perspectives sequentially, yielding unsynchronized videos with non-exact activity matches.
More recent ego-exo efforts focus on specific activities in one or two environments.
Assembly101~\citep{sener2022assembly101} and H2O~\citep{Kwon_2021_ICCV} provide time-synced ego and exo video at a lab tabletop where people assemble toy cars or manipulate handheld objects, with 53 and 4 participants, and 513 and 5 hours of footage, respectively. LEMMA~\citep{jia2020lemma} contains multi-agent, multi-task activities with 15 common daily tasks, performed by 8 individuals in 14 unique kitchens/living rooms. Homage~\citep{homage} provides 30 hours of ego-exo video from 27 participants in 2 homes doing household activities like laundry. EgoExoLearn~\citep{huang2024egoexolearn} provides 120 hours of egocentric videos emulating the human demonstration following process with exocentric demonstration videos.

Compared to any of the prior efforts, Ego-Exo4D offers an order of magnitude more participants, diverse locations, and hours of footage (\numcamerawearers participants, \numscenes unique scenes, \numlocations cities, \numhourstotal hours).  Importantly, our focus on skilled tasks takes the participants out of the lab or home and into settings like soccer fields, dance studios, rock climbing walls, and bike repair shops.  Such activities also yield a wide variety of full body poses and movements within the scene, beyond using objects at a tabletop.  This variety %
means Ego-Exo4D %
augments
existing 3D human body pose datasets~\citep{egobody,li2023ego,Joo_2017_TPAMI,egohumans,hps}.
Finally, compared to any prior ego-exo resource, Ego-Exo4D's suite of modalities %
and benchmark tasks are novel and will expand the research directions the community can take for egocentric and/or exocentric video understanding.
Table~\ref{tab:comparison_with_datasets} summarizes Ego-Exo4D's properties compared to those of existing datasets.

\vspace*{-0.1in}

\paragraph{Human skill and video learning}
Analyzing skill and  action quality has received limited attention %
~\citep{assessing-eccv2014,baller-iccv2017,action-quality-2019,skill-determination,Doughty_2019_CVPR,logo}.
Research in instructional or ``how-to" videos is facilitated by (largely exo) datasets like HowTo100M~\citep{miech19howto100m} and others~\citep{tang2020comprehensive,zhukov2019cross,youcook2,ikea-assembly}.
Challenges include grounding keysteps~\citep{zhukov2019cross, elhamifar2020self,videoclip,mil-nce,miech19howto100m,dvornik2022flow,EgoProceLECCV2022,video-distant}, procedural planning~\citep{procedure-learning-fei-fei-li, procedure2,p3iv,pdpp,procedure3, shvetsova2022everything,ko2022video,cao2022locvtp}, learning task structure~\citep{task-structure,zhou2018towards,elhamifar2020self,alayrac,ashutosh-neurips2023,zhou2023paprika}, and leveraging noisy narrations
~\citep{video-distant,mil-nce,miech19howto100m}.    A portion of Ego-Exo4D is procedural activities, but unlike the above, it offers simultaneous ego-exo  capture.
The scale and diversity of our data---including its three forms of language descriptions---widen the avenues for skilled activity understanding research. %

\vspace*{-0.1in}

\paragraph{Ego-exo cross-view modeling}
There is limited prior work on ego-exo cross-view modeling, arguably due to a lack of high-quality synchronized real-world data.  Prior work explores matching people between videos~\citep{ardeshir2016ego2top, ardeshir2018egocentric, fan2017identifying, xu2018joint,Wen_2021_ICCV} and learning view-invariant~\citep{ardeshir2018exocentric, sigurdsson2018actor, Sermanet2017TCN, yu2019see, yu2020first, sherry-neurips2023} or ego features~\citep{li2021ego}.  %
Beyond the specific case of ego-exo, cross-view methods are explored for translation
~\citep{Regmi_2018_CVPR, REGMI2019, tang2019selectiongan, ren2021crossmlp,exo2ego-eccv2024,4diff}, novel view synthesis~\citep{liu2021infinite, ren2022look, rombach2021geometry, wiles2020synsin, watson2022novel, tseng2023consistent, chan2023generative}, and aerial to ground matching~\citep{shah-aerial,deepgeo}.
Ego-Exo4D provides a testbed of unprecedented size and variety for cross-view modeling.  In addition, our ego-exo relation %
tasks (cf.~Section~\ref{sec:benchmarks}) surface new challenges in novel-view synthesis with widely varying viewpoints.

%% file: sec/3_dataset.tex
\section{Ego-Exo4D dataset}\label{sec:dataset}

Next we introduce the dataset and its scope.  Notably, the video capture was a distributed but coordinated effort performed by \numlabscollecting research labs who worked together over nearly two years to create Ego-Exo4D. 
Importantly, our data collection across the sites was a coordinated effort, with common guidelines, scenarios, and camera rigs.  In this way, the dataset is cohesive at the same time it is diverse.  

In the following, we first introduce the ego-exo camera rig and time synchronization process (Sec.~\ref{sec:rig}).  Then we overview the domains and activities that compose the dataset (Sec.~\ref{sec:domains}), followed by discussion of the participants' diverse backgrounds and expertise (Sec.~\ref{sec:participants}).

\begin{figure}[t]
    \centering
    \includegraphics[width=0.8\linewidth]{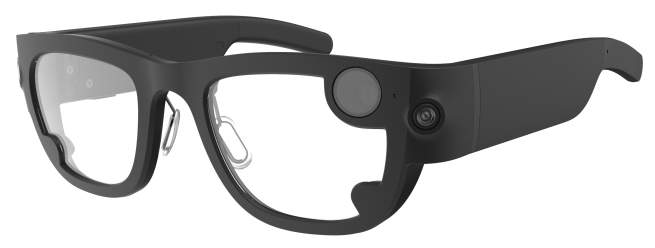}
    \caption{The Aria device used for egocentric recordings.}
    \label{fig:aria-device}
\end{figure}

\subsection{Ego-exo camera rig}\label{sec:rig}

To collect ego-exo data at a global scale, we developed a low-cost camera recording rig that was portable, auto-synchronized, captured a rich suite of sensor data, and attainable internationally.

Our solution consists of 1 Aria (see Figure~\ref{fig:aria-device}), 4 GoPros\footnote{This represents the common core of the collection rig used in all capture settings.  In certain captures, \emph{additional} exo or ego GoPros are also used.}, 1 GoPro Remote, 4 Tripods, 4 SD Cards, 4 Tripod Mount Adapters, 4 Velcro'd Battery Packs, 4 USB-A to USB-C Cables, 1 Glasses Sports Strap, 1 Smartphone, and 1 Laptop or Tablet for questionnaires. The total cost excluding the Aria/phone/laptop is under \$3,000.

\subsubsection{Aria device and sensors}

Aria is an egocentric recording device in glasses form-factor created by Meta. It is designed as a \textit{research tool} for egocentric machine perception and contextualized AI research, and %
available to researchers %
across the world through \href{http://projectaria.com}{projectaria.com}.

\label{appendix:aria:device}
The Aria device emulates future AR- or smart-glasses catering to machine perception and egocentric AI rather than human consumption. It is designed to be wearable for long periods of time without obstructing or impeding the wearer, allowing for natural motion even when performing highly dynamic activities ---such as playing soccer or dancing. It has a total weight of 75g (compared to over 150g for a single GoPro camera), and fits just like a pair of glasses. 

\begin{figure}
    \centering
\begin{minipage}{0.50\linewidth}%
\includegraphics[width=\linewidth]{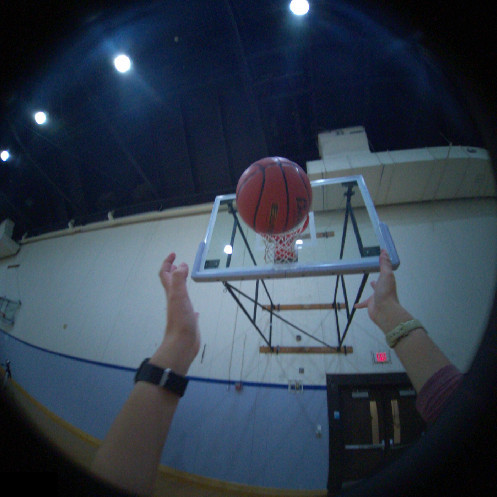}%
\end{minipage}
\begin{minipage}{0.235\linewidth}%
\includegraphics[width=\linewidth]{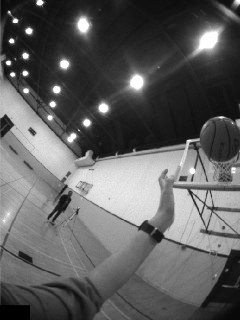}\\[0.5mm]%
\includegraphics[width=\linewidth]{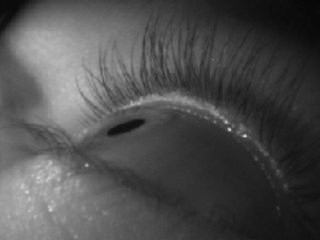}%
\end{minipage}
\begin{minipage}{0.235\linewidth}%
\includegraphics[width=\linewidth]{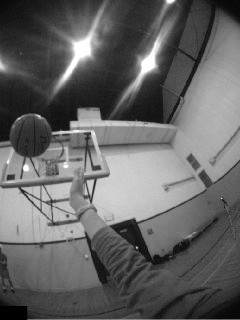}\\[0.5mm]%
\includegraphics[width=\linewidth]{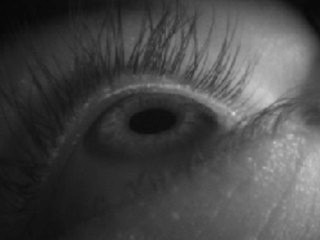}%
\end{minipage}\\%
\includegraphics[width=0.19\linewidth]{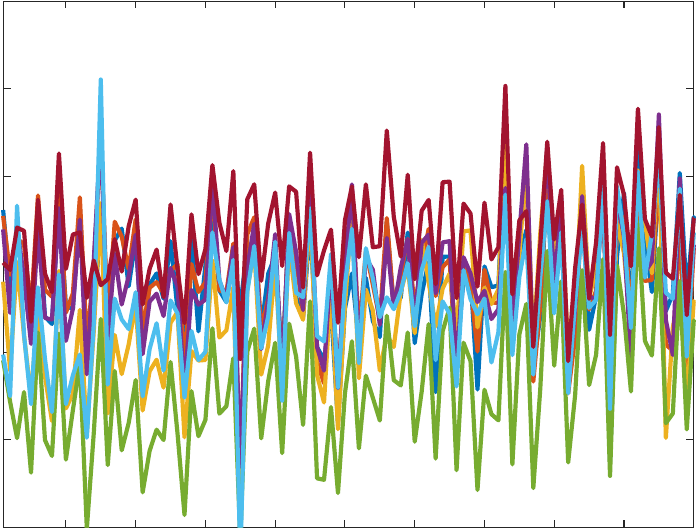}
\includegraphics[width=0.19\linewidth]{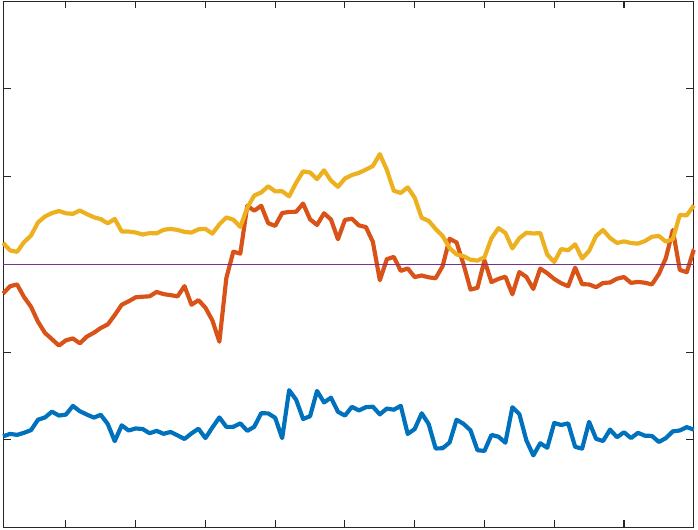}
\includegraphics[width=0.19\linewidth]{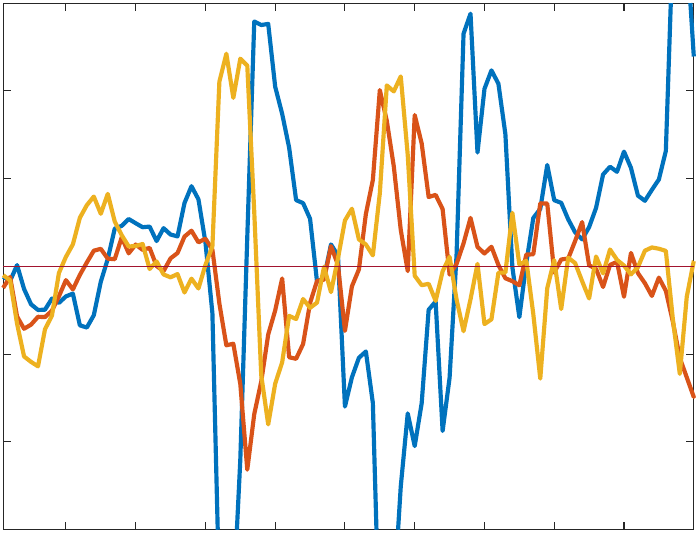}
\includegraphics[width=0.19\linewidth]{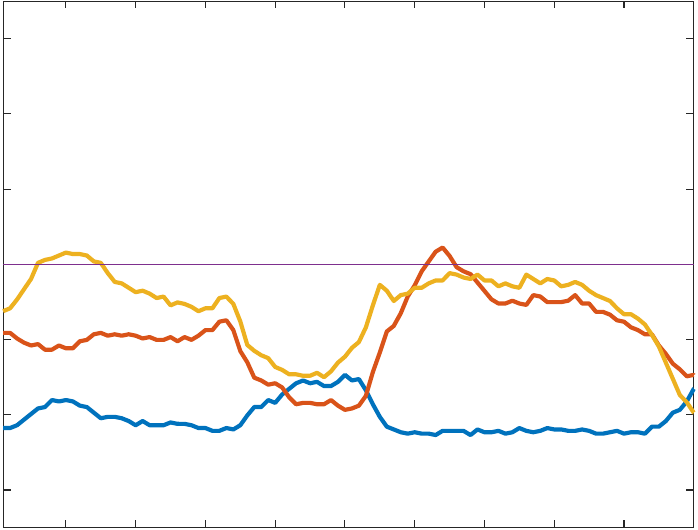}
\includegraphics[width=0.19\linewidth]{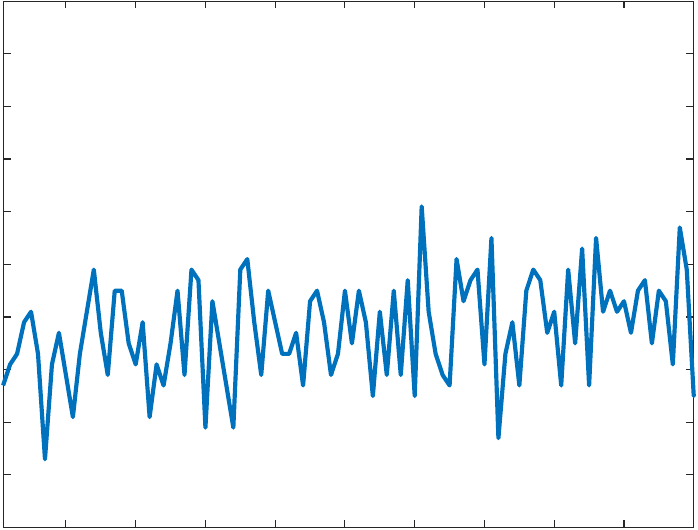}
    \caption{Sensor streams recorded by the Project Aria device. Top: RGB camera, left and right monochrome and eye cameras. Bottom: 10-second extracts from microphones, accelerometer, gyroscope, magnetometer and barometer respectively. }
    \label{fig:appendix:aria:sensors}
\end{figure}

Further, the device integrates a rich sensor suite that is tightly calibrated and time-synchronized, capturing a broad range of modalities.  See Figure~\ref{fig:appendix:aria:sensors}. For Ego-Exo4D, the following sensor configuration is used:
\begin{itemize}
    \item \textbf{One rolling-shutter RGB camera} recording at 30\,fps and $1408 \times 1408$ resolution covering a field of view of $110^{\circ}$. 
    \item \textbf{Two global-shutter monochrome cameras} recording at 30\,fps and $640 \times 480$ resolution. They provide peripheral vision, and each cover a field of view of $150^{\circ}$.
    \item \textbf{Two monochrome eye-tracking cameras} recording at 10\,fps and $320 \times 240$ resolution.
    \item \textbf{An array of seven microphones} recording spatial audio around the wearer.
    \item \textbf{Two IMUs} (800\,Hz and 1000\,Hz respectively), \textbf{a barometer} (50\,fps) and \textbf{a magnetometer} (10\,fps).
\end{itemize}
All sensor streams come with metadata such as timestamps and per-frame exposure times. All data is made available in raw form as part of the Ego-Exo4D dataset.  For convenience, we also include pre-computed slices of data that suit specific purposes, e.g., 2D gaze points, mp4s of each camera, and smaller .vrs files with a subset of sensor streams.

\begin{figure}
    \centering
\begin{minipage}{1.0\linewidth}%
\includegraphics[width=\linewidth]{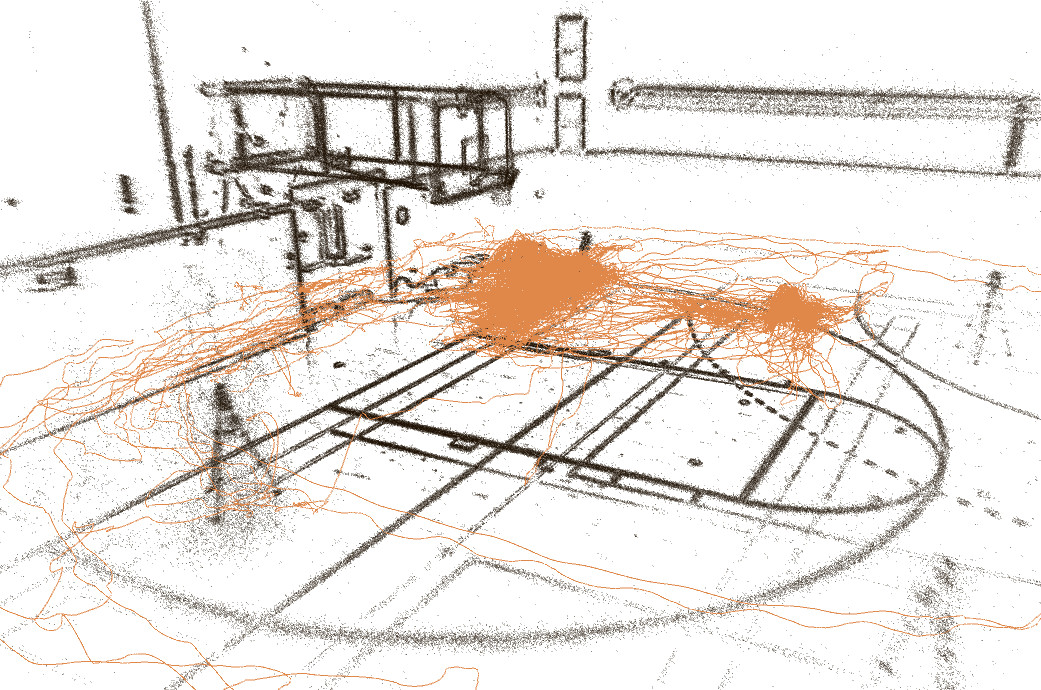}%
\end{minipage}
\includegraphics[width=0.32\linewidth]{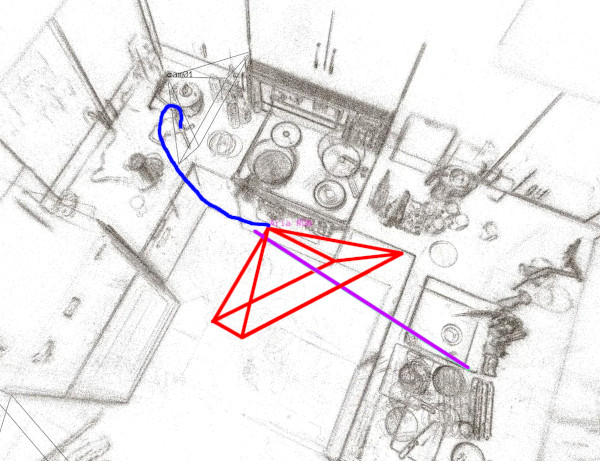}
\includegraphics[width=0.32\linewidth]{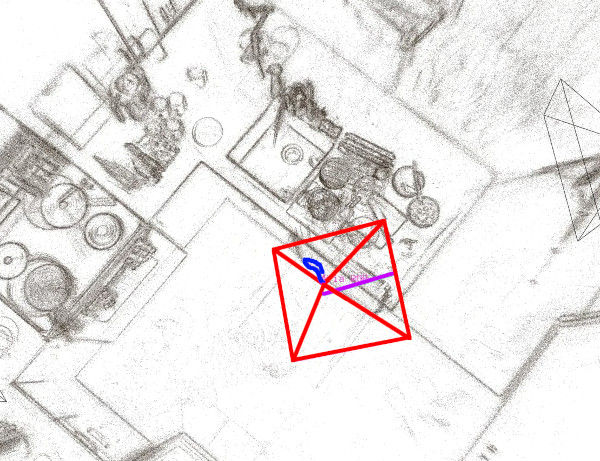}
\includegraphics[width=0.32\linewidth]{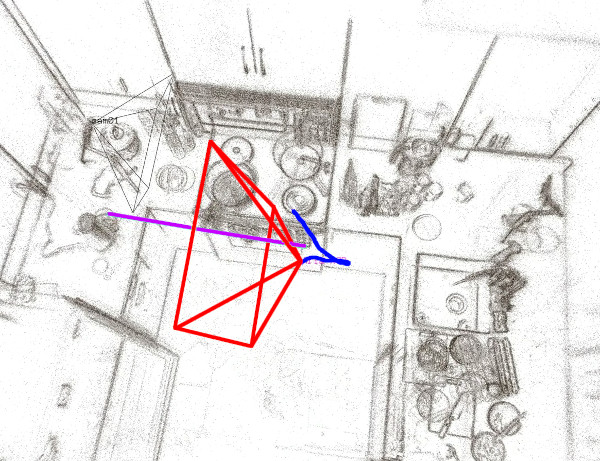}
    \caption{Aria MPS output for several recordings. Top: point cloud and estimated egocentric camera trajectory for a basketball session in %
    Chapel Hill.
    This single continuous recording is 60 minutes long, has a total trajectory length of 2188\,m, and contains 41 distinct takes. Bottom: three screenshots of a cooking recording, visualizing the current camera pose (red), eye gaze (purple), and last second of motion (blue). }
    \label{fig:appendix:aria:mps}
\end{figure}

\subsubsection{Precomputed 3D spatial signals}
\label{appendix:aria:mps}

Project Aria's machine perception service (MPS) provides software building blocks that simplify leveraging the different modalities recorded. These functionalities are likely to be available as real-time, on-device capabilities in future AR- or smart-glasses. We use the following core functionalities and include their raw output as part of the dataset. See Figure~\ref{fig:appendix:aria:mps}.  See \citep{aria_2023} for more details. %
\begin{itemize}
    \item \textbf{Calibration:} All sensors are intrinsically and extrinsically calibrated. MPS also provides time-varying online-calibration that corrects for tiny deformations due to temperature changes or stress applied to the glasses frame.
    \item \textbf{Aria 6\,DoF localization:} Every recording is localized precisely and robustly in a common, metric, gravity-aligned coordinate frame, using a state-of-the-art VIO and SLAM algorithm. This provides millimeter-accurate 6\,DoF poses for every captured frame, as well as high-frequent (1\,kHz) motion in-between camera frames. 

\item \textbf{Eye gaze:} The gaze direction of the user is estimated as a single outward-facing ray anchored in-between the wearer's eyes. We use an optional eye gaze calibration procedure, where the mobile companion app directs the wearer to gaze at a pattern on the phone screen while performing specific head movements. This information was then used to generate a more accurate eye gaze direction, personalized to the particular wearer. 

\item \textbf{Point clouds:} A 3D point cloud of static scene elements is triangulated from the moving Aria device, using photometric stereo over consecutive frames or left/right SLAM camera. The output contains both the 3D point clouds as well as the raw, causally computed, 2D observations of every point in the camera images.

\item \textbf{GoPro 6\,DoF localization:} For Ego-Exo4D, we additionally built functionality on top of the existing Aria MPS functionality, specifically to localize the static GoPro cameras. To achieve this, we use the map built with Aria's SLAM cameras, and perform 6 DoF localization of GoPro frames on the map. 
To obtain the GoPro calibration, we manually calibrated one device in the lab to obtain default parameters, and then use the P4P~\citep{kukelova2016efficient} algorithm (with RANSAC to reject matching outliers) to estimate the 6 DoF pose, as well as re-estimate the focal length to compensate for possible calibration variation between devices.
\end{itemize}

\subsubsection{Recording procedure}

Our recording procedure involves setting up the static GoPros on tripods in locations generally consistent within each scenario, conducting a walk-around with the Aria to build a basemap for 3D reconstruction and camera localization, displaying QR codes at the start/end to assist time sync, and showing a take separation QR between each take.
The camera rig was extended for certain sites with additional mounts and GoPros, discussed below.

Specifically, to sync cameras, we employ a pre-rendered sequence of QR Codes (\emph{i.e.}, QR code video) that encode a wall-clock time. We show this QR code video using the smartphone at 29fps to all cameras in sequence and exploit the difference in frame rates to finely sync the cameras. An additional stage of manual verification ensures each GoPro camera was within 1 frame (+-16.66ms) of the Aria RGB camera.
To amortize the setup and tear down time required for each recording, we record multiple `takes' (\emph{i.e.}, one instance of a certain task) back-to-back and use a `Take Separator' QR code to separate takes in post-processing.

\subsection{Domains and environments}\label{sec:domains}

\input{sec/3_dataset-domains}

\subsection{Participants}\label{sec:participants}

Next we describe the participants who wore the egocentric cameras in Ego-Exo4D.

\paragraph{Credentials and expertise}

We recruited \numcamerawearers total participants from the local communities of \numlabscollecting labs. 
All scenarios feature real-world experts, where the camera-wearer participant has specific credentials, training, or expertise in the skill being demonstrated. 
For example, among the Ego-Exo4D camera wearers are professional and college athletes; jazz, salsa, and Chinese folk dancers and instructors; competitive boulderers; professional chefs who work in industrial-scale kitchens; bike technicians who service dozens  of bikes per day.
Many of %
them have (individually) over 10 years of experience. 

Experts are prioritized given they are likely to conduct activities without mistakes or distractions, providing a strong ground truth for how to approach a given task. However, we also include capture from people with varying skill levels, as well---essential for our proposed skill proficiency estimation task (Section~\ref{sec:proficiency}).  
Notably,  Ego-Exo4D represents human intelligence in a new way by capturing domain-specific expertise---both in the video as well as the accompanying expert commentary (see Section~\ref{sec:language})---portraying the evolution of a skill from beginners to experts.

\paragraph{Demographics}

\begin{figure}[t]
\centering
    \includegraphics[width=1\linewidth]{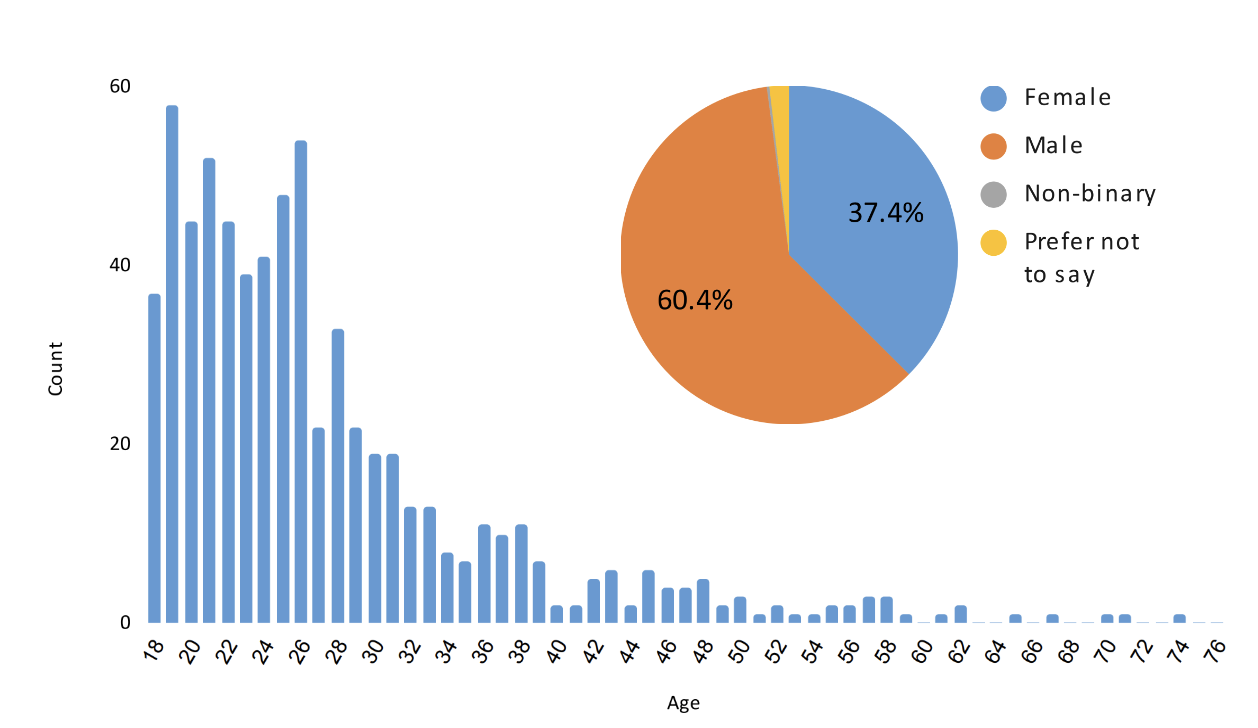}
    \caption{Camera wearer participants' self-reported demographic information (age and gender)}
    \label{fig:demographics}
\end{figure}

The camera wearers range in age from 18 to 74 years old, with 37\% self-identifying as female 60\% male and 3\% as non-binary or preferring not to say. See Figure~\ref{fig:demographics}.
In total, the participants self report more than 24 different ethnicities.\footnote{Sharing this information was optional for all research subjects. Ethnicity is reported based on location specific categories as defined by the relevant partner lab. No such information was gathered from research subjects participating in our collections in California, New York, and Pittsburgh, Pennsylvania.}   Details are in Appendix~\ref{sec:appendix-people}.

\begin{figure}[t]
    \centering
\hspace*{-0.175in}
    \begin{tabular}{c}
    \begin{tabular}{cc}
    \includegraphics[width=0.66\linewidth]{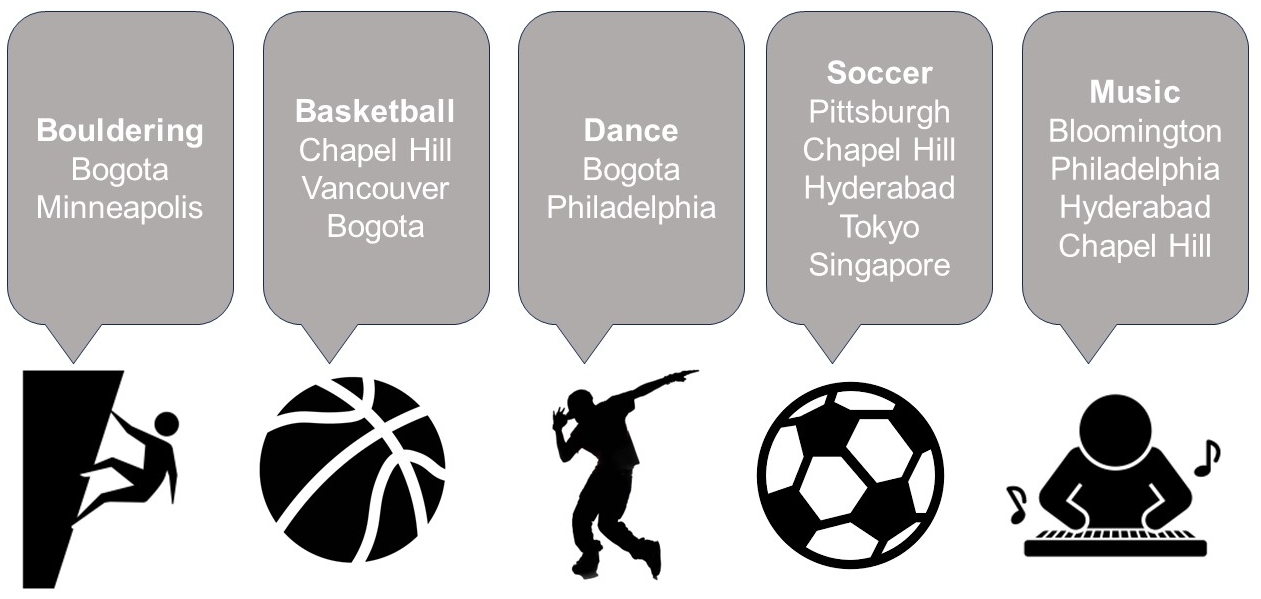}
        \includegraphics[width=0.4015\linewidth]{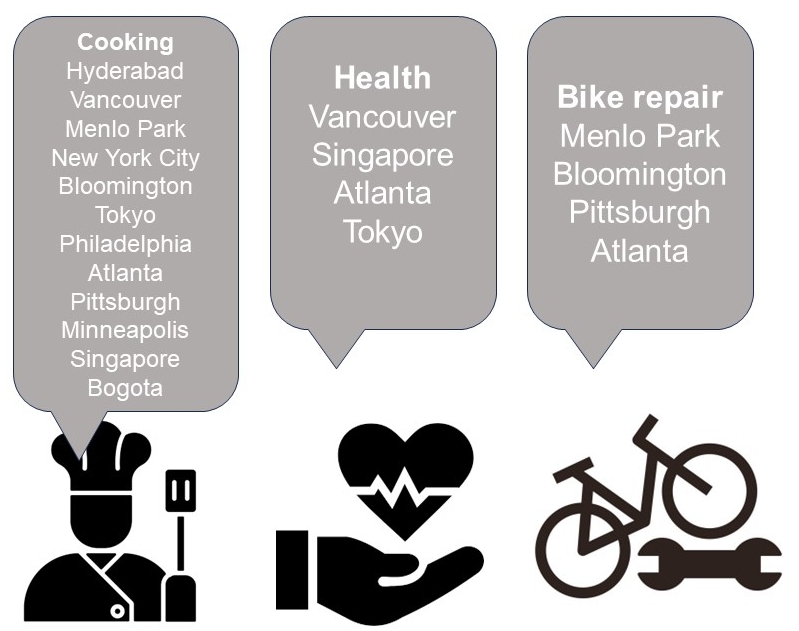}
        \end{tabular}\\
        \includegraphics[width=0.88\linewidth]{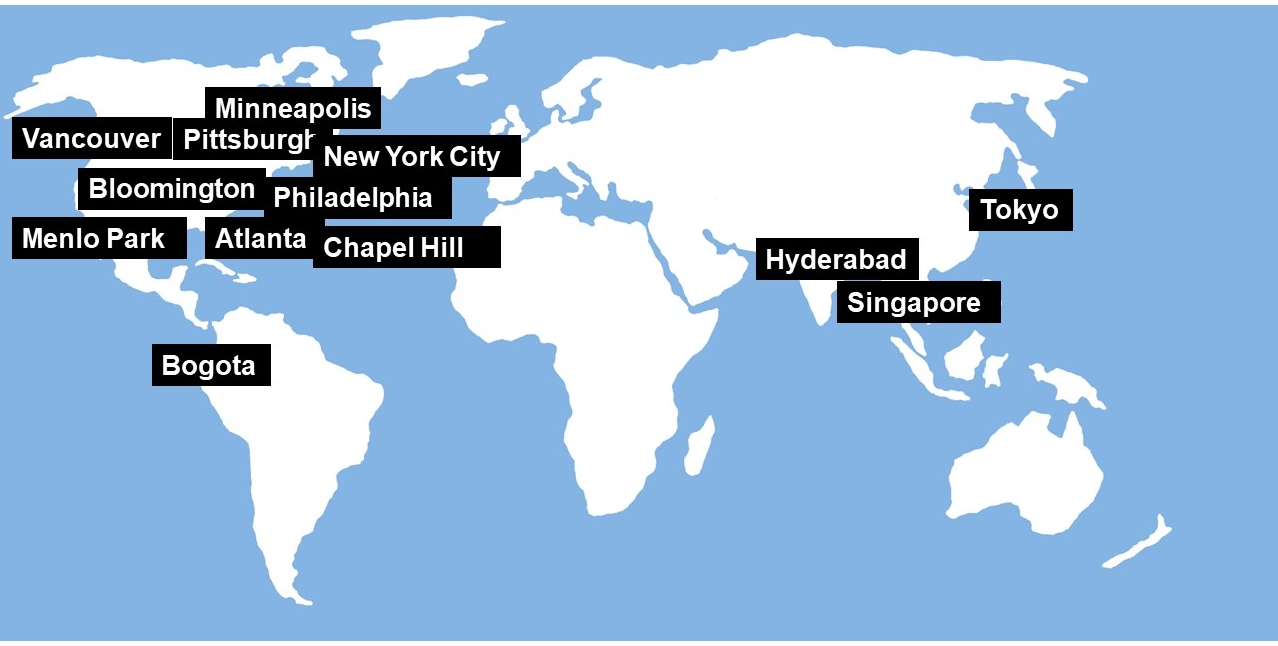}
        \end{tabular}
    \caption{Geographic coverage of Ego-Exo4D and breakdown of which scenarios are captured in which cities.  Note that even within a given city, there may be multiple sites (e.g., multiple bike repair shops or kitchens in the same city).  %
    }
    \label{fig:scenarios-per-university}
\end{figure}

\paragraph{Recruiting}

To recruit participants, each partner institution chose its own approach.  This included using campus email lists, flyers in coffee shops, word of mouth to  family and friends, online ads, posts on social media, university communication channels, temp hiring agencies, and connecting with schools, gyms, and teams, e.g., soccer schools, climbing gyms, professional and university athletics organizations.

\paragraph{Characterizing skill levels}

To ensure consistent, high quality annotations for our benchmarks (discussed below), we identified three crucial pieces of  information about the participants that would be difficult to capture with third-party annotators: the participant's \textit{skill} level, the \textit{objects} they are using, and the \textit{actions} they are completing.
We captured the participants' perceived skill level and performance using pre-task and post-task surveys, respectively, which are available with the Ego-Exo4D dataset.  

For the pre-task survey, we ask 10 questions like, ``how many years have you been doing this task?" and ``have you taught this activity to others before?" (details in Table~\ref{tab:questionnaire_qs} in Appendix~\ref{sec:appendix-people}).  These questions are designed to be more easily quantifiable than simply asking participants to self-rate their skill level.\footnote{We also obtain proficiency ratings for the participants via our expert commentators (cf.~Section~\ref{sec:commentary}).}

For the post-task survey, we ask the participant to reflect on how well they did the task, with questions like, ``what mistakes or errors did you make?" and ``did it take longer or shorter than your initial expectation and why?".    Finally, to capture the \textit{actions} and  \textit{objects} with which they interact, we ask participants to perform a round of first-person narrations called ``narrate-and-act" (detailed below in Section~\ref{sec:language}).

\subsection{Compliance with ethical standards}

Ego-Exo4D was collected following rigorous privacy and ethics standards.  This included undergoing formal independent review processes at each institution to establish the standards for collection, management, and informed consent.  Similarly, all Ego-Exo4D data collection adhered to the \href{https://www.projectaria.com/community-guidelines/}{Project Aria Research Community Guidelines} for responsible research. Since the scenarios allow for closed environments (e.g., no passerbys) nearly all video is available without de-identification. %
For information about each individual partner's protocols and restrictions, please see Appendix~\ref{sec:appendix-collection}.  
Ego-Exo4D data is gated behind a license system, which defines permitted uses, restrictions, and consequences for non-compliance.

%% file: sec/3_dataset-domains.tex
\begin{figure*}[t]
    \centering
    \includegraphics[width=1\linewidth]{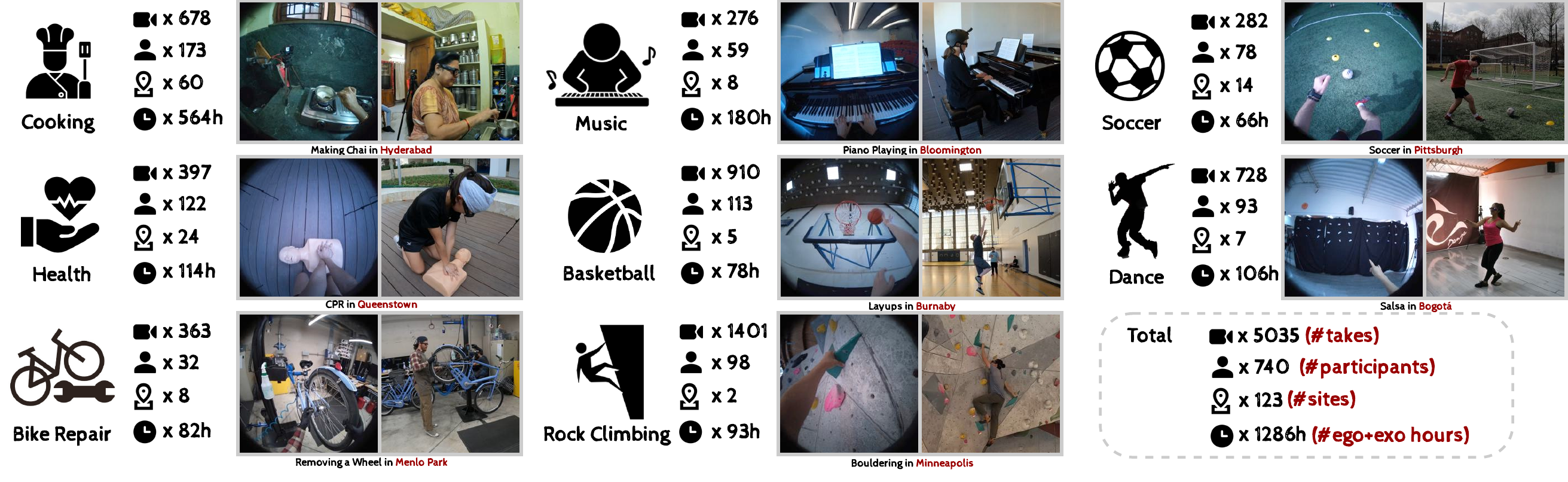}
    \caption{Ego-Exo4D captures skilled activity from \KGCR{43 tasks and 689 keysteps in} 8 domains, in a wide variety of \numscenes scenes in \numlocations cities in Japan, Colombia, Canada, India,  Singapore, and 7 US states.  Each domain is captured at multiple sites---from 2 to 64 unique locations.  In total the dataset offers \numhourstotal hours of ego+exo video comprised of \numtakes takes from \numcamerawearers camera wearers. An average take is \KGCR{2.6} minutes. %
    }
    \label{fig:scenarios_v2}
\end{figure*}

Ego-Exo4D focuses on \emph{skilled human activity}.  This is in contrast to existing ego-only efforts like Ego4D~\citep{grauman2022ego4d}, which has a broad span of daily-life activities.  We intentionally select the domains based on a few criteria: Will it illustrate skill and a variety of expertise? 
Do we have access to real-world settings and participants for that scenario?  
Is there visual variety to be expected across different instances? Will the ego and exo views offer complementary information?  Will it present new challenges unaddressed by current datasets?  %
Overall, by scoping to certain domains, we aim to build up sufficient density of data within a core set of skills for training and evaluating models.

\paragraph{Physical and procedural activities}

Intersecting these criteria, we arrived at two broad categories\footnote{Note that in general physical and procedural are not mutually exclusive labels.  An activity can both require physical skill and procedural steps.}  of skilled activity: \emph{physical} and \emph{procedural}, together comprising eight total domains.   The physical domains are soccer, basketball, dance, bouldering, and music.  They emphasize body pose and movements as well as interaction with objects (e.g., a ball, musical instrument).  The procedural domains are cooking, bike repair, and health care.  They require performing a sequence of steps to reach a goal state (e.g., a completed recipe, a repaired bike) and generally entail intricate hand-object manipulations with a variety of objects (e.g., bike repair tools; cooking utensils, appliances, and ingredients). All domains entail regular attention shifts (revealed by head pose and gaze) by the participant.
Figure~\ref{fig:scenarios_v2} summarizes the eight domains with example frames and data statistics for each.  
In Section~\ref{sec:benchmarks} we discuss how the domains relate to the proposed benchmarks.

\begin{table*}[!thb] 
\centering
\footnotesize{
\begin{tabular}{|ll|ll|}
\hline
\textbf{Procedural} & & \textbf{Physical} & \\
\hline
\emph{Cooking}: & \emph{Health}: & \emph{Music}: & \emph{Bouldering}: \\
- Omelette & - COVID test & - Violin & - V0 through V10 \\
- Scrambled eggs & - Cardiopulmonary & - Piano & \\
- Tomato and egg & Resuscitation (CPR) &  - Guitar & \\
- Sesame-ginger Asian salad & & &  \\
- Greek salad & \emph{Bike repair:} & \emph{Basketball:} & \emph{Soccer}: \\
- Dumplings & - Remove/install a wheel & - Mikan layup drill & - Freestyle dribbling \\
- Noodles & - Replace an inner tube & - Righthand reverse layup & - Freestyle juggling \\
- Pasta & - Clean and lubricate the chain & - Mid-range jump shot & - Penalty kicks \\
- Sushi roll & - Adjust rear derailleur & & \\
- Samosa &  (both limit screws & & \emph{Dance}: \\
- Coffee latte &  \& indexing) & & - Easy choreography \\
- Chai tea & & & - Advanced choreography \\
- Milk & & & \\
- Cookies & & & \\
- Brownies & & & \\
\hline
\end{tabular}}
\caption{The 43 specific activities collected for the three \emph{procedural} and five \emph{physical} domains}
\label{tab:scenarioslist}
\end{table*}

In total, we have 43 activities derived from the eight domains.   For example, cooking is comprised of 14 recipes; soccer is comprised of 3 drills, and music is comprised of 3 instruments.  See Table~\ref{tab:scenarioslist}.  Those 43 activities break down further into 689 total unique keysteps.  The length of a take ranges from 8 seconds to 42 minutes, with procedural activities like cooking having the longest sustained captures.  

\paragraph{Distribution of activities per site}

To achieve visual diversity in the data, multiple labs across our team (typically 3-5) captured each Ego-Exo4D domain.  Figure~\ref{fig:scenarios-per-university} shows the breakdown of which scenarios were captured by each partner institution as well as a map highlighting the locations of the \numlabscollecting labs involved in data collection.\footnote{An additional four institutions not shown on the map are part of the Ego-Exo4D consortium (e.g., contributing to benchmarks) but did not collect data.  They are UT Austin (USA), KAUST (Saudi Arabia), University of Catania (Italy), and University of Bristol (UK).}  The domain selection per site is based on the lab's own preferences and local opportunities to capture data of these scenes at scale.  
Cooking is our one cross-cutting domain, collected at each site. We identified cooking as a priority domain because it resonates around the world as a human need and interest. 
In total, the cooking scenario of Ego-Exo4D contains more than \KGCR{650} takes of cooking performed by more than \KGCR{170} chefs in \KGCR{60} different environments around the world, forming nearly \KGCR{100} hours of ego video alone.

\paragraph{Authentic environments for capture}

\begin{figure*}[tp]
    \centering
    \includegraphics[width=1\linewidth]{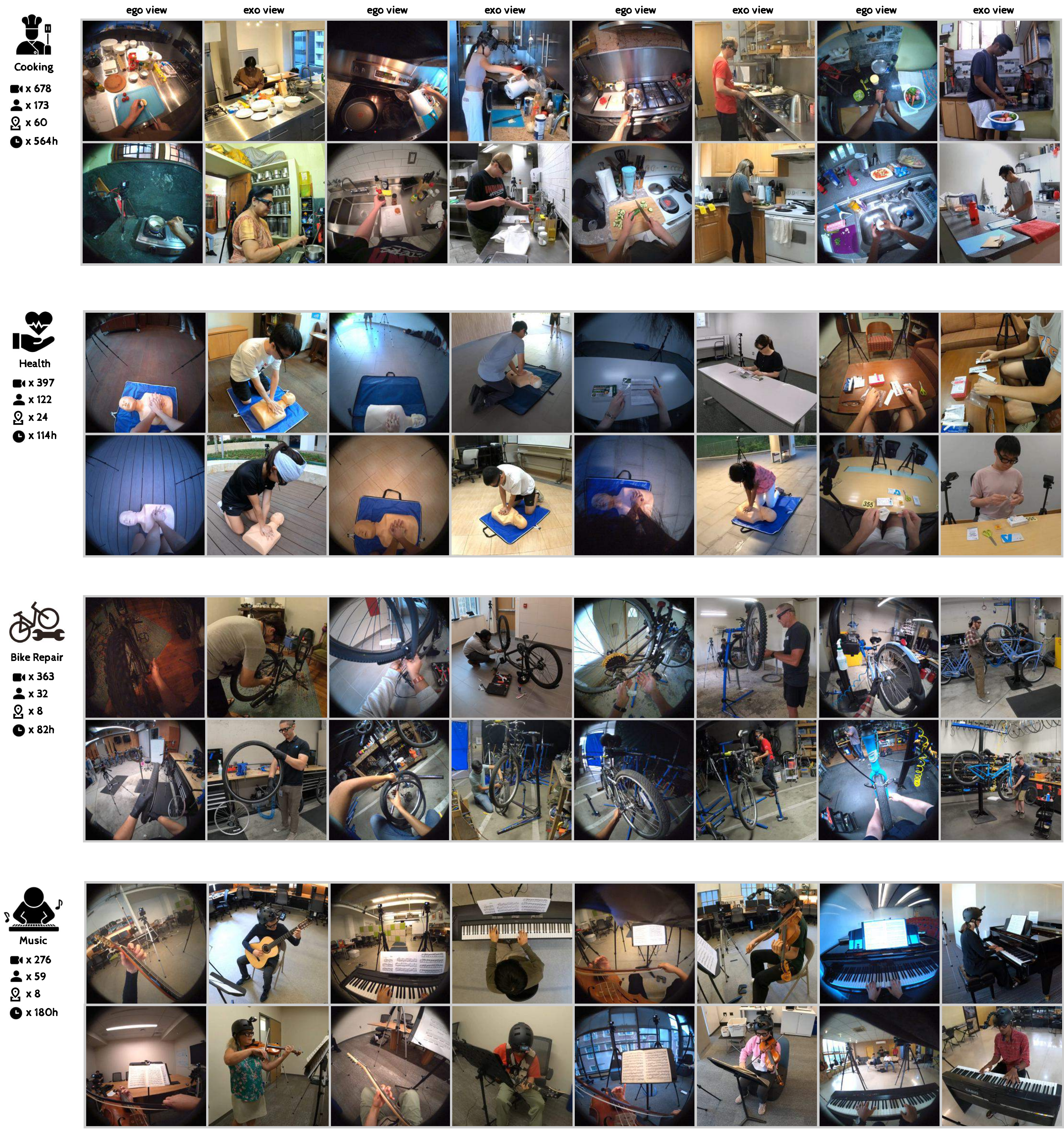}
    \caption{Ego-Exo4D captures skilled activity from 8 domains, in a wide variety of \numscenes scenes in \numlocations different cities in Japan, Colombia, Canada, India,  Singapore, and 7 US states.  Every odd column shows an ego view, and the adjacent even column shows one of its paired exo views.
    }
    \label{fig:big-scenarios-1}

\end{figure*}

\begin{figure*}[tp]
    \centering
    \includegraphics[width=1\linewidth]{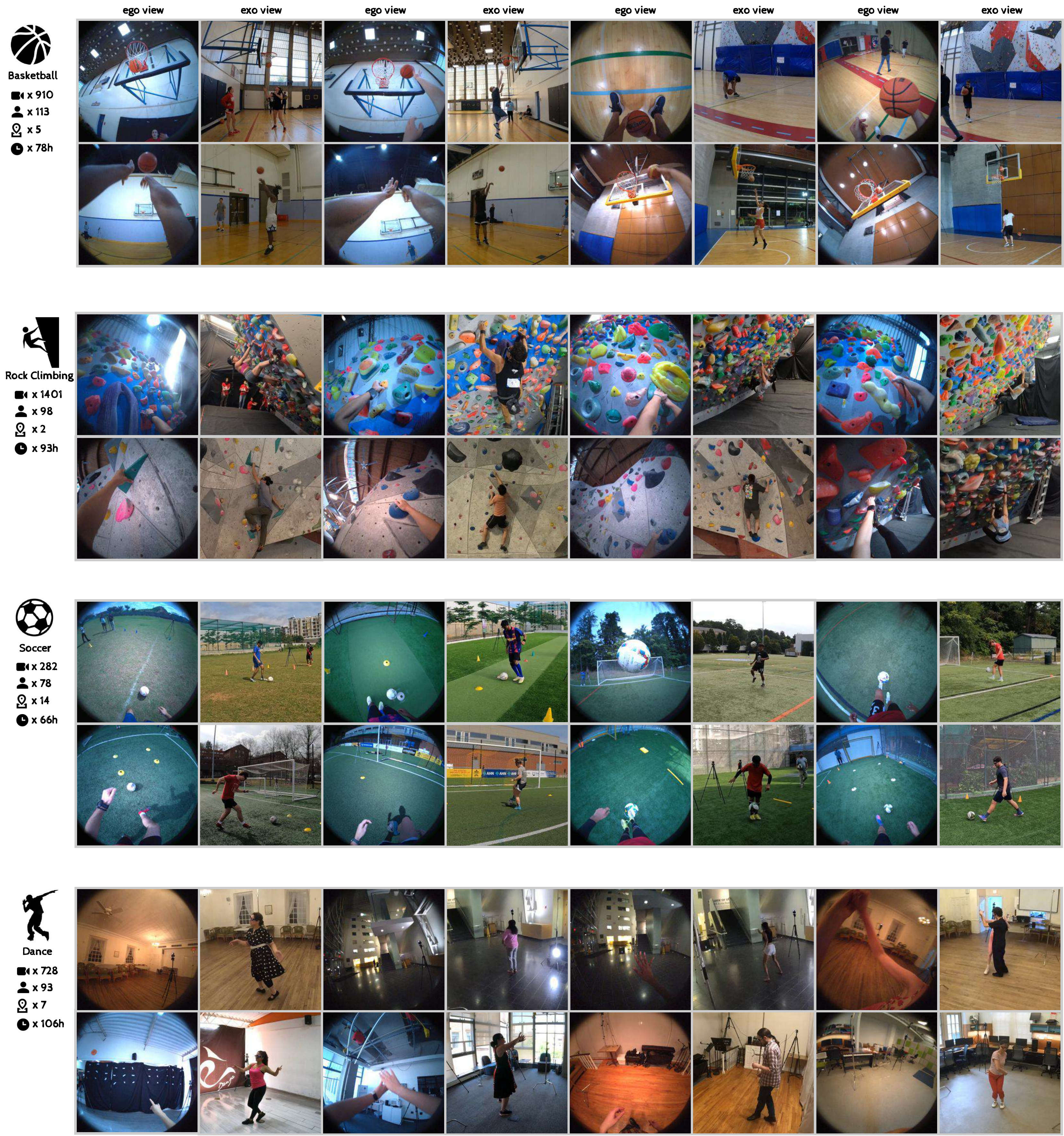}
    \caption{Ego-Exo4D captures skilled activity from 8 domains, in a wide variety of \numscenes scenes in \numlocations different cities in Japan, Colombia, Canada, India,  Singapore, and 7 US states.  Every odd column shows an ego view, and the adjacent even column shows one of its paired exo views.
    }
    \label{fig:big-scenarios-2}
\end{figure*}

The data is collected in authentic settings---such as real-world bike shops, soccer pitches, or bouldering gyms--—as opposed to lab environments.  
Since every domain is covered by more than one lab, the dataset exhibits visual variety from the different physical locations.  For example, we have videos of chefs in New York City, Vancouver, Philadelphia, Bogota, and others; soccer players in  Tokyo, Chapel Hill, Hyderabad, Singapore, and Pittsburgh.  Furthermore, even within the captures done by a single lab, there are often multiple different sites used for filming (e.g., a couple different bike shops in the same city).  Figure~\ref{fig:big-scenarios-1} and~\ref{fig:big-scenarios-2} shows example frames illustrating the variety of the sites and tasks.

\paragraph{Domain-specific collection guidelines}

To ensure consistency across the dataset, we developed data collection guidelines for each domain. These guidelines describe the recommended camera positioning, instructions for participating camera wearers, along with important context-specific considerations. For example, given privacy concerns, our health guidelines required data collection participants to discard COVID tests before results were visible. The guidelines provide general parameters from which to collect data; however, they were not rigid steps. 
Indeed, to support diversity and implementation at a global scale, there are site-specific nuances, for example, differing standards for the implementation of CPR, cultural differences in the ingredients used for different targeted dishes, and location-specific bouldering routes. 

The primary domain-specific design decision is the placement of the exocentric cameras.  The best visibility points for a given domain depends on the general scene and objects involved (open soccer field vs. cluttered kitchen with cabinets) as well as how the person interacts with the space.  For example, for soccer recordings, an exo camera placed in the goal gives great visibility of the players' shots, while in dance or music, an overhead exocentric (or downward pointing egocentric) camera captures important close-to-body detail about the participants' arms and hands.   Generally the exo cameras were placed to ensure viewpoint coverage and achieve the complementary hand-object near-field interactions as well as the participants' full-body movements.


Appendix~\ref{sec:appendix-collection} describes the data collection details that are specific to each consortium partner site, e.g., how they recruited participants, which of the domains and activities they captured, or any modalities they added on top of the common rig.

%% file: sec/4_language.tex
\begin{figure*}[t]
    \centering
    \includegraphics[width=1\linewidth]{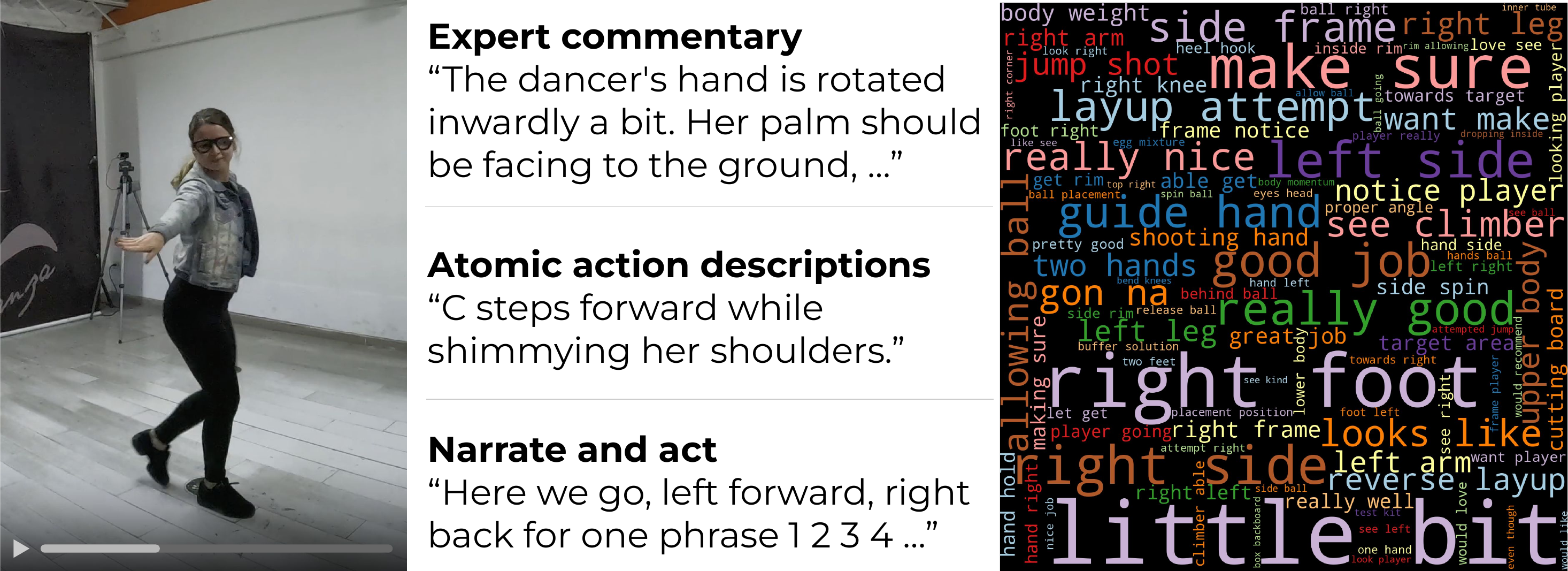}
    \caption{Ego-Exo4D offers 3 paired video-language corpora. Word cloud is from expert commentary which critiques the performance. %
    }    \label{fig:narrations}
\end{figure*}

\section{Natural language descriptions}\label{sec:language}

Ego-Exo4D also offers three kinds of paired natural language datasets, each time-indexed alongside the video: expert commentary, narrate and act, and atomic action descriptions. 
See Figure~\ref{fig:narrations} and Figure~\ref{fig:appendix:annotations:fig} for examples of each language type  highlighting their distinctions in style and point of view.

These language annotations are not steered towards any single benchmark, but rather are a general resource 
that will inspire new language-vision possibilities, such as grounding actions and objects, self-supervised representation learning, multimodal embeddings, video-conditioned language models, and skill assessment from video.  
We also anticipate the temporally grounded descriptions to be valuable for
pre-training foundation models~\citep{kevin2022egovlp, pramanick2023egovlpv2} or automated video captioning~\citep{pan2020spatio, iashin2020multi, zhao2023learning}, both of which are rapidly growing areas of research.  
Furthermore, the time-anchored aspect of the three language annotations provides the opportunity to retrieve time points in specific Ego-Exo4D videos that correspond to queried moments, actions, or phrases.  Finally, they are valuable to mine the dataset for the distribution of objects and activities present, e.g., for taxonomy formation.

\subsection{Expert commentary}\label{sec:commentary}

The first language dataset is spoken \emph{expert commentary}.  The goal is to 
reveal nuances of the skill that are not always visible to non-experts.  
We recruited \numexperts experts (distinct from the participants) to critique the recorded videos, call out strengths and weaknesses, explain how the specific behavior of the participant (e.g., hand/body pose, use of objects) affects the performance, and provide spatial markings to support their commentary.   We provide both the transcribed speech and the raw audio (interesting for its inflection and non-word utterances), as well as the experts' spatial drawings and numeric ratings of each participant's skill.  

\KGupdate{These commentaries are quite novel: they focus on \emph{how} the activity is executed rather than \emph{what} it entails, capturing subtle differences in skilled execution.  We believe this can unlock new fundamental problems (e.g., proficiency estimation below) and disruptive future applications (e.g., AI coaching).}

In the following, we describe the qualifications and background of our experts, followed by the commentary instructions and scope.

\paragraph{Experts' qualifications}

The \numexperts experts are not only well-credentialed in their areas of expertise, but also have coaching or teaching experience.  When recruiting the experts, our selection criteria focused on technical skills, communication, and performance during a live video commentating exercise.

\begin{figure}[t] 
    \centering
    \includegraphics[width=\linewidth]{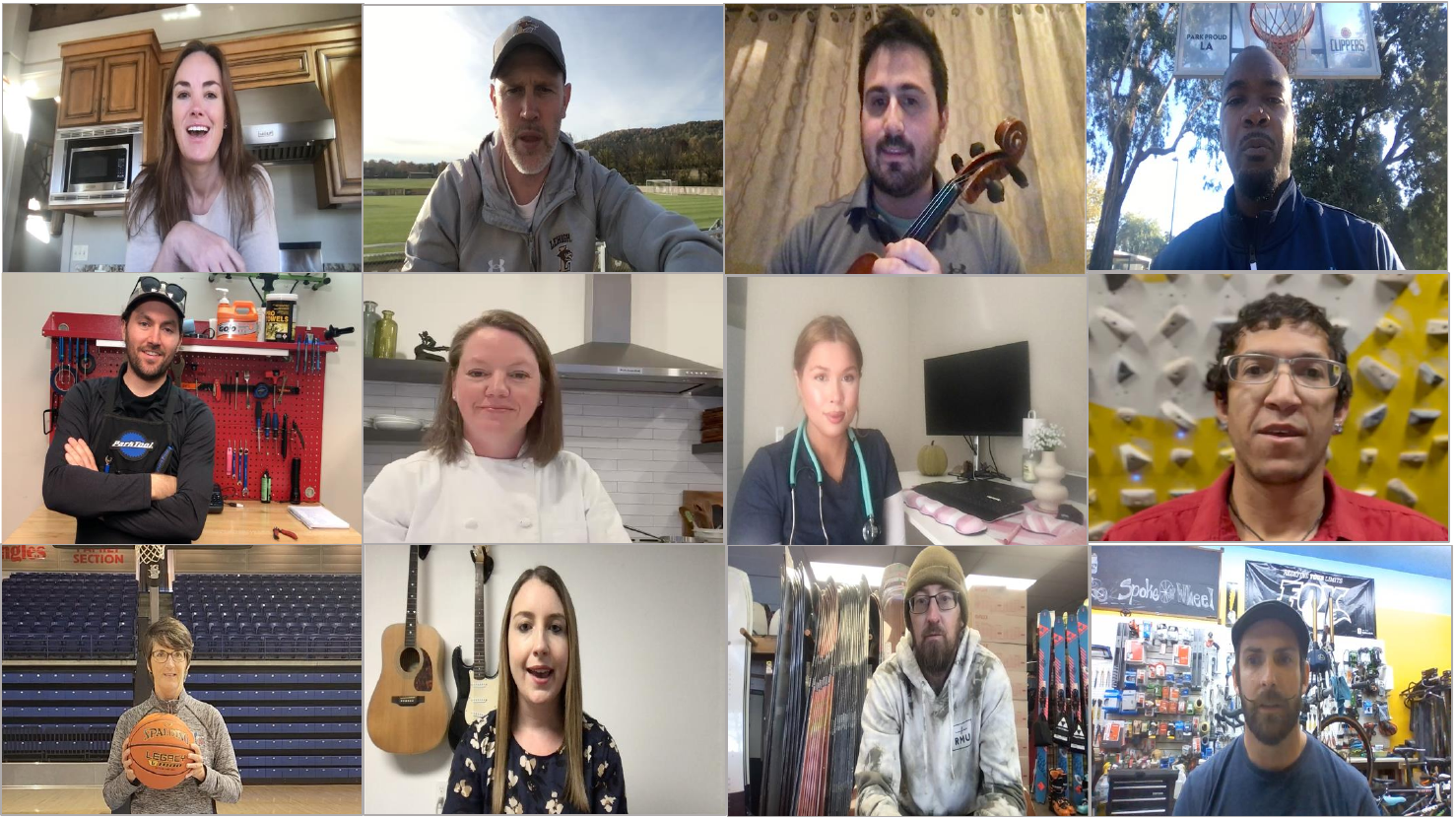}
    \caption{The \numexperts experts who perform the expert commentary language annotations are highly trained in the domain they are commentating, and they often have professional coaching or teaching experience.}
    \label{fig:expert-faces}
\end{figure}

On average, 90\% of the recruited experts possess more than 10 years of professional experience and all have served during this time in the capacity of a coach, instructor or mentor. All experts further have either an advanced degree in their domain of focus or an industry certification. Certification authorities include the US Soccer Federation, the American Culinary Federation, USA Climbing, the American Red Cross, Trek Bikes, and New York State’s Initial Certification in Teaching Dance, among others.   Multiple individuals were recruited across each domain, with the goal of generating language and expertise diversity. 
Due to employment considerations, all experts are residents of the United States.  See Figure~\ref{fig:expert-faces}.

\paragraph{Expert commentary guidelines}

Experts are provided with two time-synchronized videos of each Ego-Exo4D skills demonstration---one showing the egocentric view and another providing a single exocentric perspective specifically selected by annotators as the view that provides the best visibility on the scene (see Sec.~\ref{sec:atomic}. Experts are first asked to watch the video in full without commenting to gain an understanding of the skills demonstration and plan out important points to note in their commentary. 

\begin{figure*}[!thb]   %
    \centering
    \includegraphics[width=\textwidth]{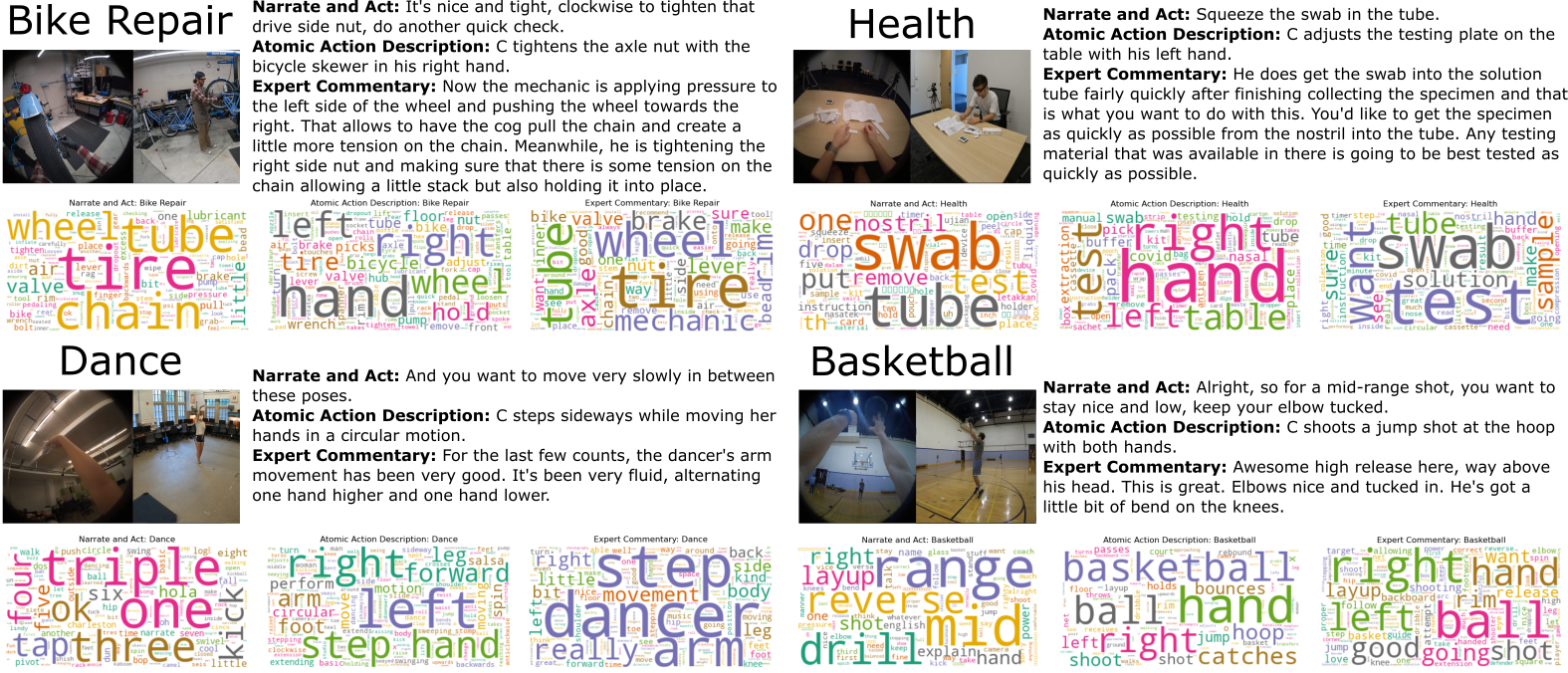}
    \caption{Examples of the three different language annotation styles: narrate and act, atomic action descriptions, and expert commentaries from four of the scenarios (bike repair, health, dance, and basketball). We also include word clouds which highlight the differences in vocabularies per scenario. In narrate and act text, we see how the participant briefly describes what they are doing and why, whereas the atomic action descriptions provide strictly a statement about the visible actions.  The expert commentary offers an expert's critique of what is shown, commenting on strengths and weaknesses and explaining how the participant's actions affect their performance.
    }
    \label{fig:appendix:annotations:fig}
\end{figure*}

Then, the experts watch the video and pause every time they have a comment, typically 7 times per minute of video.  The experts are encouraged to focus on critiques and teaching advice, as opposed to simply describing what the participant is doing.  We record their spoken language descriptions of what is most effective or ineffective about the camera wearer’s actions, the quality of the execution, and mistakes they see.  All commentary is time-anchored and retrospective, focusing on insights and perspectives relating to actions visible up until that point in the video. We choose to collect commentary as verbal recordings in order to maintain the naturalness of the performance descriptions and do so quickly. Each piece of spoken commentary is unbounded in length, and averages 4 sentences. 
 We transcribe the commentaries automatically with OpenAI's Whisper for automatic speech recognition~\citep{radford2023robust}. 
Figure~\ref{fig:appendix:annotations:fig} shows example commentaries; more are available in Table~\ref{tab:example-commentary} in Appendix~\ref{sec:appendix-language}.

\begin{figure}[t] %
    \centering
    \includegraphics[width=\linewidth]{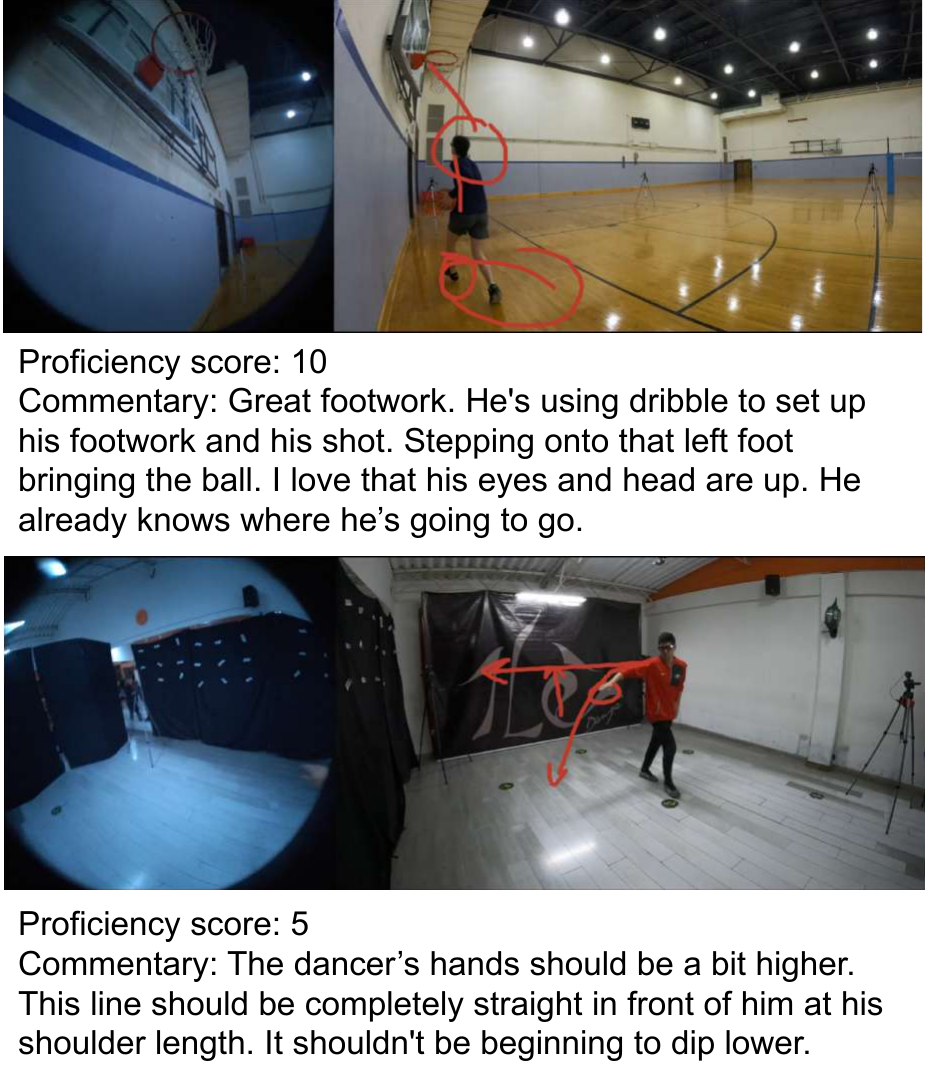}
    \caption{Two examples of expert commentary and proficiency scores, along with spatial drawings (red) done by the expert to augment their spoken comments.}
    \label{fig:expert_commentary_drawing}
\end{figure}

Aside from the spoken commentary, the experts also provide spatial drawings and proficiency scores.  During commentary, experts had the option to use a “telestrator” tool to enhance their commentary with freehand sketches to spatially localize information or otherwise help explain a point (see Figure~\ref{fig:expert_commentary_drawing}).   
They also provide an overall proficiency rating on each video, assessing how well the task was performed with a short written justification.  They score the video on a scale of 1 (least skilled) to 10 (most skilled). In many cases, experts coordinated within their domain group to calibrate this scoring.

Each video has expert commentary by 2-5 distinct experts, offering a variety of perspectives for the same content.  
In total, we have \numexpertcommentaries pieces of time-stamped, video-aligned commentary, the result of more than 6,000 hours of work by the \numexperts experts.
 Overall, we believe the commentary is a unique window into the skilled actions that (through language) surfaces many subtleties about the actions not evident to the untrained eye.

\subsection{Narrate and act}

The second language dataset consists of \emph{narrate-and-act descriptions} provided by the participants themselves. 
They are in the style of a tutorial or how-to video, where the participant explains what they are doing and why. They are reminiscent of the narrations provided in Internet how-to videos, but with less 
with less stylization and without any professional post-production editing.\footnote{For some activities which were more physically intense, such as dancing or bouldering, we asked participants to instead narrate either just before or just after the action to reduce the difficulty of doing this live.}
See Figure~\ref{fig:appendix:annotations:fig} for examples; more are in Table~\ref{tab:example-commentary} in the Appendix.

 Unlike the third-party expert commentary above, these are first-person reflections on the activity given by the people doing them.   Generally the commentary is richer in constructive feedback about the quality of the activity, whereas the narrate-and-act narrations are interesting for their simultaneous nature and first-person analysis of what the participant is doing. 
The behavior in this extra take is expected to differ from that of the non-narrated tasks, in that 
it is likely that the participant will complete the scenario more slowly than normal to concentrate on explaining what they were doing.
These narrations are available for about 10\% of all takes in Ego-Exo4D, since we wanted participants to execute the tasks without pausing for the bulk of the recordings. 
They can potentially be used for multimodal learning as is currently explored in the literature with how-to video narrations
 ~\citep{video-distant,mil-nce,miech19howto100m,ashutosh-neurips2023}.

\subsection{Atomic action descriptions}\label{sec:atomic}

The third language dataset consists of \emph{atomic action descriptions}.  Whereas the  commentary and narrate-and-act language reveals spoken opinions and reasons for the actions (the ``why and how"), this stream of text is specifically about the ``what".  Inspired by Ego4D's \emph{narrations}~\citep{grauman2022ego4d}, \KGCR{these are short statements written by third-party (non-domain expert) annotators, timestamped for every atomic action performed by the participant for all videos in the dataset, for a total of 432K sentences.} %
This data is valuable for mining for taxonomies of objects and actions in the data, indexing the videos with keywords for exploring the dataset, and for future research in video-language learning, as has been quite successful for the Ego4D narrations~\citep{kevin2022egovlp,hiervl,pramanick2023egovlpv2}.

\paragraph{Annotation description} We present each take to the annotators as a collaged video consisting of the egocentric view, left and right grayscale SLAM, four or five fixed-position exocentric cameras, and single-track composite audio; for a subset of videos, a helmet-mounted GoPro view is also available.
Annotators are asked to provide a play-by-play description of what happens, as seen across \emph{any} of the views. 
Potential contents include actions by the camera wearer, other individuals interacting with the camera wearer, and relevant environmental events.

Each narration is atomic and time-anchored: as much as possible, each narration should be limited to one verb and have a single associated timestamp, roughly within a second of its occurrence in the video.
For consistency across narrations, and consistency with Ego4D's narrations, the camera wearer in each take is referred to as ``C" (e.g., ``C picks up a wrench.''). 
Other individuals are referred to by other letters (e.g., ``Man X kicks the soccer ball back to C.''); these letter labels are not necessarily consistent across takes, but refer to the same individual within a take.
Many videos are narrated by two independent annotators, and we make both available.
Figure~\ref{fig:appendix:annotations:fig} shows examples, and more are in Table~\ref{tab:example-commentary} of the Appendix.  See Table~\ref{tab:atomic_stats} in the Appendix for atomic action descriptions summary statistics.

\paragraph{Visibility labels} Because of the multi-view nature of the Ego-Exo4D capture rig, certain actions or events may not be visible across all camera feeds.
While we hope Ego-Exo4D leads to increased attention toward multi-view learning, many existing systems fundamentally assume a single view at a time; if a camera does not have a view of the narrated action or event, this may lead to a confusing learning signal, or pose an impossible ask for a model to infer.

Thus, we also ask that the annotators answer two additional question per narration: 1) an indicator of whether the narration is visible from the egocentric camera, and 2) which (if any) of the static exocentric cameras provide the best view.  
If there are multiple equally good views, annotators are free to pick any.
In particular, we found this \emph{best exocentric view} helpful for other Ego-Exo4D annotation efforts: the narration visibility tags played a role in exocentric view selection for both the correspondences benchmark (Section~\ref{sec:relation}) and expert commentary (above), and frame selection for hand and body pose (Section~\ref{sec:egopose}).

\paragraph{Comparing the language annotations}

How do the statistics from all three forms of language differ?
Overall, expert commentary tends to use a much larger vocabulary and more lengthy statements, since commentators are giving more elaborate statements of advice and explanation.  The temporal density of the atomic action descriptions is greater than the other two forms, since the annotators are pausing to describe every single action of the camera wearer.  Narrate-and-act comments use a vocabulary size in between the other two, reflecting the more free-form speech (compared to the written atomic actions) is used.  Across the different scenarios, the trends are mostly similar, with the most noticeable differences being the temporal density; it is particularly high for both cooking and soccer.  In the former, there are many procedural steps, whereas in the latter there are many instances of the drill being executed. 

See Figure~\ref{fig:appendix:annotations:statistics:fig} in the Appendix for the detailed statistics, and Figure~\ref{fig:appendix:annotations:wordcloud:fig} in the Appendix for word clouds per scenario and annotation type highlighting the differences in vocabulary and word frequency.

%% file: sec/5_benchmarks.tex
\section{Ego-Exo4D benchmark tasks}\label{sec:benchmarks}

\begin{table*}[t]
\small
\centering
\renewcommand{\arraystretch}{1.25}
\resizebox{\linewidth}{!}{
\begin{tabular}{|c|c|cc|}
\hline
\multirow{2}{*}{\bf Benchmark} & \multirow{2}{*}{\bf Annotation Type} &
\multicolumn{2}{c|}{\textbf{Ego-Exo4D (v2)}} \\
 &  & {\bf Num Takes}& {\bf Annotations} \\
\hline
\multirow{3}{*}{Relations} & \multirow{3}{*}{Manual} & \multirow{3}{*}{1335} & 5566 objects \\
 & & & 742K ego masks \\
 & & & 1.1M exo masks \\
\hline 
\hline
\multirow{3}{*}{Keystep recognition} & \multirow{3}{*}{Manual} & \multirow{3}{*}{1088} & 17 activities, 664 keysteps \\
 & & & 27.6K ego segments (87h) \\
 & & & 143K ego+exo segments (454h) \\
\hline
\multirow{2}{*}{Procedure understanding} & \multirow{2}{*}{Manual} & \multirow{2}{*}{\arxivb{628}} & 6 activities, 186 keysteps \\
 & & & \arxivb{8.6K segments (30h)} \\
\hline
\hline
\multirow{5}{*}{Proficiency estimation} & \multirow{2}{*}{Semi-automatic} & \multirow{2}{*}{2987} & {2987 proficiency scores} \\
& &  & (demonstrator)  \\
& \multirow{3}{*}{Manual} & \multirow{3}{*}{912} & 19K ``good'' segments \\
& &   & 20K ``tips'' segments \\
& &   & (demonstration) \\
\hline
\hline
\multirow{2}{*}{Ego pose (Body)} & \multirow{1}{*}{Automatic} & 2559 & 9.2M 3D / 46.87M 2D \\
 & \multirow{1}{*}{Manual} & 1358 & 376K 3D / 2M 2D \\
\hline
\multirow{2}{*}{Ego pose (Hand)} & \multirow{1}{*}{Automatic} & 976 & 4.3M 3D / 21M 2D \\
 & \multirow{1}{*}{Manual} & 458 & 68K 3D / 340K 2D \\
\hline
\end{tabular}
}
\caption{Summary of annotation statistics for the different benchmark tasks of Ego-Exo4D.}
\label{tab:annotation_stats}
\end{table*}

Our second major contribution is to define the core research challenges in the domain of egocentric perception of skilled activity, particularly when ego-exo data is available for training (if not testing).
To that end, we devise a suite of  foundational benchmark tasks organized into four task families: relation (Sec.~\ref{sec:relation}),  recognition (Sec.~\ref{sec:recognition}), proficiency (Sec.~\ref{sec:proficiency}), and ego-pose (Sec.~\ref{sec:egopose}).  

For each task, we provide not only suitable multimodal data, but also
high quality annotations that allow training and evaluating models, as well as baselines that provide a starting point from which the research community can build.  Table~\ref{tab:annotation_stats} overviews the annotations provided with Ego-Exo4D, and Table~\ref{tab:comparison_with_datasets} summarizes key distinctions with existing datasets.  

We ran the first formal teaser Ego-Exo4D challenge in 2024, and will launch the full suite of challenges and leaderboards next year.

In the following sections, for each benchmark task we provide 1) the motivation and applications of solving that task, 2) the formal task definition, 3) key related work, 4) a description of the annotations, 5) an overview of the metrics we use to evaluate the task, 6) the design the baseline models, and 7) their results on the released dataset.\footnote{Note that all results are from ``v2" of Ego-Exo4D released in March 2024.  The smaller v1 is now considered obsolete and should not be used for future publications and comparisons.}

Important: To ensure fair comparisons in any future work using Ego-Exo4D, researchers need to account for 1) the precise task input-output definitions and 2) the train/test/val splits available with the annotations.  Specifically, for each task below, when formally defining the inputs and outputs, we also explicitly specify which inputs are \emph{excluded} from use, if any.

\begin{figure*}[!t]
    \centering
        \includegraphics[width=\linewidth]{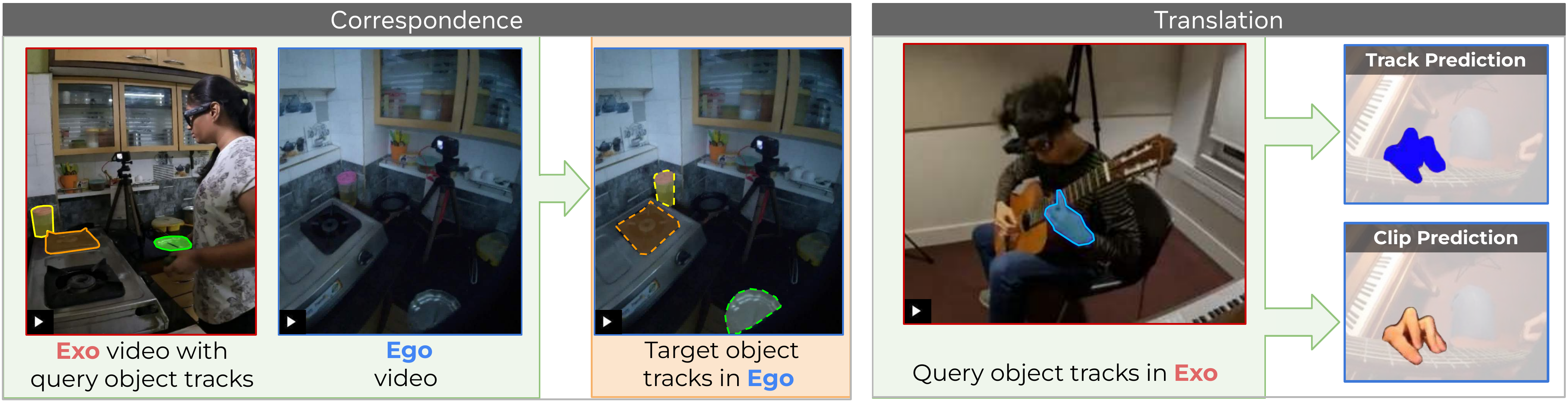}
    \caption{The ego-exo relation family consists of the tasks of correspondence (left) and translation (right). %
    }%
    \label{fig:relation_definition}      
\end{figure*}

\subsection{Ego-exo relation}\label{sec:relation}

Our ego-exo \emph{relation} tasks deal with relating the video content across the extreme ego-exo viewpoint changes.  They take the form of object-level matching (correspondence) and synthesis of one view from the other (translation).   See Figure~\ref{fig:relation_definition}.

\subsubsection{Ego-exo correspondence}
\input{sec/benchmarks/correspondence-benchmark}

\subsubsection{Ego-exo translation}
\input{sec/benchmarks/translation-benchmark}

\subsection{Ego-exo keystep recognition}\label{sec:recognition}

\KGnew{This family of tasks centers around recognizing the keysteps of a procedural activity and modeling their %
dependencies. Specifically, there are three tasks: fine-grained keystep recognition (Sec.~\ref{sec:keystep_finegrained}), efficient multimodal keystep recognition (Sec.~\ref{sec:keystep_energy_efficient}), and procedure understanding (Sec.~\ref{sec:keystep_task_graph}).  We refer to the family of tasks as ``ego-(exo)" since exo may be available at the time of training but not inference.}  See Figure~\ref{fig:recognition_tasks}.

\subsubsection{Fine-grained keystep recognition} \label{sec:keystep_finegrained}
\input{sec/benchmarks/keysteps-benchmark}

\subsubsection{Energy-efficient multimodal keystep recognition} \label{sec:keystep_energy_efficient}
\input{sec/benchmarks/multimodal-benchmark}

\subsubsection{Procedure understanding} \label{sec:keystep_task_graph}
\input{sec/benchmarks/taskgraph-benchmark}

\subsection{Ego-exo proficiency estimation}\label{sec:proficiency}
\input{sec/benchmarks/proficiency-benchmark}

\subsection{Ego pose}\label{sec:egopose}
\input{sec/benchmarks/bodypose-benchmark}

%% file: sec/benchmarks/correspondence-benchmark.tex
\paragraph{Motivation}
Establishing object-level correspondences between ego and exo viewpoints 
would allow AI assistants to
provide visual instructions by matching third-person observations of objects from instructional videos
to those in the user's first-person view.
Compared to the general correspondence problem,  our setting requires tackling a number of challenges: extreme viewpoint differences, high degrees of object occlusion, and many small objects (e.g., cooking utensils and bike repair tools).

\paragraph{Task definition}
Given a pair of synchronized ego-exo videos and
a sequence of query masks of an object of interest in one of the videos,
the task is to identify the corresponding mask for the same object in each synchronized frame of the other view, if visible.  See Figure~\ref{fig:relation_definition}, left. %
The task can be posed with query objects in either the ego or exo video, with both directions presenting interesting challenges (e.g., high degree of occlusion in ego views, and small object size in exo views).
This task is especially challenging in our dataset, since we have to handle long videos with an average length of 3 minutes, as well as very small objects with areas of only a few pixels. 

\NEW{\KGnew{Importantly,} the input to the model \KGnew{\emph{excludes}} semantic labels or names for the objects, camera pose information relating the two views, and IMU or active range sensor measurements.
We do not use such information as we want to encourage the development of methods for open-world correspondence, not relying on predefined sets of objects or inputs that require non-consumer camera devices.}

\paragraph{Related work}
Related tasks
are image-level sparse correspondence given query points (instead of object masks)~\citep{jiang2021cotr} and image-level object co-segmentation~\citep{vicente2011object} for jointly segmenting semantically similar objects. %
Our task goes beyond static object correspondence, since the interplay between human pose and object state changes during manipulation 
necessitate using temporal context and tracking %
as the query object can be highly occluded or blurry~\citep{tang2023egotracks}.

\paragraph{Annotations}
We annotate pairs of temporally synchronized egocentric and exocentric videos with segmentation masks for selected object instances from six scenarios: \emph{Cooking}, \emph{Bike Repair}, \emph{Health}, \emph{Music}, \emph{Basketball} and \emph{Soccer}.
We exclude \emph{Bouldering} and \emph{Dance} from this benchmark as they have limited diversity of objects.
We focus on objects used by the camera-wearer at any point \EM{during the execution of the activity and that are visible in both views for at least some frames of the sequence}. These  masks allow us to define object-level correspondence between the views.
We used a multi-stage annotation process for annotating paired ego-exo videos.
\NEW{There are 1.8M masks annotated at 1fps covering 5.6k objects from 1335 takes. Overall, an average of 5.5 objects are annotated with correspondences between the two views in each take, with each object tracked for an average of 173 frames (excluding frames with occlusions).}
See Appendix \ref{sec:appendix:relation} for details and statistics.

\paragraph{Metrics}
We adopt the following metrics in our evaluation:
\begin{enumerate}
\item \emph{Location Error} (LE), which we define as the normalized distance between the centroids of the predicted and ground-truth masks.
\item \emph{Intersection Over Union} (IoU) between the predicted and ground-truth masks.
\item \emph{Contour Accuracy} (CA)~\citep{perazzi2016benchmark}, which measures mask shape similarity after translation is applied to register the centroids of the predicted and ground-truth masks.
\item \emph{Visibility Accuracy} (VA)~\citep{brodersen2010Balanced}, which evaluates the ability of the method to estimate the visibility of the object in the target view, as in practice it may often be occluded or outside the field of view. We measure this performance using balanced accuracy. Note that, in contrast to the previous metrics that compare segmentation masks at frames where the object is visible in both views, this metric is computed based on all frames with query masks.
\end{enumerate}

\paragraph{Baselines}
Finding object mask correspondences across pairs of videos is an under-explored area in video understanding.
Therefore, we investigate two diverse baseline approaches for our ego-exo correspondence task: (a) a \emph{spatial model} that tackles the correspondence problem independently at each time point, and (b) a \emph{spatio-temporal model} that takes into account the history of predicted correspondences.
\begin{itemize}
  \item \textit{Spatial baseline model.}  This model receives as inputs an egocentric frame, the associated exocentric frame, and a query object segmentation mask in one of the views. It then outputs the mask in the other view (if the object is visible in that view). It can be thought of as a generalization of query-point correspondence approaches proposed for sparse image correspondence~\citep{jiang2021cotr}. We implement this baseline in the form of a Transformer-based image correspondence model, {\em XSegTx} (Cross View Segmentation Transformer), which extends SegSwap~\citep{shen2022learning}, a method originally proposed for image co-segmentation, i.e., for segmenting common objects in a pair of images.\\
  \item \textit{Spatio-temporal baseline model.} The spatio-temporal model receives as input the pair of ego-exo video clips as well as an object segmentation track in one of the views, and outputs segmentation masks in the other view for the frames that the object is visible in both views. It can be thought of as performing generalized tracking across views. We build our baseline model on top of XMem~\citep{cheng2022XMem}, a model originally proposed for tracking a specific target object given its segmentation mask in the first frame. In particular, our baseline model, called {\em XView-XMem}, adapts XMem to track the object across different views given ground-truth segmentation masks for one of the views in each frame.
\end{itemize}
See Appendix \ref{sec:appendix:correspondence} for implementation details. 

\paragraph{Results} We benchmark our XSegTx and XView-XMem baseline models on the test set in Table~\ref{tab:correspondence_results_val}. We experiment with two settings: providing the ground-truth object track in the exo view (exo query mask) and predicting it in the ego view, and vice versa.  %

First, we observe that exploiting temporal cues 
helps with tackling the object correspondence task as shown by the significant increase in performance achieved by the spatio-temporal baselines (ST type) compared to the spatial ones (for example, IoU improves from $13.88$\% to $22.14$\% in the Ego$\rightarrow$Exo setting).

Second, we can see a big difference in performance between the Ego$\rightarrow$Exo and Exo$\rightarrow$Ego settings for all the baselines. In particular, models perform worse when the sequence of query masks is provided for the egocentric video and the model needs to predict query masks in exocentric video. This might be due to the heavy occlusion and very small size of objects in the exocentric views, making segmentation very challenging. While predicting a very tiny mask in the exo view can be very difficult, models can reason about the type and rough location of the object from a tiny mask in the exo view and thus accurately detect and segment it in the ego view, where it is much larger.

However, all our baselines achieve a performance inferior to $23\%$ IoU in the Ego$\rightarrow$Exo setting and inferior to $24$\% IoU in the Exo$\rightarrow$Ego setting. This shows the challenging nature of the task and the dataset. We note that our dataset includes a great degree of object shape variation and high number of very small objects which are very difficult to model.

We also show some qualitative results in Fig.~\ref{fig:corres-qualitative}. As we can see, the spatial baseline (XSegTx) struggles to track the same object throughout the video. For example, in the bottom example, XSegTx alternates between predicting one and two object masks whereas the spatiotemporal baseline (XView-XMem) reliably tracks a single object throughout the sequence, showing the importance of exploiting temporal cues in the data.
See Appendix \ref{sec:appendix:correspondence} for more analysis and visualizations of the results.

\begin{table*}[ht]

\centering
\resizebox{\linewidth}{!}{%
\begin{tabular}{l l l c c c c c}
    \toprule
    Query Mask & Method & Type & Bal. Acc.$\uparrow$ & IoU$\uparrow$  & Location Score$\downarrow$ & Contour Acc.$\uparrow$\\
    \midrule
    Ego & XSegTx (random weights) & S & 50.00 & 0.48 & 0.118  & 0.014 \\
    Ego & XSegTx & S & \underline{66.31} & 18.99 & \underline{0.070} & \underline{0.386} \\
    Ego & XMem (w/o finetuning) & ST & 64.39  & \underline{19.28} & 0.151 & 0.262 \\
    Ego & XView-Xmem (w/ finetuning) & ST & 61.24 & 14.84 & 0.115 & 0.242 \\
    Ego & XView-Xmem (+ XSegTx) & ST & \textbf{66.79} & \textbf{34.90} & \textbf{0.038} & \textbf{0.559} \\
    \midrule
    Exo & XSegTx (random weights) & S & 50.00 & 1.08 & 0.203  & 0.024 \\
    Exo & XSegTx & S & \textbf{82.01} & \textbf{27.14} & \textbf{0.104} &\textbf{ 0.358} \\
    Exo & XMem (w/o finetuning) & ST & 60.35 & 16.56 & 0.160 & 0.204 \\
    Exo & XView-Xmem (w/ finetuning) & ST & \underline{61.72} & 21.37 & 0.139  & 0.269 \\
    Exo & XView-Xmem (+ XSegTx) & ST & 59.71 & \underline{25.00} & \underline{0.117} & \underline{0.327} \\
    \bottomrule
\end{tabular}
}
\caption{Baseline evaluation for the ego-exo correspondence benchmark on test set. Best results are reported in bold, whereas the second best results are underlined.}
\label{tab:correspondence_results_val}
\end{table*}

\input{sec/appendices/figs/correspondence/qualitative}

%% file: sec/appendices/figs/correspondence/qualitative.tex
\begin{figure*}
\centering
\setkeys{Gin}{width=\linewidth}
\begin{tabularx}{\textwidth}{l|XXXXX}
\toprule
Ego GT & \includegraphics{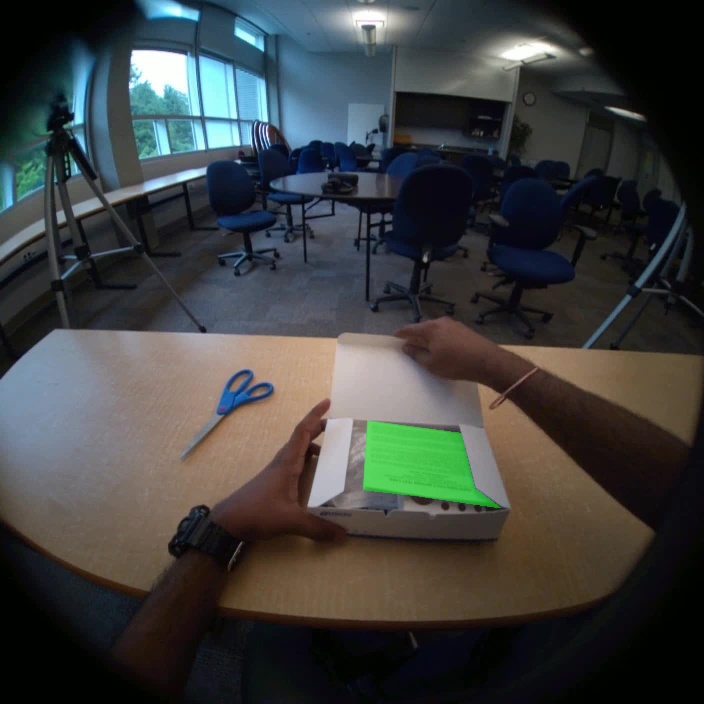} & \includegraphics{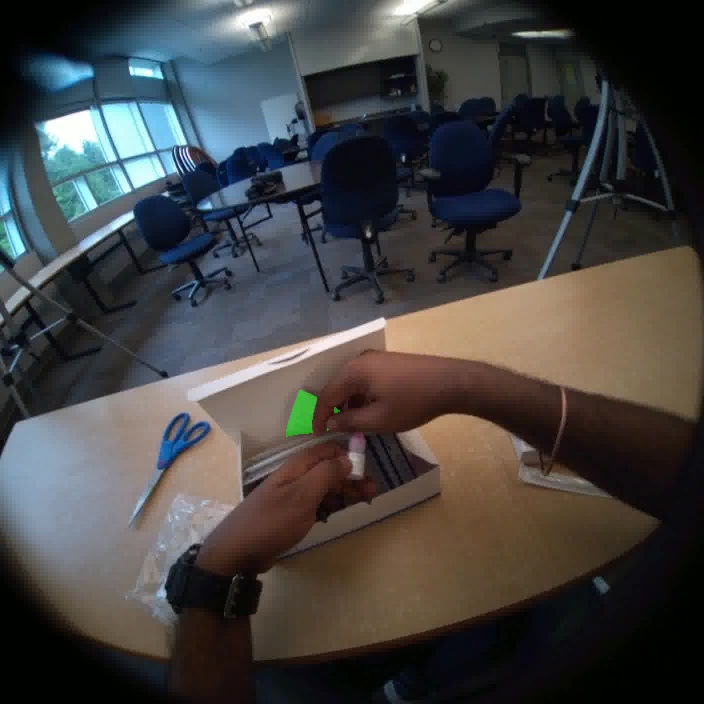} & \includegraphics{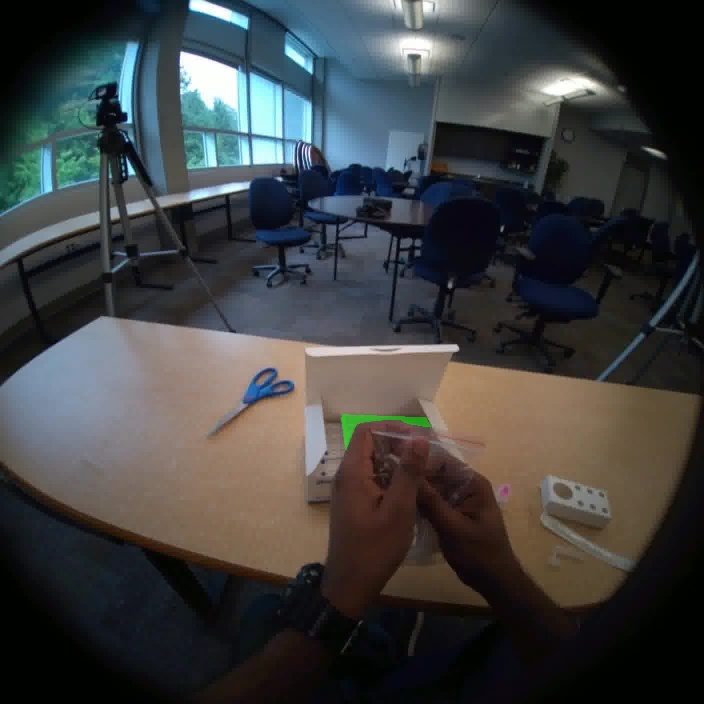} & \includegraphics{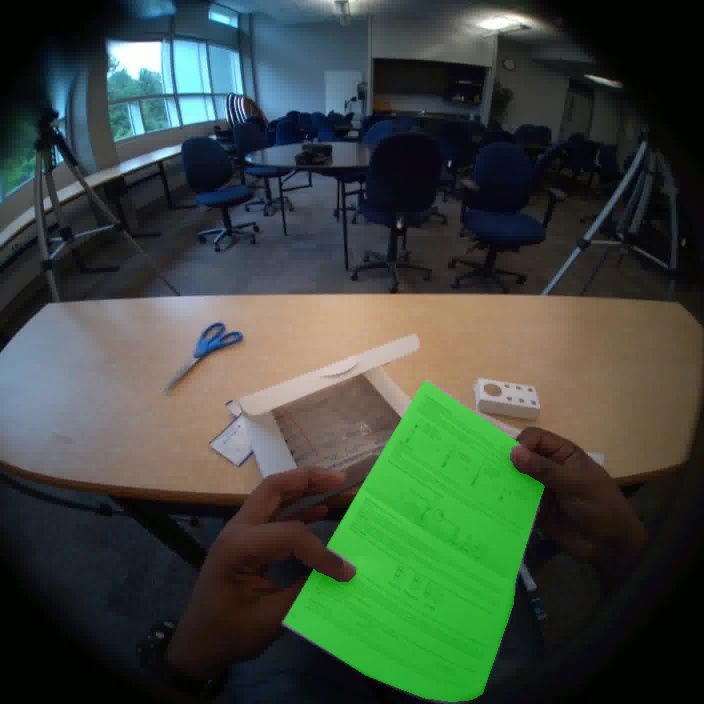} & \includegraphics{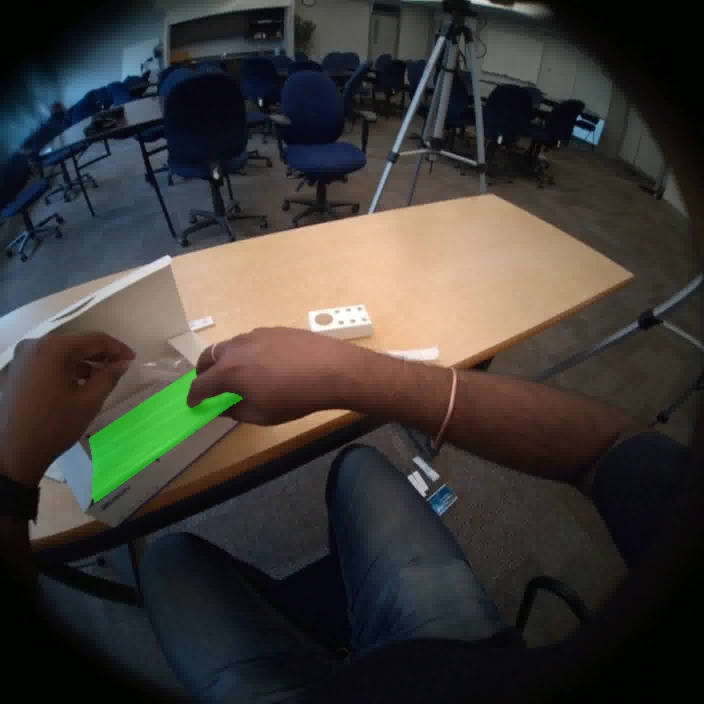} \\

Exo GT & \includegraphics{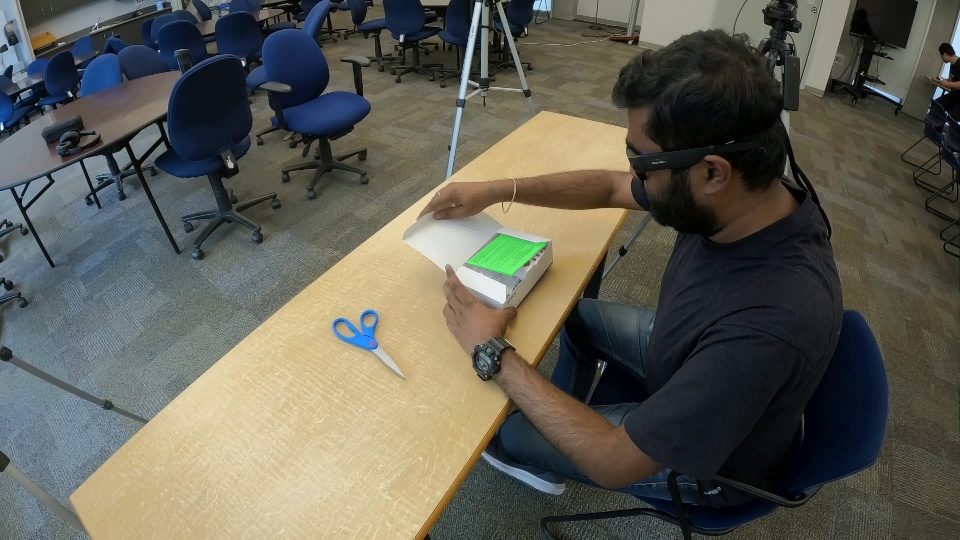} & \includegraphics{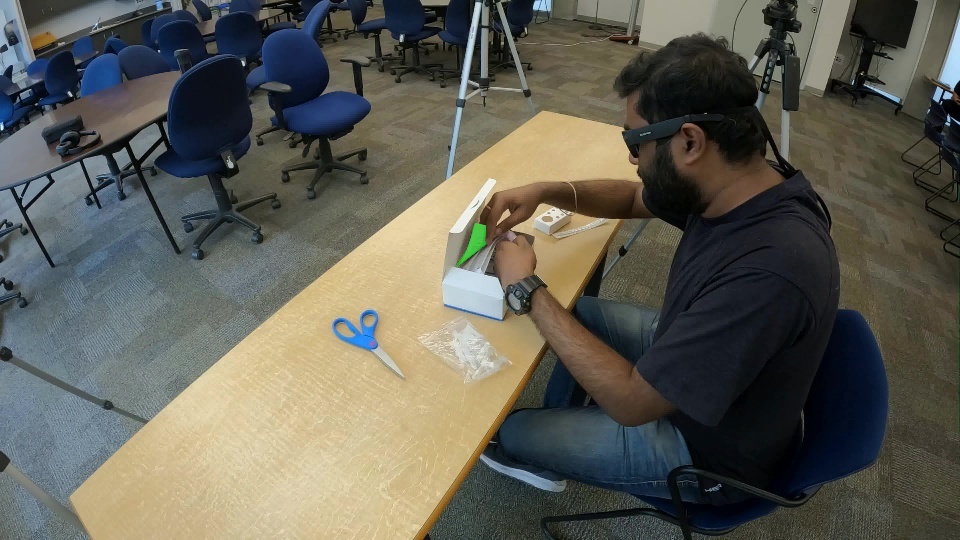} & \includegraphics{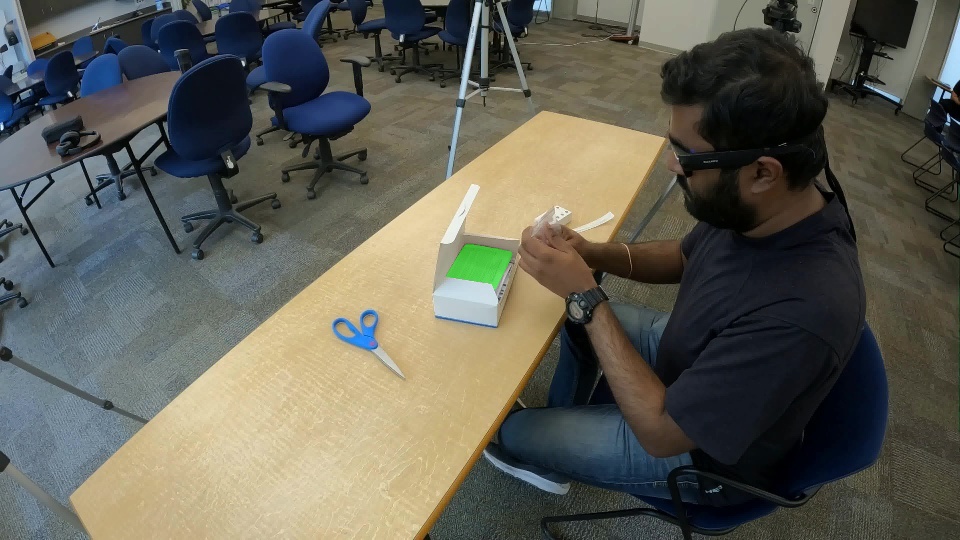} & \includegraphics{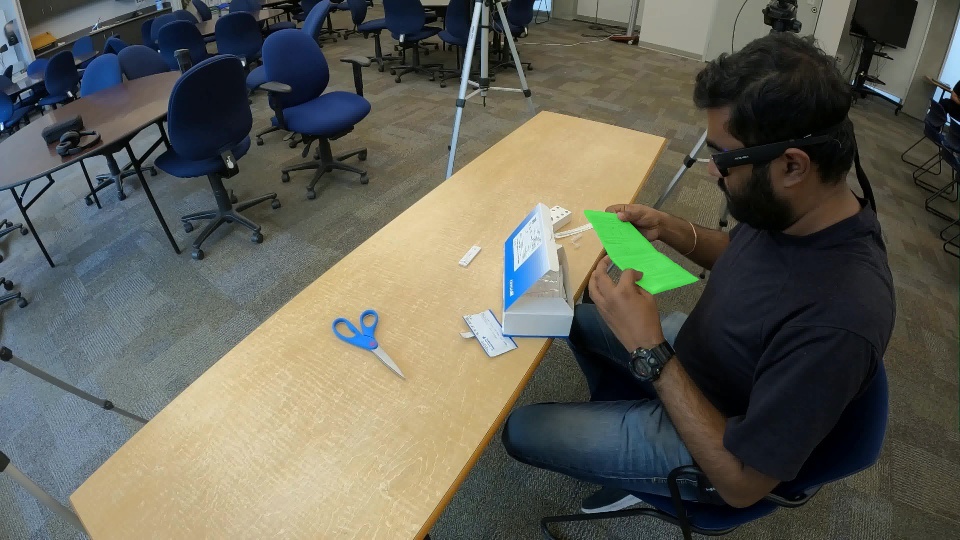} & \includegraphics{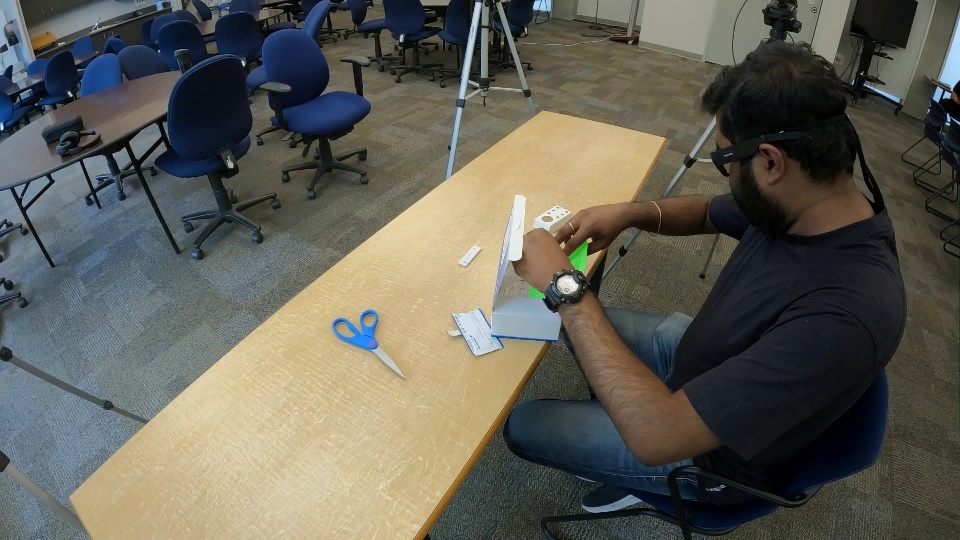} \\

XSegTx & \includegraphics{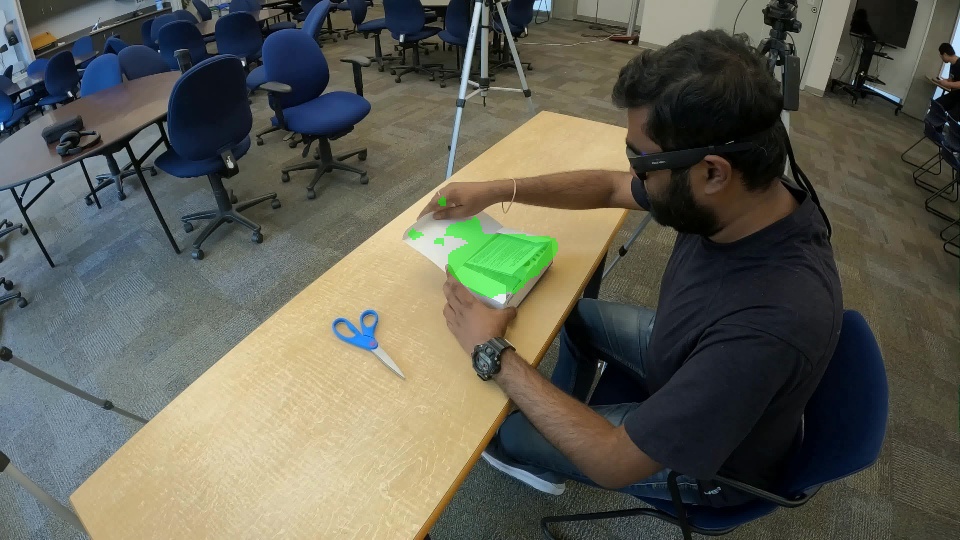} & \includegraphics{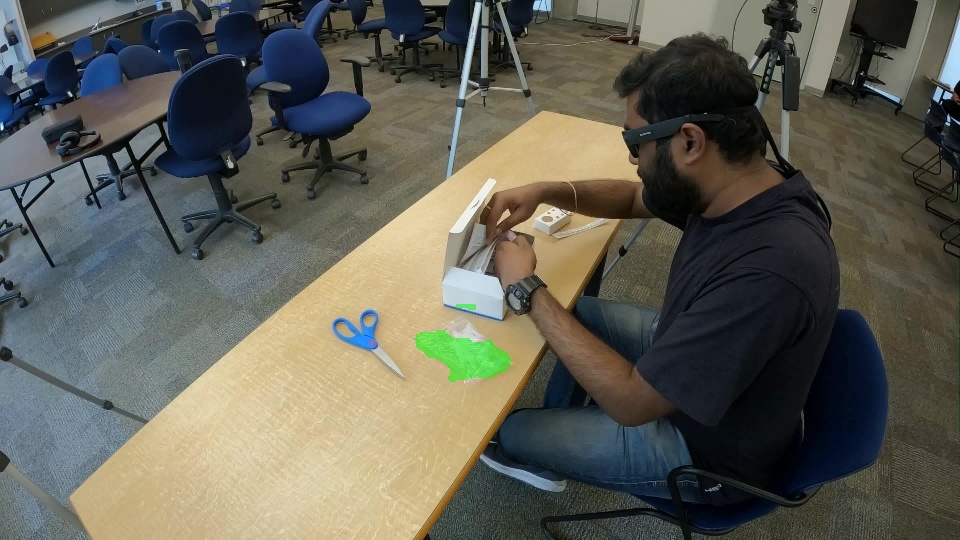} & \includegraphics{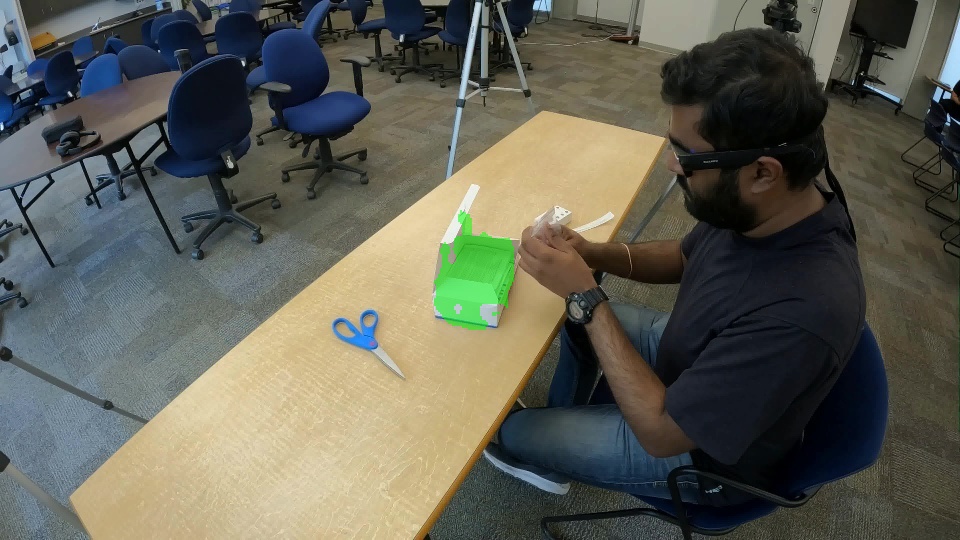} & \includegraphics{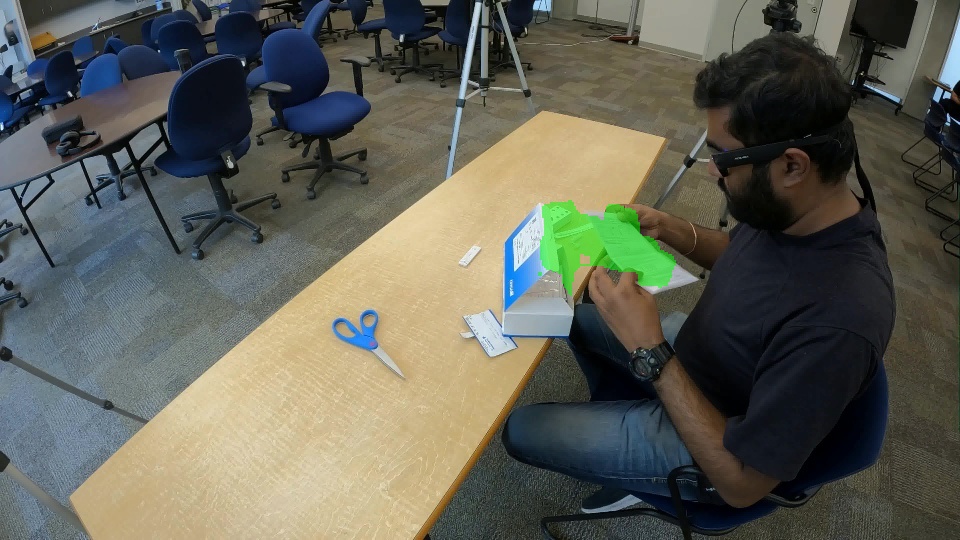} & \includegraphics{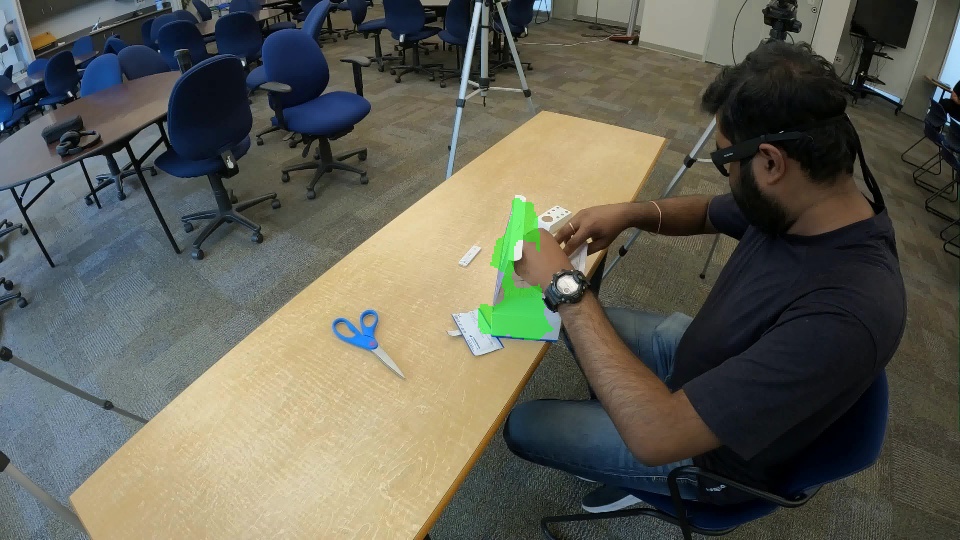} \\

XView-XMem & \includegraphics{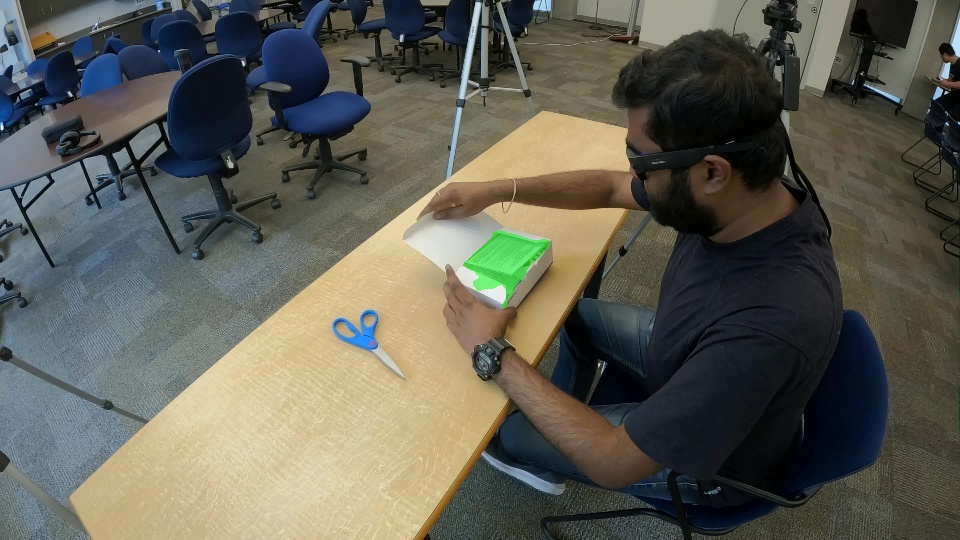} & \includegraphics{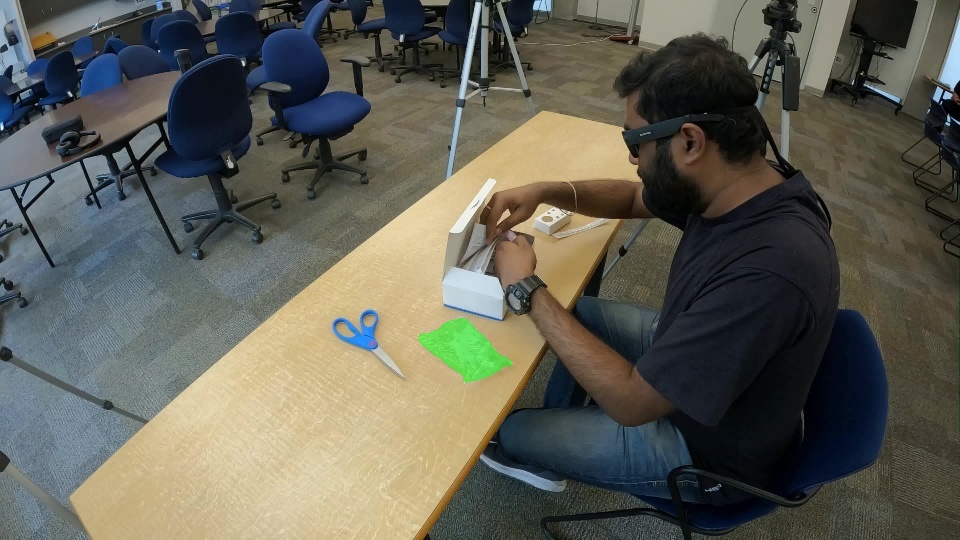} & \includegraphics{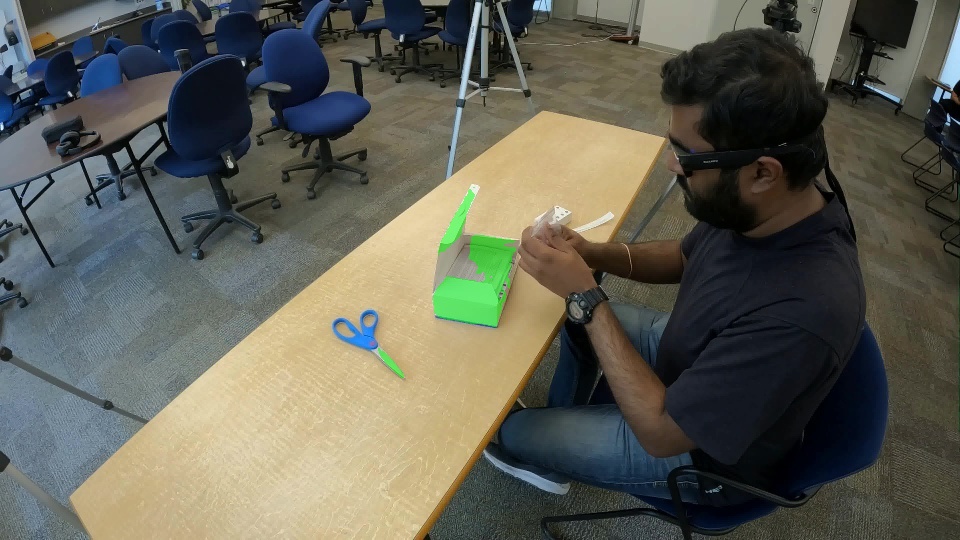} & \includegraphics{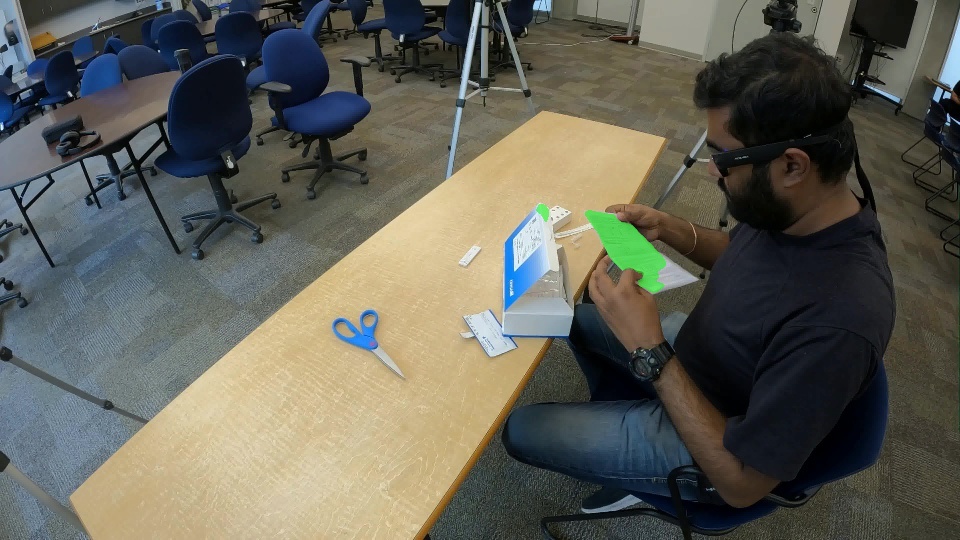} & \includegraphics{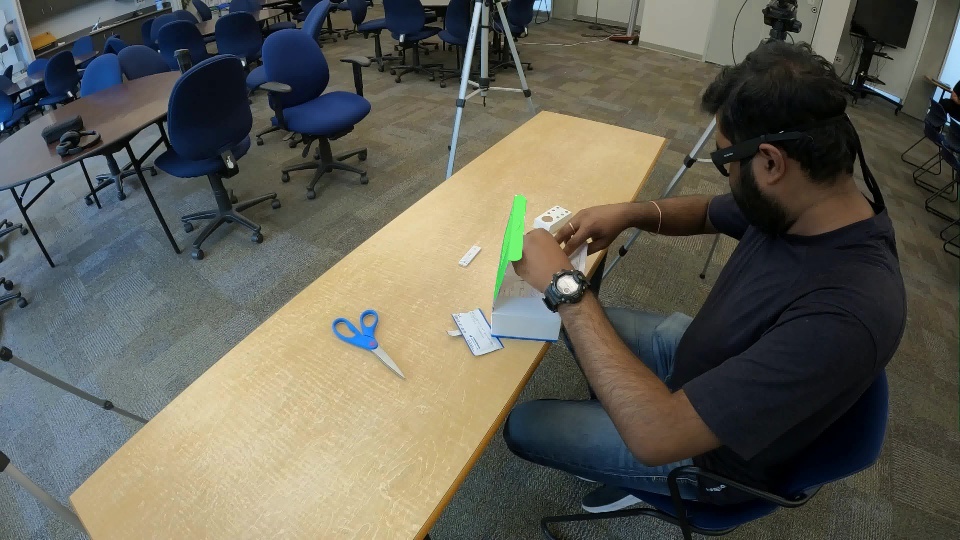} \\
\midrule
Ego GT & \includegraphics{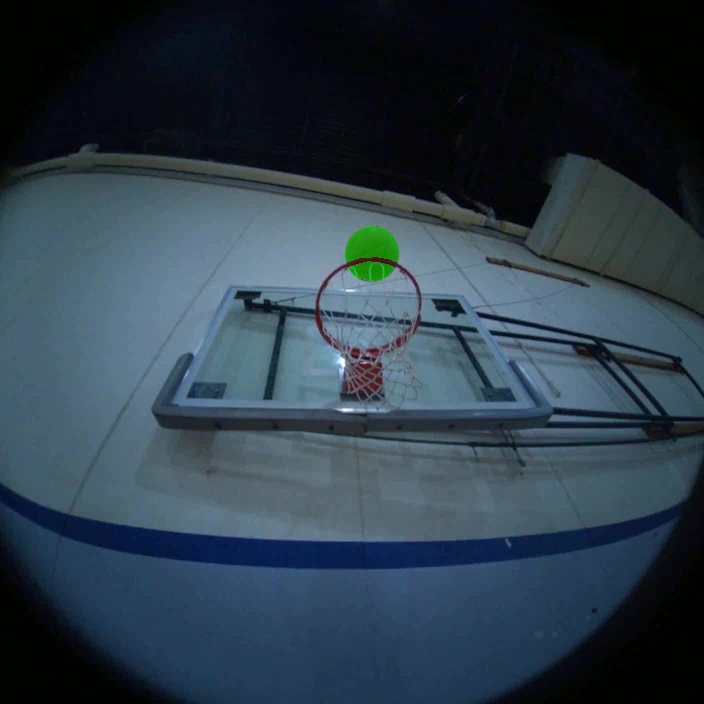} & \includegraphics{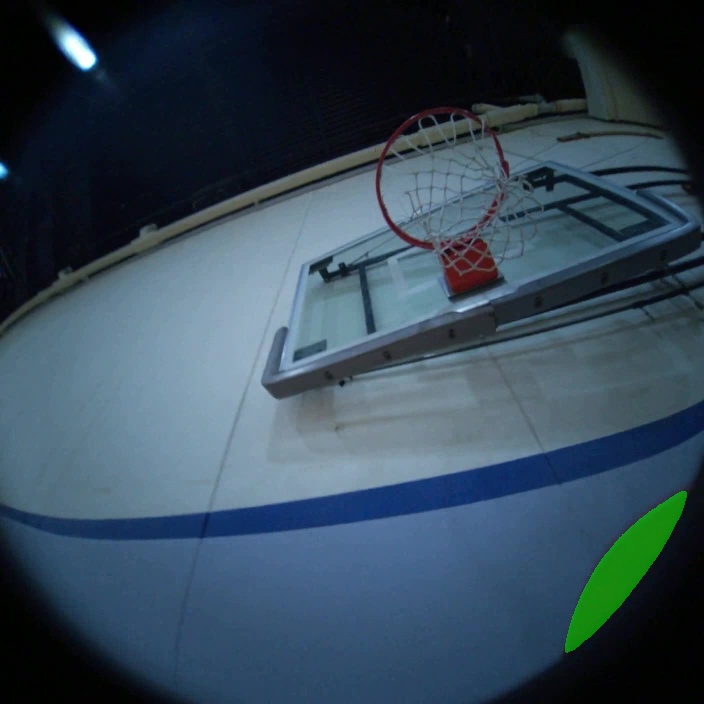}  & \includegraphics{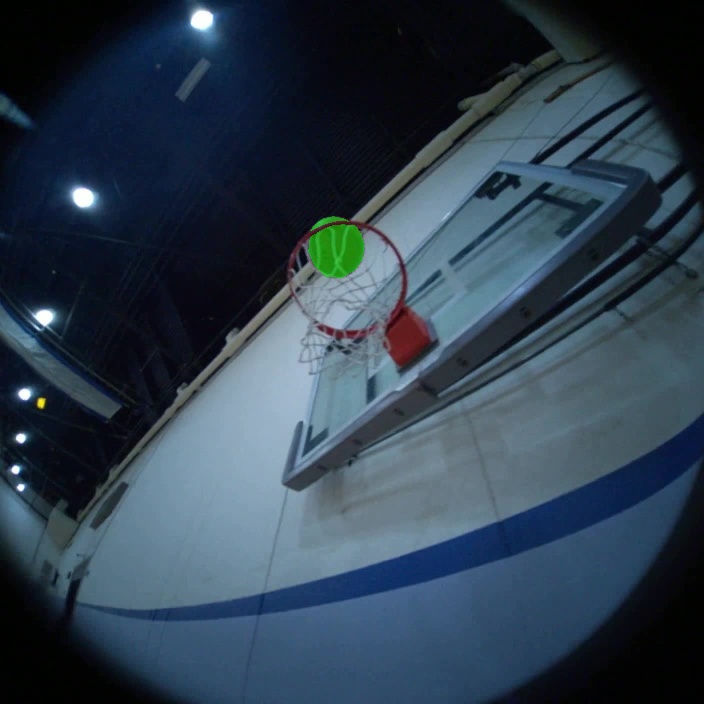}  & \includegraphics{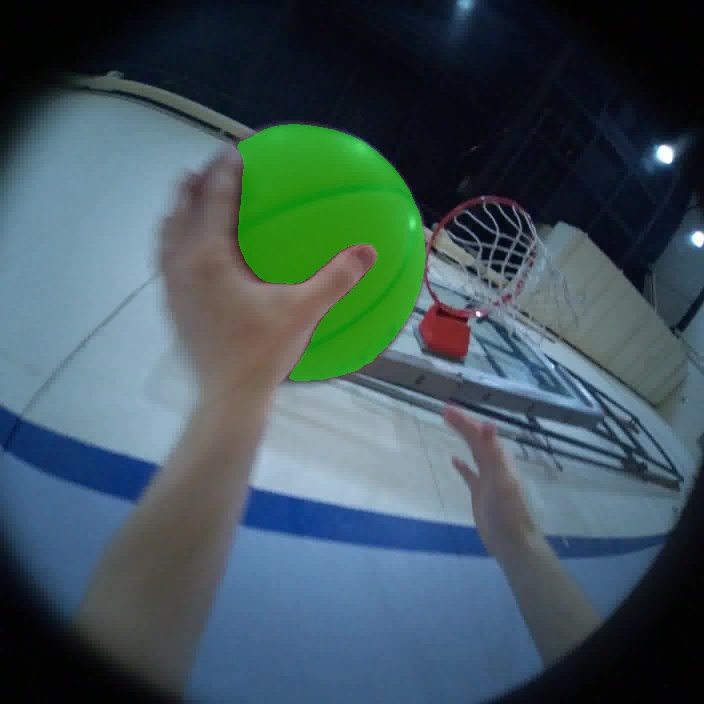}  & \includegraphics{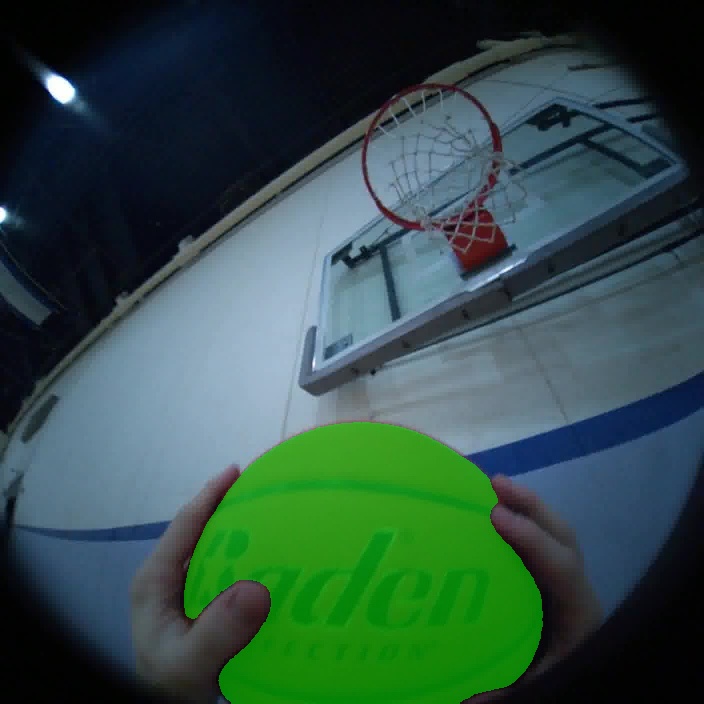}  \\

Exo GT & \includegraphics{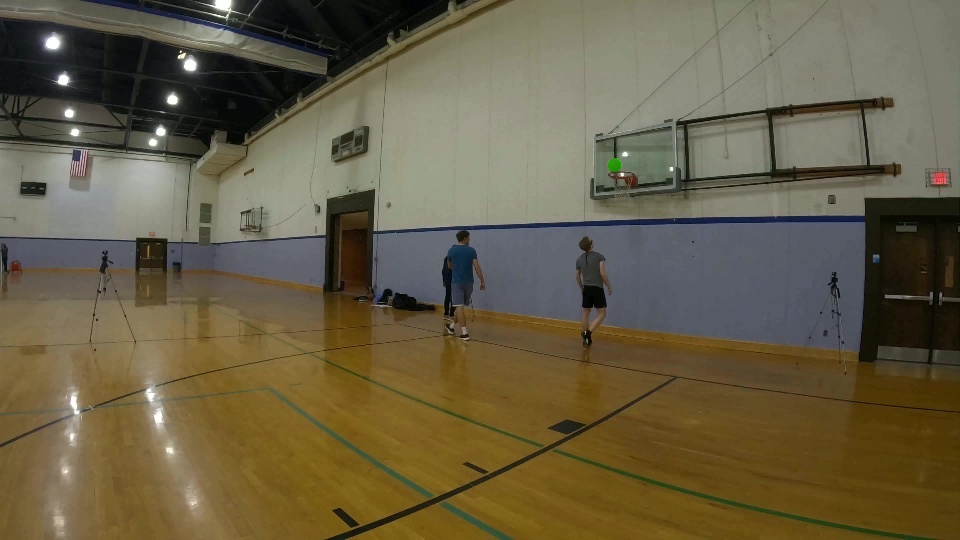} & \includegraphics{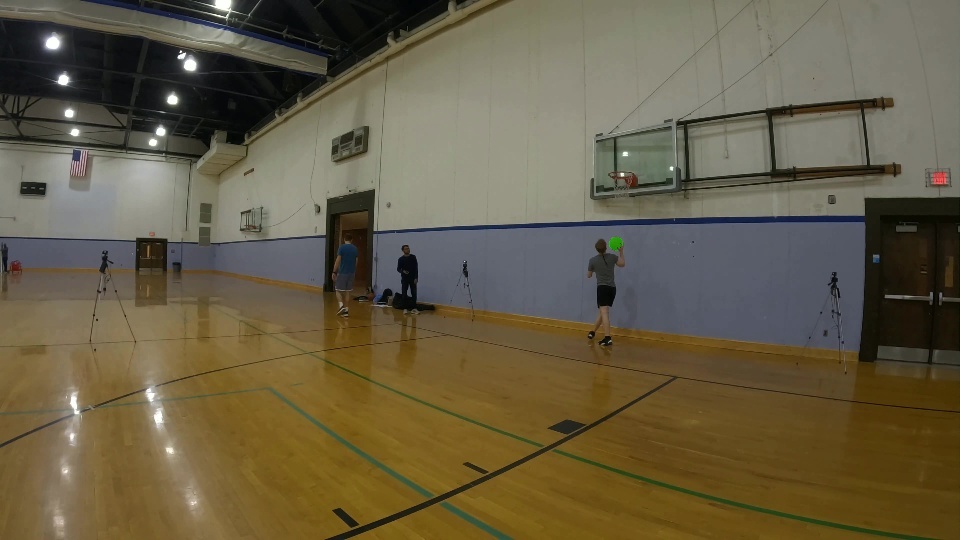}  & \includegraphics{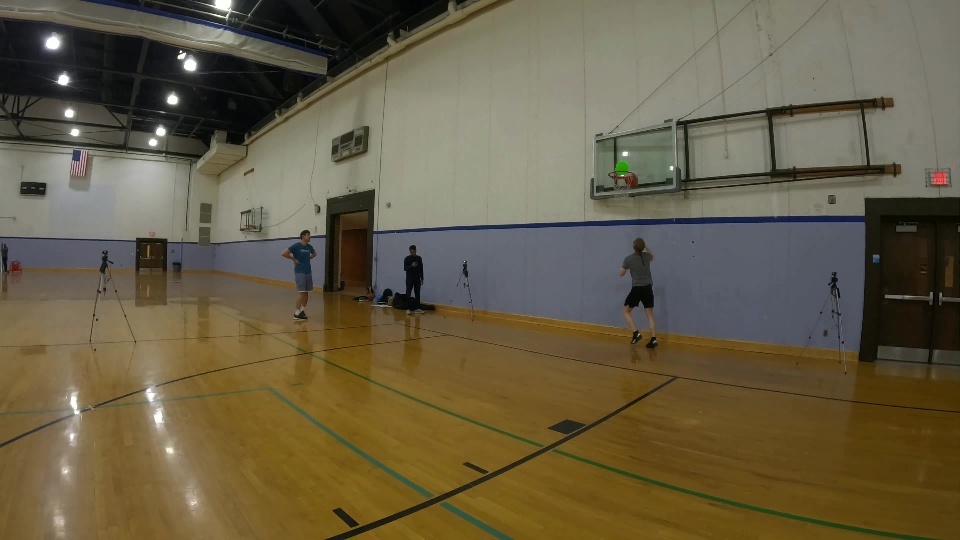}  & \includegraphics{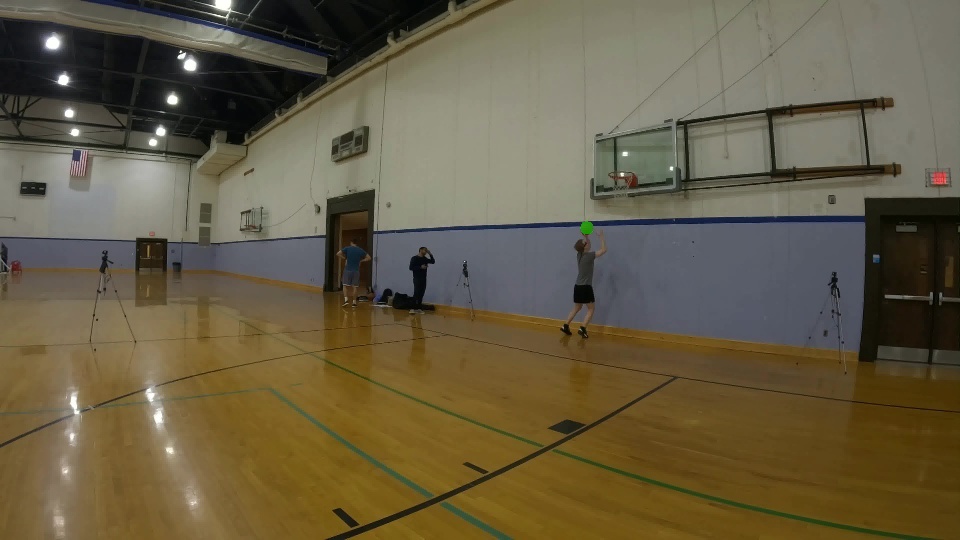}  & \includegraphics{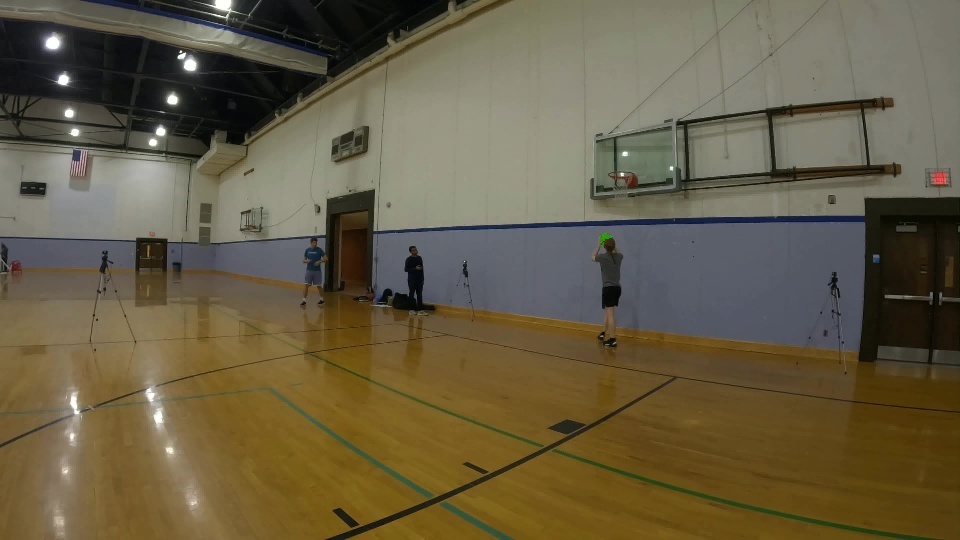}  \\

XSegTx& \includegraphics{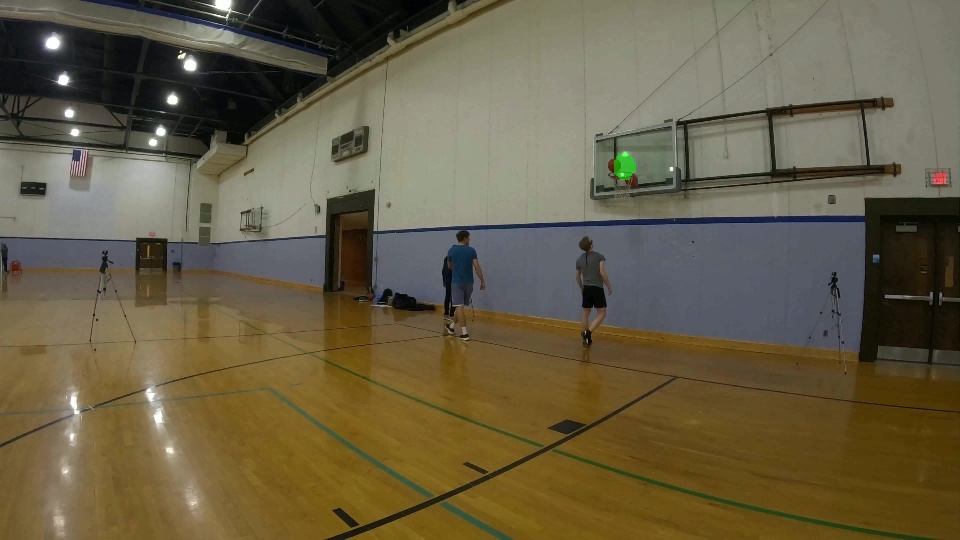} & \includegraphics{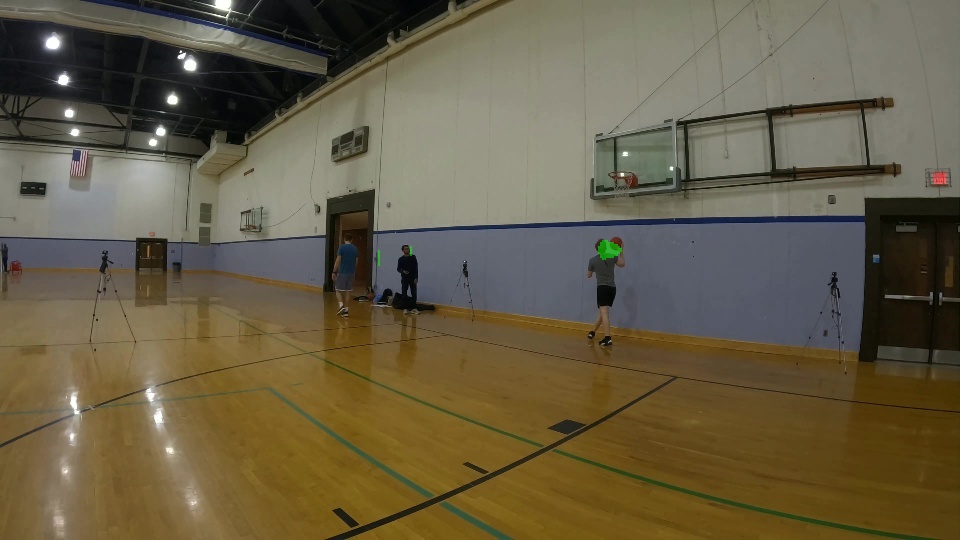}  & \includegraphics{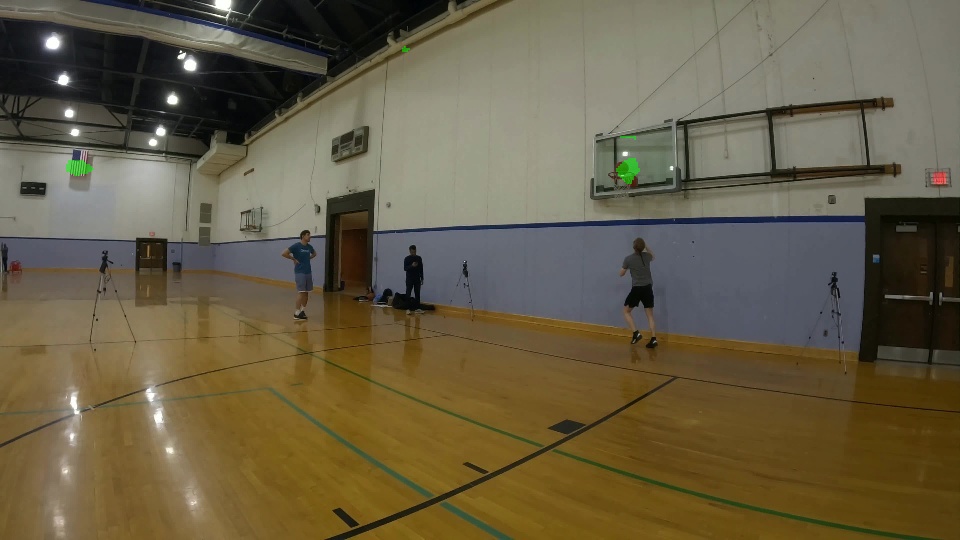}  & \includegraphics{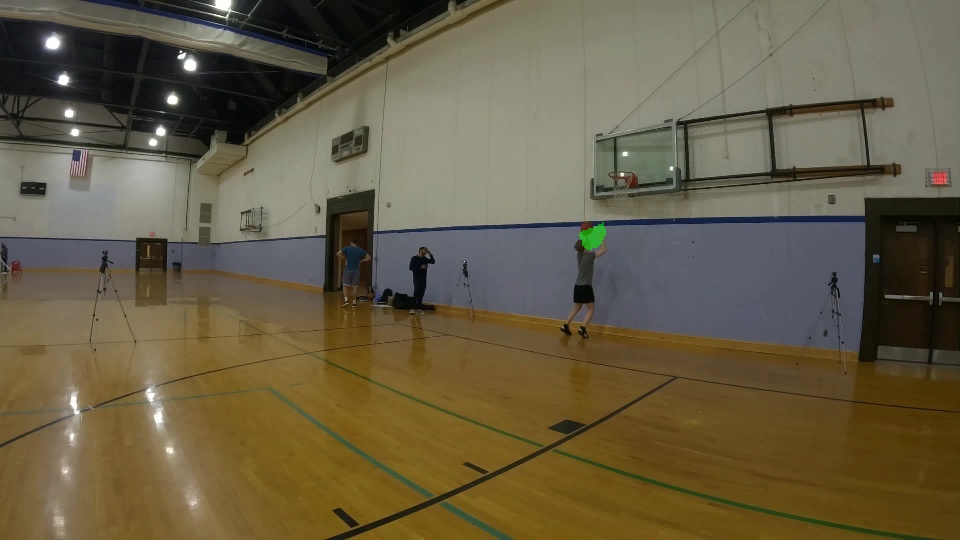}  & \includegraphics{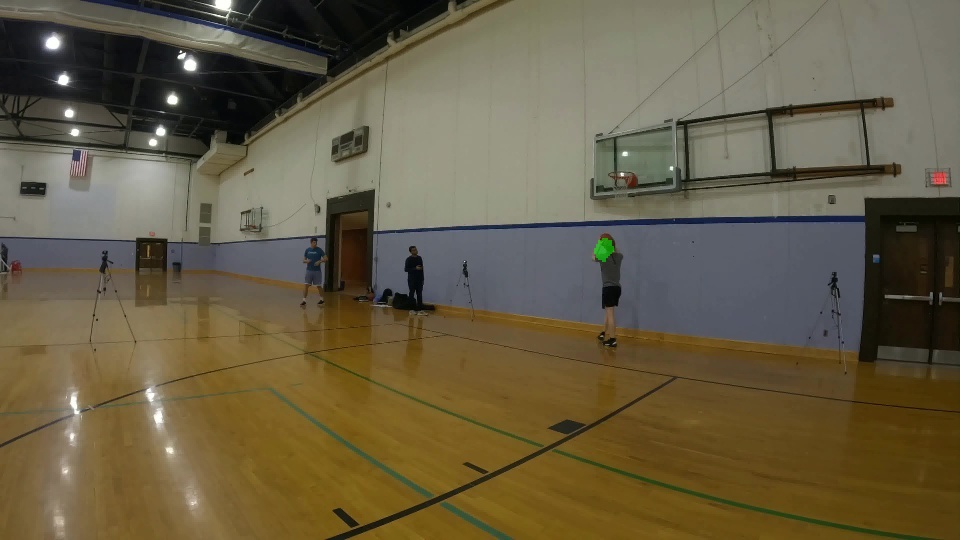}  \\

XView-XMem & \includegraphics{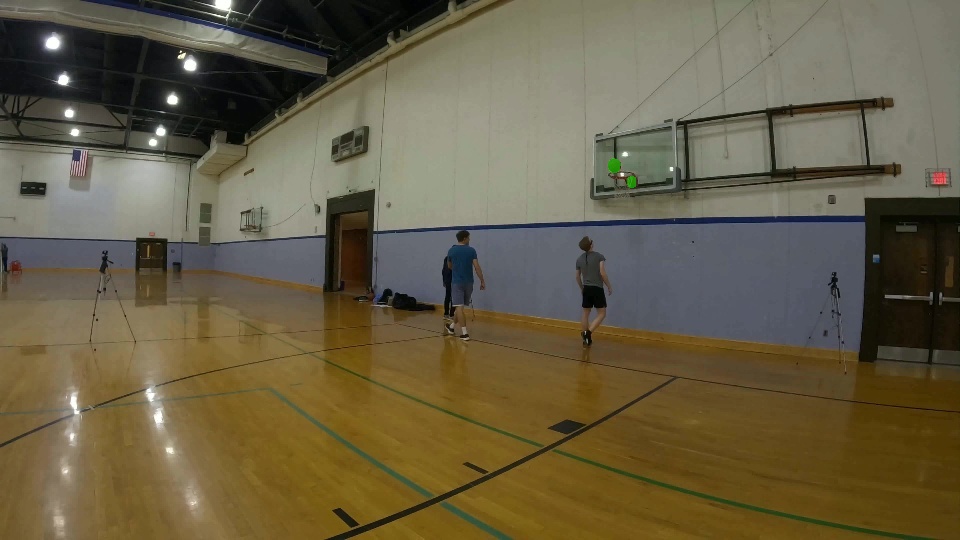} & \includegraphics{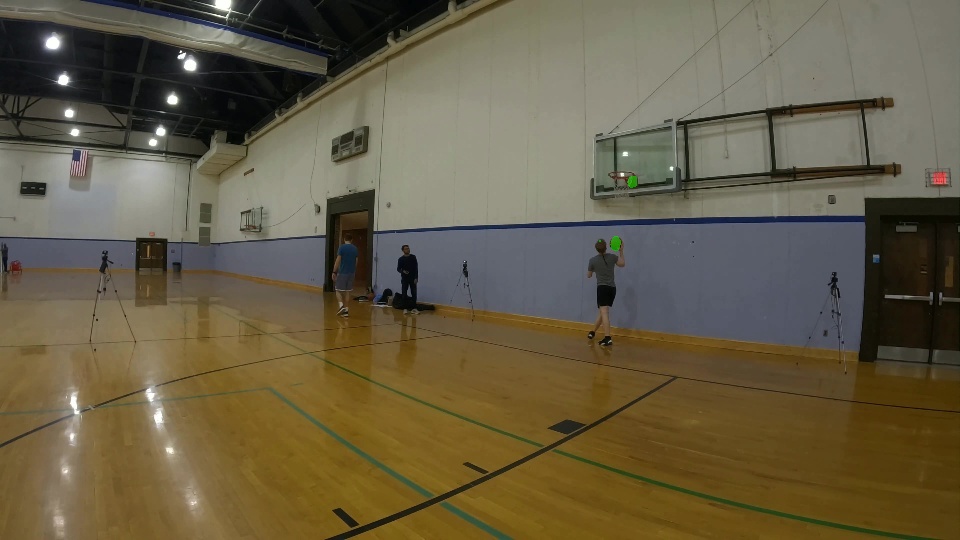}  & \includegraphics{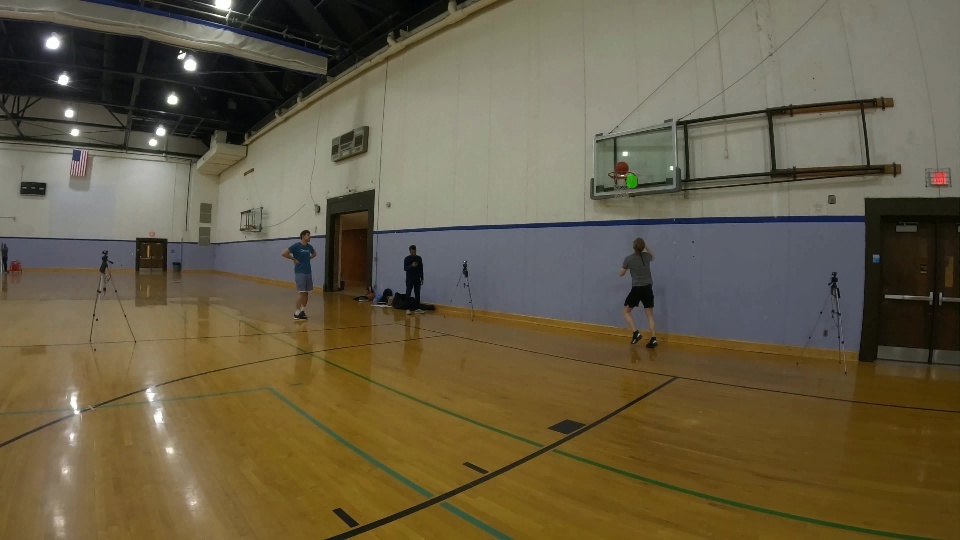}  & \includegraphics{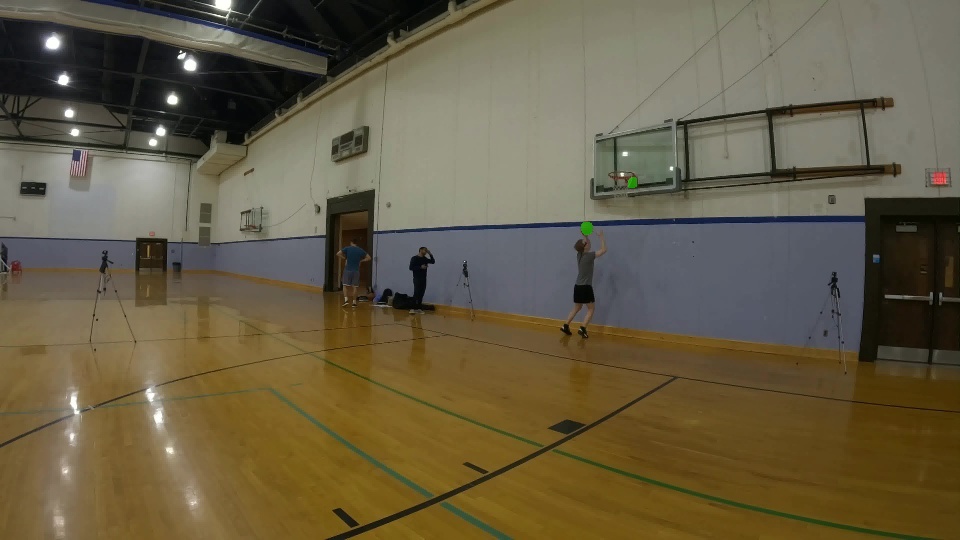}  & \includegraphics{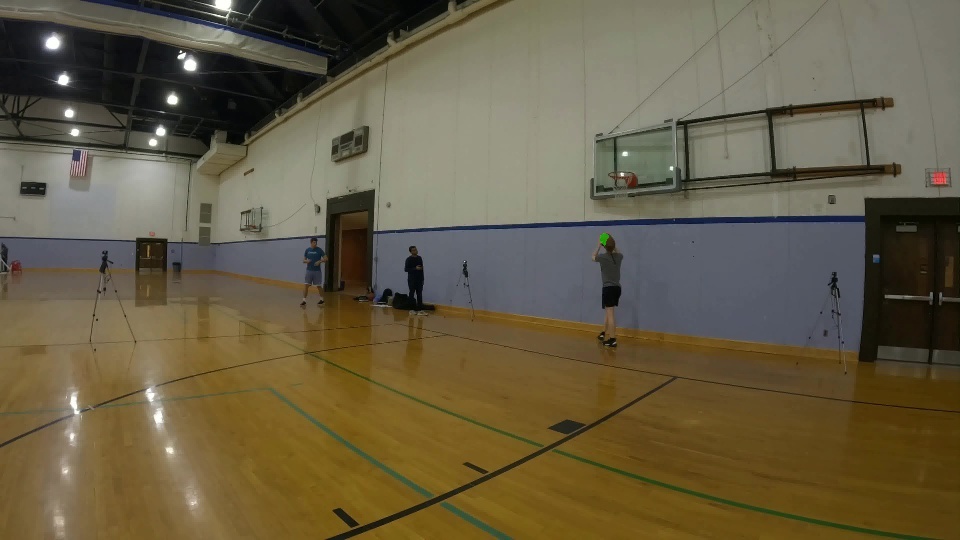}  \\

\midrule
Ego GT & \includegraphics{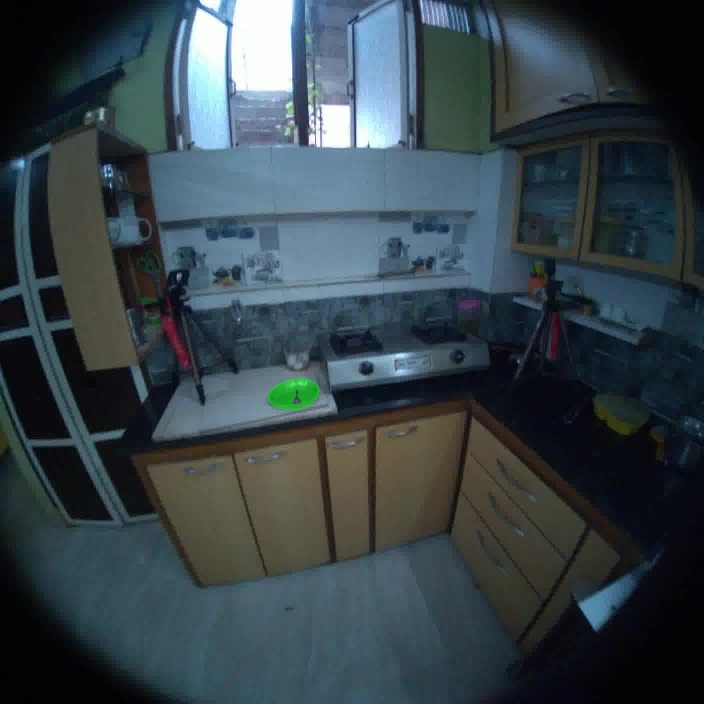} & \includegraphics{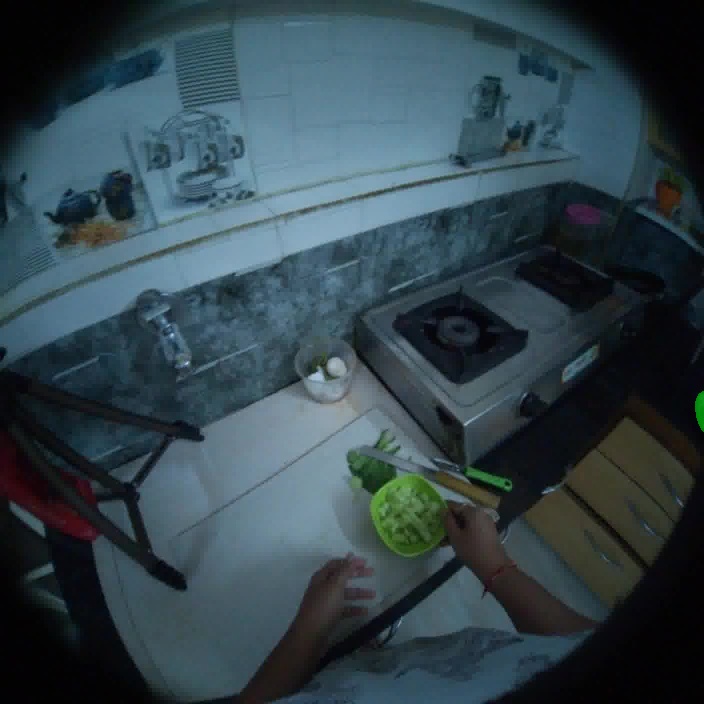} & \includegraphics{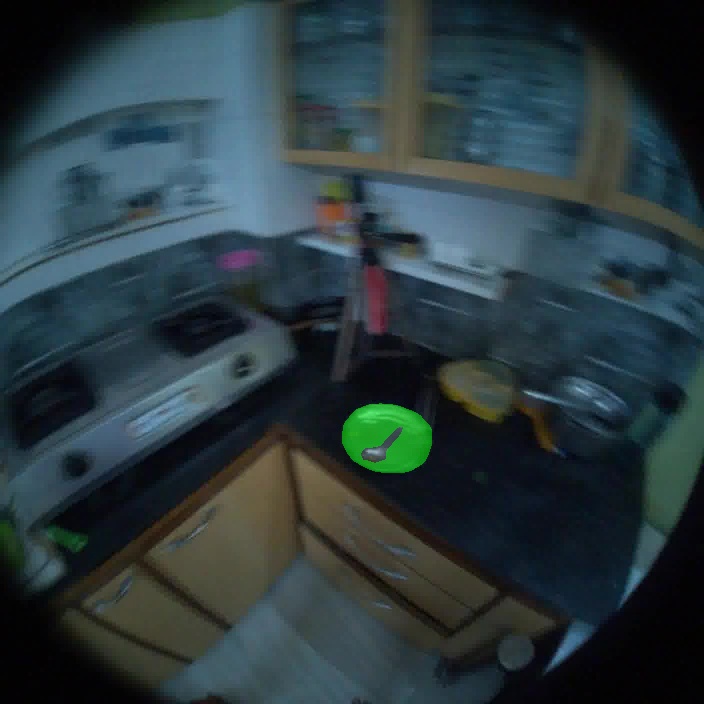} & \includegraphics{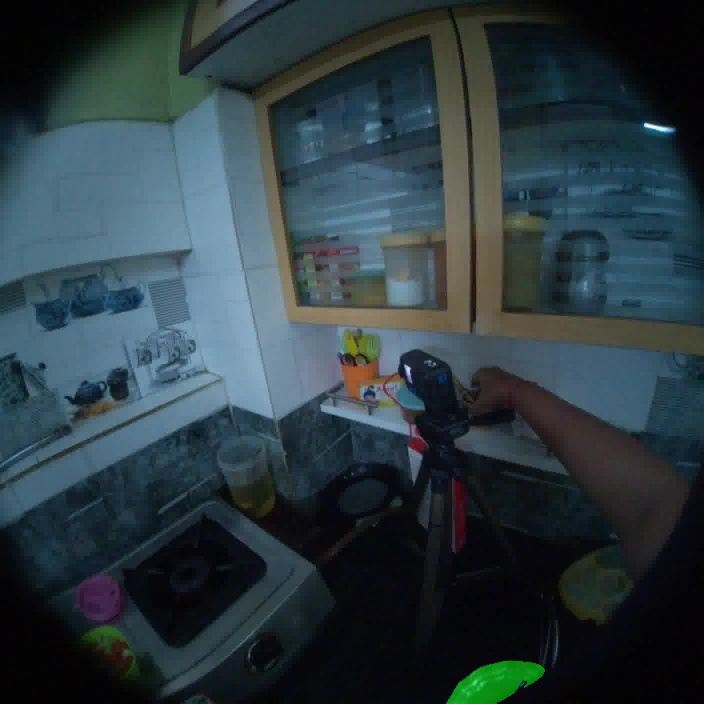} & \includegraphics{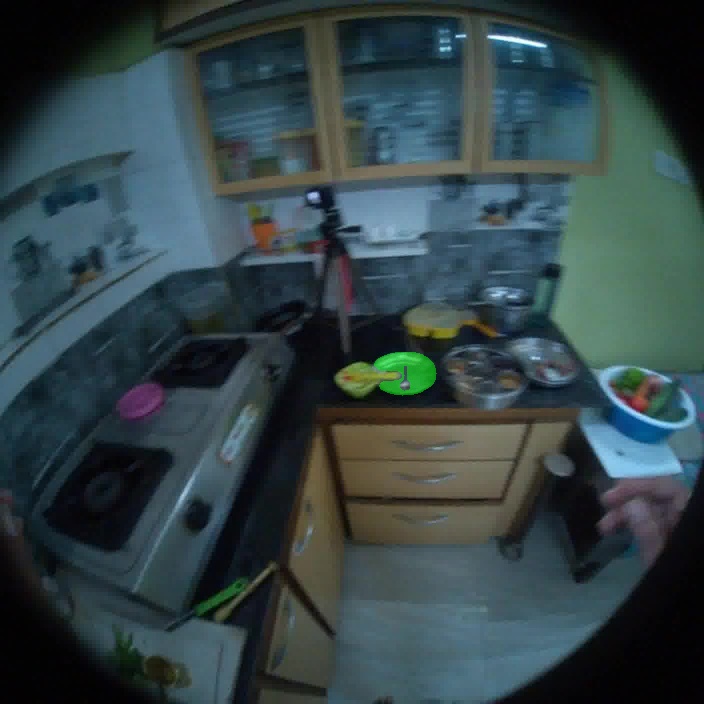} \\

Exo GT & \includegraphics{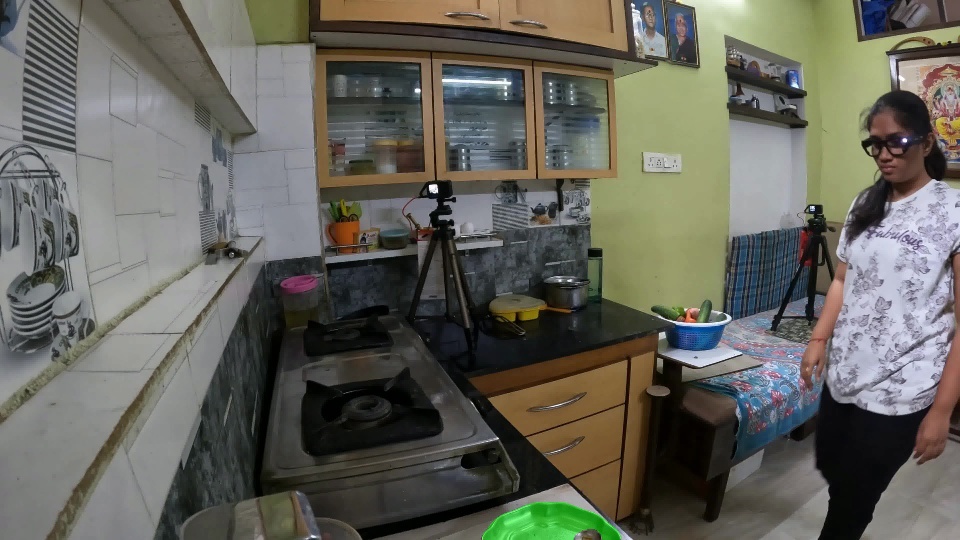} & \includegraphics{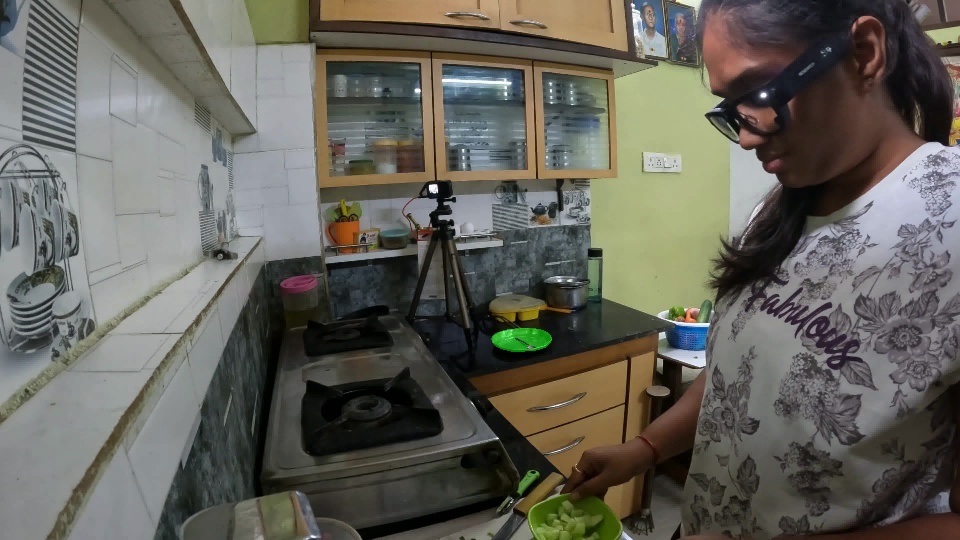} & \includegraphics{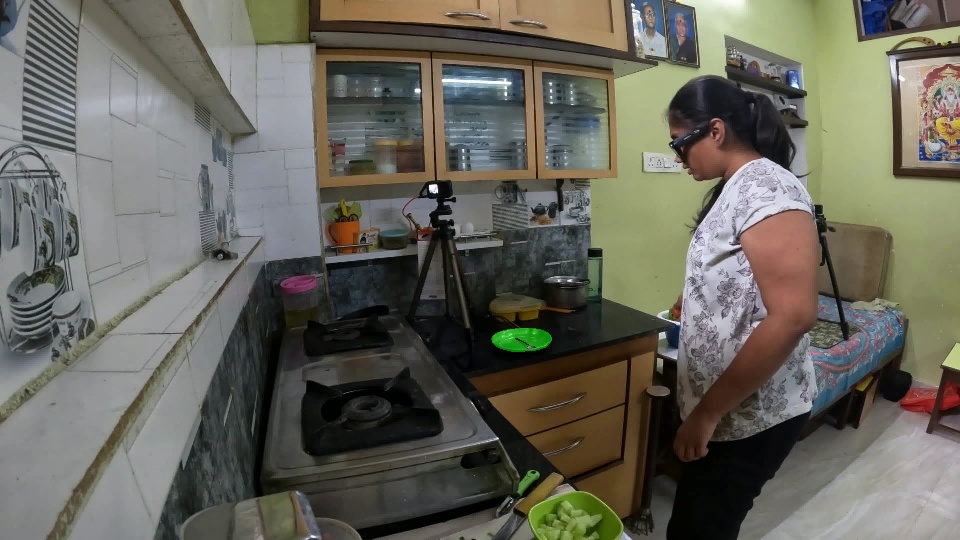} & \includegraphics{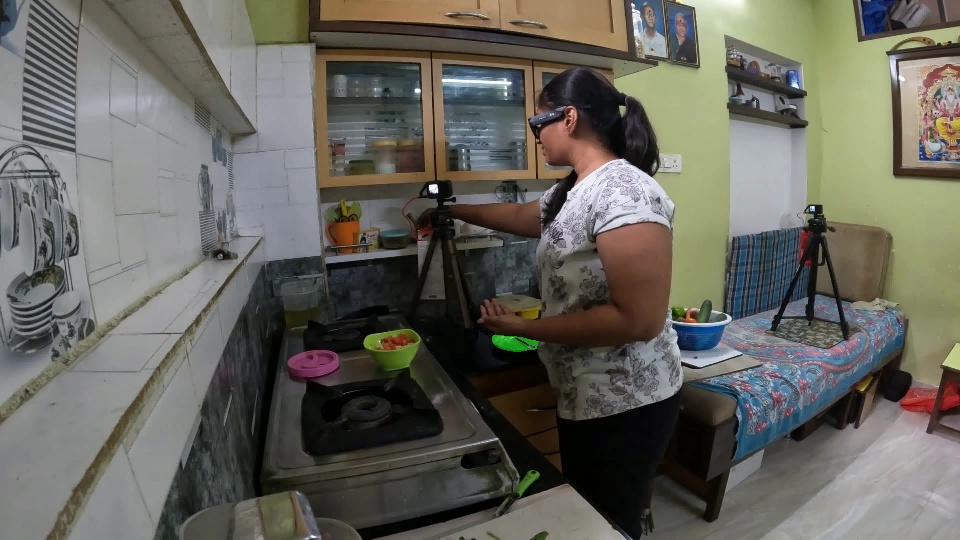} & \includegraphics{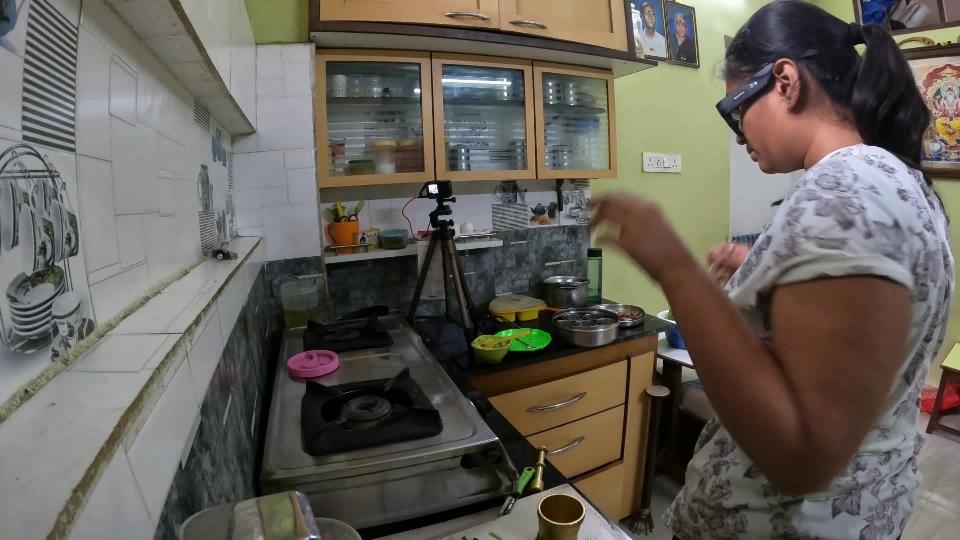} \\

XSegTx & \includegraphics{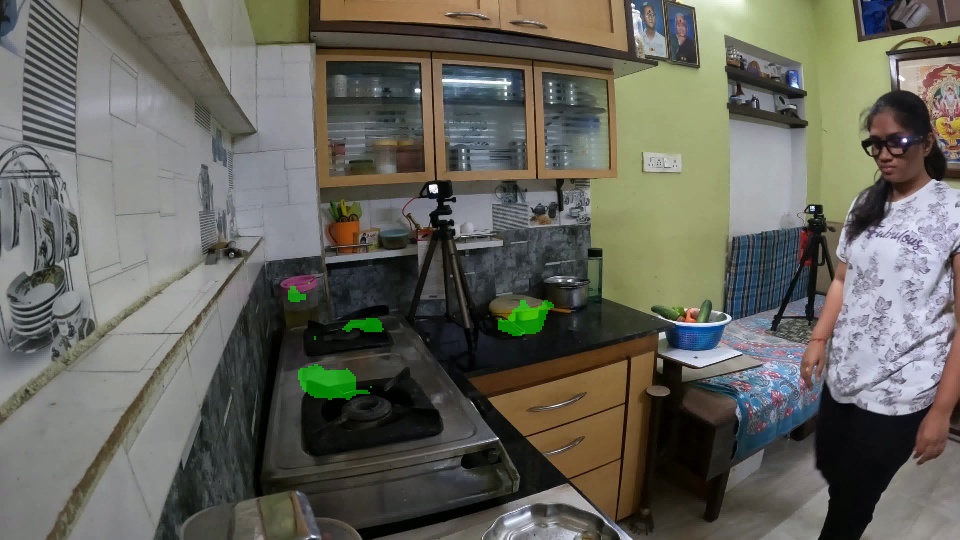} & \includegraphics{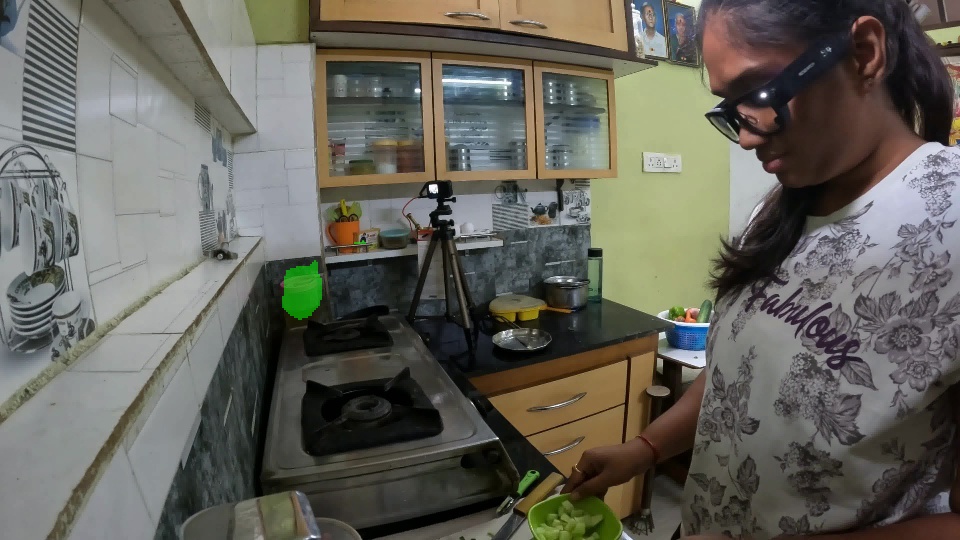} & \includegraphics{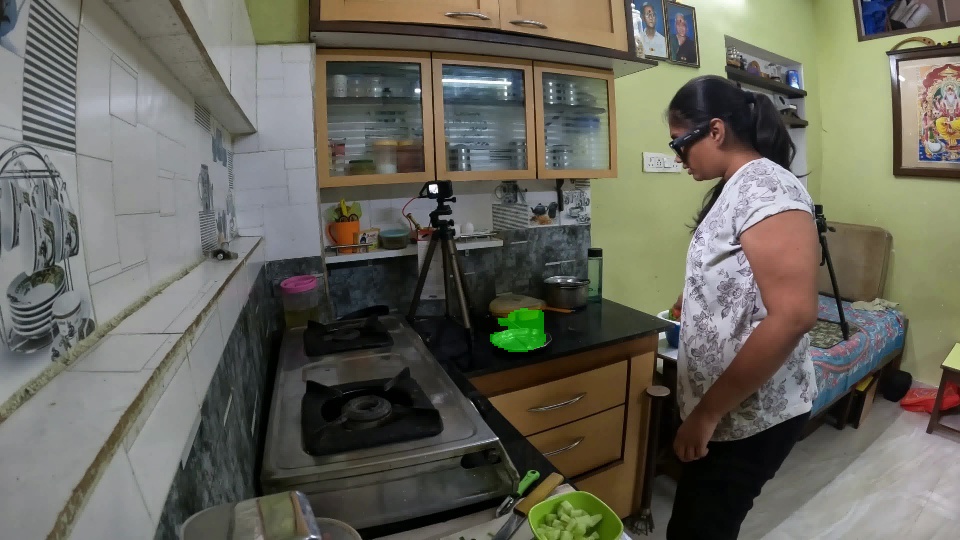} & \includegraphics{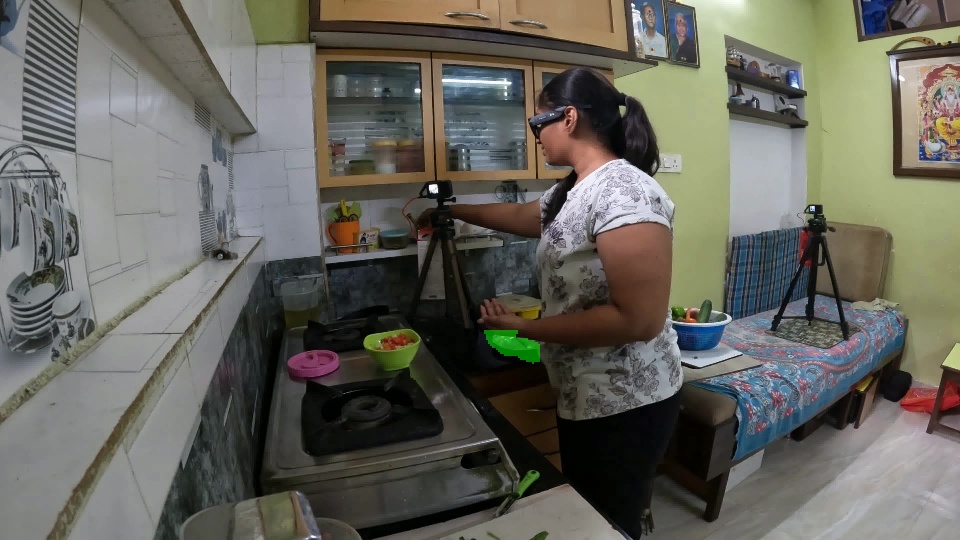} & \includegraphics{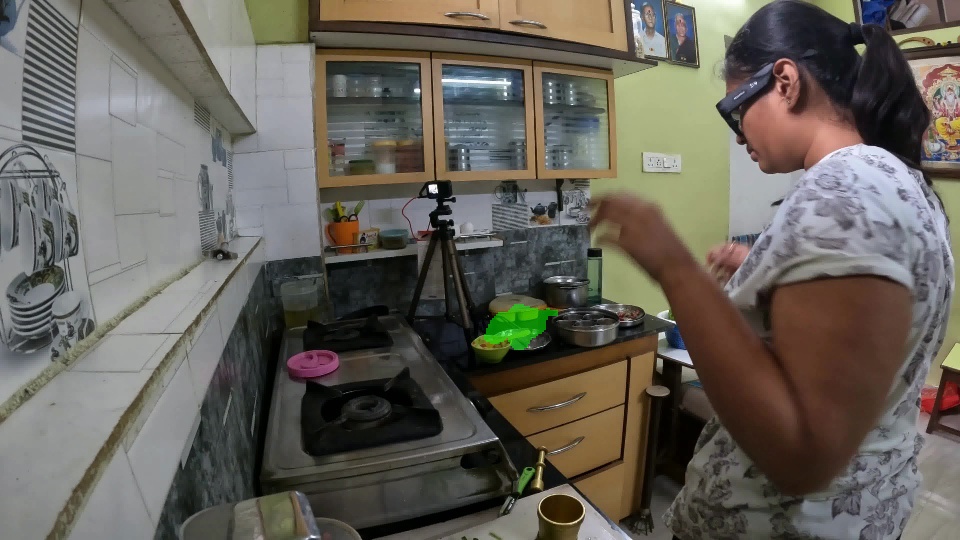} \\

XView-XMem & \includegraphics{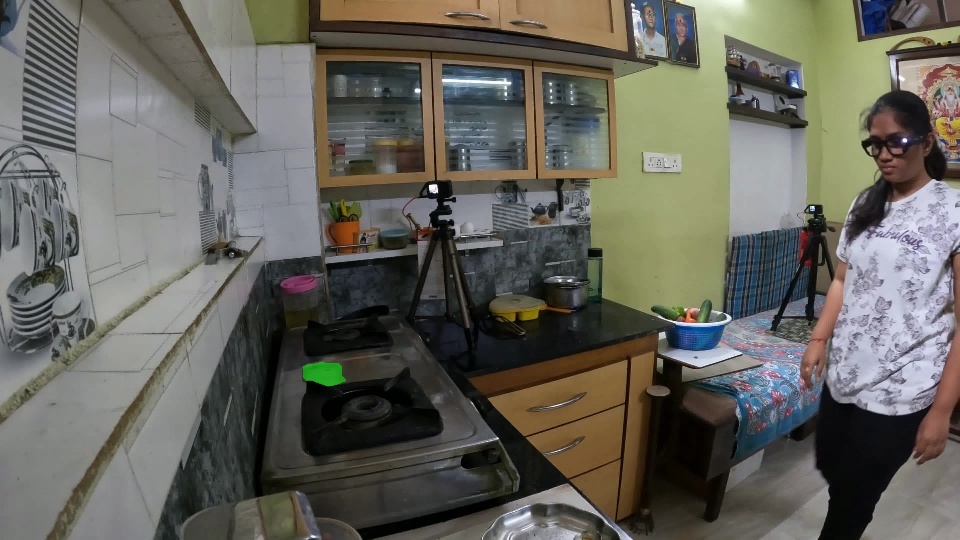} & \includegraphics{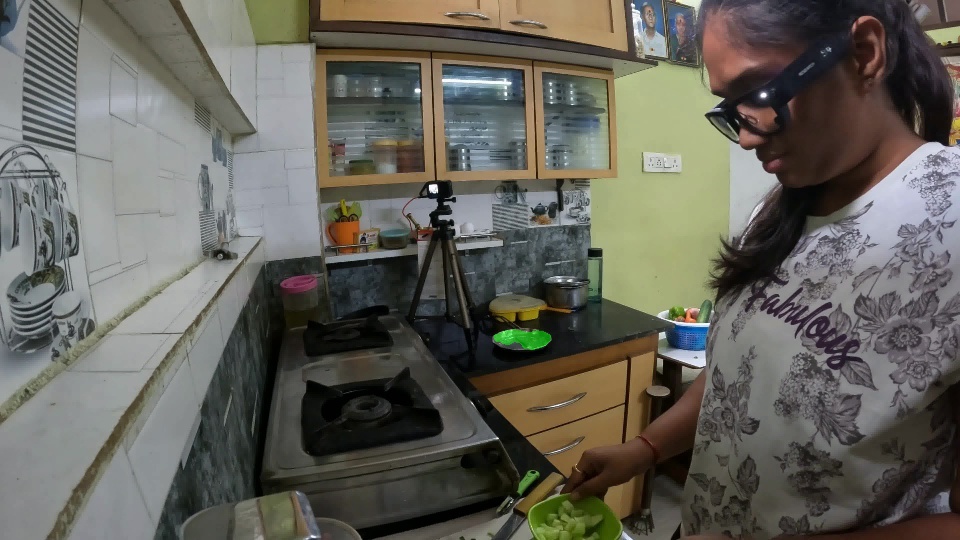} & \includegraphics{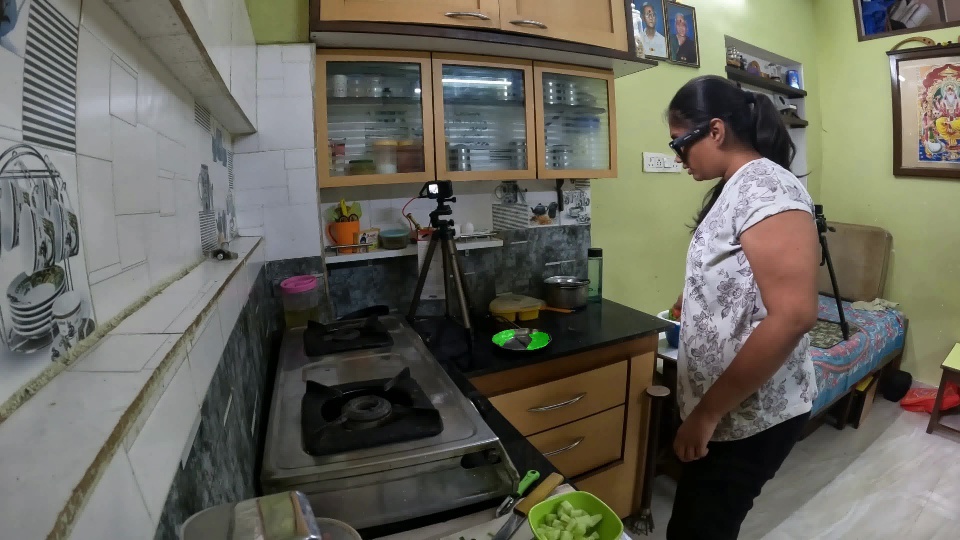} & \includegraphics{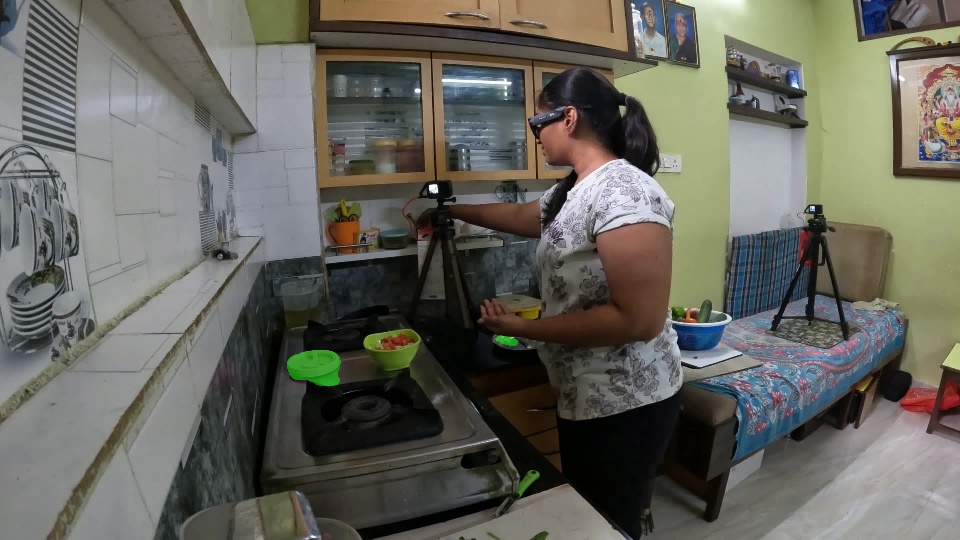} & \includegraphics{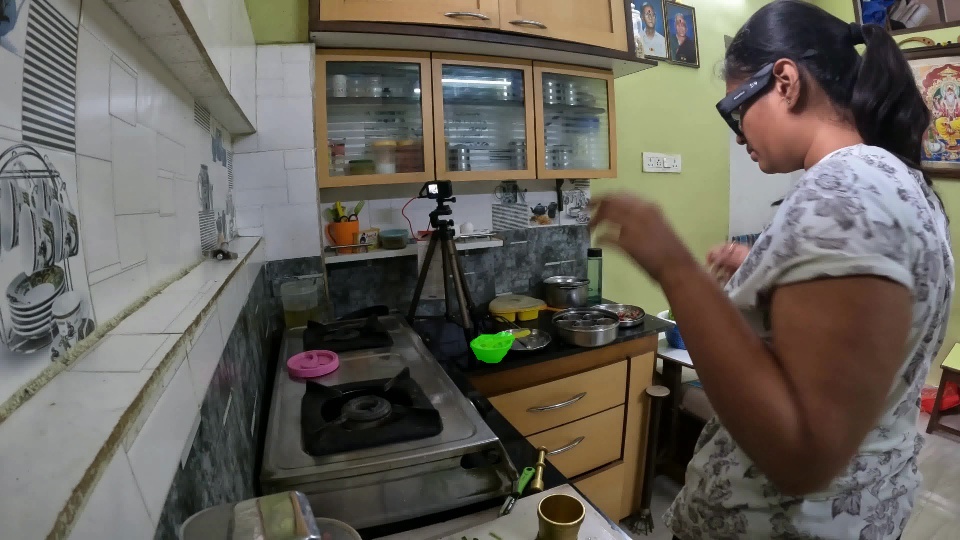} \\

\bottomrule
\end{tabularx}
\caption{Qualitative results for the different ego-exo correspondence baselines.}
\label{fig:corres-qualitative}
\end{figure*}

%% file: sec/benchmarks/translation-benchmark.tex
\paragraph{Motivation}

The second of the two ego-exo relation tasks is \emph{ego-exo translation}.  Our translation task entails synthesizing a target ego clip from a given exo clip. This problem may be viewed as a form of ego-exo correspondence with missing information: given the masks of an object in the exo clip, its corresponding masks and pixel values must be generated in the {\em unobserved} ego clip. We note that this problem cannot be solved by image-based rendering or via geometric transformations, since the dramatic differences in ego-exo viewpoints cause the two cameras to capture different regions of the same objects.

We believe this problem will drive novel research %
for combining recognition and object synthesis. For example, in Figure~\ref{fig:relation_definition} (right), the approach must perceive the hand as the generation target and 
make effective use of the hand's object-specific shape and appearance priors in order to synthesize the ego view of the fingertips---which are not visible in the exo clip. Furthermore, this task will stimulate advances in visual odometry, as the method must be able to infer the ego camera pose from the third-person clip. 

Ego-exo translation also holds strong application potential, as it may unlock the ability to generate first-person renderings of videos that were originally captured from a third-person perspective. For example, it may enable AR coaching, with objects and interactions lifted from a third-person instructional videos and re-synthesized from the camera-wearer perspective to better guide the user in the execution of complex activities. Ego-exo translation may also be used to generate abundant first-person training data from existing large-scale collections of third-person videos in order to train robot perception models.

\paragraph{Task definition}

The translation benchmark focuses on generating information in the egocentric view given the exocentric view.
We decompose ego-exo translation into two separate tasks: {\em ego track prediction} and {\em ego clip
generation} (Figure~\ref{fig:relation_definition}, right). 
Ego track prediction estimates the segmentation mask of an object in the {\em unobserved} ego frames given the object masks in the observed exo clip. Ego clip generation %
entails generating the image values (i.e., RGB) within the given ground-truth ego mask by making use of the exo clip and the object masks in those frames.  This decomposition effectively splits the problem into two tasks: 1) predicting the location and shape of the object in the ego clip, and 2) synthesizing its appearance given the ground-truth position. For both subtasks, the input exo clip consists of 5 frames evenly sampled from a time span of 5 seconds.

We believe that the decoupling of these two tasks will promote faster progress on the individual sub-problems and facilitate understanding of the key challenges in each of them. 
For each, we consider a variant where the pose of the ego camera with respect to the exo camera is available to use at inference time. This simplifies the problem but reduces the applicability of the method, since this information is typically not available for arbitrary third-person videos. Finally, we note that while the opposite direction of translation (i.e., ego-to-exo) could be considered, here we focus on the task of ego generation because of its higher value for robotics and AR applications.

\NEW{Note that we restrict the input to include only the exo view and the object masks in order to promote the design of methods that can translate arbitrary third-person video into an egocentric one. Thus, the input \KGnew{\emph{excludes}} depth maps, 3D point clouds, IMU, or SLAM, which would simplify the task at the expense of general applicability, since these signals are typically not available for in-the-wild video. The only exception is a variant of the task where the ego camera pose for all frames of the clips is given as input. We consider this formulation in order to estimate a sort of  ``upper bound'' on translation performance under the unrealistic assumption of known ego-exo camera relation.}

\paragraph{Related work}

Ego-exo translation relates to cross-view image synthesis~\citep{Regmi_2018_CVPR, tang2019selectiongan, Lu_2020_CVPR}. Within this genre, the problem of exo-to-ego generation was recently introduced for both images~\citep{liu2020pgan} and video~\citep{romy-exo2ego,liu2021stagan,4diff}, and approached using GANs or diffusion conditioned on the input view. \KGupdate{Our work not only formalizes this task with ample data, but its formulation also draws attention to the need for a \emph{semantic} basis to new view synthesis across extreme view changes.}

\paragraph{Annotations}

Translation uses the same annotations as the correspondence task discussed above in Section~\ref{sec:relation}.

\paragraph{Metrics}

We adopt a diverse set of metrics to assess the different aspects of the generated translation.  As for the task of correspondence, we use {\em Visibility Accuracy} (VA) to evaluate the ability of the method to predict the visibility of the target object in the ego view but this time given only exo frames as input. We consider the visibility prediction correct if and only if either of these two conditions are met: (1) the predicted mask is empty when the object is invisible in the ego view, or (2) the predicted mask is non-empty when the object is visible in the ego view. 
Furthermore, we adopt the following metrics defined for correspondence to gauge the performance of  {\em Ego Track Prediction}: 1) {\em Location Error} (LA) 2) {\em Intersection Over Union} (IoU) and 3) {\em Contour Accuracy} (CA)~\citep{perazzi2016benchmark}. \NEW{The IoU and CA are calculated after registering the centroids of the predicted mask and the ground-truth ego mask, in order to gauge mask prediction independent of location error.}
To evaluate {\em Ego Clip Generation} we use two popular image quality metrics (SSIM, PSNR~\citep{hore2010SSIM}) and three perceptual metrics (DISTS~\citep{ding2020dists}, LPIPS~\citep{zhang2018lpips} and CLIP similarity~\citep{radford2021clip}).

\paragraph{Baselines}
\label{sup:translation_baselines}
For track prediction, we implement the GAN-based method pix2pix~\citep{pix2pix2017} and {the NeRF-based method GNT~\citep{gnt2023}}. For clip generation, we employ the GAN-based method pix2pix~\citep{pix2pix2017} and the diffusion model DiT~\citep{Peebles2022DiT}. It is worth noting that, as discussed below, we introduce specific modifications to adapt these methods to our task requirements. \NEW{All baselines utilize exo images and masks, with only the GNT model making use of the extra input of ego camera pose.}\\

\noindent \textit{Ego Track Prediction} involves generating segmentation masks for the egocentric view based on the exocentric video clip and the exocentric object masks. We consider the following two baselines for this task:

\begin{itemize}%
\itemsep0em 
\item \textbf{pix2pix-mask}. We modify the generator of pix2pix to have inputs and outputs of 4 channels. Specifically, the exo frame and the exo mask are concatenated as the inputs while the 4-channel outputs are ego frame (3 channels) and ego mask (1 channel). The ego frame is supervised with the losses used in pix2pix. We use the bootstrapped cross-entropy loss~\citep{reed2014bootce} and the dice loss~\citep{sudre2017dice} for mask prediction. \\

{
\item \textbf{GNT-mask}.
We adopt the Generalizable NeRF Transformer (GNT)~\citep{gnt2023} as another baseline leveraging the camera poses. %
In our adapted version, the image encoder takes a 4-channel image (exo frame and mask) as inputs to predict the ego frame and ego mask. Formally, during the training of our GNT-mask, for each point $x$ and viewing direction unit vector $d \in \mathbb{R}^3$, the ray transformer $f$ in GNT predicts two key attributes:
RGB Color ($c$) and
Object Existence Score ($e$), in which $e$ signifies the probability of an object being present at point $x$.
During rendering, the volumetric radiance field encoded by the ray transformer can then be rendered into a 2D image as well as a 2D object mask.\\ %
}
\end{itemize}

\noindent \textit{Ego Clip Generation} requires producing pixel values representing the target object in the egocentric view. To achieve this, we leverage 6 different input images for each frame: exo frame, exo mask, exo object crop, cropped exo mask, ego mask and cropped ego mask. The cropped exocentric and egocentric masks are generated by considering a bounding box to isolate the relevant portions of the exocentric and egocentric masks, respectively. The ``exo object crop" refers to the RGB image obtained by cropping out the relevant region using the cropped exocentric mask. We resize these 6 images to the same size ($256\times 256$). We evaluate two baselines for this task:

\begin{itemize}%
\itemsep0em 
\item \textbf{DiT-pix}. We adopt the Transformer-based diffusion model DiT~\citep{Peebles2022DiT}.
We predict the ego object crop by conditioning the DiT on the 6 input images in two manners. Initially, these six images are concatenated along the channel dimension and subsequently combined with the noisy ego object crop, forming the input to DiT. Additionally, two ResNet-50 architectures encode the six images into low-dimensional features, which are then incorporated into each layer of DiT via AdaLN~\citep{perez2018adaln}.\\ %

\item \textbf{pix2pix-pix}. We adopt pix2pix~\citep{pix2pix2017} for clip generation as well by concatenating the 6 images along the channel dimension as inputs to the pix2pix model.\\
\end{itemize}

All of the above-mentioned baselines perform image-to-image generation. We implement also clip-to-clip variants of these methods by taking multiple frames as inputs and predicting results for all frames jointly. For pix2pix, we achieve this by replacing the original 2D-Conv with 3D-Conv, and 2D-BatchNorm with 3D-BatchNorm. For DiT, we use space-time divided attention as in TimeSformer~\citep{bertasius2021space}.\\

\begin{table*}
\small
\setlength{\tabcolsep}{2pt}
    \caption{Results on exo to ego translation task.}
    \begin{subtable}{.5\linewidth}
      \centering
      \captionsetup{width=.9\linewidth}
        \caption{Evaluation of translation baselines for the subtask of ego track prediction.}
        \label{tab:res:track_prediction}  
         \begin{tabular}{lcccc}
\toprule
 Method  & \makecell[c]{Ego \\ Cam. Pose}& \makecell[l]{Location \\ Error$\downarrow$} & \makecell[l]{Contour \\ Acc.$\uparrow$}  &  \makecell[l]{IoU $\uparrow$} \\
pix2pix-mask &  No & 21.6 & 4.5 & 5.3 \\
$+$multi-frame & No & 20.1 & 3.1 & 3.5 \\
\midrule
GNT-mask & Yes & \textbf{19.6} & \textbf{15.5} & \textbf{10.3} \\
  \end{tabular}
    \end{subtable}%
    \begin{subtable}{.5\linewidth}
      \centering
      \captionsetup{width=.9\linewidth}
        \caption{Evaluation of translation baselines for the subtask of ego clip generation.}
        \label{tab:res:clip_generation}  
         \begin{tabular}{lccccc}
\toprule
 Method & SSIM $\uparrow$ &PSNR $\uparrow$ & DISTS $\downarrow$ & LPIPS $\downarrow$ & CLIP $\uparrow$\\
pix2pix-pix &  0.42 & 16.4 & 0.36 & 0.50 & 79.8 \\
DiT-pix & 0.59 & 16.1 & 0.31 & 0.46 & 81.9 \\
  \end{tabular}
    \end{subtable} 
\end{table*}

\paragraph{Results}

We employ the validation set for the purpose of selecting optimal checkpoints and hyper-parameters, which are subsequently evaluated on the test set.

{For the task of Ego Track Prediction, 
 \NEW{both pix2pix-mask and GNT-mask perform poorly in estimating the object visibility, achieving Visibility accuracy around 50\%, i.e., same as random guess (50.0\% for GNT-mask and 56.2\%  for pix2pix-mask, on the v2 test set). However, the ResNet-50 trained exclusively to attend to this binary classification achieves a VA of 82.9\% on the v2 test set.} %
We assess mask quality by considering distance (Location Error) and similarity metrics (IoU and Contour Accuracy) between predicted and ground-truth masks \NEW{after registration}. The 3D-aware NeRF-based baseline, GNT-mask, outperforms the implicit baseline, pix2pix-mask, overall. However, it does so by exploiting the ego camera pose as additional input. It is noteworthy that both baselines perform poorly on this task, likely due to the inherent challenges in correctly predicting the location and shape of the target object in the ego view, probably due to the fact that it often has diminutive size in the exo view.}

\begin{figure*}
    \centering
    \includegraphics[width=\linewidth]{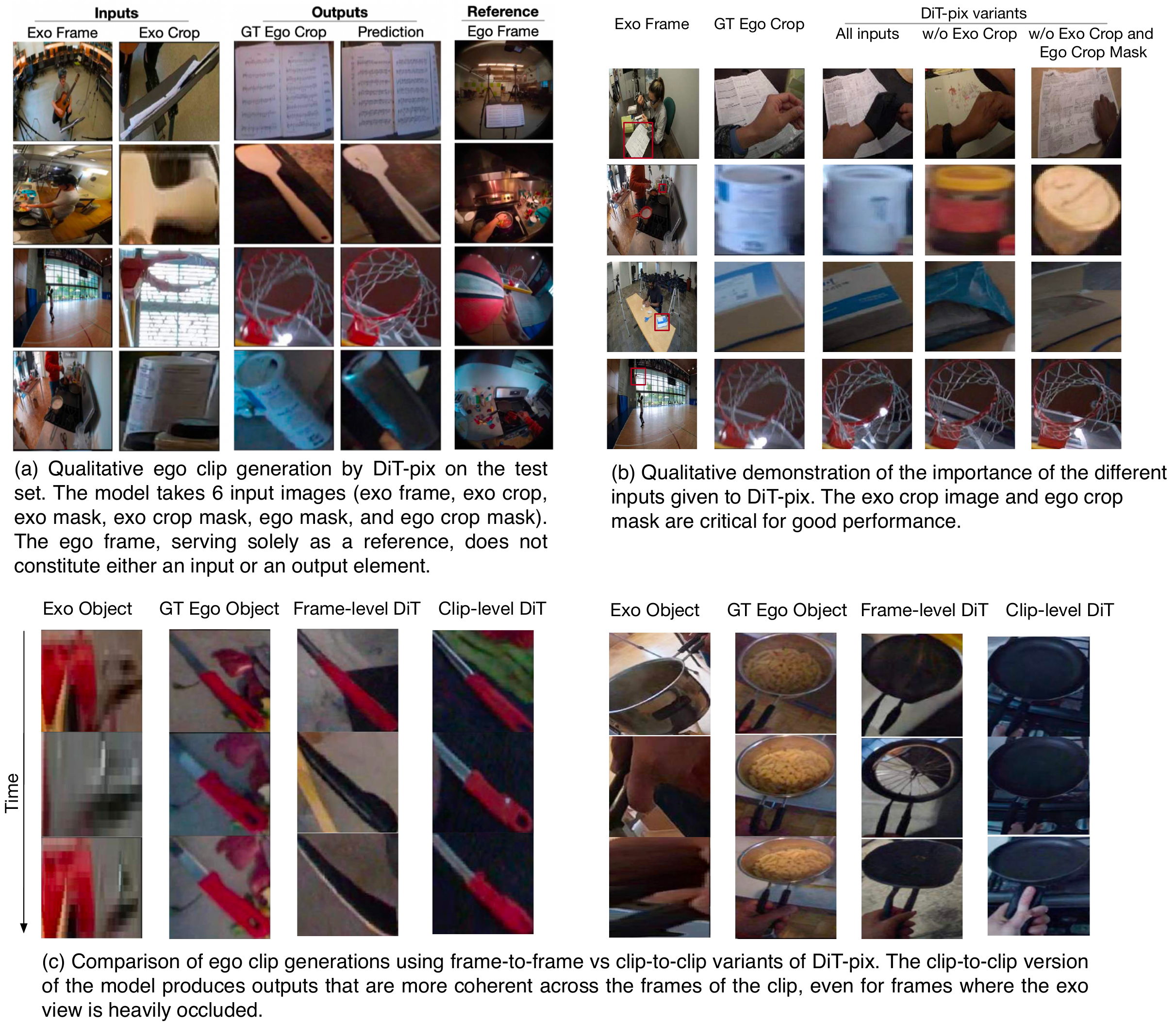}
    \caption{Qualitative results for exo-to-ego translation task.}
    \label{fig:translation}
\end{figure*}

In the case of Ego Clip Generation (Table~\ref{tab:res:clip_generation}), the Diffusion model DiT-pix demonstrates superior performance across all metrics compared to the GAN-based pix2pix-pix. Qualitative results (Figure~\ref{fig:translation}a) illustrate that DiT-pix can generate highly photorealistic images, aligning closely with the ground-truth in most instances. However, there are occasional cases (the last 2 rows) where the shape of the object is accurately generated, but the texture deviates slightly. 

{ We further verify the importance of each input in
Figure~\ref{fig:translation}b. Without exo object crop as input, the model fails to correctly infer the color and texture of the target object in the ego view. This result is expected as the source objects often represent a very small region of the entire exo frame. Additionally, without the ego crop mask as input, the model predicts the orientation of the object incorrectly. These observations highlight the importance of the cropped inputs.} 

\NEW{We can observe in Table~\ref{tab:res:track_prediction} that multi-frame (i.e., clip-to-clip) prediction does not provide a quantitative advantage over frame-to-frame prediction. Yet, we noticed that the multi-frame variant often yields generations that are more consistent across frames, even for frames where the exo view is heavily occluded, as can be seen in Figure~\ref{fig:translation}c. This is reasonable as a clip-level model can more effectively learn about the target object from multiple frames and fill-in information that is missing in individual exo frames.}

Please see Appendix \ref{sec:appendix:translation} for a break down of ego-exo translation results across different scenarios.

%% file: sec/benchmarks/keysteps-benchmark.tex
\begin{figure*}[t]
    \centering
    \includegraphics[width=\linewidth]{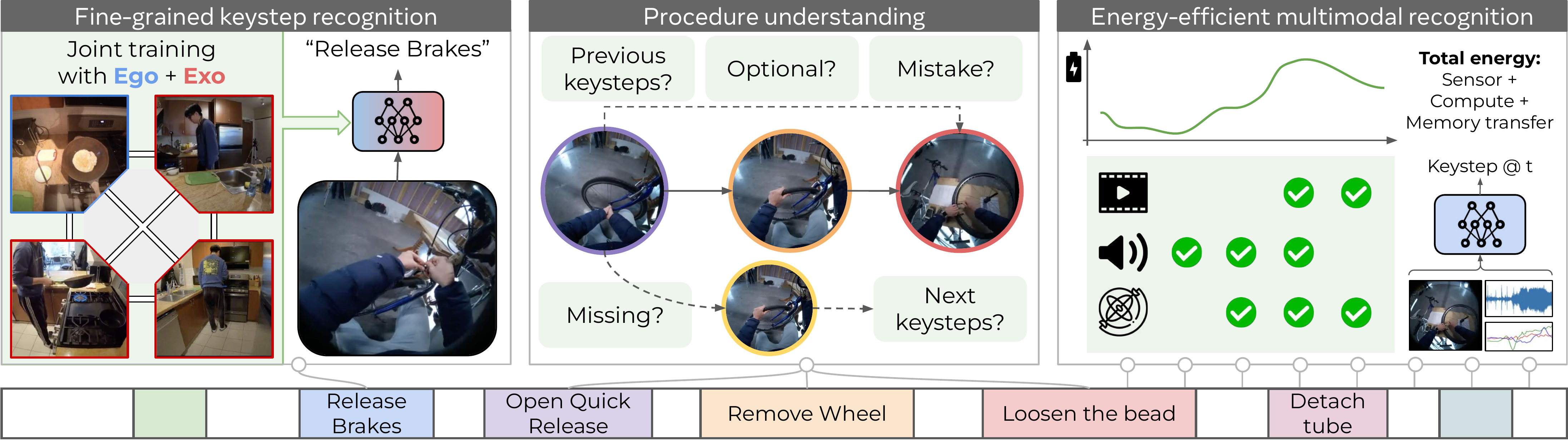}
    \caption{\textbf{Ego-exo keystep recognition.} This family of tasks consists of fine-grained recognition (left, Section~\ref{sec:keystep_finegrained}), procedure understanding (center, Section~\ref{sec:keystep_task_graph}), and energy-efficient multimodal recognition (right, Section~\ref{sec:keystep_energy_efficient}).} 
    \label{fig:recognition_tasks}
\end{figure*}

\paragraph{Motivation}
Recognizing the step a camera wearer is performing is non-trivial: keysteps in the same activity may look similar and may involve differentiating subtle differences in hand-object interactions with heavy occlusions and head motion.  %
Models with access to multiple views during training can leverage their complementarity to account for the deficiencies of each one, by learning viewpoint invariant representations or distilling multi-view signals into a single model (e.g., human hands from ego; body pose from exo). 

\paragraph{Task definition}
We consider trimmed video clip classification as the keystep recognition task. At training time we are given a labeled collection $\mathcal{D}$ of ego-exo video clips: $\mathcal{D} = \{(\mathcal{V}^{(1)}_{ego},\mathcal{V}^{(1)}_{{exo}^{1-M}}, y^{(1)}), \hdots, (\mathcal{V}^{(N)}_{ego},\mathcal{V}^{(N)}_{{exo}^{1-M}}, y^{(N)}) \}$ where $y^{(n)}$ denotes the keystep label of the $n$-th sample. The video clips are manually trimmed from long procedural videos to contain only the keysteps to recognize. At test time, given {\em just} the ego view of a trimmed clip $\mathcal{V}_{ego}$, the model must predict its keystep label $y$.

Classification of trimmed video clips is a problem formulation commonly adopted in action recognition benchmarks~\citep{kinetics,ucf,goyal2017something}. However, our task differs from action recognition in three fundamental aspects. First, it targets fine-grained keystep recognition rather than classification of coarse activities. We note that this adds significant complexity, since different keysteps of an activity often involve manipulating the same objects in the scene (e.g., folding the bedsheet and smoothing out the bedsheet) and are consequently difficult to tell apart. Second, different keysteps may be represented over largely different time spans (e.g., the average time span for ``kneading dough'' is 87.3 seconds, in stark contrast with ``getting salt'', which averages at 3.6 seconds), thus requiring analysis at different levels of temporal granularity. The third key difference is the potential to leverage contextual cues available in exocentric videos during training to improve the prediction accuracy on egocentric videos. 

\NEW{Note that at test time, the input to the model includes just the ego-view videos (RGB only). 
Exo-view videos, activity and scenario names, narrations, audio and associated metadata such as eye gaze, 3D point clouds, camera pose, and IMU information are \KGnew{\emph{excluded} as inputs for inference}. Intuitively, these additional modalities could provide valuable contextual cues, such as environmental awareness from exo-view videos, semantic meaning from narrations, or attentional signals from eye gaze, which could help the model better understand the visual content of the ego-view videos and improve its keystep recognition performance. We encourage exploring their potential utility in training to leverage these benefits, but for the purpose of evaluation, we restrict the input to RGB video only at test time to ensure our approach remains vision-centric.%
}

\paragraph{Related work}
Keystep recognition has been studied in first-person~\citep{sigurdsson2018charades,goalstep,EgoProceLECCV2022,ragusa2021meccano} or third-person~\citep{htstep,tang2020comprehensive,zhukov2019cross,zhou2018towards,ashutosh-neurips2023} videos; however, limited work considers both views together.  %
Prior work considers cross-view learning with unpaired %
videos~\citep{ardeshir2018exocentric,li2021ego,sherry-neurips2023} and view-invariant feature learning on paired videos~\citep{sigurdsson2018actor}. %
In contrast, we explore keystep recognition in large-scale, procedural activities with fully synchronized training videos.

\paragraph{Annotations}

\begin{figure*}[!h]
    \centering
    \includegraphics[width=\linewidth]{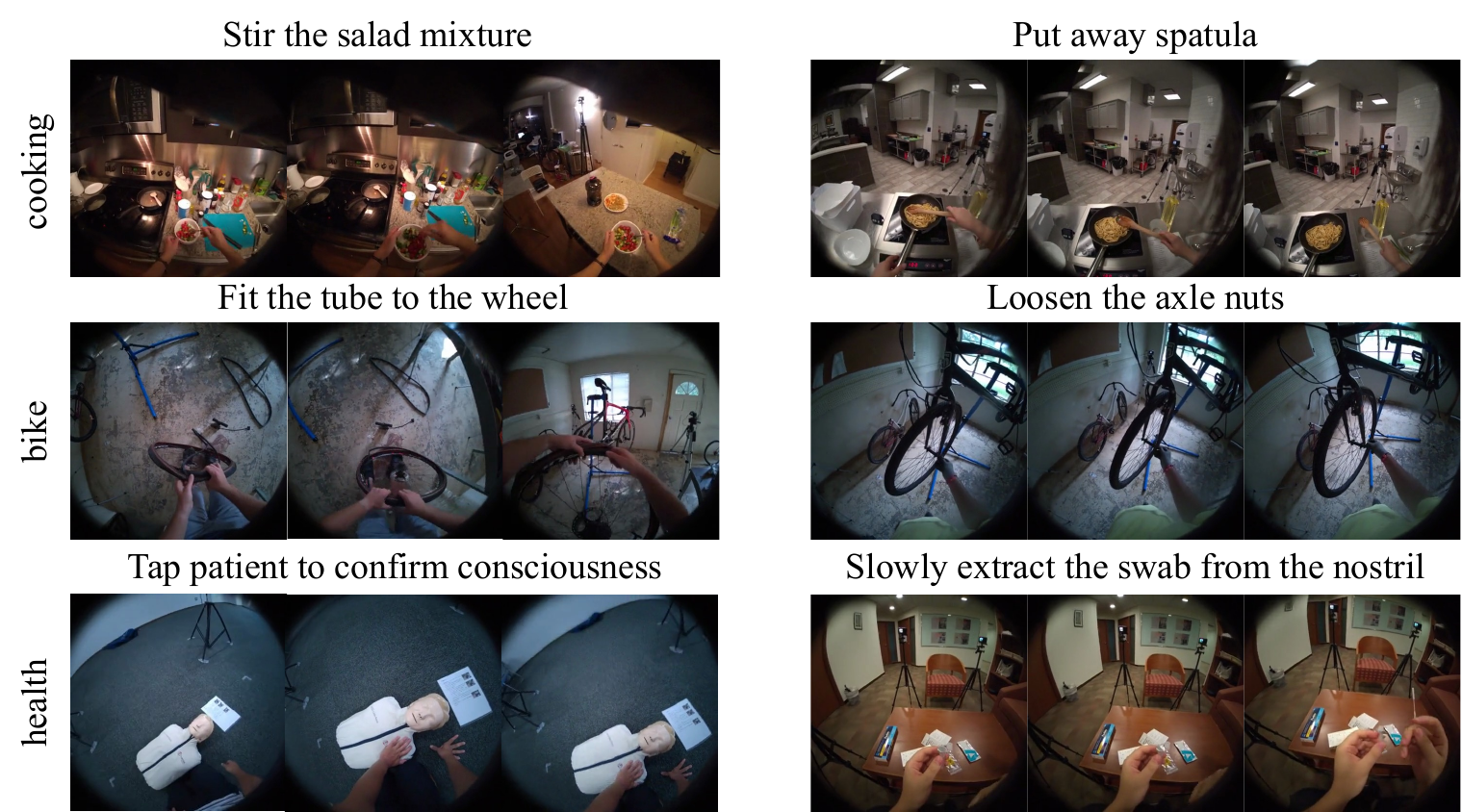} 
    \caption{Example keysteps from cooking, bike repair, and health scenarios. Keystep labels are displayed above each frame sequence.}
    \label{fig:keystep_example}
\end{figure*}

We annotate videos featuring any of the three procedural activities (i.e., cooking, bike repair, health) with temporal segments of \textit{keysteps}, i.e., actions that contribute  towards the completion of a procedural task. Each keystep annotation contains the start and end timestamps, a category label, a natural language description \TNIJCV{(e.g., ``add dried herbs''or ``fit the tire onto the bike'')}, and a flag indicating whether the keystep is essential or optional for task completion. To accurately model the hierarchical nature of the activities, we also develop a hierarchical keystep taxonomy concurrently with the annotation process\TNIJCV{, in an iterative, data-driven manner. In total, we annotate 143,442 segments, spanning 664 keysteps across 17 activities. Figure \ref{fig:keystep_example} shows example keystep annotations, highlighting the challenges of fine-grained keystep recognition where subtle differences in hand-object interactions and contextual cues are crucial for distinguishing between activities. Complete details on the annotation interface and taxonomy development are in Appendix~\ref{sec:appendix:keystep_recognition}.}

\begin{figure*}[!th]
    \centering
    \includegraphics[width=\linewidth]{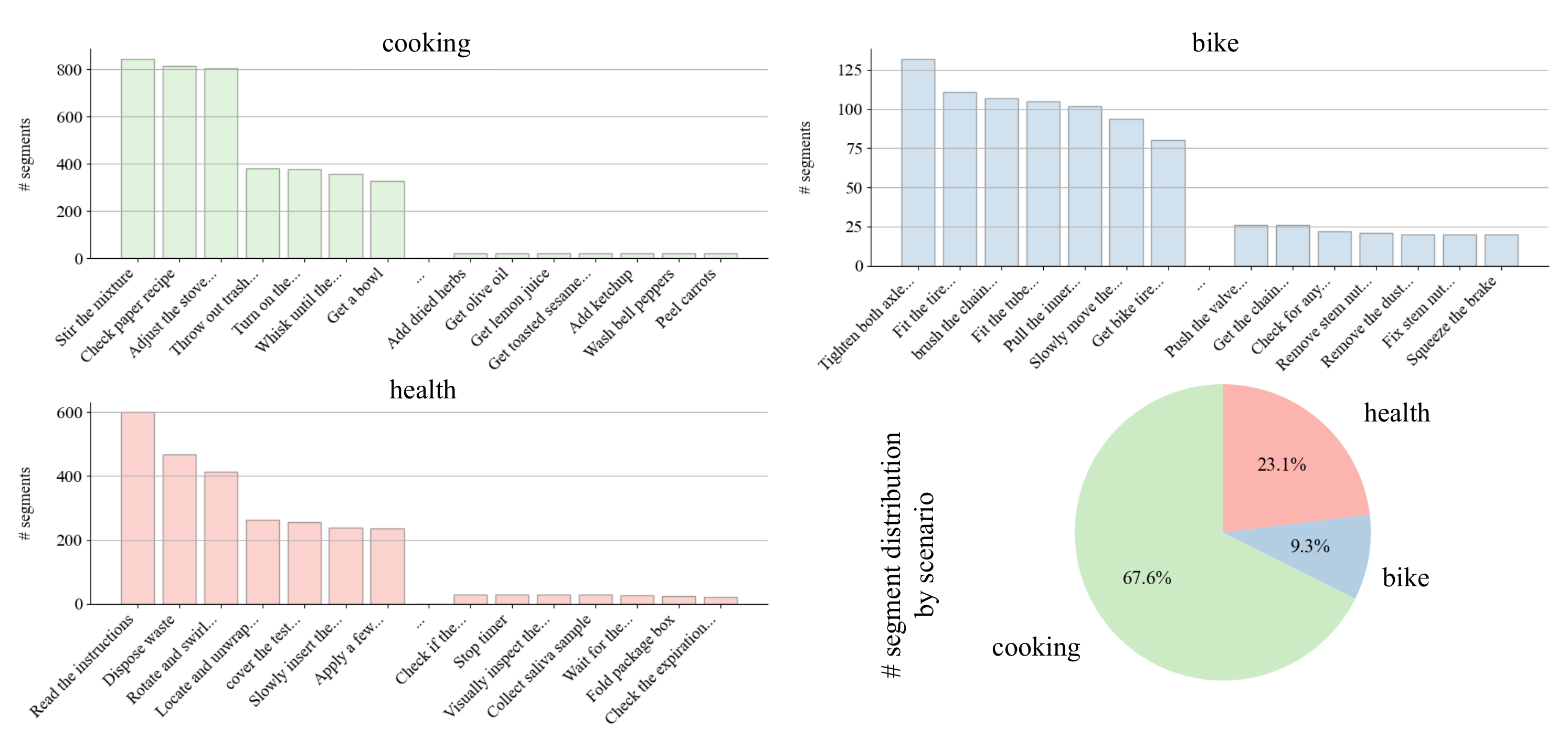} 
    \caption{Keystep distribution in our dataset for each procedural scenario: cooking, bike repair, and health.}
    \label{fig:baseline_keystep_distribution}
\end{figure*}

\paragraph{Metrics}
\TNIJCV{We report top-1 accuracy for evaluation.} Since the keysteps in our dataset exhibit a very long-tailed distribution, we set a cutoff threshold at 20 samples per keystep, limiting our analysis to 278 unique keysteps as shown in Figure \ref{fig:baseline_keystep_distribution}. \NEW{Some of these keysteps are illustrated in Figure \ref{fig:keystep_example}}. \TNIJCV{Dataset split details are in Appendix~\ref{sec:appendix:keystep_recognition}.}  

\paragraph{Baselines}
To understand the best strategy for egocentric keystep recognition with paired ego-exo training data, we consider a diverse set of baselines approaches, including methods for action classification, video representation learning, and ego-exo transfer. 
\begin{itemize}
    \item \textbf{Action classification.} As a prototypical example of this classic genre, we select a TimeSformer~\citep{bertasius2021space} model initialized with the checkpoint pretrained on the large-scale third-person action dataset Kinetics-600~\citep{kinetics} due to its strong performance in various video understanding tasks.\\
    \item \textbf{Video-language pretraining.} We adopt the EgoVLPv2 framework \citep{pramanick2023egovlpv2} and pre-train the model jointly on the Ego4D \citep{grauman2022ego4d} (which contains only ego views) and the Ego-Exo4D datasets (which encompasses both ego and exo views). We balance the number of samples between these two datasets by augmenting Ego-Exo4D with LaViLa-style \citep{zhao2023learning} narration rephrasing.\\ %
    \item \textbf{View-invariant learning.} A two-stage training approach is employed. In the first stage, we utilize all available (ego, exo) video pairs in the dataset for training a view-invariant (VI) encoder. The training objective is a clip-level contrastive loss~\citep{oord2018representation}, aiming at identifying the synchronized (ego, exo) pairs as positive, and the non-synchronized pairs as negative. In the second stage, this pretrained model is further trained with a classification loss, aligning with the clip-level classification nature of the downstream task. Note that to align with the clip-level classification task, our contrastive loss operates at the clip-level, rather than at the frame-level as was done in view-invariant loss proposed in~\citep{sermanet2017time,sigurdsson2018actor}. \\
    \item \textbf{Viewpoint distillation.} This also adopts a two-stage training approach. In the first stage, we train a multi-view teacher that takes both ego and exo views as input. In the second stage, a single-view ego student is trained, distilling knowledge~\citep{hinton2015distilling} from the multi-view teacher to encapsulate information from both views.\\
    \item \textbf{Ego-exo transfer.} Here we follow the methodology proposed in Ego-Exo~\citep{li2021ego} which uses egocentric pseudo-labels to pre-train the network. We employ a masked autoencoder (MAE)~\citep{videomae} backbone, initialized from a Kinetics checkpoint, and the pseudo-labels provided from the Ego-Exo checkpoint to fine-tune with two auxiliary heads (Object-Score and Interaction-Map).  We then further finetune the model with a classification loss for fine-grained keystep recognition.

\end{itemize}
For the first two baselines (which utilize pretrained checkpoints from well-established benchmarks), two training settings are further examined: one using only the ego view for the classification loss and the other utilizing both ego and exo view videos, with the training objective being the sum of ego view and exo view classification losses. %
\TNIJCV{Implementation details are in Appendix~\ref{sec:appendix:keystep_recognition}.}

\begin{table*}[ht]
\centering
    \small
    \setlength{\tabcolsep}{5pt}
\begin{tabular}{lc|cc}
& & \multicolumn{2}{c}{Ego Accuracy (\%)} \\ \hline
Method    &  Train data & Val & Test \\ 
\hline
TimeSFormer~\citep{bertasius2021space} (K600) & ego & 35.13 & 35.93\\
TimeSFormer~\citep{bertasius2021space} (K600) & ego,exo & 32.68 & 31.04 \\
EgoVLPv2~\citep{pramanick2023egovlpv2} (Ego4D) & ego & 36.51 & 37.55 \\
EgoVLPv2~\citep{pramanick2023egovlpv2} (Ego4D) & ego,exo & 35.84 & 36.59 \\
EgoVLPv2~\citep{pramanick2023egovlpv2} (EgoExo4D) & ego & 36.04 & 37.72 \\
EgoVLPv2~\citep{pramanick2023egovlpv2} (EgoExo4D) & ego,exo & \underline{39.10} & 38.76 \\
VI Encoder~\citep{oord2018representation} (EgoExo4D) & ego,exo & \textbf{40.34} & \textbf{41.53} \\
Viewpoint Distillation~\citep{hinton2015distilling} & ego,exo & 38.19 & \underline{39.49} \\
Ego-Exo Transfer MAE~\citep{li2021ego} & ego,exo & 37.17 & 36.58\\ 
\hline
\end{tabular}
\caption{Top-1 accuracy of keystep recognition on val and test data. The pre-training dataset is denoted in parentheses. VI is short for view-invariant.} 
\label{tab:keystep}
\end{table*}

\paragraph{Results} 
Table \ref{tab:keystep} reports the Top-1 accuracy for ego classification on both validation and test sets. Among all the baselines, the VI Encoder emerges as the top performer, achieving a test accuracy of 41.53\%. It is closely followed by Viewpoint Distillation and EgoVLPv2 pretrained on EgoExo4D, which attain test accuracies of 39.49\% and 38.76\% respectively.  
These results open discussion on how to effectively utilize exo videos during training to enhance ego keystep recognition during test time. 

First, we note that different approaches respond differently to the addition of exo-view videos during training. Specifically, while the TimeSFormer (K600) exhibits a degradation when the exo classification loss is integrated into the objective (i.e., test accuracy drops from 35.93\% to 31.04\%, EgoVLPv2 pretrained on EgoExo4D benefits from the introduction of exo-view videos (i.e., test accuracy improves from 37.72\% to 38.76\%. This enhancement is also evident in the VI encoder and viewpoint distillation when compared to TimeSFormer (K600) that only utilizes ego-view videos for training. These observations suggest that certain baselines are better equipped at leveraging exo information during training to improve ego keystep recognition.

\TNIJCV{Finally, in Appendix Figure~\ref{fig:baseline_keystep_detailed}, we show a breakdown of performance by viewpoint. In short, we find that ego views are more informative for steps involving manipulation of small objects, like `cut carrots' and `unpack the new tube', while exo views show advantages for keysteps like `have a conversation asking different questions'.} Overall, we posit that the endeavor to enhance view-invariant learning and to more effectively harness the complementary information from exo views for ego keystep recognition remains an open avenue. Our findings underscore the need for further investigation and innovation in this domain.

%% file: sec/benchmarks/multimodal-benchmark.tex
\paragraph{Motivation}
Current activity detection models assume access to densely sampled clips from the full video and ample computational resources to process them. These assumptions are incompatible with real-world devices (e.g., mobile phones, AR glasses) where the camera is not always on and the compute budget is limited by battery life. This task focuses on building energy-efficient video models to pave the way for feasibility on real-world hardware.  

\paragraph{Task definition}
Whereas the keystep recognition task (presented in Sec.~\ref{sec:keystep_finegrained}) entails classifying keystep video clips in batch without regard for energy costs, in this task, the goal is to perform \emph{online} classification of keysteps in a streaming egocentric multi-modal video, within an energy budget. We consider an ego video {$\mathcal{V}_{ego}$} of arbitrary length $T$ comprising a stream of $K$ different sensory modalities (e.g., RGB images, audio, IMU, etc.). At each time step $t$, where $1 \leq t \leq T$, the video consists of  samples for each available modality, such that 
${\mathcal{V}_{ego}^t = \{S^t_1,  \ldots, S^t_K\}}$, where $S^t_j$ denotes the sample {at time $t$} for the $j^{\text{th}}$ modality.

Given {$\mathcal{V}_{ego}$}  and an energy budget $B$, our task is to learn a model $\mathcal{F}$ that maximizes the overall keystep recognition performance across the full video while also ensuring that the combined energy for sensing and running model inference does not exceed $B$. $F$ consists of a sensor triggering policy $\mathcal{F}^{\mathcal{P}}$ and a keystep prediction model $\mathcal{F}^{\mathcal{K}}$. At every step $t$, the policy $\mathcal{F}^{\mathcal{P}}$ decides which sensor(s) to activate and sample from, in order to produce the model's current observation ${O^t}$, such that $O^t \subseteq \{S^t_1, \ldots, S^t_K\}$. Given $O^t$, the keystep predictor $\mathcal{F}^{K}$ outputs its estimate of the ground truth keystep for the current step.

The energy budget accounts for the cost of operations in each model forward pass, the cost of moving intermediate activations in and out of memory and the cost of the continuous operation of sensors, each having different cost profiles (e.g., IMU and audio sensors are relatively cheaper to operate than camera sensors). 
Note that the sensor triggering policy may be static (e.g., sample video at 4 frames per second (fps), keep audio/IMU off; sample 1 fps video, keep IMU always on) or dynamic (e.g., depending on the audio, decide whether to trigger video capture). We keep our task definition general, allowing the challenge to admit a wide variety of recent approaches ranging from pure video-based efficient backbone architectures \citep{feichtenhofer2020x3d} to multi-modal triggering approaches and, naturally, a combination of them.

Note that at test time, the input to the model can only include current and past observations as our task is strictly an online recognition task. However, we encourage exploring modalities beyond those considered in our experiments, e.g., IMU or camera poses inferred from video, audio, and IMU.

\paragraph{Related work}
Prior work on efficient models considers light-weight architectures~\citep{feichtenhofer2020x3d,vasu2023mobileone,howard2017mobilenets,zhang2018shufflenet,tan2019mnasnet,mehta2021mobilevit}, efficient input processing~\citep{gao2020listen,korbar2019scsampler,ghodrati2021frameexit,meng2020ar,tan2023egodistill}, or inference optimizations~\citep{iandola2016squeezenet, esser2019learned, polino2018model, zhu2017prune,blockdrop}.
In all cases, they optimize computation (FLOPs), parameter count, or prediction throughput (FPS), which in isolation are insufficient to characterize running on real-world devices. To address this, we propose the first benchmark for \emph{energy-efficient} video recognition that is tied to real-world, on-device constraints, and measures total power consumed.

\paragraph{Annotations}
This task uses the same egocentric videos and annotations as keystep recognition. However, in addition to the raw RGB video, it uses the audio stream (and potentially other sensors) as another sensor modality.

\paragraph{Measuring energy consumption} 
Accurately measuring energy consumption of models is crucial for their use in AR/VR devices~\citep{abrash2021creating,chen2019eyeriss}. The energy used comes from a complex interplay of sources including sensors, compute, communication, data processing, memory transfer (SRAM and DRAM), and leakage -- many of which are typically ignored when building \emph{efficient} computer vision models, despite their large energy consumption (e.g., memory transfer accounts for over 50\% of the total power~\citep{yang2022three}). 

We consider three key factors when modeling energy consumption following prior work~\citep{sze2020evaluate}. (1) Compute energy: the cost of each model forward pass as a function of the number of operations (MACs). (2) Memory transfer energy: the cost associated with memory read-write operations for storing intermediate activations and model outputs. (3) Sensor triggering energy: the cost associated with turning on / off and continuous operation of sensors (camera, audio, IMU). For a model that processes an observation $O^t$, the total energy consumed can then be formulated as:
\begin{eqnarray}
    \label{eq:energy}
    E(O^t) &=& \alpha * C(O^t) + \beta * M(O^t) + \nonumber \\
    & & \sum_{j=1 \ldots K} \gamma_{j} * \mathbbm{1}(S_j \in O^t)
\end{eqnarray}
\noindent where $C(O^t)$ corresponds to the total number of multiply-add operations computed during the forward pass (in MAC/s), $M(O^t)$ corresponds to the total memory transferred to/from DRAM (in MB/s), and $S_j \in O^t$ corresponds to whether the $j$-th sensor is active. Finally, $\alpha,\beta,\gamma_j$ are weighting factors that measure the contribution of each energy source.
We select these weighting parameters to reflect real-world AR/VR hardware capabilities. Namely, $\alpha$ = 4.6 pJ/MAC~\citep{sze2020evaluate,desislavov2023trends}; $\beta$ = 80 pJ/byte~\citep{horowitz20141}; $\gamma_{rgb}$ = 15 mW and $\gamma_{audio}$ = 0.5 mW~\citep{liu20204}.

\TNIJCV{We adapt off-the-shelf profiler software built for PyTorch to compute the quantities in Eqn.~\ref{eq:energy} -- the energy consumption expressed as power (mW). Details about the profiler are in Appendix~\ref{sec:appendix:energy_efficient}}.

\paragraph{Metrics}
Following prior work~\citep{de2016online}, we evaluate online keystep detection performance using 
per-frame calibrated mean average precision (mcAP), which accounts for the imbalance in the keystep labels in our dataset.
We measure energy consumption in mW as described above. There is a natural trade-off between efficiency and better performance. Thus, we evaluate models in two tiers by setting a budget for the power consumption in each tier, namely 20 mW for the \emph{high-efficiency} tier and 2.8W for the \emph{high-performance} tier\TNIJCV{, selected based on existing efficient architectures. More details about the tiers are in Appendix~\ref{sec:appendix:energy_efficient}.}

\paragraph{Baselines}
We provide a family of (less/more expensive) keystep prediction models for solving the task. Each model has a unimodal or audio-visual feature encoder followed by a keystep classification head. \TNIJCV{Experimental setup and implementation details are in Appendix~\ref{sec:appendix:energy_efficient}.}

\begin{itemize}
\item \textbf{X3D-XS~\citep{feichtenhofer2020x3d}.} This is a vision-only model comprising the X3D-XS feature encoder, which progressively expands the feature size and representational capacity of its layers, and later contracts them for achieving better performance-efficiency trade-off. \\

\item \textbf{LaViLa~\citep{zhao2023learning}.} This is another vision-only model where the visual feature encoder is trained through CLIP-style video-language pre-training.\\

\item \textbf{Light-ASDNet~\citep{liao2023light}.} This is an audio-only model that represents audio as spectrograms and efficiently encodes them by splitting 2D convolutions into 1D convolutions along the spectrogram temporal dimension~\citep{liao2023light}.\\

\item \textbf{Audio-Visual Late Fusion (AV-LF).} This is an audio-visual model that does late fusion of visual features (encoded with X3D-XS or LaViLa) and audio features from Light-ASDNet by using linear layers.

\end{itemize}

\begin{figure*}[ht]
\captionsetup[subfloat]{}
\centering
\begin{subfigure}[t]{0.45\textwidth}
\includegraphics[width=\textwidth]{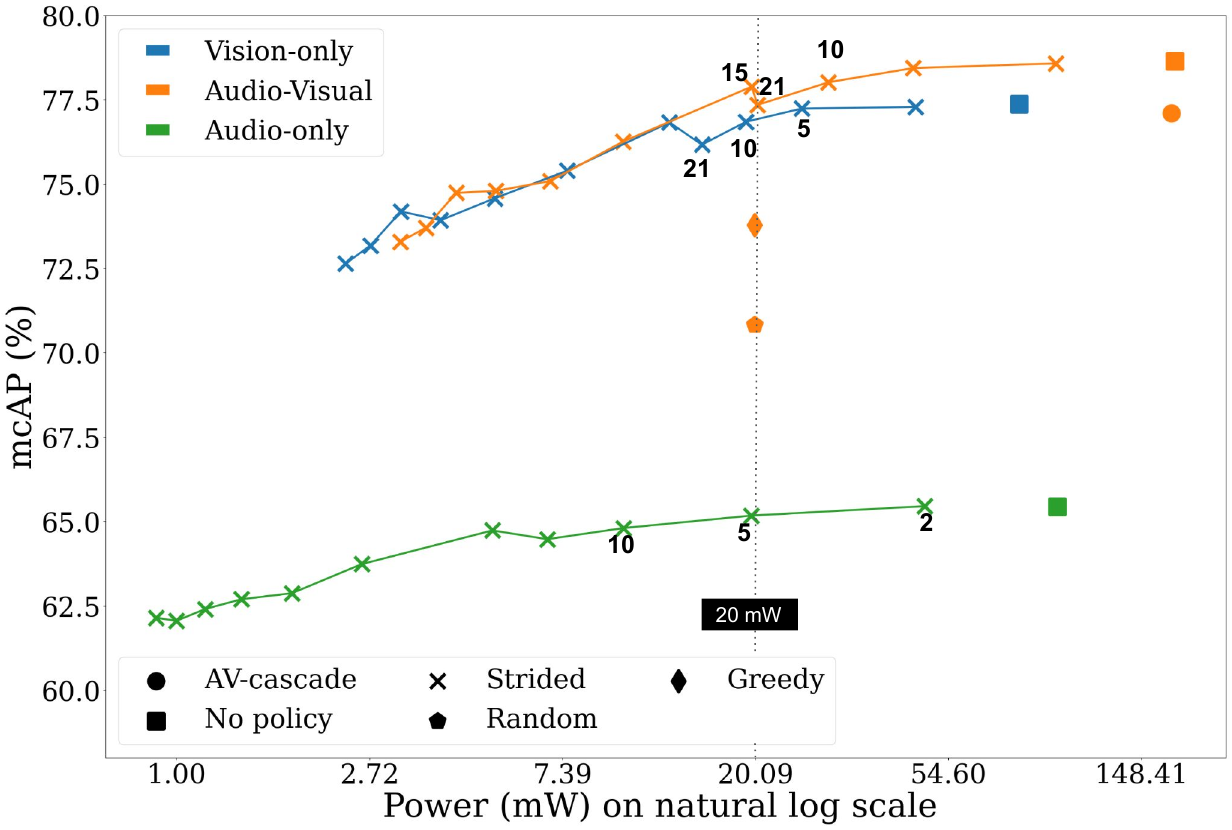}
  \caption{\emph{High-efficiency} tier (budget = 20 mW)}
  \label{subfig:eeAll_x3d}
\end{subfigure}%
~
\begin{subfigure}[t]{0.45\textwidth}
\includegraphics[width=\textwidth]{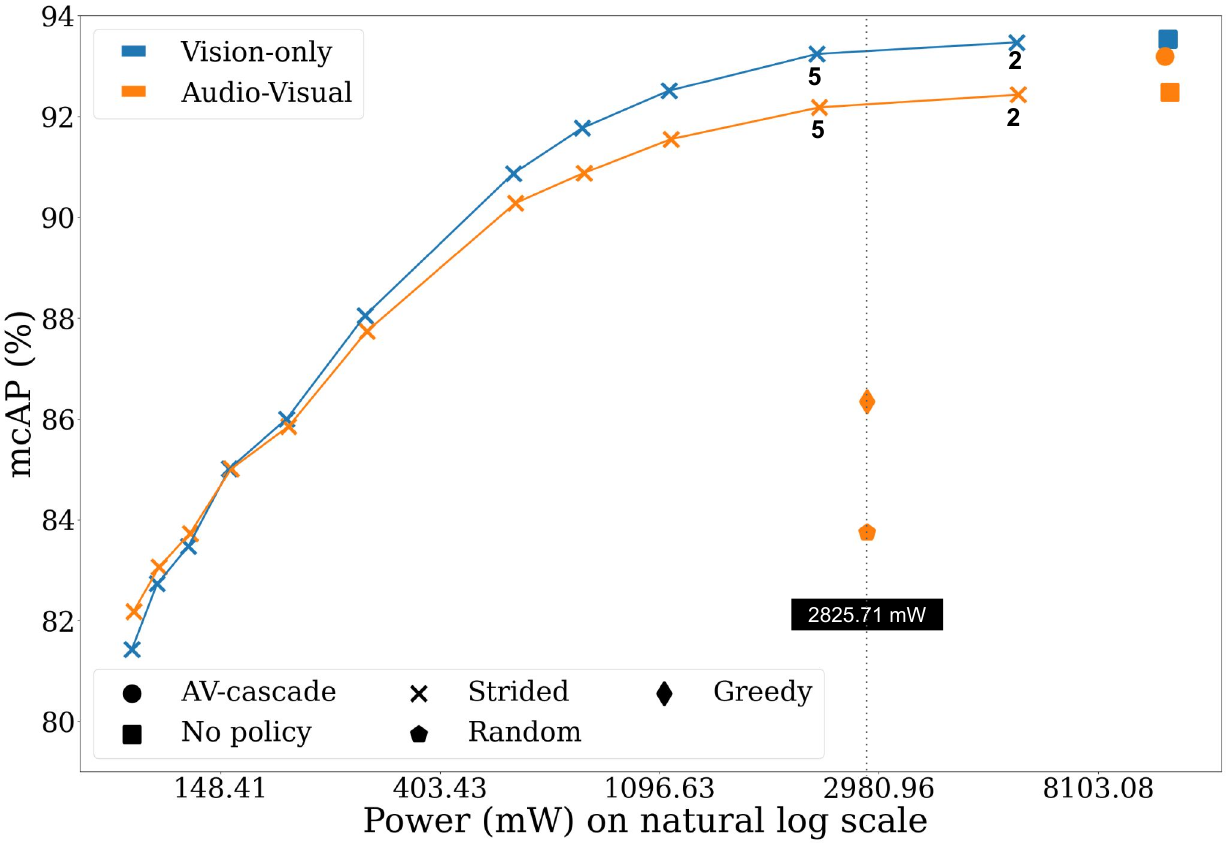}
  \caption{\emph{High-performance} tier (budget = 2825.71 mW)}
  \label{subfig:eeAll_lavila}
\end{subfigure}

\small
\caption{\SM{Keystep prediction performance (mcAP) vs. total power consumption with different 
prediction backbones and sampling policies for both \emph{high-efficiency} (left) and \emph{high-performance} (right) tiers. For the models using a fixed stride, we show their stride value in text if their total energy consumption is close to the budget.}
}
\label{figure:ee-all}
\end{figure*}

To improve the energy efficiency of the aforementioned keystep predictors, we employ the following baseline policies for determining when to sample or skip each modality:

\begin{itemize}
\item \textbf{Fixed stride.} This is a policy that samples the input (video or audio) every $s$ prediction steps. We evaluate different $s$ values, where $s$ ranges from 2-150 steps.\\ 

\item \textbf{AV-LF + greedy.} This is a policy that greedily uses  up the budget by sampling both audio and vision as early as possible, and uses the AV-LF backbone for keystep prediction.\\

\item \textbf{AV-LF + random.} This is a policy that randomly samples or skips the audio and/or visual inputs until it runs out of budget, and uses the AV-LF backbone for prediction.\\

\item \textbf{Audio-Visual (AV) Cascade.} This is a policy that initially uses the Light-ASDNet model to predict the keystep, and switches over to the LaViLa model if the audio-based prediction confidence is below a confidence threshold of 0.5.

\end{itemize}

\begin{table}[!t]
\small
\begin{tabular}{cc}
\begin{subtable}{\columnwidth}
\resizebox{\columnwidth}{!} {
\begin{tabular}{lc|cc}
Method      & Modality & mcAP (\%) $\uparrow$   & Power (mW) $\downarrow$    \\ \hline
Light-ASDNet~\citep{liao2023light} + $s=5$ & A &  65.18    & 19.67 \\
X3D-XS~\citep{feichtenhofer2020x3d} + $s=10$ & V & 76.85    & \textbf{19.14}            \\
AV-LF w/ X3D-XS + $s=15$  &   AV   & \textbf{77.89}    &   19.70        \\
\end{tabular}
}
\caption{\emph{High-efficiency tier} (budget = 20 mW).}
\label{tab:tier1_best}
\end{subtable}
\\ 
\begin{subtable}{\columnwidth}
\resizebox{\columnwidth}{!} {
\begin{tabular}{lc|cc}
Method      & Modality & mcAP (\%) $\uparrow$  & Power (mW) $\downarrow$    \\ \hline
Lavila~\citep{zhao2023learning} + $s=5$ & V & \textbf{93.24 } &   \textbf{2245.66}        \\
AV-LF w/ Lavila  + $s=5$ &   AV       & 92.18 & 2274.40    \\
\end{tabular}
}
\caption{\emph{High-performance} tier (budget = 2.8W).}
\label{tab:tier2_best}
\end{subtable}
\end{tabular}
\caption{Keystep prediction results.}
\end{table}

\paragraph{Results}
In Fig.~\ref{subfig:eeAll_x3d}, we plot the recognition mcAP of all models against their total power consumption for the \emph{high-efficiency} tier. We can see that combining vision and audio is better than using only vision or audio. Thus suggests that the two modalities carry complementary cues that are useful for the task. However, all vision-only models outperform their audio-only counterparts, which indicates that vision is the most critical modality for the task. The raw backbones generally perform better than the models using a sampling policy, but at the cost of requiring higher energy, making them impractical to use in online settings. Among the models that use a fixed stride, a lower stride generally improves the performance while hurting energy efficiency. %
Using the greedy or random policy with AV-LF leads to a sharp decline in performance compared to using a fixed stride, showing that sampling very early or randomly in the episode is suboptimal for our online recognition task. AV-cascade also performs worse than most audio-visual models while also requiring more energy, possibly because the audio backbone often outputs wrong but over-confident predictions that prevent switching over to the more reliable vision backbone when required.  

For easy reference, in Table~\ref{tab:tier1_best} we report the recognition performance and total power consumption of our best uni-modal and audio-visual models within budget for the \emph{high-efficiency} tier.

In Fig.~\ref{subfig:eeAll_lavila}, we plot the recognition mcAP of all models against their total power consumption for the \emph{high-performance} tier. Different from the {high-efficiency} tier, the audio-visual backbone \NEWRESULT{generally} performs worse than the vision-only backbone, possibly because the LaViLa features are strong enough by themselves, and fusing them with audio features through the simple mechanism of linear late fusion reduces their expressivity. Otherwise, the overall behavior of different sampling policies is 
similar across the two tiers. We report the recognition performance and total power consumption of the best uni-modal and audio-visual models within the \emph{high-performance} budget in Table~\ref{tab:tier2_best}.

\TNIJCV{Finally, in Appendix Figure~\ref{figure:del_perf}, we present a breakdown of performance by keystep labels and the behavior of audio- and vision-only models across them. In short, we find audio-only models have an affinity for sounding actions like \emph{stir fry egg mixture} and \emph{cut butter}.}

%% file: sec/benchmarks/taskgraph-benchmark.tex
\paragraph{Motivation}
The procedure understanding task consists in inferring the underlying structure of a procedure from the observation of natural videos of subjects performing the procedure.

The real-world motivation for our procedure understanding task has basis in augmented reality (AR), robotics, and more in general in assistive systems. Indeed, automatically understanding the \emph{structure} of a procedure from video, e.g., inferring keystep orderings and preconditions, will allow to assist or guide users carrying out the procedure through AR or to allow robots to learn from human demonstrations. 
Beyond recognizing the current keystep, an assistive system could verify that some mandatory keysteps are missing, suggest possible future ones, and detect procedural mistakes. Similarly, robots could learn the structure of a procedure from human demonstrations. Mining the structure of procedures has been shown useful for planning~\citep{chang2020procedure,bi2021procedure} and improving keystep recognition~\citep{ashutosh-neurips2023,zhou2023procedure} keystep discovery~\citep{EgoProceLECCV2022}, and for procedural mistake detection~\citep{seminara2024differentiable}.

\paragraph{Task definition}
Figure~\ref{fig:recognition_tasks} (center) illustrates the proposed task.
Given a video segment $s_i$ and its segment history $S_{:i-1}=\{s_1, \ldots, s_{i-1}\}$, models have to 1) determine \textit{previous keysteps} (i.e., keysteps which should be performed before $s_i$); infer if $s_i$ is 2) \textit{optional} (i.e., it can be omitted without compromising the correct execution of the procedure) or 3) a \textit{procedural mistake} (a keystep which should not have been performed in that moment due to missing pre-conditions); 4) predict \textit{missing keysteps} (i.e., key-steps which should have been performed before $s_i$); and 5) forecast \textit{next keysteps} (i.e., key-steps for which dependencies are satisfied and hence which could be executed next).

The task is weakly supervised, with two versions based on the level of supervision:
1) instance-level: video segments and their keystep labels are available during training and inference, similar to an action recognition task; 2) procedure-level: unlabeled video segments and a taxonomy of procedure-specific keystep names are given for training and inference. Note that, being weakly supervised, in both cases, explicit information on the structure of the procedure---such as the occurrence of mistakes or lists of pre-conditions---are not available for training.
Also note that, when the procedure-level supervision is considered, the input to the model \emph{excludes} keystep labels both at training and test time. At both the procedure and instance levels of supervision, models are required to process the video in a causal fashion, meaning that predictions made at time $t$ only depend on observations made at time $t’<t$.

\paragraph{Related work}
Prior work focusing on procedural understanding learns an explicit graph \citep{jang2023multimodal,xu2020benchmark,soran2015generating} as ground truth or uses a task graph for representation learning \citep{ashutosh-neurips2023,task-structure,zhou2023paprika} and short-term step understanding \citep{dvornik2022flow,ashutosh-neurips2023,zhou2023paprika}. Other work \citep{sener2022assembly101,ding2023every} studies mistake detection in a supervised setting. We are the first to propose procedural understanding to evaluate the long-term structure of the task in a weakly-supervised setting.

\paragraph{Annotations}
For this task, we considered the following procedures: \textit{i.e., Covid-19 Rapid Antigen Test, Fix a Flat Tire - Replace a Bike Tube, Remove a Wheel, Install a Wheel, Clean and Lubricate the Chain and First Aid - CPR}. These scenarios represent structured activities with clear procedural constraint, yet allow a certain degree of variability in correct task executions.
For each of the considered procedures, we manually labeled task-graphs as structures encoding the keystep orderings leading to a correct execution of the procedure (detailed below). A task graph is meant as a way to encode all orders of keysteps which lead to a correct execution of the task.

\begin{figure}
    \centering
    \includegraphics[width=\columnwidth]{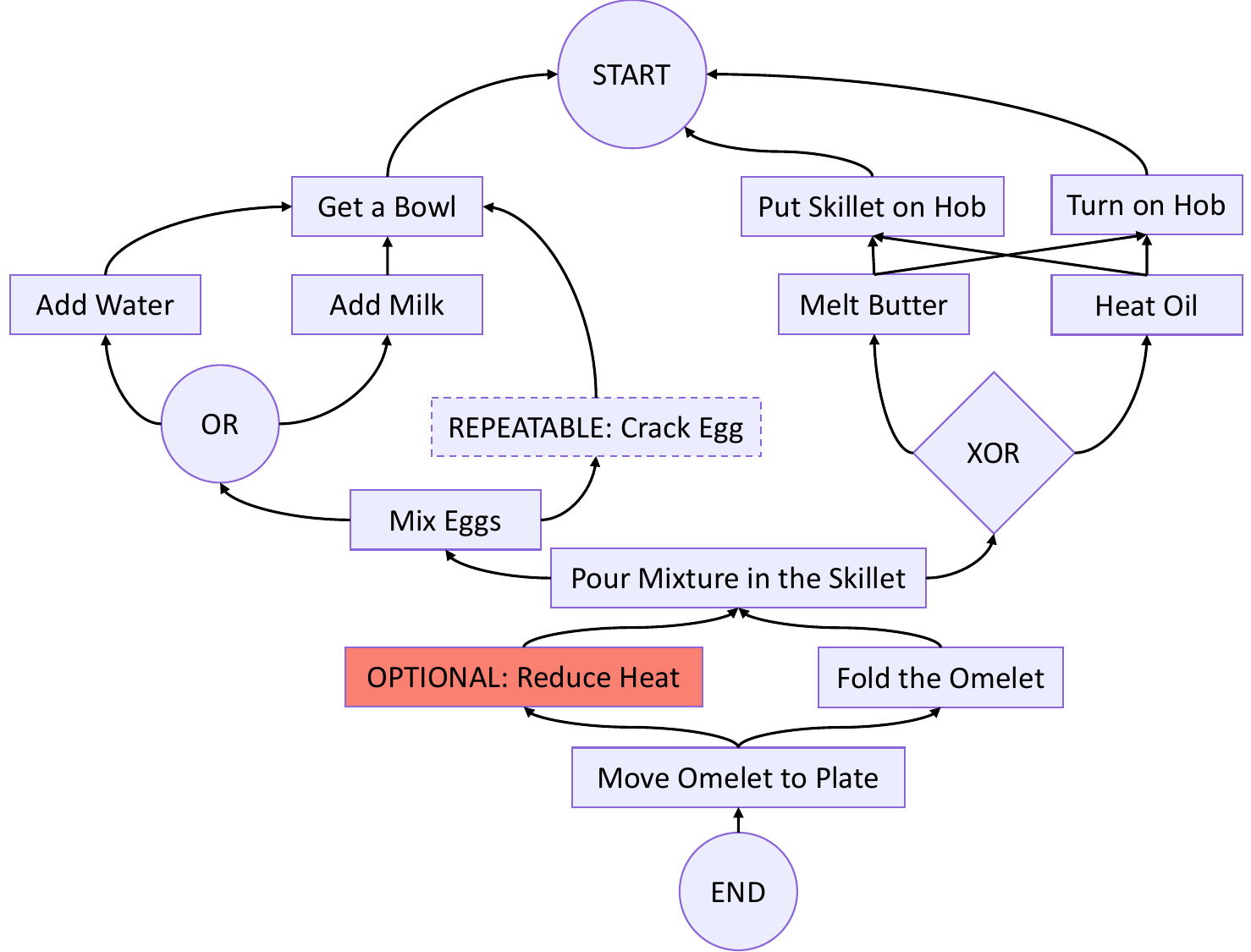}
    \caption{Example task-graph of a "Cooking Omelet" procedure. 
    }
    \label{fig:task_graph}
\end{figure}

\textit{Task-graphs.} We define a task-graph as a directed graph in which nodes represent keysteps and directed edges represent dependencies. For instance, in the example task-graph reported in Figure~\ref{fig:task_graph}, the ``Add Milk $\to$ Get a Bowl'' structure denotes that keystep ``Get a Bowl'' has to be executed before keystep ``Add Milk''. 
If a keystep has more than one dependency, all of them need to be satisfied. For instance, both ``Put Skillet on Hob'' and ``Turn on Hob'' need to be executed before ``Heat Oil''.
Besides directed edges, task-graphs also contain ``OR'' and ``XOR'' structures, which combine dependencies logically, as well as ``optional'' and ``repeatable'' node attributes. 
For instance ``Mix Eggs'' can be performed if either ``Add Water'' or ``Add Milk'' (or both) are executed, whereas ``Pour Mixture in the Skillet'' requires either ``Melt Butter'' or ``Heat Oil'' to be executed, but not both. Repeatable nodes (e.g., ``Crack Egg'') can be repeated as long as their outgoing nodes (pre-conditions) are satisfied and incoming nodes (future nodes) are not executed. For instance, one could keep cracking eggs as long as the bowl is in place, but not after mixing the eggs. An optional node (e.g., ``Reduce Heat'') can be omitted, but, if included, it needs its preconditions to be correctly satisfied.

\textit{Task-graph construction.} We first familiarized ourselves with the procedural tasks by watching videos with  annotated keysteps. We then initialized task graphs with procedural dependencies obtained from keystep annotations through the following procedure:  a) a directed graph is first generated from the observed keystep transition frequencies; b) edges of the transition graph are filtered based on transition probabilities using a threshold parameter which is manually tuned for each scenario; c) edge directions are inverted to convert frequent transitions into dependencies. These initial graphs were then refined and manually corrected. 

\textit{Segment-level annotations.} 
A task graph is a global representation of a procedure including information on dependencies and partial orderings of keysteps. Since our task is defined at the keystep level, we need to ``project'' the constraints expressed by the task graph onto keystep video segments, which we do with the following procedure.

Let $S = \{s_1,\ldots,s_n\}$ be a labeled sequence of keysteps in a given video. We denote with $y_i$ the annotated keystep label of segment $s_i$ and with $Y_{:i}=\{y_1,\ldots,y_{i}\}$ the sequence of labels up to the $i$-th keystep. Using these keystep annotations, each segment $s_i$ is automatically matched to a task-graph and augmented with the following attributes: 1) a list of \textit{previous keysteps}---these are the in-neighbors of the matched node, 2) \textit{optional} labels---directly derived from the optional node attribute, 3) a \textit{procedural mistake} label---this is set to ``true'' if the in-neighbors of the matched node do not correspond to segments in the history $Y_{:i}$, 4) the list of \textit{missing keysteps}---the in-neighbors of the matched node not listed in $Y_{:i}$, and 5) the list of \textit{next steps}---nodes for which in-neighbors appear in $Y_{:i}$. Non-repeatable nodes are listed only if they do not appear in $Y_{:i}$.

Given the weakly supervised nature of the task, we only release keystep level annotations on the validation set, while annotations on the training set are not shared nor used for the development of the baselines, and test labels are private, with evaluations on the test set possible by submitting predictions to a server.
Also note that we do not release the labeled task graphs to avoid leaking test and training labels.

\begin{figure*}[!t]
    \centering
    \fbox{(a)\includegraphics[height=4.30cm]{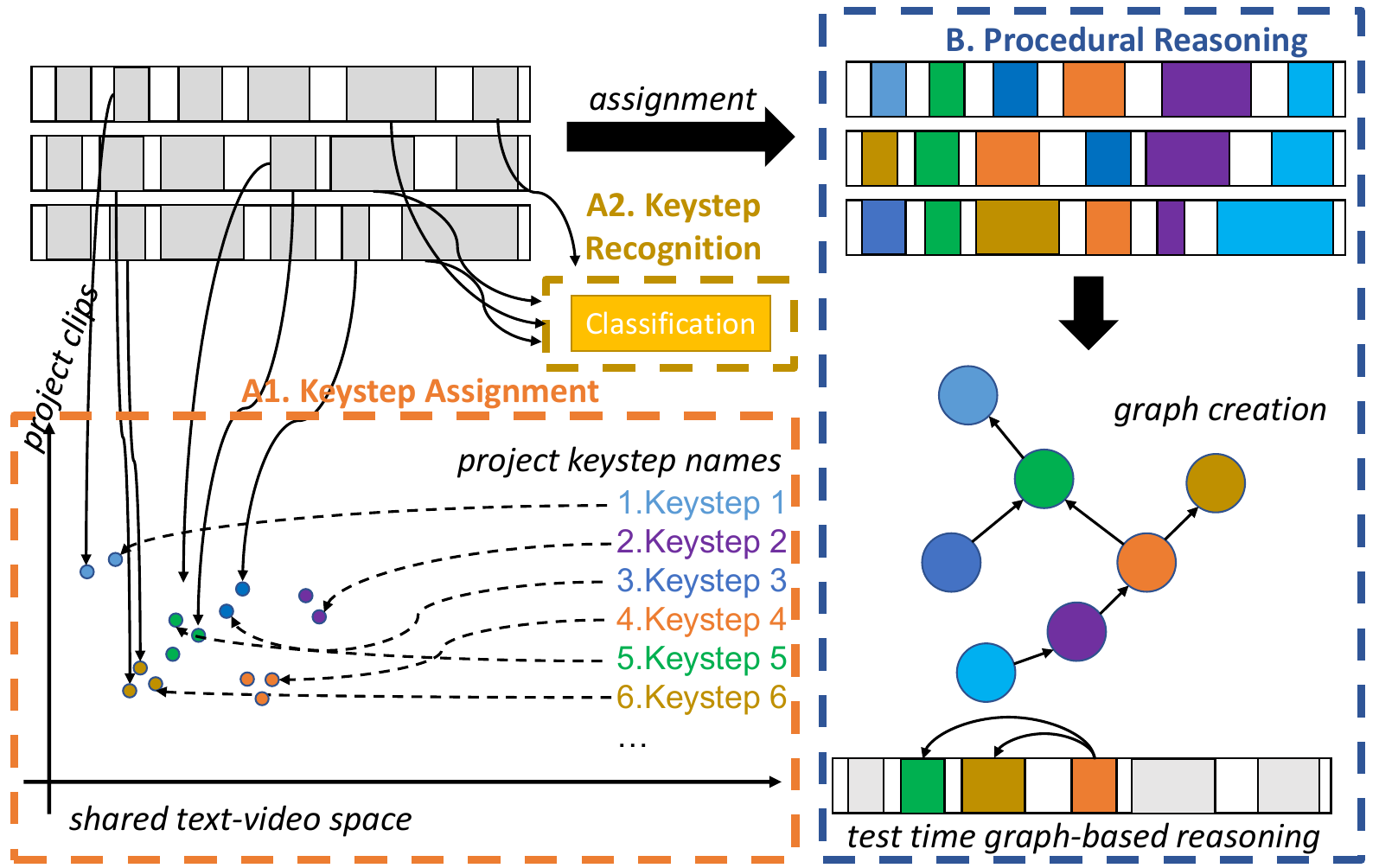}}\hfill
    \fbox{(b)\includegraphics[height=4.30cm]{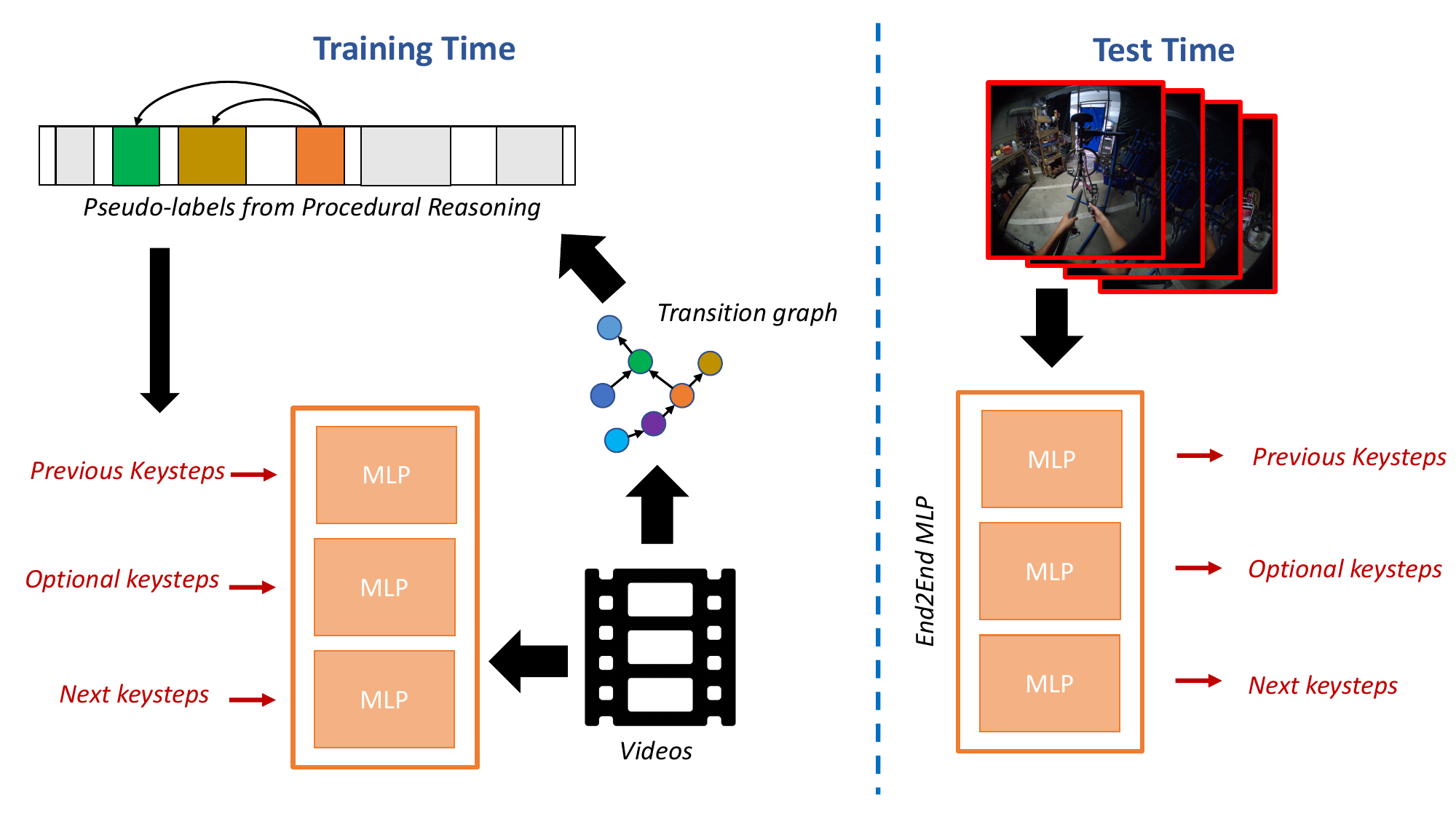}}
    \caption{Overview of the two procedure understanding approaches considered in our evaluation: (a) graph-based baselines for procedure understanding rely on a Keystep Assignment or a Keystep Recognition and a Procedural Reasoning component; (b) the architecture of our end-to-end baseline.}
    \label{fig:baselines}
\end{figure*}

\paragraph{Metrics}
We consider the task of determining lists of keysteps as a detection task with an imbalance between positives (the keysteps to be detected, e.g., preconditions) and negatives (the keysteps which are not to be detected, e.g., keysteps which are not preconditions) and evaluate all methods using the calibrated Average Precision (cAP)~\citep{de2016online}. Note that, according to this measure, a random baseline would on average achieve a performance of $50\%$.

\paragraph{Baselines}
We consider graph-based and end-to-end baselines. Graph-based baselines (see Figure~\ref{fig:baselines}(a)) include two main components: a segment-keystep assignment module (A1) which provides a pseudo-labeling of video segments based on a pre-trained video-language model, and a procedural reasoning module (B) which makes predictions based on a transition graph built either from ground truth (for instance-level supervision) or pseudo-labels (for procedure-level supervision). Additionally, we provide a baseline where the keystep assignment step is replaced with label predictions from the Keystep Recognition task (A2). Note that in the training set, segments with a confidence score below 20\% have been discarded.

End-to-end baselines (see Figure~\ref{fig:baselines}(b)) are trained to predict the same results as graph-based baselines directly from video, with the aim to obtain a compact algorithm which does not explicitly make use of a graph. The end-to-end architecture consists of three MLPs designed to predict previous, optional, and future keysteps. Each MLP has six heads, one for each considered scenario.

Additional baseline implementation details are provided in Appendix \ref{sec:appendix:proc_understanding}.

\paragraph{Results}
Table~\ref{tab:baseline_results} reports the results obtained by our baselines %
and compares them against those produced by a ``uniform'' baseline, predicting previous/optional/mistakes/missing/next keysteps with equal probabilities. 
Results show that the graph-based baseline relying on ground truth annotations significantly outperforms the uniform baseline for most of the tasks, excluding future keystep predictions. This suggests that even simple keystep co-occurrences are informative to some degree of the overall structure of the procedure. 

The limited performance gains on future keystep prediction highlight the complexity of the task and the need for further research.  The end-to-end model trained with instance-level supervision achieves lower or similar performance, trading accuracy for test-time efficiency, due to the absence of an explicit graph.
Procedure-level baselines achieve lower results because they do not rely on ground truth labels. The keystep prediction approach achieves better results compared to the keystep assignment mechanism for all tasks, except for optional keysteps. Despite our efforts, performance is below the uniform baseline, indicating that there is room for future investigations.

\begin{table*}[]
\small
\resizebox{\textwidth}{!}{
\begin{tabular}{lll|l|ccccc}
Supervision & Baseline & Keystep Labels & Inf. Set & Prev. Keysteps & Opt. Keysteps & Proc. Mistakes & Miss. Keysteps & Fut. Keysteps \\ \hline
- & Uniform Baseline & - & Val/Test & 59.18/59.13 & 56.71/56.73 & 60.54/60.66 & 65.58/65.64 & 65.65/65.65 \\ \hline
Instance-Level & Graph-Based & Ground Truth & Val/Test & 82.49/82.32 & 58.95/62.10 & 73.19/73.06 & 84.29/82.63 & 63.48/62.82 \\
Instance-Level & End-to-End & Ground Truth & Val/Test & 62.05/62.05 & 51.85/61.39 & 56.75/52.07 & 60.11/61.77 & 60.35/59.25 \\ \hline
Procedure-Level & Graph-Based & Keystep Assignment & Val/Test & 54.26/53.43 & 49.86/52.36 & \underline{56.46}/\underline{57.81} & \underline{60.97}/53.92 & 52.50/53.54 \\
Procedure-Level & End-to-End & Keystep Assignment & Val/Test & 55.37/54.82 & \textbf{52.12}/\underline{60.78} & 52.84/54.73 & 56.11/53.75 & \textbf{58.88}/\underline{57.47} \\
Procedure-Level & Graph-Based & Keystep Prediction & Val/Test & \textbf{64.56}/\textbf{66.22} & 49.51/49.00 & \textbf{61.15}/\textbf{58.59} & \textbf{61.50}/\textbf{64.18} & \underline{57.87}/\textbf{58.34} \\ 
Procedure-Level & End-to-End & Keystep Prediction & Val/Test & \underline{57.43}/\underline{57.92} & \underline{51.54}/\textbf{61.01} & 51.68/54.92 & 54.99/\underline{55.15} & 57.35/56.92 \\ \hline
\end{tabular}
}
\caption{Results for the procedure understanding task. Best results are reported in bold, the second best results are underlined. All results are in percentage.}
\label{tab:baseline_results}
\end{table*}

%% file: sec/benchmarks/proficiency-benchmark.tex
\paragraph{Motivation}

Going beyond recognizing what a person is doing, this task aims to infer the user's skill level. Such an ability could lead to novel coaching tools that let people learn new skills more effectively, or new ways to \emph{evaluate} human performance in domains like sports or music.

\paragraph{Task definition} 

\begin{figure}[t]
    \centering
    \includegraphics[width=\linewidth]{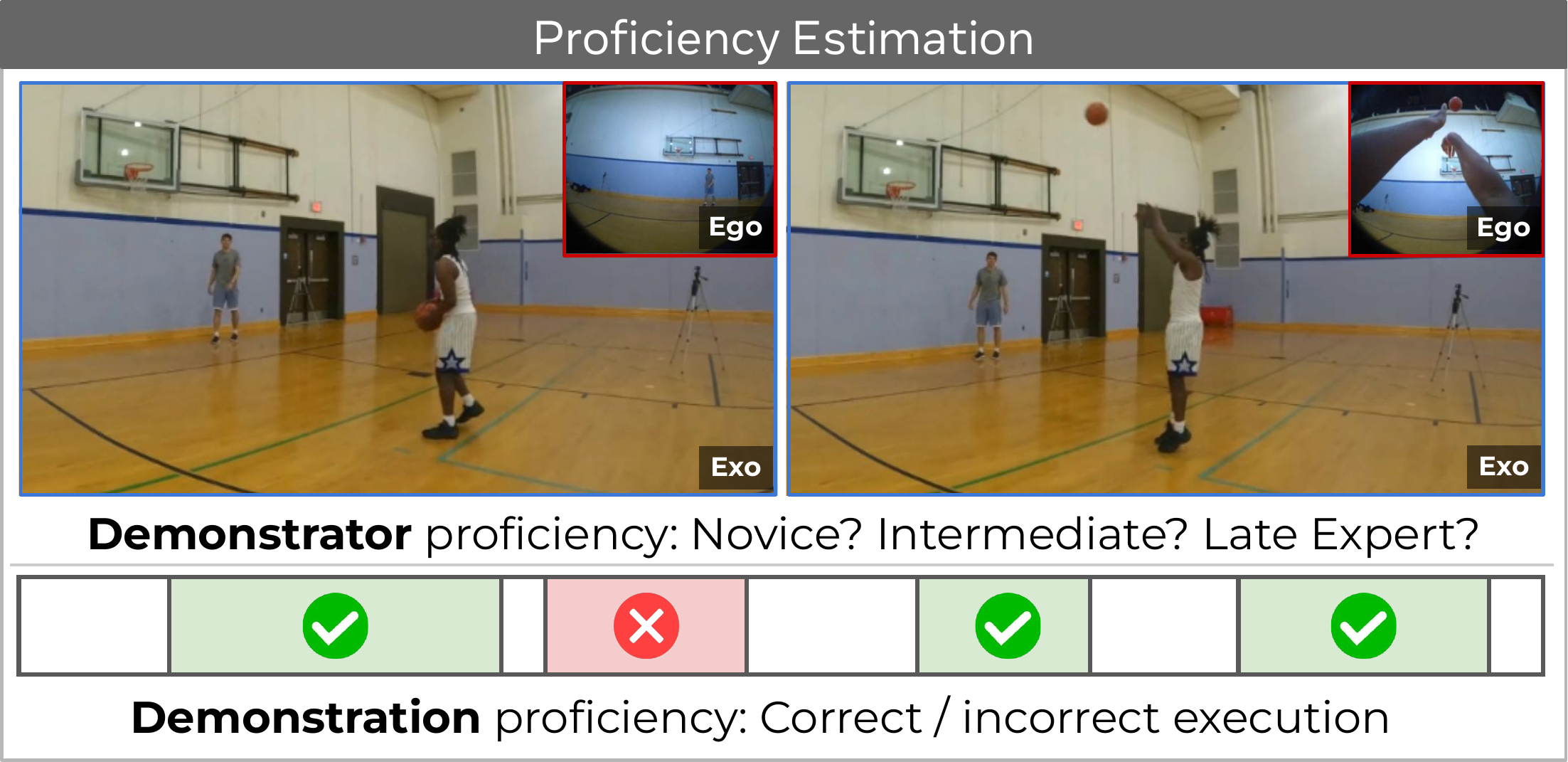}
    \caption{Demonstrator and demonstration proficiency estimation.\vspace{-0.15cm}} 
    \label{fig:proficiency_task}
\end{figure}

We consider two variants: (1) \textit{demonstrator} and (2) \textit{demonstration} proficiency estimation. Both tasks consider one egocentric and (optionally) $M$ exocentric videos of a demonstrator performing a task, which are synchronized in time, as their inputs: $\mathcal{V} = \left(\mathcal{V}_{ego},\mathcal{V}_{exo}^{1}, \cdots, \mathcal{V}_{exo}^{M}\right)$. See Figure~\ref{fig:proficiency_task} for an illustration. We provide more details for each variant below. \vspace{-0.05in}\\

\noindent {\em Demonstrator proficiency estimation:} The goal is to estimate the demonstrator's skill level from one or more task demonstrations. It is formulated as a video classification task with the following classes: (novice, early expert, intermediate expert, late expert). \vspace{-0.05in}\\

\noindent {\em Demonstration proficiency estimation:} Given a single task demonstration, the goal is to identify parts of the video where the task execution was good (i.e., `good executions') or needs further improvement (i.e., `needs improvement'). It is formulated as a temporal localization task, where we localize instances of `good executions' and `needs improvement' throughout the task demonstration. Formally, we can express the demonstration proficiency estimation function $h$ as $\hat{G}, \hat{I} = h(\mathcal{V})$, where $\hat{G} = \big\{t^g_1, t^g_2, \cdots, t^g_{|G|}\big\}$ are the timestamps where the participant shows good task execution, and $\hat{I} = \{t^i_1, t^i_2, \cdots, t^i_{|I|}\}$ are the timestamps where the participant needs to improve their skill level. Note that parts of the video that do not reveal the participant's skill are left unlabeled. \\

\noindent Both tasks inherently benefit from multi-view data. Egocentric video captures fine-grained information about the hand pose and object interactions, which can be critical in tasks that such as cooking (e.g., chopping vegetables) and music (e.g., placement of fingers on the guitar). On the other hand, the exocentric videos provide broader information about the demonstrator’s body pose, which can be highly indicative of proficiency in tasks that require extensive physical motion such as basketball, soccer, and dancing. 

Note that the input to the model excludes textual descriptions/narrations of the activity, audio, gaze sensor readings, and any subject information, which would simplify the task significantly at the expense of usability since these signals are typically not available for in-the-wild video. Our formulation encourages the development of proficiency estimation methods from visual cues. %

\paragraph{Related work} Prior work uses egocentric~\citep{baller-iccv2017, Doughty_2019_CVPR} or exocentric~\citep{parmar2017learning, mtlaqa,10.1007/978-3-030-00937-3_25}
views for proficiency estimation in sports~\citep{baller-iccv2017,mtlaqa,assessing-eccv2014}, health~\citep{Liu_2021_CVPR, 10.1007/978-3-030-00937-3_25,Zhang_2013_CVPR, DBLP:journals/corr/ZiaSBSE17}, and others~\citep{Doughty_2019_CVPR, Yu_2021_ICCV}. We propose the first multi-view egocentric and exocentric proficiency estimation benchmark. Unlike prior work, our benchmark spans diverse, day-to-day physical and procedural scenarios and includes temporally localized annotations of (in)correct executions. 

\paragraph{Annotations}
We now describe the annotation procedure for the two proficiency estimation tasks. \vspace{-0.1in}\\

\noindent {\em Demonstrator proficiency estimation.}  We assign four proficiency labels (novice, early expert, intermediate expert, late expert) to each person performing activity demonstrations (one label per person). Most levels correspond to experts since Ego-Exo4D videos are dominantly targeted towards expert participants who can perform the task successfully (see Appendix~\ref{sec:appendix-collection}). Four proficiency classes makes the task challenging but still approachable.\footnote{Subtle variations between five or more levels of proficiency can be insufficiently observable from vision alone, and even difficult for expert annotators to reach consensus.} We derive annotations for this task from participant surveys and expert commentary. Please see Appendix~\ref{sec:appendix:proficiency_estimation} for more details, including a visualization of the proficiency score distribution for each scenario (Figure~\ref{fig:appendix-demonstrator-scores}). 

We split our dataset into train/val/test splits based on the common split shared across benchmarks. The dataset statistics are shown in Table~\ref{tab:appendix-proficiency-stats}. Note that we exclude the bike repair and health scenarios from the demonstrator proficiency task. The distribution of participants for bike repair is heavily skewed towards late experts. The predominant activity in the health collection is COVID testing, where skill levels are hard to determine due to the simplicity of the task.\vspace{-0.1in}\\

\begin{table}[t]
\centering
\footnotesize
\begin{tabular}{@{}lcccccc@{}}
\toprule
                          & \multicolumn{3}{c}{Demonstrator} & \multicolumn{3}{c}{Demonstration} \\
\multicolumn{1}{c}{}      & Train & Val & Test & Train & Val & Test \\ \midrule
Basketball                & 575   & 143 & 167  & 146   & 47  &  19  \\
Bike repair               &  -    & -   &  -   &  41   &  9  &  15  \\
Cooking                   & 200   & 53  & 86   &  80   & 24  &  39  \\
Dance                     & 380   & 127 & 148  &  80   & 35  &  27  \\
Health                    &  -    &  -  &  -   &  42   & 12  &  16  \\
Music                     & 138   & 36  & 71   &  94   & 32  &  35  \\
Rock Climbing             & 561   & 159 & 230  &  65   & 15  &  22  \\
Soccer                    & 129   & 43  & 66   &   8   &  3  &   6  \\ \midrule
\multicolumn{1}{r}{Total} & 1983  & 561 & 768  & 556   &177  & 179  \\ \bottomrule
\end{tabular}
\caption{\small Distribution over video takes in proficiency estimation benchmark.}
\label{tab:appendix-proficiency-stats}
\end{table}

\noindent {\em Demonstration proficiency estimation.} We leverage temporally localized annotations that include the timestamps of steps demonstrated in the video as well as the proficiency category for each demonstrated step instance (i.e., good execution or needs improvement). For this task, we consider all 8 scenarios, as shown in Table~\ref{tab:appendix-proficiency-stats}. We derive annotations for this task from expert commentary, where task experts carefully analyze videos and provide timestamped commentary on the participant's performance (see Section~\ref{sec:commentary}). In particular, given a single timestamped comment from an expert, we annotate whether the comment describes a good execution and/or provides tips for improving the participant's skill level. See Table~\ref{tab:demonstration_proficiency_annotations} in Appendix~\ref{sec:appendix:proficiency_estimation} for example annotations. These annotations are then associated with the timestamp provided with each comment to obtain a list of timestamps for good executions $\{t_1^g, t_2^g, \cdots\}$ and tips for improvement $\{t_1^i, t_2^i, \cdots\}$ in each video. Overall, the demonstration proficiency estimation task consists of 556 train / 177 val / 179 test videos (see Table~\ref{tab:appendix-proficiency-stats} for a breakdown per scenario). 

\paragraph{Metrics}
For demonstrator proficiency estimation, we measure performance using top-1 classification accuracy. For demonstration proficiency estimation, we measure the temporal localization performance using a modified mean average precision (mAP). Unlike prior temporal action localization methods which use temporal IoU between segments, we use the $L_1$-distance between the predicted and ground-truth timestamps to measure mAP. Therefore, we define mAP based on $L_1$-distance (in seconds) thresholds.

\paragraph{Baselines}
Next, we define the baselines for each task.\vspace{-0.05in}\\

\noindent {\em Demonstrator proficiency estimation:} We adopt TimeSformer~\citep{bertasius2021space} for our experiments. We train one model on the egocentric view (``ego model"), and a separate model on all 4 exocentric views (``exocentric model"). The models are trained to classify individual clips using the cross-entropy loss. At inference time, we perform late fusion to incorporate information from both egocentric and exocentric video streams. We average the softmax predictions across both egocentric and exocentric models to obtain the final video label prediction. We also average results over three spatial crops during inference following prior work~\citep{bertasius2021space}.\vspace{-0.05in}\\ 

\noindent {\em Demonstration proficiency estimation:} We adopt ActionFormer~\citep{zhang2022actionformer}, a video action localization model for our experiments. Unlike traditional action localization, we infer only a single timestamp since our annotations contain only a single point in time for each good execution or tip for improvement. We accordingly adapt ActionFormer for timestamp regression and define $\mathcal{L}_1$-distance based mAP metrics. We train our models with Omnivore features~\citep{omnivore} extracted from overlapping time intervals in the video. For the experiments involving multiple views (i.e., multiple exo views or ego + exo views), we simply concatenate the features for all views at each time step.\\

\noindent Please see Appendix~\ref{sec:appendix:proficiency_estimation} for additional implementation details about the baselines.

\begin{table}[t]
\footnotesize

\centering
\setlength{\tabcolsep}{1.5pt} %
\begin{tabular}{@{}ccccccccc@{}}
\toprule
                                         &             & \multicolumn{6}{c}{Accuracy} \\
                                         &             & \multicolumn{2}{c}{Ego} & \multicolumn{2}{c}{Exos} & \multicolumn{2}{c}{Ego + Exos} \\
Method                                   & Pretraining & Val & Test & Val & Test & Val & Test \\ 
\cmidrule(r){1-2} \cmidrule(lr){3-8}
Random                                   & -           & 26.4 & 26.4 & 26.4 & 26.4 & 26.4 & 26.4 \\
Majority-class                           & -           & 32.3 & 42.4 & 32.3 & 42.4 & 32.3 & 42.4 \\
TimeSFormer                              & -           & 40.6 & 33.9 & 39.0 & \textbf{47.5} & 39.9 & 45.7 \\
TimeSFormer                              & K400        & \textbf{47.2} & 44.1 & 37.8 & 47.0 & 40.3 & 46.1 \\
TimeSFormer                              & HowTo100M   & 45.1 & 36.7 & 39.8 & 46.6 & \textbf{43.7} & 47.0 \\
TimeSFormer                              & EgoVLP      & 44.7 & 43.8 & \textbf{40.5} & 44.5 & 39.4 & 43.5 \\
TimeSFormer                              & EgoVLPv2    & 46.7 & \textbf{50.4} & 37.0 & 47.0 & 37.1 & \textbf{48.7} \\
\midrule
\multicolumn{8}{c}{\textbf{Inference with multiple takes per demonstrator}} \\
TimeSFormer                              & EgoVLPv2    & 48.3 & 51.0 & 36.0 & 47.3 & 43.1 & 49.1 \\

\bottomrule
\end{tabular}
\caption{\small \textbf{Demonstrator proficiency estimation.} We report top-1 accuracies for various baselines on the demonstrator proficiency estimation task. Our learned models use the TimeSFormer architecture~\citep{bertasius2021space}.}
\label{tab:demonstrator_proficiency_results}
\end{table}

\paragraph{Results}

Our Ego-Exo4D dataset has $5$ views (1 egocentric view, and $M=4$ exocentric views). We run the proficiency tasks in two settings: one where the exo view is available at test time, and one where it is not.  For the latter, benchmarking baseline models with only the egocentric view is important when the target is augmented reality applications like wearable headsets and mobile robotics. For the former, results with benchmarking with both egocentric and exocentric views are helpful to capture the multi-view aspect of the problem. \vspace{-0.1in} \\

\noindent {\em Demonstrator proficiency estimation.} We present results for demonstrator proficiency estimation in Table~\ref{tab:demonstrator_proficiency_results}. We include two na\"ive baselines to account for biases in the dataset. The random baseline uniformly samples one skill level at random. The majority-class baseline predicts the majority class within each scenario. TimeSFormer trained from random initialization outperforms the na\"ive baselines by a significant margin, demonstrating the ability of learned methods to quantify skill levels from videos. Ego videos are sufficient to achieve good performance in most cases, while the exo videos are beneficial in tasks such as bouldering, highlighting the complementary nature of the ego and exo viewpoints. Initializing TimeSFormer using pre-trained weights improves over random initialization, particularly on ego videos. Furthermore, fusing the predictions from the ego view and exo views does not improve performance, likely due to the simplicity of late fusion. In the last row of Table~\ref{tab:demonstrator_proficiency_results}, we further report results when providing multiple demonstrations from a participant for evaluating TimeSFormer. This matches or outperforms evaluating TimeSFormer on a single demonstration, highlighting the potential for obtaining more accurate skill estimates by studying multiple demonstrations. We further study scenario-specific performance of the baselines in Appendix~\ref{sec:appendix:proficiency_estimation}. We find that egocentric views are beneficial for scenarios such as cooking that require close-up views of hands and objects, whereas exocentric views are more useful for scenarios such as bouldering that require body-pose information. Overall, our benchmark presents new challenges for video-based skill understanding and our results highlight the difficulty of the task, suggesting good scope for improvement in future work. \vspace{-0.05in} \\

\noindent {\em Demonstration proficiency estimation.} We present results for the demonstration proficiency estimation task in Table~\ref{tab:demonstration_proficiency_results}. We include three na\"ive baselines along with ActionFormer~\citep{zhang2022actionformer}. The ``Random tips/good exec." baseline randomly predicts a tip or a good execution label every $5.97$ seconds, i.e., the average temporal span between adjacent annotations in our dataset. The ``Uniform tips" baseline predicts a tip for improvement label every $5.97$ seconds. The ``uniform good exec." baseline predicts a good execution label every $5.97$ seconds. We evaluate ActionFormer models trained on ego only, exo only and ego + exo views. All na\"ive baselines perform poorly on this task. The learned ActionFormer baseline outperforms the na\"ive baselines by a good margin. However, the absolute mAP scores are fairly low, suggesting that the task is very challenging and has a significant scope for improvement in methods.

\begin{table*}[t]
\centering
\footnotesize
\setlength{\tabcolsep}{2.5pt} %
\begin{tabular}{@{}c|ccc|ccc|ccc@{}}
\toprule
\multicolumn{10}{c}{Val/test results} \\
\midrule
                                          & \multicolumn{3}{c|}{Ego} & \multicolumn{3}{c|}{Exos} & \multicolumn{3}{c}{Ego + Exos}\\

Method                                    & mAP$_{0.25}$  & mAP$_{1.0}$   & Avg.          & mAP$_{0.25}$  & mAP$_{1.0}$   & Avg.          & mAP$_{0.25}$  & mAP$_{1.0}$   & Avg.          \\ \midrule
Random                                    & 0.48/0.45     & 5.23/4.72     & 2.46/2.20     & 0.48/0.45     & 5.23/4.72     & 2.46/2.20     & 0.48/0.45     & 5.23/4.72     & 2.46/2.20     \\
Uniform tips                              & 0.49/0.45     & 5.28/5.18     & 2.48/2.39     & 0.49/0.45     & 5.28/5.18     & 2.48/2.39     & 0.49/0.45     & 5.28/5.18     & 2.48/2.39     \\
Uniform good exec.                        & 0.43/0.46     & 4.79/4.62     & 2.27/2.17     & 0.43/0.46     & 4.79/4.62     & 2.27/2.17     & 0.43/0.46     & 4.79/4.62     & 2.27/2.17     \\
ActionFormer~\citep{zhang2022actionformer} & \textbf{0.95/1.04} & \textbf{6.33/7.56} & \textbf{3.27/3.87} & \textbf{1.08/1.14} & \textbf{7.50/7.36} & \textbf{3.84/3.87} & \textbf{0.97/1.14} & \textbf{7.03/7.90} & \textbf{3.57/4.04} \\ \bottomrule
\end{tabular}

\caption{\small \textbf{Demonstration proficiency estimation benchmark.} We report the mean average precision ($\%$) for various baselines on the demonstration proficiency estimation task for the val and test splits. mAP$_{k}$ is measured at an $L_1$-distance threshold of $k$ seconds. The average mAP (Avg.) measures the mAP averaged across $k=\{0.25, 0.5, 1.0\}$ seconds.}
\label{tab:demonstration_proficiency_results}
\end{table*}

%% file: sec/benchmarks/bodypose-benchmark.tex
\begin{figure*}[t]
    \centering
    \includegraphics[width=\linewidth]{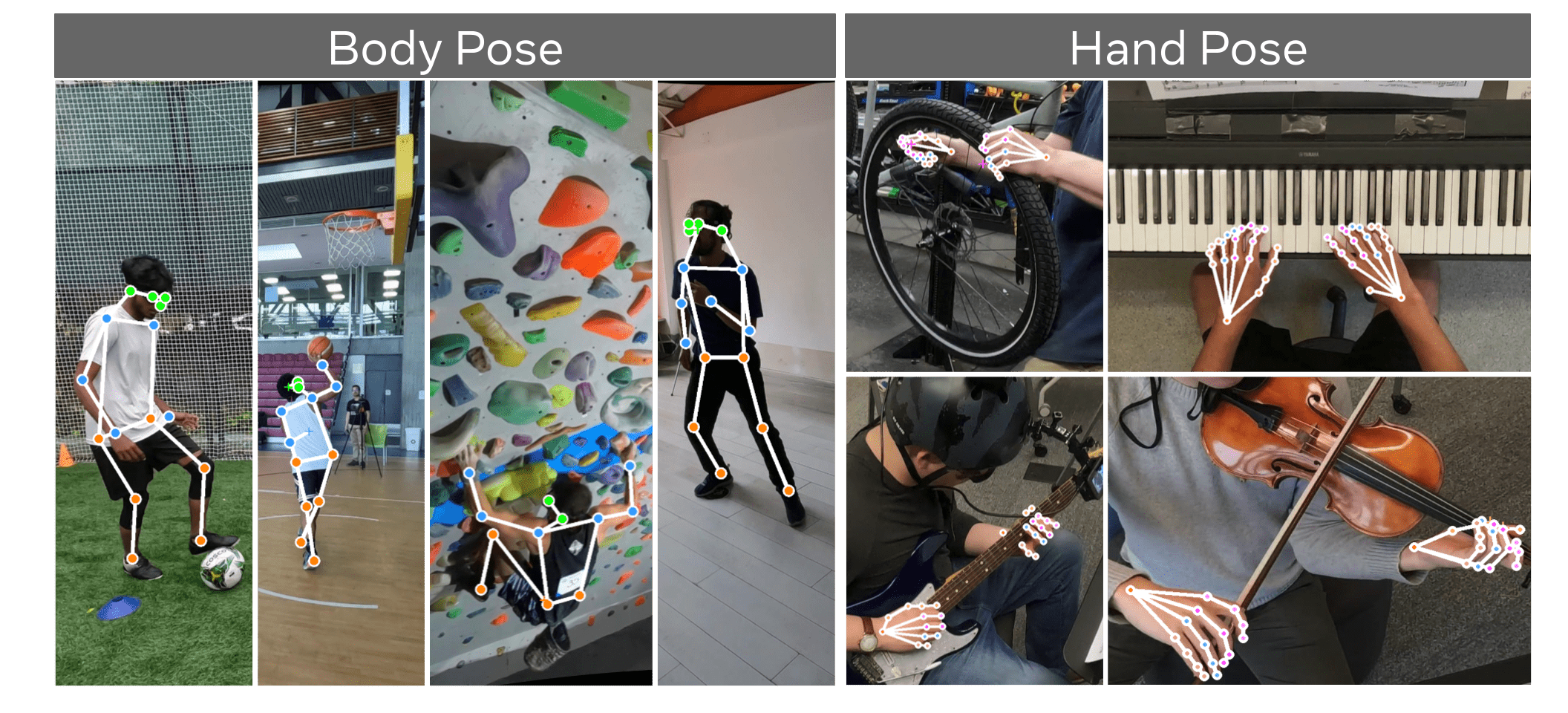}
   
    \caption{Hand and body keypoints for ego-pose estimation.%
    }
    \label{fig:pose_task}
\end{figure*}

\paragraph{Motivation}

Having presented benchmark tasks about ego-exo relation, recognition, and proficiency assessment, we now define the final family of tasks centered on body and hand pose.  
This family of tasks is motivated by recovering the skilled body movements of participants, even in the extreme setting of monocular ego-video input in dynamic environments, as shown in Figure \ref{fig:pose_task}. Estimating the physical state of a person's body---the 3D positions of the arms, legs, hands---from the ego view is essential for wearable AI systems that can support human activity. Challenges include subtle and flexible movements, frequent occlusion, and body/hand parts out of view.   

For each scenario, we invite experts as the participants to enhance the complexity and variety of the motions captured. As an example, expert musicians typically demonstrate more advanced and varied finger techniques ($300+$) compared to beginners or intermediate players ($<100$) during the recordings. Such complexity enables the model to (1) extract more representative latent features, and (2) learn subtle patterns and relationships that might be missed in a more homogeneous dataset, for both estimation and prediction tasks.

\paragraph{Task definition}
The ego pose benchmark is divided into two separate tasks: \textit{body pose estimation} and \textit{hand pose estimation}. 

In our \textit{body pose estimation} task, the goal is to estimate the 3D human pose sequence $\mathcal{P}=\{ \mathcal{P}_1,..., \mathcal{P}_T\}$ using either an egocentric RGB video input sequence $\mathcal{V}_{ego}=\{ \mathcal{V}_1,..., \mathcal{V}_T\}$, an IMU sensor sequence $\mathcal{H}_{imu}=\{ \mathcal{H}_1,..., \mathcal{H}_T\}$, or both, where $1 \leq t \leq T$ is the given time window remapped to make the starting timestamp to be $1$, and $P_t \in \mathcal{R}^{17\times3}$ correspond to the $17$ joints following the MS COCO convention in 3D. $T$ can have different values depending on the length of the particular annotated clip. Note that at test time, we only estimate the error across the visible annotated joints at frame $t$.

The \textit{ego hand pose} task entails predicting the three-dimensional coordinates of the camera wearer's hands.  Given an egocentric frame, the goal is to estimate the 3D joint location for the hands that are (at least) partially visible in the ego view. The output is parameterized as 21 3D joints per hand following the MS COCO dataset convention~\citep{lin2014microsoft}. Frames from the ego view are extracted and undistorted for both training and evaluation. The 2D hand bounding boxes are generated by projecting the 3D hand joints onto the 2D image planes and subsequently enclosing these projections.

Since the Ego Pose benchmark is aimed at promoting the development of methods that perform body pose estimation solely from first-person raw video or IMU data, the input \emph{excludes} egocentric modalities that would unfairly simplify the task (e.g., audio captured from a wearable camera, eye gaze), as well as exocentric video or any signals that can be extracted from it.

\paragraph{Related work}
 Limited prior work explores 3D body pose from a wearable camera. Some methods assume no body visibility~\citep{jiang2017seeing,Yuan_2018_ECCV,Yuan_2019_ICCV, Luo2021DynamicsRegulatedKP,li2023ego}, while others assume partial observability by modifying cameras to capture the body~\citep{rhodin2016egocap, Tome_2019_ICCV,xu2019mo,ahuja2019mecap,hwang2020monoeye}. Our dataset can be used for both paradigms. 
 
 Existing hand pose datasets use constrained environments~\citep{simon2017hand,moon2020interhand2} with simple hand motion
~\citep{hampali2020honnotate, Kwon_2021_ICCV, ohkawa2023assemblyhands}, whereas we include diverse real-world scenarios with skilled hand motions, e.g., with expert musicians and bike mechanics.

\paragraph{Annotations}
The \textit{3D human body pose} annotation process consists of two main stages: (1) automatic ground truth generation, and (2) manual multi-view keypoint annotation/correction. Through this process we derive 3D keypoint annotations for approximately \KGCR{14M frames}.

In the automatic ground truth generation phase, we use off-the-shelf models \citep{mmpose2020} to predict the 2D bounding boxes from each of the exocentric views. Since there could be multiple people in the scene and we only want to consider the one wearing the egocentric camera, we project the 3D headset location from the MPS output to select which box corresponds to the camera wearer. Then, we run an off-the-shelf 2D human keypoint detector \citep{mmpose2020} for each bounding box to obtain the 2D keypoints. Finally, we run 3D triangulation with RANSAC to minimize the reprojection errors to obtain the 3D keypoints for the camera wearer. In the manual annotation phase, we import the undistorted frames and the reprojected 2D keypoints into our multi-view annotation interface. 

The \textit{3D human hand pose} annotation process also consists of two stages, i.e., the automatic ground truth generation and the manual multi-view keypoint annotation.  Compared to the body pose, the main difference in automatic ground truth generation is that we also detect hand keypoints from the egocentric frame, and we use the result from the whole body pose estimation to infer the hand locations when there are multiple people in the scene. Similarly, for manual annotation, besides the exocentric frames, we also show the annotators the egocentric frames to allow them to annotate/correct hand keypoints. For each annotated joint in manual annotations, we provide the number of views used for triangulation as the indicator of the confidence for the provided ground truth data. Meanwhile, the correction the annotators make for hand joints on ego images can serve as the indicator to understand the difficulty for hand reconstruction from the given ego view.

\KGCR{Ego-Exo4D offers the largest available manually annotated body pose (376K 3D/2M 2D) and hand pose (68K 3D/340K 2D) annotations.  Along with this, we also provide 9.2M/47M (body) and 4.3M/21M (hand) automatically generated groundtruth 3D and 2D poses, totaling about 13.M frames 
In total, we have approximately 14M frames of 3D ground truth (GT) and pseudo-GT combined across body and hands.
To our knowledge, this represents the largest collection of body pose annotations in the literature, whether for ego or exo video.

How good is the auto GT?  Between manual and automatic annotations, the body and hand MPJPEs are 3.33 cm and 1.87 cm, respectively, much smaller than the best baseline methods.  It is important to note that Ego-Exo4D tackles real-world scenarios with five or fewer cameras rather than controlled environments. This introduces challenges like increased occlusions from body and objects along with limited view and resolution of hands from distant cameras. Despite this, our auto generation pipeline surpasses baselines, showcasing robustness and efficacy. Experiments below further show performance boosts across baselines when using automatic ground truth, demonstrating its effectiveness.
Note that automatic GT and manual GT are not mutually exclusive, and people can choose whether/how automatic GT is used for training.}

\paragraph{Metrics}
To evaluate the performance of \textit{body pose estimation} approaches we calculate the Mean Per Joint Position Error (MPJPE) in centimeters (cm), and the Mean Per Joint Velocity Error (MPJVE) in meters per second (m/s). 

The \textit{ego hand pose} baselines are evaluated according to  both the MPJPE and the PA-MPJPE metrics. The MPJPE measures absolute Mean Per Joint Position Error, while the PA-MPJPE calculates the average 3D joint errors after performing Procrustes Alignment on hand poses. Both metrics are reported in millimeter (mm) unit. 

\paragraph{Baselines}
We evaluate three state-of-the-art baseline methods for the \textit{body pose estimation} task. Moreover, to gauge the performance of deep-learning-based methods, we create a static pose baseline, which consists of fixing the 3D human body pose prediction to be the average pose in the training set and translating it according to the IMU sensor. Thus, the fixed prediction matches the camera location at each frame.

\begin{itemize}
\itemsep0em 
\item \textbf{Kinpoly.} Kinpoly \citep{Luo2021DynamicsRegulatedKP} proposes to use a simulated humanoid to track head pose and create full-body motion based on action types. Based on the input head-pose and action type, Kinpoly synthesizes realistic human pose and human-object interactions inside a physics simulator. Different from kinematic-based methods that directly output joint angles or positions for pose estimation, Kinpoly outputs joint torques as the final product and controls a simulated humanoid for pose estimation.\\ %

\item \textbf{EgoEgo.} EgoEgo \citep{li2023ego} uses a two-step approach for egocentric body pose estimation, by estimating the head pose from the egocentric video first, and then using a diffusion model to generate the full body motion sequence based on the head pose sequence. For head pose estimation, it obtains the initial head pose trajectory using DROID-SLAM \citep{teed2021droid}, and then uses learning-based methods to correct the head pose, including a GravityNet to estimate the additional rotation and a HeadNet with optical flow features as input to estimate the scaling factor to the trajectory. The full body pose is generated with a modified version of DDPM \citep{ho2020denoising} that is conditioned on head pose and trained on AMASS \citep{AMASS:2019}. We show the evaluation of the conditional diffusion part here.\\

\item \textbf{\arxivb{Location-based.}} This baseline is inspired by state-of-the-art methods that use transformer-based models for body pose estimation from sparse inputs~\citep{castillo2023bodiffusion, jiang2022avatarposer}. We adapt these methods to utilize 3D positions as opposed to the traditional parametric body model. During the training phase, the model was subjected to $40,000$ iterations, using the Adam optimizer with a learning rate of $1e^{-4}$. The window size for temporal analysis was set at 40 frames, and we minimized the Mean Squared Error (MSE) loss between predicted poses and ground truths. As for the input, our model receives a sequence of head poses captured by the device.

\end{itemize}

We implemented and/or trained four baseline models for the \textit{ego hand pose estimation}. To estimate the 3D hand joint from monocular 2D ego view images, 2D heatmaps can be explicitly estimated and lifted to 3D space, or 3D joints can be directly estimated from extracted 2D features. The feature extractor backbone could be CNN-based or transformer-based.
The proposed baseline methods cover different choices of model designs. 
All the baseline models work on single frame images without temporal information. The baseline models are trained on manual or manual+automatic annotations, and are only evaluated on manual annotations. 

Notably, most baseline methods generate hand mesh as final results in their original paper. We modified them to be trained and supervised only on 2D/3D hand joints (not on hand mesh) to fit the benchmark. 

\begin{itemize}
\itemsep0em 
\item \textbf{THOR-net.} THOR-net~\citep{aboukhadra2023thor} uses Keypoint-RCNN as the feature extractor to obtain 2D information and derive 2D hand keypoints heatmaps explicitly. The method then lifts 2D estimates to the 3D space using GraFormer~\citep{zhao2022graformer}, which is a model consisting of Graph Convolutional layers and Attention layers. We use only the 2D-to-3D pose GraFormer branch in THOR-net to adapt the method to our task. The training takes around 4 hours for the manual dataset on a GeForce RTX 4090 Graphics Card, and around 10 hours for the dataset combining manual and automatic annotations. \\

\item \textbf{HandOccNet.} HandOccNet~\citep{park2022handoccnet} uses a ResNet50\citep{he2016deep}-based FPN~\citep{lin2017feature} to extract 2D features. The method then uses two Transformer-based modules:  Feature Injecting Transformer (FIT) to inject hand information
into occluded region, and Self-Enhancing Transformer (SET) to further refine the 2D features. The method proposes a regressor based architecture to produce 2D keypoints, MANO~\citep{MANO:SIGGRAPHASIA:2017} pose, and MANO shape parameters to predict joints and vertices. To accommodate our baseline, only 2D keypoints and 3D joints location losses are used in the training phase. The training takes around 2 hours for the manual dataset
on 8 NVIDIA V100 Graphics Cards. \\

\item \textbf{POTTER.} POTTER~\citep{zheng2023potter} proposes Pooling Attention Transformer (PAT) to extract 2D visual features, which significantly reduces the memory and computational cost without sacrificing performances. The method then applies a mesh regression head HybrIK~\citep{li2021hybrik} to generate 3D joint and mesh results.  The training takes around 43 minutes for manual dataset, and around 4 hours for manual+auto dataset on a GeForce RTX 4090 Graphics Card.\\ 

\item \textbf{METRO.} METRO~\citep{lin2021end} extracts a CNN-based global image features. The method then uses a transformer encoder to jointly model vertex-vertex and vertex-joint interactions, and outputs 3D joint coordinates and mesh vertices simultaneously. Since the training of METRO strongly depends on hand mesh supervision, which is not present in the annotations, we borrowed the checkpoint trained on FreiHand~\citep{kolotouros2019learning} dataset and run the inference only, without training it on our benchmark. 
\end{itemize}

\paragraph{Results}
Table~\ref{tab:resultsV2_body_pose_with_gt_headset_pose} shows the evaluation results of all the baseline approaches for the \textit{body pose estimation }task. \NEW{First, note that the static pose baseline obtains a significantly higher MPJPE than all the other approaches. This finding suggests that the poses across different scenarios in the dataset are extremely diverse. Thus, attempting to have the same static pose for all test cases is unfeasible.} In contrast, the proposed baseline implementations achieve notable enhancements in performance. Table~\ref{tab:res:body_pose_per_scenario} shows the performance of each method per scenario. While these developments are promising, we believe that further refinement is possible, especially in lower body pose estimation and to ensure temporal consistency in predictions.

\begin{table}[]
\centering
\resizebox{\columnwidth}{!}{
\begin{tabular}{@{}llccc@{}}
\toprule
\multicolumn{1}{c}{\multirow{2}{*}{Method}}  & \multicolumn{2}{c}{Validation} & \multicolumn{2}{c}{Test}  \\ \cmidrule(l){2-5} 
\multicolumn{1}{c}{} & MPJPE & MPJVE & MPJPE & MPJVE  \\
\midrule
Static pose & 254.29 & - & 215.87 & - \\
EgoEgo  & 24.53 & 0.78 & 26.38 & 0.66  \\
Kinpoly  & 21.66 & 0.86 & 24.36 & 0.65 \\
\arxivb{Location-based}  & 20.73 & 0.74 & 18.51 & 0.64 \\ \bottomrule
\end{tabular}}
\caption{\textbf{Results for the 3D human body pose benchmark.} We report the Mean Per Joint Position Error in cm and the Mean Per Joint Velocity Error in m/s for all the baseline approaches.}
\label{tab:resultsV2_body_pose_with_gt_headset_pose} 

\end{table}

\begin{table}
\small
  \centering
  \begin{tabular}{lcccc}
\toprule
 Scenario & EgoEgo & Kinpoly & \arxivb{Location-based}\\
\midrule
Basketball & 21.36 & 24.98 & \arxivb{19.89} \\
Soccer  & 23.08 & 19.09 & \arxivb{16.62} \\
Bike repair & 30.18 & 25.19 & \arxivb{20.61} \\
Cooking & 23.71 & 20.80 & \arxivb{12.65} \\
Health & 32.57 & 29.23 & \arxivb{11.63} \\
Dance & 20.93 & 18.03 & \arxivb{21.15} \\
Music & 33.81 & 30.30 & \arxivb{15.00} \\
\bottomrule
  \end{tabular} 
  \caption{\textbf{Body pose estimation Test results per scenario.} We report the Mean Per Joint Position Error in cm.}
  \label{tab:res:body_pose_per_scenario}  
\end{table}

\begin{figure}[t]
    \centering
    \includegraphics[width=0.5\textwidth]{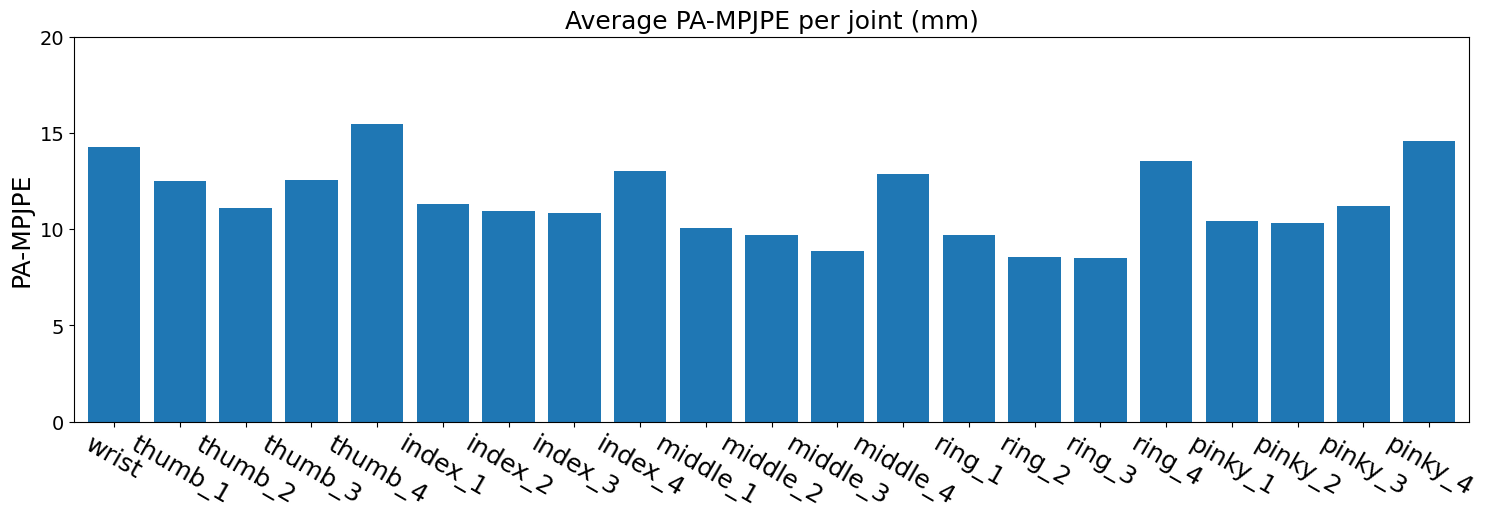}
    \caption{Average PA-MPJPE for each joint. Results generated from POTTER~\citep{zheng2023potter} evaluation. }
    \label{fig:hand_PA_MPJPA_per_joint}
\end{figure}

We report the MPJPE and PA-MPJPE of the baseline models for the \textit{body pose estimation} task in Table~\ref{tab:hand_baseline_results}, and their corresponding parameter numbers and multiply-accumulate operations (MACs) in Table~\ref{tab:hand_baseline_para_MACs}. 
We further analyze the error distribution across different hand joints. Figure~\ref{fig:hand_PA_MPJPA_per_joint} shows that the thumb finger and finger tips tend to have larger errors, most likely because they are occluded or invisible more often. 

\begin{table}[]
\small
\resizebox{\columnwidth}{!} {
\begin{tabular}{c|cc|cc}
\hline
           & \multicolumn{2}{c|}{Manual}           & \multicolumn{2}{c}{Manual+Auto}      \\ \hline
           & \multicolumn{1}{c|}{MPJPE} & PA-MPJPE & \multicolumn{1}{c|}{MPJPE} & PA-MPJPE \\ \hline
METRO*     & \multicolumn{1}{c|}{-}      &     20.61     & \multicolumn{1}{c|}{-}      &    20.61      \\ \hline
THOR-net   & \multicolumn{1}{c|}{51.24}    &       17.99    & \multicolumn{1}{c|}{47.64}      &     17.61     \\ \hline
HandOccNet & \multicolumn{1}{c|}{-}      &     17.22     & \multicolumn{1}{c|}{-}      &      13.56    \\ \hline
POTTER     & \multicolumn{1}{c|}{30.57}      &     11.14     & \multicolumn{1}{c|}{28.94}      &    11.07      \\ \hline
\end{tabular}
}
\caption{MPJPE and PA-MPJPE in mm for ego hand pose baseline models. * denotes methods {\em not} trained on the benchmark. }
\label{tab:hand_baseline_results}
\end{table}

\begin{table}[]
\small
\begin{tabular}{cc}
\begin{subtable}{\columnwidth}
\resizebox{\columnwidth}{!} {
\begin{tabular}{m{2cm}|m{2cm}|m{2cm}|m{2cm}}
\hline
          & THOR-net\newline\citep{aboukhadra2023thor} & HandOccNet\newline\citep{park2022handoccnet} & POTTER\newline\citep{zheng2023potter} \\ \hline
Params (M) &     59.5     &     37.22       &    14.5    \\ \hline
MACs (G)   &    123.6     &  15.5      &    5.2    \\ \hline
\end{tabular}
}
\caption{Number of parameters and MACS for the different ego hand pose baselines.  \note{consider joining up the smaller tables into a singl figure with subparts for flow.}}
\label{tab:hand_baseline_para_MACs}
\end{subtable}
\\ 
\begin{subtable}{\columnwidth}
\begin{tabular}{c|c|c|c|c}
\hline
\# visible views &  3     & 4     & 5     & 6     \\ \hline
PA-MPJPE (mm)         & 14.01 & 12.15 & 11.03 & 10.02 \\ \hline
\end{tabular}    
\caption{PA-MPJPE for joints that are visible in different number of views (including ego and exo views). Results generated from POTTER~\citep{zheng2023potter} evaluation. }
\label{tab:hand_num_views}
\end{subtable}
\end{tabular}
\caption{Analysis for the hand pose benchmark. }
\end{table}

\begin{table}[]
\centering
\small

\end{table}

For each annotated joint, the manual annotations keep record of the number of views where the joint is visible. The visible 2D observation is then used for triangulation in 3D ground truth generation. This can be taken as an indicator of the uncertainty of the ground truth, and the difficulty level for the estimation of the joint (usually, a joint visible by fewer views indicates that it is more entangled with objects or other part of the hand). Table~\ref{tab:hand_num_views} shows that the PA-MPJPE decreases as the visible number of views increases.  \NEWRESULT{To guarantee the ground truth accuracy, all experiments are performed only on joints at least visible in 3 cameras. }

%% file: sec/6_conclusion.tex
\section{Conclusions}

Ego-Exo4D provides a robust data collection pipeline, a dataset of unprecedented scale and realism, and a benchmark suite for ego-exo skilled activity understanding and video learning.
We propose a replicable pipeline to collect synchronized multi-exo cameras along with egocentric data in diverse settings indoors and outdoors.
The setup was replicated across country boundaries to collect a homogeneous dataset.
This offers, for the first time, collection of ego-exo data outside mocap suits or lab settings, capturing the participants where they naturally carry out their skilled activities---e.g. chefs in their kitchens, dancers in their studios, and football players on the pitch.

Eight compelling domains were selected with diverse skilled activities.
We divide these domains into physical skills---those that particularly require strengthening, flexing or training the human body to carry out a skill, e.g. dancing; and procedural skills---those that one masters through the usage of tools to manipulate the surrounding environment, e.g. cooking.
Both types of skilled activities (physical and procedural) have never been explored jointly. 
By bringing the two types of skilled activities to a common dataset and benchmark, Ego-Exo4D will address the ever-lasting promise of assistive technologies, beyond a single domain or application.

The dataset comes with a suite of benchmarks, models, evaluation scripts, web-based visualizer, and baselines, to assist the research community in exploring and building on the challenges posed by Ego-Exo4D.

One of the challenges of Ego-Exo4D, and consequently its limitations, is the difficulty in optimizing the positions of exo cameras in the various settings.
Often, the action is occluded by the person in the majority of the exo camera due to the standard and static positioning of these cameras.
This impacted the annotations at times, and we opt to manually select suitable exo cameras during annotations.
Additionally, the data is long-tailed due to the natural durations of activities.
For example we have 9x more hours of cooking than soccer, as preparing a meal takes much longer than a soccer drill.
The tasks also differ in their skill challenge; for example, learning to shoot the basketball into a hoop requires a lot more training and expertise than learning to carry out a COVID test.
This diversity, while part of daily activities, could introduce challenges to current model training.

In addition to the three foundations of this dataset: pipeline, dataset, and accompanying benchmarks, two additional research gems should be highlighted for interesting future directions.

First, the videos are accompanied by three levels of linguistic descriptions: (i) fine-grained narrations of ongoing actions, (ii) descriptions of the activity by the participants themselves reflecting on their expertise and why they perform in a certain manner---we refer to this as `act and narrate', as well as (iii) commentary from expert tutors. These are people trained to teach or evaluate the skill of others. The commentary is temporally synced with the action and enriched by spatial highlights to pay attention to particular ways in performing skills, both showcasing excellence as well as points for improvement---we refer to this as `expert commentary'.
Ego-Exo4D thus offers the first resource of its kind to compare how actors and observers reflect, similarly or distinctly, on their skills.

Second, Ego-Exo4D offers for the first time the chance to study detailed hand-pose including hand-object interactions and full body pose in one dataset. A few recent works have showcased the potential of combining both in synthetic~\citep{Tendulkar_2023_CVPR} or controlled~\citep{GRAB2020} settings. Ego-Exo4D can be used as a base to take these directions further into real-world recordings.

Ego-Exo4D is a massive step towards a holistic understanding of the individual camera wearer---particularly their personal goals to advance their skill levels for their job or hobby.
Models that understand one's skill will offer the ultimate assistive companion, as noted by the survey paper~\citep{plizzari2024outlook}.
Such a companion can offer actionable personalised feedback~\citep{ashutosh2024expertafexpertactionablefeedback}, then continuously monitor and quantify that feedback's impact on the camera wearer's skill over time, towards life-long skill mastering.
Beyond skill understanding in video, Ego-Exo4D serves as a resource to deepen research in general 3D vision (environment reconstruction, camera relocalization, and others), video-language learning (grounding actions and objects, multimodal representation learning, language generation), and traditional exocentric activity understanding.

%% file: sec/7_contribution.tex
\section*{Contribution statement} 

\begin{small}
    This project is the result of a large collaboration between many institutions over the last two years. Initial authors represent the leadership team of the project. Kristen Grauman initiated the project, served as the technical lead, initiated the recognition and proficiency benchmarks and expert commentary, and coordinated their working groups. Andrew Westbury served as the program manager and operations lead for all aspects of the project. Lorenzo Torresani led development of the capture domains, initiated the relation and ego-pose benchmarks, and coordinated their working groups.  Kris Kitani led development of the multi-camera rig and supported the Ego-Exo4D engineering team on all aspects of the data annotation and organization.  Jitendra Malik served as a scientific advisor.  Authors with stars ($^\ast$) were key drivers of implementation, collection, and/or annotation development throughout the project.  Authors with daggers ($^\dagger$) are faculty and senior researcher PIs for the project.  The Appendices detail the contributions of individual authors for the various benchmarks, data collection, and annotation pipelines.
\end{small}

\section*{Acknowledgements}

\begin{small}
We gratefully acknowledge the following colleagues for valuable discussions and support of our project: Vittorio Caggiano, Sarah Carroll, Il\'{e} Danza, Ahmad Darkhalil, Zona de Bloque, Alex Dinh,  Rene Martinez Doehner, Ivan Cruz, Matt Feiszli, Vance Feutz, Kelly Forbes, Rohit Girdhar, Pierre Gleize, Andr\'{e}s Hern\'{a}ndez, Shun Iwase, Hanxiao Jiang, Armin Kavian, Bolin Lai, Vivian Lee, Brighid Meredith, Ashley Massie, Natalia Neverova, Manohar Paluri, Joelle Pineau, Artsiom Sanakoyeu, Paresh Shenoy, Jiaray Shi, Jiasheng Shi, Gaurav Shrivastava, Mitesh Singh, Manasi Swaminathan, Arjang Talattof, Ali Thabet, Laurens van der Maaten, Andrea Vedaldi, and Tobby Zhu.

We also sincerely thank the 52 experts who contributed to the expert commentary for their expertise and support; they are listed individually in Appendix~\ref{sec:appendix-language}.
Thank you to the Common Visual Data Foundation (CVDF) for hosting the Ego-Exo4D dataset.    Finally, thank you to the \numcamerawearers participants who contributed to this dataset and shared their skills in video.

UT Austin is supported in part by the IFML NSF AI Institute.
University of Catania is supported in part by the project Future Artificial Intelligence Research (FAIR) – PNRR MUR Cod. PE0000013 - CUP: E63C22001940006. Luigi Seminara is supported by PNRR PhD scholarship “Digital Innovation: Models, Systems and Applications” DM 118/2023.
Simon Fraser University is supported in part by the Canada Research Chairs Program (CRC-2019-00298) and NSERC Discovery (2019-06489)
Georgia Tech is supported in part by NSF (\#2144194) and NSF-GRFP
Indiana University is supported in part by NSF DRL-2112635 (AI Institute for Engaged Learning). 
Univ. of Bristol is supported in part by EPSRC UMPIRE (EP/T004991/1) and EPSRC PG Visual AI (EP/T028572/1). Z. Zhu is supported by UoB-CSC Scholarship.
University of Tokyo is supported in part by JST ASPIRE Grant Number JPMJAP2303 and JSPS KAKENHI Grant Numbers JP24K02956.
\end{small}

%% file: sec/appendices/A_Aria.tex
\clearpage
\section{Camera setup and recording details} \label{sec:appendix-aria}

\subsection{Time sync}

\begin{figure}[t]
    \centering
    \includegraphics[width=1\linewidth]{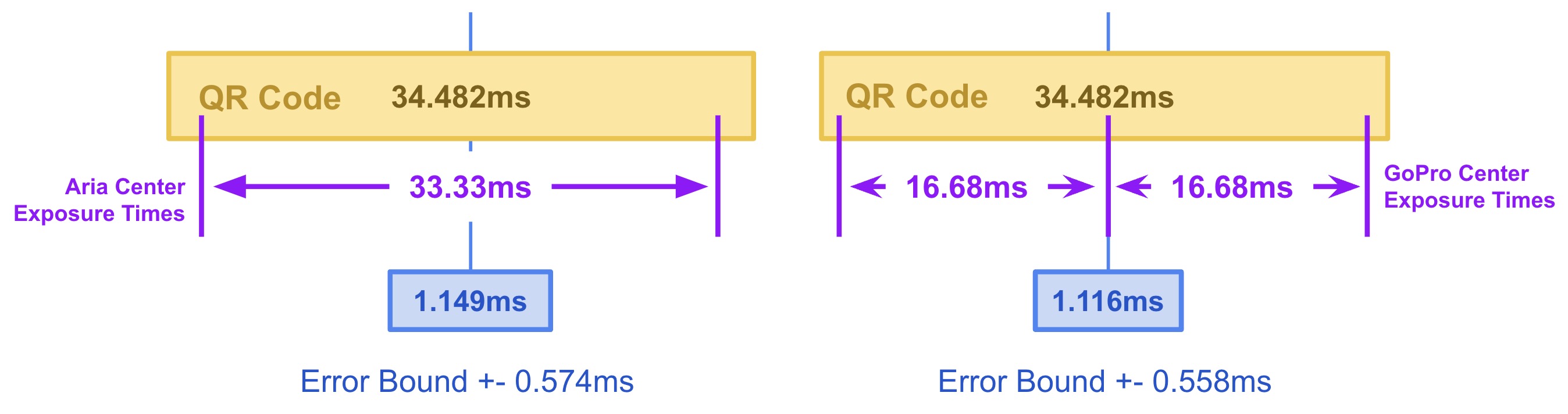}
    \caption{With a QR code timer playing back at exactly 29fps, cameras with evenly spaced center-exposures can be precisely time-localized to the QR timer with these multi-QR patterns.}
    \label{fig:timesync-timing}
\end{figure}

To sync cameras, we employ a pre-rendered sequence of QR Codes (\emph{i.e.}, QR code video) that encode a wall-clock time. We show this QR code video using the smartphone at 29fps to all cameras in sequence and exploit the difference in frame rates to finely sync the cameras. In theory, the QR code decoded on a frame that captures a QR change is likely the one that was visible during that frame's center of exposure. With a single QR, the camera's center of exposure time could be anywhere within the 34{.}48ms that the QR is shown. However, with two consecutive frames with the same QRs, we can localize that time 
down to $\pm 0.574$ms. The same approach yields $\pm 0.558$ms for the 59fps GoPros given 3 consecutive frames (see Figure \ref{fig:timesync-timing}),  providing sub-frame synchronization accuracy.

\KGCR{We manually verified that each GoPro camera was within 1 frame (+-16.66ms) of the Aria RGB camera by %
visually comparing them at single-frame moments (e.g., contact frames) using a synced video collage at the start and  end of each capture.  We checked points near the start and end of each capture under the logic that sync is a linear mapping and camera clock speed is mostly constant, so if the error is +- 1 frame at the start and +- 1 frame at the end, it will be +- 1 frame throughout.}

An `audio sync' fingerprint was played at the start and end of each capture to synchronize audio streams but has not been used.

\paragraph{Challenges and workarounds}

In practice, \KGCR{$\sim$70\%} of recorded captures yielded frame-accurate sync through our automated pipeline. Inaccurate sync causes included observed issues (\emph{e.g.}, phone changing orientation mid-playback, video playback interruptions) and suspected ones (\emph{e.g.}, videos not playing back at precisely 29fps, center exposure times not being evenly spaced). To recover these captures, we employed a manual sync procedure wherein people manually selected frame timestamps that should be aligned based on precisely time-localizable events, \emph{e.g.}, a lighter first sparking, a soccer ball making contact with a cleat, or a hand beginning a fast slide down the neck of a guitar. This unblocked the remaining \KGCR{$\sim$30\%} of captures at the cost of less accurate sync.

\paragraph{Alternatives}

We explored and disqualified other sync options---notably using Timecode with TentacleSync or Ultrasync. Both of these solutions use LTC to encode a 1fps timestamp into the audio channel of a connected device. Using them with GoPros would cost us the stereo-audio modality, which we opted to keep to support audio-based research areas. We additionally lacked an ergonomic input solution for Aria to use while recording, so that mandated non-intrusive sync solutions.

\subsection{Take separation}

To amortize the setup and tear down time required for each recording, we record multiple `takes' (\emph{i.e.}, one instance of a certain task) back-to-back and use a `Take Separator' QR code (different from the time sync QR code video) that is identified in post-processing to auto-separate each take. This enables us to scale up recording---particularly for the physical scenarios where a single take can be less than a minute long. Data collectors track metadata for each take, identifying them by index and marking data such as participant ID (anonymous unique identifier), task (\emph{e.g.}, making tea, making cucumber salad, performing CPR), and whether the take should be dropped (\emph{i.e.}, if it is just setup time between activity enactments).

\begin{figure*}[t]
\begin{center}
    \centering
    \includegraphics[width=1\linewidth]{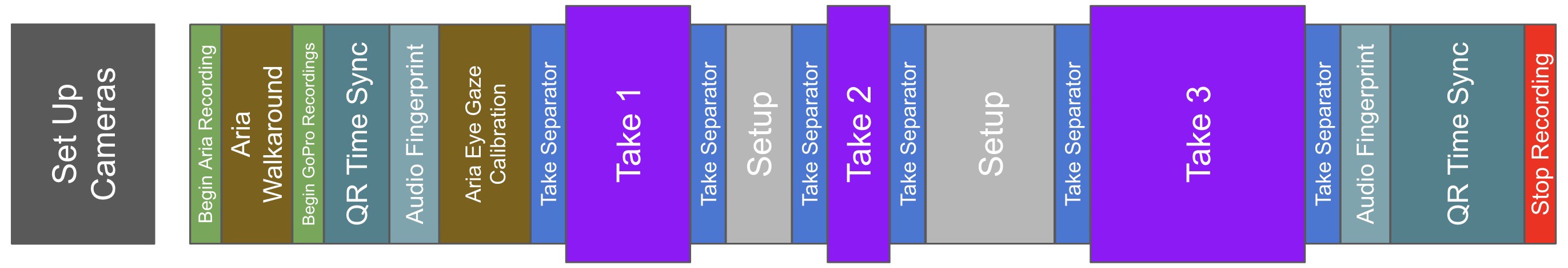}
\end{center} 
\caption{Overview of the recording procedure}\label{fig:recording}
\end{figure*}

\subsection{Recording procedure}

Our rig setup procedure entails setting up the stationary exo cameras in the recording environment and displaying QR codes to perform time sync and then take separations.  
Figure~\ref{fig:recording} overviews our recording procedure.

\begin{enumerate}
    \item Position tripods, power on GoPros, and set camera angles to ensure maximum human coverage.
    \item Begin Aria recording via smartphone.
    \item Conduct a walk-around with the Aria glasses to build a basemap for 3D reconstruction and camera localization. Match the viewpoint of each GoPro camera by positioning the Aria directly in front of its lens.
    \item Start QR Timesync Video off-screen. Show QR video to Aria RGB camera.
    \item Use GoPro Remote to begin GoPro recording. Show QR video to each GoPro camera. Play Audio Sync fingerprint from the center of the space.
    \item Pass Aria glasses to (new) participant. Perform Eye Gaze Calibration via the Aria app. Show `Take Separator QR' to one GoPro and begin the take. Show `Take Separator QR' to one GoPro after the take is complete and repeat this step for each participant/take. Do not repeat gaze calibration if the participant has not changed.
    \item Play Audio fingerprint from the center of the space. Restart the QR Timesync Video off-screen. Show it to the Aria RGB camera, then each GoPro. Stop recording on all cameras.
\end{enumerate}

The core camera rig was extended to handle onsite requirements and regional challenges. The team at Universidad de los Andes introduced a top-down (ceiling mounted) GoPro for dance, which was adopted by the team at the University of Pennsylvania with an overhead pole mount. The teams at University of Pennsylvania, IIIT-Hyderabad, and Indiana University added an additional egocentric, head-mounted GoPro.

\subsection{Aria post-processing} First, the Aria Machine Perception Services (MPS) pipeline is invoked for \textit{each full Aria recording}---these  typically are about 20 minutes to 1 hour long and can include several takes, the hand-over in-between takes, as well as some other set-up steps. This is followed by localizing all GoPro videos of that scene as described above, and finally followed by time-synchronization across Aria and the GoPro cameras, as well as take-separation, as described below.

There are total of \numrecordings Aria recordings processed by MPS---containing the total \numtakes takes in the dataset.
95.9\% of these recordings have successful Aria localization throughout the whole recording, with only 3.5\% containing a partial tracking failure (leading to short gaps in the 6DoF trajectory). Three (0.6\%) recordings failed completely. The most common failure reason is physical shock on the glasses, for example when the glasses are accidentally dropped on the ground or the table. 

Furthermore we attempted to localize a total of \numgoprorecordings GoPro recordings, 91.4\% of which are successfully localized. Similar to the Aria recordings, GoPro's are localized on a \textit{recording} level rather than on a \textit{take} level. This helps in particular with very short takes as are common during physical activities---as there otherwise would not be sufficient visual overlap across Aria and GoPro perspectives.
The most dominant reason for GoPro localization failure occurs when the GoPro is pointed to an texture-less area (e.g. a white table) which lacks the necessary visual features to perform localization.  As the GoPro's are static, this cannot be compensated for by device motion as is the case for the moving Aria device.

Technical documentation and open-source tooling for Aria recordings and MPS output is available on Github\footnote{\href{https://github.com/facebookresearch/projectaria_tools}{https://github.com/facebookresearch/projectaria\_tools}} and the associated documentation page\footnote{\href{https://facebookresearch.github.io/projectaria_tools/docs/intro}{https://facebookresearch.github.io/projectaria\_tools/docs/intro}}. It includes both python and C++ tools to convert, load, and visualize data; as well as sample code for common machine perception and 3D computer vision tasks.

%% file: sec/appendices/A_collection.tex
\section{Data collection}\label{sec:appendix-collection}

Twelve research labs came together for nearly two years to create Ego-Exo4D.  Importantly, our collection across the sites was a coordinated effort, with common guidelines, scenarios, and camera rigs.  In this way, the dataset is cohesive at the same time it is diverse.  In this section we describe the data collection details that are specific to each partner site, e.g., how they recruited participants, which of the 8 scenarios they captured, or any modalities they added on top of the common rig.  

Figure~\ref{fig:scenarios-per-university} shows the breakdown of which scenarios were captured by each partner institution as well as a map highlighting the locations of the 12 labs involved in data collection.  Note that an additional four institutions not shown on the map are part of the consortium (e.g., contributing to benchmarks) but did not collect data.  They are UT Austin (USA), KAUST (Saudi Arabia), University of Catania (Italy), and University of Bristol (UK).

\begin{figure}[t]
  \centering
  \begin{subfigure}[b]{\linewidth}
  \centering
    \includegraphics[width=0.48\linewidth]{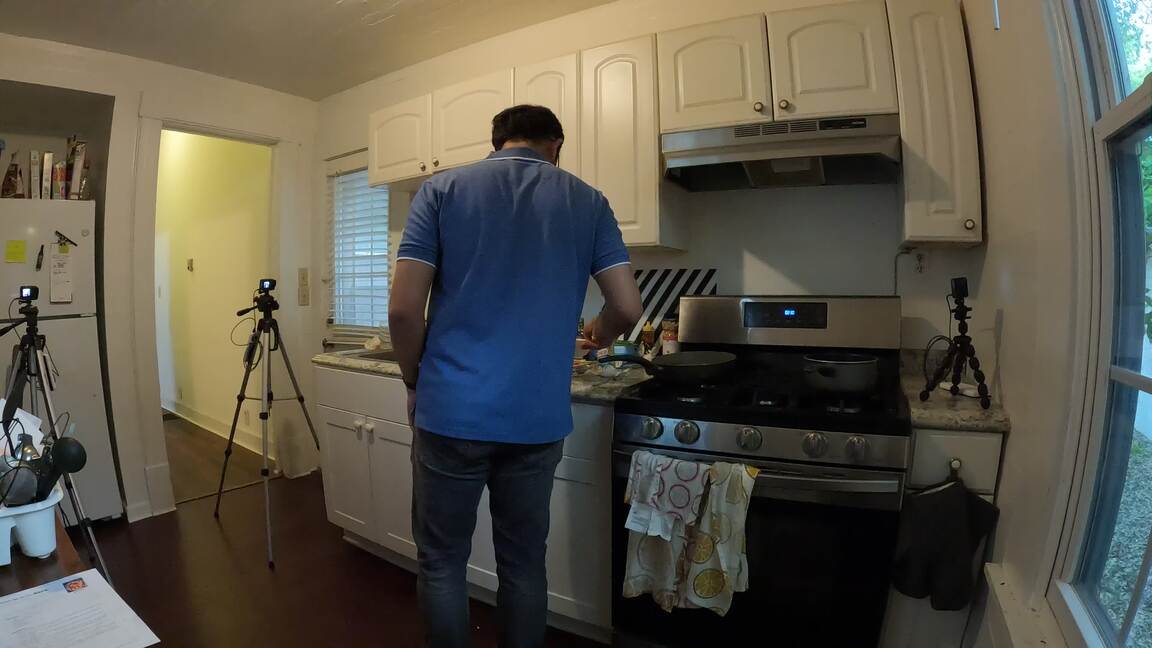}%
    ~
    \centering
    \includegraphics[width=0.48\linewidth]{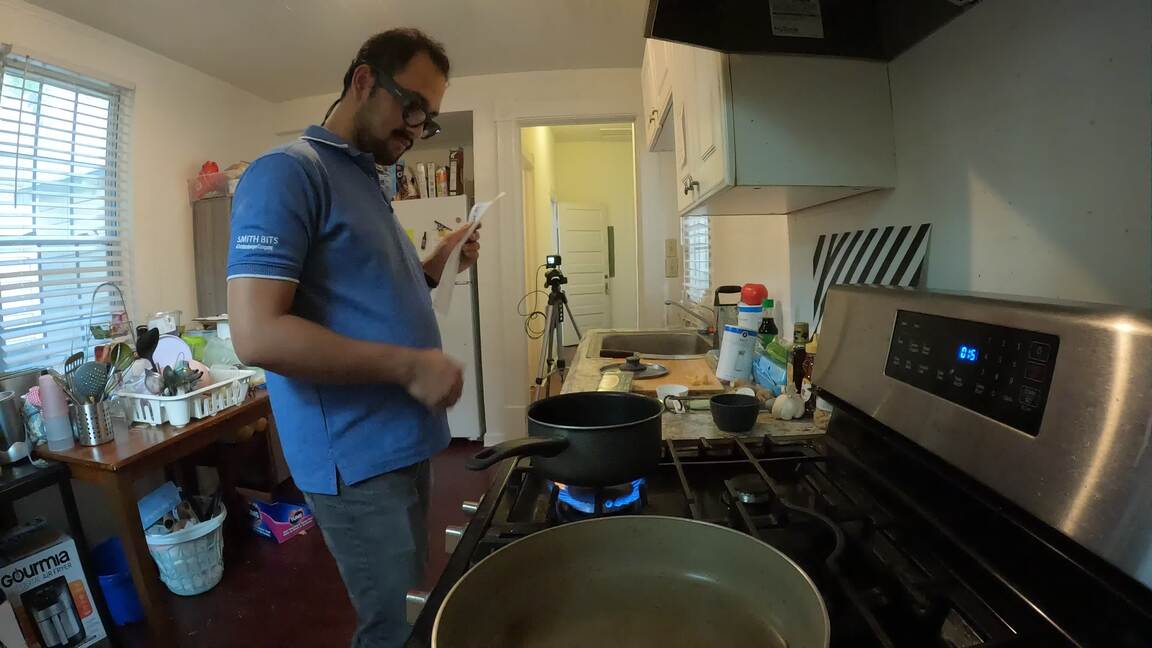}
    \caption{Cooking Scene}
    \label{cookscene}
  \end{subfigure}
  
  \begin{subfigure}[b]{\linewidth}
  \centering
    \includegraphics[width=0.48\linewidth]{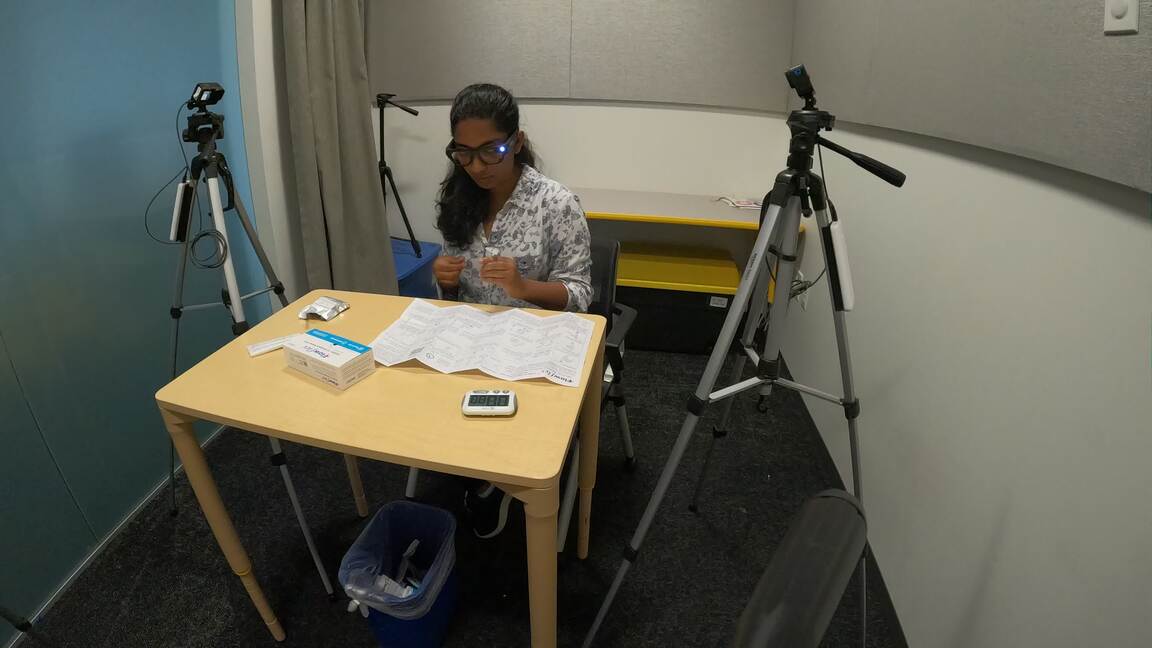}%
    ~
    \centering
    \includegraphics[width=0.48\linewidth]{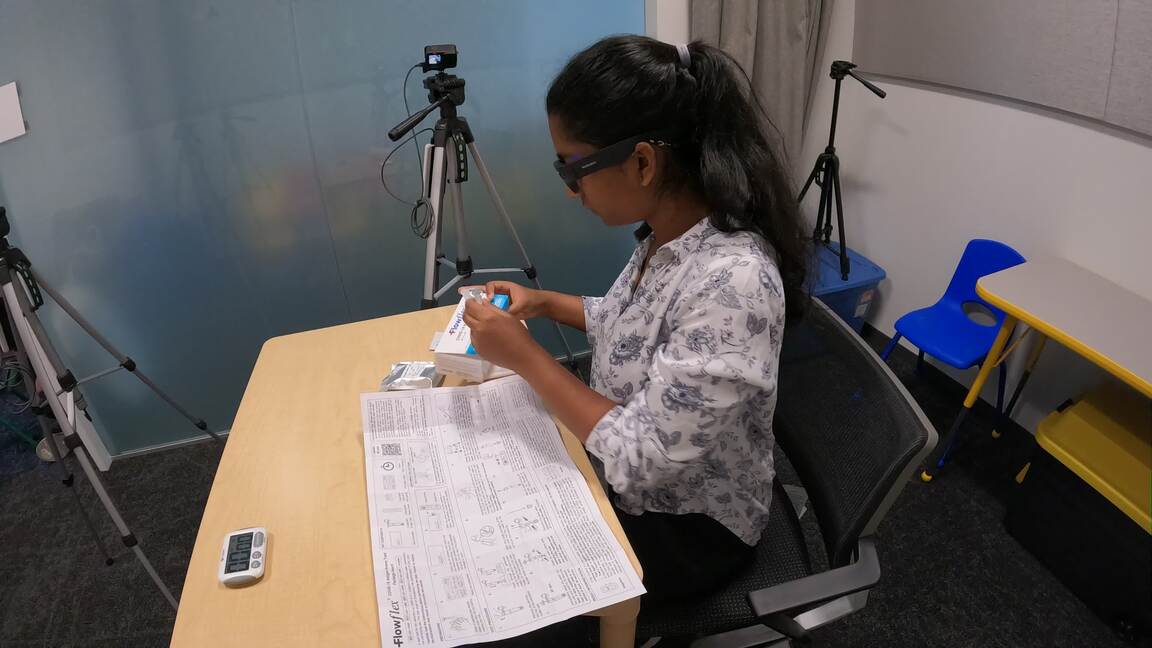}
    \caption{COVID Test Scene}
    \label{covid_scene}
  \end{subfigure}

  \begin{subfigure}[b]{\linewidth}
  \centering
    \includegraphics[width=0.48\linewidth]{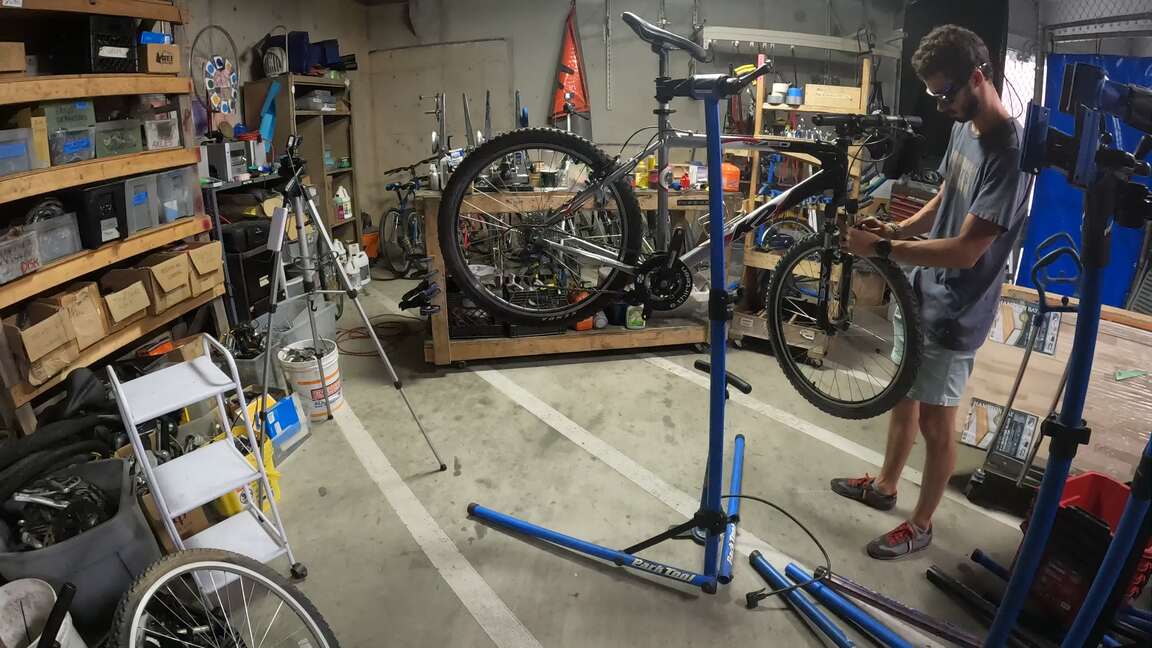}%
    ~
    \centering
    \includegraphics[width=0.48\linewidth]{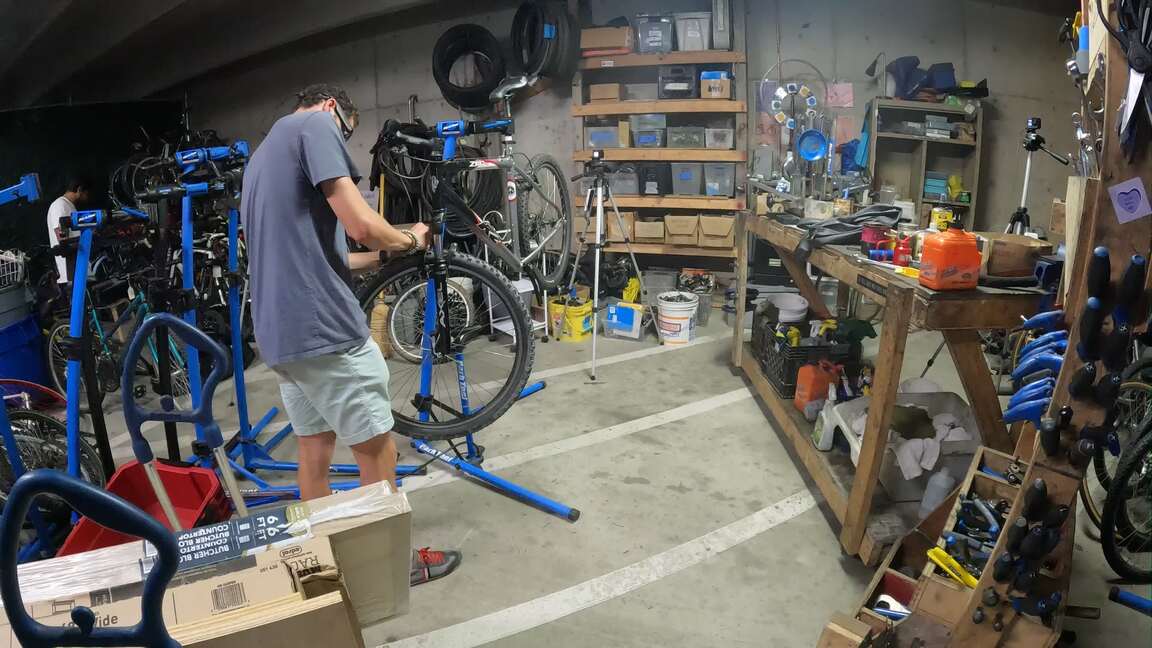}
    \caption{Bike Repair Scene}
    \label{bikeshop}
  \end{subfigure}

  \caption{Views from two different cameras for each scenario collected in Atlanta, GA, USA.}
  \label{Scene cam set up}
\end{figure}

\subsection{Carnegie Mellon University}

\input{sec/appendices/collection/cmu}

\subsection{FAIR, Meta}
\input{sec/appendices/collection/fair}

\subsection{Georgia Tech}
\input{sec/appendices/collection/gatech}

\subsection{IIIT-Hyderabad}

\input{sec/appendices/collection/iiit}

\subsection{Indiana University}
\input{sec/appendices/collection/iu}

\subsection{National University Singapore}
\input{sec/appendices/collection/nus}

\subsection{Simon Fraser University}
\input{sec/appendices/collection/sfu}

\subsection{Universidad de los Andes}
\input{sec/appendices/collection/losandes}

\subsection{University of Minnesota}
\input{sec/appendices/collection/umn}

\subsection{University of North Carolina}
\input{sec/appendices/collection/unc}

\subsection{University of Pennsylvania}
\input{sec/appendices/collection/upenn}

\subsection{University of Tokyo}
\input{sec/appendices/collection/tokyo}

%% file: sec/appendices/collection/cmu.tex
Carnegie Mellon University focused on three skill-based activity scenarios: (1) soccer, (2) bike-repairs, (3) cooking. The exocentric cameras for our collections, four in total, were arranged approximately in a square configuration at a consistent height to capture the full range of the activity. Notably, for the soccer activities, an additional exocentric viewpoint was positioned inside the goal post to offer a more comprehensive perspective on the participants.

\textbf{Soccer} 
In the soccer scenario, we collaborated with professional players from the Pittsburgh Riverhounds team, representing the experts, and students from Carnegie Mellon University (CMU) as the beginners. We captured the soccer scenario across 4 different locations. The drills featured a variety of movements such as dribbling, goal kicks, and juggling, with each participant performing for a minimum of 3 minutes. This scenario resulted in roughly 4 hours of egocentric footage and 18 hours from exocentric perspectives, encompassing 32 participants in total. 

\textbf{Bike repair}
In the bike-repair segment, our experts were seasoned mechanics with over a decade of experience from Allegheny county. To ensure authenticity, we visited each mechanic in their respective shops to allow usage of their own tools and setup. Four tasks were captured for each bicycle, and we ensured bicycle diversity by selecting different sizes, shapes, colors, and makes. The tasks include tire removal, tube change/ inflation, tire reassembly, and clean/lube chain. This yielded 3 hours of egocentric recordings and 12 hours of exocentric footage, encompassing 22 different bicycles. 

\textbf{Cooking}
For the cooking section, we documented a professional chef in his traditional kitchen environment. Our dish of choice was scrambled eggs, and to inject variety, the chef prepared it using different techniques. This segment summed up to an hour of egocentric recordings and 4 hours from exocentric viewpoints.

All recordings were conducted in Pittsburgh, PA, USA, strictly adhering to CMU’s Institutional Review Board (IRB) guidelines. Every participant was briefed about the recording process, and prior to their involvement, a signed consent form was obtained.

%% file: sec/appendices/collection/fair.tex
We collected 119 total takes of skills demonstrations in New York and three different locations in California. We focused on cooking and bike repair, bringing in a skilled workforce of chefs and bike technicians that serve major kitchens and repair shops in the area. We used the unified camera rig of 1 Aria and 4 GoPros without any additional sensors. 

\textbf{Bike Repair}
Our skilled mechanics performed four different bike repairs for a total of 102 takes. We focused specifically on wheel repairs (removing and installing the wheel \& flat repairs). While we strive for diversity in terms of the model of bikes, a majority of those in the dataset are drawn from standard fleet bike models, which contain identical parts and components.  The location featured in the dataset is a well-equipped, industrial scale bike shop. 

\textbf{Cooking} 
Our chefs recorded five different recipes as part of 17 unique takes, including salads, egg dishes, and Asian garlic noodles. Locations featured in the dataset are three different professional kitchens used to prepare and serve hundreds of people each day.  %

Internal documentation and processes ensured all participants provided informed consent to appear in the dataset and participation was strictly voluntary.   

In total, we were able to mobilize five chefs and four bike mechanics. %
Participating chefs and bike technicians are highly skilled, with all research subjects reporting that they do the activity shown in the dataset daily or weekly. Similarly, eight research subjects have more than 10 years professional experience.

%% file: sec/appendices/collection/gatech.tex
Collection at Georgia Tech focused on the Health, Cooking, and Bike Repair scenarios. Across these 3 scenarios, 279 takes were captured with 34 unique participants. For all scenarios, the unified camera rig was positioned such that 2 exocentric cameras would ensure capture of the participant's hands, and the other 2 exocentric cameras would capture the participant's full body and the full environment. 

Participants were recruited from different sources including flyers, campus organizations, email lists, and word of mouth. Five of these participants completed data collections for 2 scenarios (4 participated in Health and Cooking, and 1 participated in Cooking and Bike Repair). Potential participants were provided with the study description and consent form prior to scheduling a recording session. At the beginning of each session, study personnel walked through the consent form with the participant, and answered any questions. The participant then reviewed and signed the consent form to confirm participation in the study. 

The recording environment differed by scenario and included participants' homes, campus meeting rooms, and an on-campus bike shop. Fig \ref{Scene cam set up} shows a sample environment and camera setup for each of the Health, Cooking, and Bike Repair scenarios.  Further details of the data collection specific to each scenario is provided below.

\paragraph{Health}
Participants for the Health Scenario took COVID rapid test kits while seated at a table. Recordings were captured in 2 different on-campus meeting rooms. Participants were recruited through campus email lists and flyers in local coffee shops. Each recording session lasted approximately 40-60 minutes and consisted of a participant completing 5-7 test kits, using 2-4 different types of test kits. 7 different types of COVID test kits were used across the full collection. In total, 96 takes were recorded from 16 unique participants.

\paragraph{Cooking}
Participants for the Cooking Scenario prepared dishes from three recipes: Asian Salad, Tomato \& Eggs, and Garlic Noodles in their home kitchens. Participants were recruited via mailing lists of local apartment complexes, contacting participants from prior research studies, and word of mouth. Each recording session lasted 2-3 hours, capturing 3-6 takes of a recipe being cooked from start to finish. Participants cooked 2-3 of the recipes during their session, depending on dietary restrictions and preferences. Participants were provided with ingredients and a paper copy of the recipe, and used equipment from their own kitchen to prepare the food. In total, 71 takes from 15 unique participants were captured. %
The takes were about evenly distributed among the three recipes.  
Recordings were completed in 10 unique kitchen environments.

\paragraph{Bike Repair}
Participants for the Bike Repair Scenario performed repairs including taking off a wheel, putting on a wheel, replacing a tube, and cleaning a dirty chain. We recruited skilled participants from a campus bike repair organization. There were 8 unique participants, who each completed 1-3 recording sessions. Each session lasted 40-60 minutes and captured 5-7 takes of individual bike repairs. In total, 112 takes were recorded. %
showing the distribution across repair tasks. One session of 6 takes was recorded in a participant's home, while the rest were recorded in the campus organization's bike shop space, which is shown in Fig \ref{bikeshop}. 
Due to the organization's access to a large quantity of used bicycles, there is large diversity in the make and model of bicycles across takes. 

The study protocol was reviewed and approved by our Institutional Review Board (IRB).

%% file: sec/appendices/collection/iiit.tex
\begin{figure}
    \centering
    \includegraphics[width=\columnwidth]{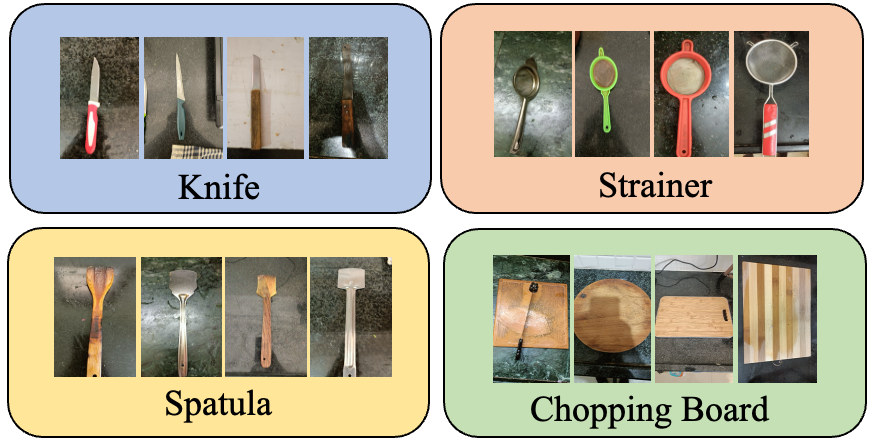}
    \caption{In Hyderabad, India, cooking was captured in different kitchens with socio-economic diversity.  We observe that the same kitchen tools appeared in different shapes. }
    \label{fig:Obj-IIITH}
\end{figure}

In Hyderabad, we contributed to three scenarios - (a) cooking, (b) soccer, and (c) music. We formulated a data collection strategy tailored to the specific scenario, as outlined below.

Our primary objective was to comprehensively capture body and hand movements, along with their interactions with objects, during the execution of the activities. In general, we adhered to the standard camera setup instructions. Nonetheless, we incorporated an extra exo-camera for capturing soccer activities in order to enhance the overall coverage of the event. Additionally, for music activities, we introduced a head-mounted Go-Pro camera. 
 
 This decision stemmed from the observation that expert musicians frequently do not directly look toward their instruments while playing. Consequently, the head-mounted camera guarantees continuous visibility of both the hands and the musical instruments, providing an ego-view perspective.

 The collection in India was done during the peak summer, and this led us to a challenging situation where the cameras frequently shut down due to overheating. To address this, we mostly avoided capturing multiple takes in one capture and placed the cameras into an ice chest box in between the captures to cool them down.
 
\textbf{Cooking}
For cooking, we reached out to people located in Hyderabad with varying socio-economic backgrounds and explained the data collection plan, and goals. We also requested them to engage their family members as well as friends in this data capturing process. Finally, we recorded the videos with the 41 informed participants capturing in a diverse set of 19 kitchens, geographically well-apart in and around Hyderabad, resulting in a rich dictionary of kitchen utensils (see Figure~\ref{fig:Obj-IIITH}), narrations in four different languages. Additionally, we made an effort to ensure a balanced representation of genders in our overall data collection process.

\textbf{Soccer}
For soccer, we reached out to three different soccer training schools in Hyderabad with the overall recording plan and process. They helped us in recruiting local soccer teams who play professional tournaments and practice almost everyday. We also recruited few players from our university soccer teams. In total, we recorded 49 participants, `performing dribbling, juggling, and penalty-kick activities.

\textbf{Music}
For the music scenario, we contacted one music school and recruited 4 musicians from them having at least 3 years of experience of playing either the piano, guitar, or both instruments. To add diversity, the musicians were asked to play western as well as Indian pieces. 

Our collection protocol was reviewed and approved by %
our university's Institutional Review Board (IRB). The primary conditions set forth by the IRB encompass the following aspects: (a) participants with 18+ age  are deemed suitable for inclusion in the project, (b) participants have provided explicit consent for their facial and vocal presence to be featured in the released videos, (c) participants have willingly agreed to take part without receiving any immediate financial incentives from the videos, and (d) participants have the autonomy to engage in the activities in an environment of their choice. The participants were given the detailed descriptions of the project beforehand and requested to sign the consent form.  Each participant received compensation as part of the process. 

We selected participants from a wide range of age groups, spanning from 18 to 61 years old, to introduce an additional layer of diversity. Moreover, the participants were from diverse professional backgrounds (e.g., coach, software engineer, data annotators, project managers etc.). Before sharing, we carefully examined each video to ensure there was no sensitive content.

%% file: sec/appendices/collection/iu.tex
We focused on  cooking, bicycle maintenance, and music scenarios.  All activities were collected using the unified camera rig, including additional sensors in specific scenarios. For cooking, the 4 GoPros were placed 90 degrees apart from each other, with 2 placed close to the participants to capture hands and objects and 2 placed further to capture the overall scene.  In music,  4 GoPros were placed in front of the player, approximately 45 degrees apart from each other. In addition, we attached an additional GoPro HERO10 camera to the participant’s head (using a helmet), tilted down roughly 80 degrees to capture hand movements. In bike repair environments, the 4 GoPros were placed 90 degrees apart from each other, of which 1 GoPro was placed close to the bike, 1 GoPro was placed close to the workbench and tools, and 2 GoPros were placed further away from the participants to capture the overall scene.

\textbf{Cooking}
For cooking, we had a total of 18 participants collect 72 takes and 20.5 hours of video. For 15 of the participants, we used a commercial test kitchen at our university. We purchased all of the ingredients and kitchen equipment ahead of time and had them ready when each participant arrived. We asked them to make four dishes (chai tea, sesame-ginger salad, tomato and eggs, and noodles) and provided printed recipes for these dishes. The remaining three participants chose to record in home kitchens, and the four dishes they made varied based on their preferences (one participant made omelet, cucumber salad, noodles, and chai tea, another made scrambled eggs, sesame-ginger salad, sushi rolls, and brownies, and the third made scrambled eggs, cucumber salad, noodles, and milk tea). Due to concerns about food safety, we discarded (composted) the cooked dishes instead of allowing the participants to eat them. 

\textbf{Music}
For music, we had a total of 17 participants collect 60 takes and 6.5 hours of video. Participants were recruited based on their self-assessed proficiency in one of three instruments: piano, violin, or guitar. We recorded in 4 different locations including two studios, an office, and an auditorium that had a piano. Participants were instructed to play scales and arpeggios (2 mins), sheet music provided by us (3 mins), freeplaying (10 mins), and then recall and talk about any mistakes that were made during the playing and what could be improved (2 mins). 

\textbf{Bike repair}
For bike repair, a total of 13 participants recorded 108 takes and about 8 hours of video. We initially planned to hire professional bike technicians, but it was very difficult to recruit them in our relatively small city. Instead, we recruited more generally, looking for participants with (self-assessed) proficiency to do four basic bike maintenance tasks: removing a wheel, changing an inner tube, reinstalling a wheel, and cleaning and lubricating the chain. Most of the takes were recorded in a small house that is used for storage by our university’s landscaping staff, and provided a realistic garage-like environment. We provided participants with a bike rack and supplies including bike tubes, pumps, tools, chain cleaner and lubricant, and gloves. To achieve diversity in different bikes and bike types, we asked participants to bring their own bike when possible, and we also provided 4 bikes (one of which belonged to one of the authors and the other three which we bought at a salvage shop). Most participants performed takes on about 3 bikes. One participant chose to record in an apartment, and one recorded in a hallway in a university building instead of the garage due to scheduling conflicts. 

Our protocol was reviewed and approved by our university’s Institutional Review Board. For each potential participant in each scenario, we first scheduled an online introduction meeting to tell them about the study and answer their questions and concerns. If they were interested, we agreed on the activity they would perform and when and where to meet for recording. We also sent them the informed consent form to give them sufficient time to review. On the recording day, we first asked them to sign the consent form, and then started recording their activities. All activities were recorded in an enclosed space to make sure that no one else accidentally entered the field of view of the cameras. We also ensured that the space did not have privacy-sensitive content, and we instructed participants not to use their phones or other devices that might show private content. 

Within a few days, we securely sent the videos to the participant so that they could review the video and ensure that they were comfortable sharing it with others. They also completed a brief online demographic study, and then were sent an incentive payment in the form of an electronic Amazon.com gift card. We made clear to participants that if they were not comfortable sharing their video, we would destroy it and they would still receive their incentive payment, although none of the participants chose this option. We gave the participants US\$20 in gift cards for each hour of their time spent recording (with a minimum of \$20, and partial hours rounded up to the nearest \$5). We gave an additional \$20 gift card to reimburse travel costs for those who came to our facilities to record (e.g. in our kitchen, bike repair shop, or on-campus studio or auditorium). For cooking and bike repair, we gave an additional \$20 gift card to participants who provided their own ingredients or bike maintenance supplies, to defray these expenses. 

We recruited participants in the Bloomington, Indiana, USA area through online email advertisement, word of mouth, physical flyers, and posting on social media. We recruited participants who were 18 years of age or older, had self-assessed expertise in the activities as described above and could perform the tasks without wearing prescription glasses (which could interfere with the Aria’s gaze tracking).

%% file: sec/appendices/collection/nus.tex
In Singapore we focus on the following scenarios: soccer, health-related activities including COVID-19 ART testing and Cardiopulmonary Resuscitation (CPR), and cooking.  In total, our collected data encompasses around 26 hours of egocentric videos and 117 hours of exocentric videos. These videos spread across 327 takes. In general, we adhered to the standard camera placement guidelines; however, for each scenario, we fine-tuned the position of the exo cameras based on practical considerations. For instance, in a small kitchen for cooking, we positioned the camera on the table to broaden its field of view.

\textbf{Soccer} 
For soccer, we conduct recordings at a university sports field. Our participants were primarily sourced through referrals provided by skilled participants recruited through online calls for participants. Additionally, during outdoor recording sessions, we occasionally invited surrounding bystanders to participate.

\textbf{Health}
For health activities, we recorded in vacant classrooms, meeting rooms, and outdoor fields. CPR sessions are captured either in a yoga classroom or in a quiet, empty outdoor field. For recruitment, we circulated online calls for participants and then, for skilled activities like CPR, we collaborated with experts to organize training courses. Participants would participate in these courses and were trained to be proficient and then conducted recording afterwards.  

\textbf{Cooking} 
As for cooking, which requires a kitchen, we used the kitchen in our lab mates' apartments and arrange other participants to go there.

Our data capture has been approved by %
our university's Institutional Review Board (IRB). The main requirements include that participants: (1) agreed to take part in the study, (2) agreed to donate their speech, image, video, IMU, and 3D scan data for the purposes of this research, (3) agreed that their face, tattoos, and voice may appear in the data, (4) have the right to withdraw their recorded data at any time.

In Singapore, high temperatures often pose the challenge of camera overheating, particularly for GoPro cameras, which can lead to protective shutdowns and interrupt data collection. To mitigate this, we place small ice cubes wrapped in wet wipes on the GoPro cameras to help cool it down during recording. Furthermore, we attempted to schedule our participants' recordings in the evening or during an overcast day.

Our data pool comprises contributions from about 93 meticulously selected participants, ensuring a proficient completion of the recordings. Particularly in soccer, most participants have extensive experience and were members of their school or college soccer teams.

%% file: sec/appendices/collection/sfu.tex
We captured three types of scenarios in a variety of environments: cooking, basketball, and COVID-19 testing. In total, 88 participants carried out activities in the three scenarios we collected in a total of 61 data capture sessions, resulting in 519 activity takes.

We used the unified camera rig and followed the general collection guidelines with a number of small adjustments to facilitate scenario-specific capture. In kitchen and health scenarios where the participant interacts with small objects in tabletop height settings, the placement of exocentric cameras was optimized in a ``two near, two far'' setup to provide for visibility of the small objects and hands while also capturing the overall human pose during the activity.

\textbf{Cooking}
The cooking scenario was captured in a decentralized fashion by going to the participants' own residences and asking them to cook in their kitchen. This allowed for diversity in the environment as well as in the participant during data capture.  Our data capture sessions resulted in 112 cooking takes.

\textbf{Basketball} 
Collection for the basketball scenario was done in a ``round robin'' fashion to reduce player-to-player overhead. We targeted a spectrum of experience levels, for example going from university basketball team players who compete at the national level to more amateur basketball players who only have played basketball occasionally. We collected 355 takes of basketball activities. 

\textbf{Health} 
Following the standard data collection guidelines for health activities, we gathered 52 takes of health activities. 

We followed the institutional research board (IRB) process at our institution to acquire approval for the participant recruitment strategy, study setup, and participant consent acquisition forms. All participants consented to their data being collected and distributed for research purposes. Participants have the right to request that their data be withheld from inclusion in the dataset.

We recruited participants by word of mouth, reaching out to specific clubs and groups for some of the activities, and more generally through advertisement using university-affiliated communication channels.

%% file: sec/appendices/collection/losandes.tex
\begin{figure}
    \centering
    \includegraphics[width=\columnwidth]{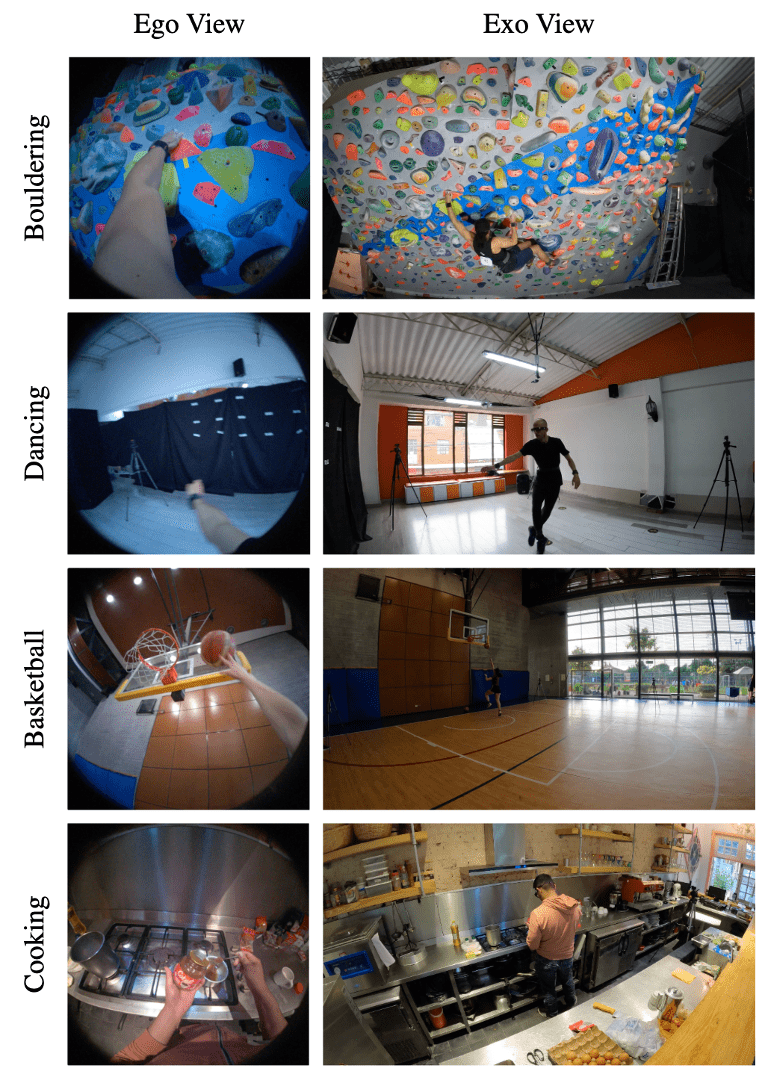}
    \caption{\NEW{Egocentric and one exocentric view for each of the recorded scenarios in Bogota, Colombia.}}
    \label{fig:stats-uniandes}
\end{figure}

We collected around 40 hours of video spanning four distinct scenarios that encompassed three physical activities (basketball, bouldering, and dancing) and one procedural activity (cooking). Figure \ref{fig:stats-uniandes} \NEW{shows examples of the diverse scenarios that we collected.} In total, we collected 2062 takes across all the activities. We used the unified camera rig with additional activity-specific sensors as described below. 

\paragraph{Bouldering} 
We partnered with a local climbing gym, which serves as a teaching and competition center in Colombia. We used the gym as the recording location and recruited participants who practice or teach bouldering there. Our focus was to recruit participants with four different levels of expertise: beginner, intermediate, advanced, and professional climbers.
We hired expert route setters to design 33 climbing routes. These routes varied from beginner (V1) to expert level (V7). For data collection, each participant attempted to complete seven routes, having 3 minutes to make as many attempts as possible for each route. The routes were selected considering the expertise level of each participant. We located four exo cameras to capture each take; two horizontal cameras were facing the climbing wall, and the other two vertical cameras were on each side of the wall. Thus, the four cameras captured a complete view of the climbing wall and the participant's movements at every moment. We gathered 1251 takes for the bouldering scenario from 40 participants. We ensured ethnic, age, and expert-level diversity across the takes.

\paragraph{Dancing} 
We collaborated with a salsa dance academy to use as a recording location and to help with participant recruitment. We recruited students from three expertise levels: beginners, intermediate, and advanced. According to the expertise level, each dancer performed different choreographies. Beginners recorded a single choreography, while intermediate and advanced participants recorded an additional one according to their expertise. Each attempt lasted one minute, and the dancer performed from six to ten attempts. The choreographies were designed by professional dancers who teach at the academy. We used five exo cameras: four forming a square, defining the dancing area, and the fifth camera placed on the ceiling. Given the salsa dance's velocity and the movements' complexity, this fifth exo camera gave a crucial point of view for further analysis. We gathered 600 takes from 40 participants across the three expertise levels.

\paragraph{Basketball} 
We collected data from the professional women's team and students from a basketball class at our University. Each participant performed six to ten attempts for each basketball exercise. We collected all captures at the basketball court at our University's Sports Center. For this setup, we used four exo cameras around the basketball ring, ensuring a complete view of each exercise. For this scenario, we collected 167 takes from 38 participants.

\paragraph{Cooking} 
We rented a professional kitchen equipped with all the necessary utensils to perform the captures. We focused on collecting data from two types of recipes: a dish with egg and a drink. Each participant could choose between cooking an omelet, scrambled eggs, tomato and eggs, and coffee latte or tea for the drink. Each participant was free to choose how to complete each recipe. Thus, our takes show diverse ways to prepare each recipe. For this setup, we placed four exo cameras around the kitchen, all facing the user, to capture the whole kitchen without losing any detail of the person making the recipe. We placed two cameras on a counter facing the kitchen and the other two on each side of the kitchen. We collected 44 takes for the cooking scenario from 20 participants.

 The Institutional Review Board (IRB) of our university reviewed and approved our study protocol. All participants signed a consent form before participating in the study.

We partnered with professional training centers for physical activities that helped us recruit volunteers with different expertise levels. These volunteers were previously familiarized with the activities and the environment where the captures occurred. In addition, we recruited family members, friends, and acquaintances of students and faculty members of our research group for cooking.

%% file: sec/appendices/collection/umn.tex
 
Collection at the University of Minnesota focused on two main scenarios: Bouldering and Cooking. A total of 249 takes with 53 unique participants were collected. We collected all data using the unified camera rig with no additions.

\paragraph{Bouldering}
The bouldering activity was collected at a local bouldering gym, focusing on a wall with 14 different routes ranging in difficulty from beginner to expert. We collected 210 takes from 42 unique participants. Participants were asked to climb four to five routes of their choice, with the ability to take breaks within or between takes. Expert climbers who felt comfortable with the routes were able to narrate their approach and climb in real time. As participants were able to choose routes freely, our five exo-cameras were set up to accommodate the entire wall.

\paragraph{Cooking}
Cooking activity was collected on-site at each individual’s home kitchen. Five exo cameras were set up in each kitchen to maximize coverage of both the participant and the environment. We captured 9 unique kitchen environments with 14 unique participants whose skill levels ranged from cooking novice to commercial chef. Participants focused on three recipes each (scrambled eggs, Greek salad, and pasta noodles from scratch), which were performed back-to-back on the day of recording.

Our data collection protocol was reviewed and approved by the Institutional Review Board at our university. At every take, the study personnel provides a guidance to a participant through the consent form prior to participation, ensuring the participant understands the purpose of the study and all risks involved, with each participant receiving payment proportional to their contribution.

Participants were recruited via word of mouth, campus organizations, and digital flyers which were distributed via local social media (Facebook) communities.

%% file: sec/appendices/collection/unc.tex
Throughout our data collection at UNC, we focused on three skill-based activity scenarios: (1) basketball, (2) soccer, and (3) music drills. We used three unique environments (i.e., a basketball gym, a soccer field, and a music studio) to capture the data for each scenario. All recordings took place on the UNC campus. UNC's Institutional Review Board (IRB) reviewed and approved our study protocol. All participants signed a consent form before participating in the study.

To recruit participants, we used an online research study database, where participants from the local area could sign up to perform our study. We recruited participants willing to perform skill-based activities such as basketball, soccer, or music drills regardless of their skill level. Additionally, to recruit a more skilled group of participants, we contacted expert musicians from UNC's School of Music and athletes from UNC's basketball and soccer teams.

In total, we collected approximately 19 hours of egocentric and 76 hours of exocentric video data spanning approximately 548 takes of activity demonstrations from 56 participants (41 male, 15 female). Among the 56 participants, 44 were aged 18-25, 10 aged 25-50, and 2 aged 50-75. Furthermore, 26 participants had more than 10 years of experience in the scenario they chose to perform (e.g., basketball, soccer, music), 13 participants had 1-10 years of experience, and 17 had less than 1 year of experience. We used standard camera placement guidelines and the same recording devices described above.

\paragraph{Basketball} All participants performed three basketball drills: Mikan Layup, Reverse Layup, and Mid-range Jumpshooting, for 388 takes. We recruited 11 expert players from the university team with 10+ years of experience. To improve the participant skill diversity in the dataset, we also recruited novice players with less than 1 year of playing experience. The location of data collection was a university basketball gym.  

\paragraph{Music} For the music scenario, we asked all 9 participants to play 5 minutes of scales and arpeggios and 10 minutes of free play for 27 takes. All of our participants were recruited from the university music club and considered themselves as experts at playing their respective instruments. The instruments featured in our collected dataset were piano, trombone, trumpet, and saxophone. The Ego-Exo4D music guidelines called for just piano, violin, and guitar, but we found it necessary to expand this list in order to gather data for this domain. All data was recorded in a university music room.

\paragraph{Soccer} For soccer, we focused on three drills: dribbling, juggling, and penalty shots for 133 takes across 12 unique participants. 7 of these participants were experts with 10+ years of experience, whereas the remaining 5 participants were casual soccer players. All videos were collected at a university soccer field.

Our study protocol was approved by the Institutional Review Board (IRB). All participants signed a consent form before participating in the study.

%% file: sec/appendices/collection/upenn.tex
\begin{figure*}[h!]
    \centering
    \includegraphics[width=\textwidth]{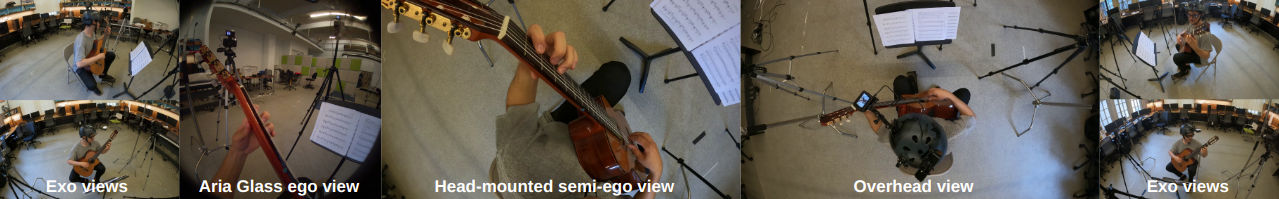}
    \caption{The Aria Glass ego view, head-mounted semi-ego view, overhead view and other static exo views in playing guitar in Philadelphia, PA, USA.}
    \label{fig:capture_illustration}
\end{figure*}

The University of Pennsylvania focused on capturing videos of experts of various levels playing musical instruments, dancing, and cooking. Over the spring and summer of 2023, UPenn captured 521 usable takes across 95 participants for the consortium's collections with up to 7 views.

One primary goal of this project is to capture detailed body movement, especially hands, across ego view and exo views.  We work to ensure highly engaged experts enjoy demonstrating their full skill capability.

The hand-object/instrument interaction region is the key to understanding human activities and evaluating their skills. 
Comprehensive hand pose information is especially important for the full analyses of scenarios collected at UPenn, especially the music scenario, where slight differences in finger motion result in entirely different performances. 

We also observed that experts had a tendency to not need to look at their hands during play. Thus, we found the initial data capture using the general camera setup to lack crucial visual information in such scenarios due to: \\
(1) (in ego view) limited field of view of Aria glasses, and skilled experts don't need to look at their hands, \\
(2) (in exo views) frequent occlusion and self-occlusion caused by participants' motion. 

We added two cameras to maximize the view coverage.

\noindent \textbf{Head-mounted Camera}:
The head-mounted camera on a helmet angled downwards to capture the hand/body region:
(1) (ego) it follows the subject's body motion faithfully, and (2) (exo) it is designed to focus on the hand-object interaction region with much less self-occlusion.  Empirically, we found this additional camera is crucial for capturing guitar, violin, and cooking scenarios.   

\noindent \textbf{Overhead Camera}: We replace the head-mounted camera with an overhead camera in 
(1) piano scenarios, where the overhead camera can have similar performance, and 
(2) dance scenarios, where the helmet can dramatically worsen the experience and performance of the participants. 

We believe the goal is not to maximize the number of hours captured but to have the participants show 
(1) diverse techniques to build models for the scenarios, and (2) unique techniques to demonstrate their skill levels.

To get the most representative recordings of the participants, we aim to maximize their engagement during the data capture. Specifically, we (1) walk through the whole process with the participants before the data capture to familiarize them with the setup (2) let them choose their favourite music piece to play or dance with in music and dance scenarios; and (3) have a narrate and act section for the musicians to demonstrate how they feel about their performance. 

\paragraph{Music }For musical instrument playing, classified as a “physical” activity, we captured takes of musicians (1) warming up (scales and or Etudes) (2) sight-reading simple sheet akin to Suzuki Practice books or Etudes exercises, and (3) freeplaying. We captured takes of violin, piano, and guitar, with a duet between a cello and a violin for one trial. Participants were recruited from a diverse pool of musicians, spanning the Penn Orchestra, local music schools, and independent music students. This pool's experience ranged from professional instructors and performers to complete newbies. We totaled 275 takes over 37 participants. Notably, during the shooting, we observed that participants were particularly uncomfortable with the helmet used to mount the GoPro; it interfered with their head movements and the bow sometimes ended up knocking against the mounted GoPro. To combat this, we added additional cushioning to depending on the subject's head shape and broke sessions into chunks to allow for breaks.

\paragraph{Cooking }Cooking, categorized as "procedural", consisted of preparing four dishes: an egg dish, a salad, a noodle dish, and a dessert. The group of participants consisted primarily of Penn students with experience ranging from amateurs to hobbyists. Professionals were unavailable due to scheduling conflicts. We totaled 81 takes over 20 participants. The entire filming process was undertaken within a three-week span, primarily at the apartment of one of the team's participants. This location expedited our data collection for this task by providing a stove and fridge for regular use.

\paragraph{Dancing }Dance captures, classified as a "physical", consist of four takes of dancers performing dance routines to a song. The dance types recorded included Lindy-Hop Jazz, Bollywood, Latin, and Chinese Folk Dance; across these genres, we totaled 165 takes over 38 participants. The Lindy-Hop Jazz dancers came from the Jazz Swing Attacks, a dance club in Philadelphia. Contact was established via Instagram, and data, collected weekly over a month. This group contained a balanced mix of experienced instructors and beginner dancers. The Bollywood dancers, the Drexel's Philly Maza, were recorded in the Drexel Engineering Building. They compete nationally but routinely train beginner recruits. The Chinese Folk Dancers were members of the local Great Wall Chinese School's dance club and independent student volunteers with prior competitive dance training. These were captured in the SIG Lab for collections.

All participants were confirmed to be at least 18 years of age by the time of participation and gave written consent for participating in these data collection trials. The consent form, in compliance with IRB guidelines, but gives participants the choice to back out. The collected information on basic demographic information should not be used to identify participants individually. All other data collected per participant (prior experience with task, average times spent per session, etc) could not be used to identify participants.

%% file: sec/appendices/collection/tokyo.tex
In Tokyo, we collected video data for three scenarios: cooking and health for procedural activities and soccer for physical activities. We followed the standard camera configuration and calibration procedure of the Ego-Exo4D dataset for all scenarios. In the following paragraphs, we will describe the specifics of each scenario, particularly the unique aspects of our data gathering.

\paragraph{Cooking} 
We recruited 12 Japanese participants living in the Tokyo area through a temporary employment agency. The gender and age of the participants were balanced to collect diverse behavior patterns. The participants cook three days or more each week in their daily lives. Each participant prepared three dishes: an omelet, a white radish \& lettuce \& tomato \& cucumber salad, and a sushi roll. We recorded both versions with and without narrations for each dish and participant. A one-page summary of each recipe was provided before data collection and was shown during video recording so that the participants could prepare the dishes smoothly, and the procedure of each recipe should be consistent between the participants. 

All video recording of the cooking scenario was done in a rental kitchen studio equipped with an island kitchen and all the necessary kitchenware for four consecutive days. The studio is situated in a busy location in downtown Tokyo, and some external noises, like ambulance sirens, were audible during the recordings. We collected 68 takes from 12 participants. Of the 68 takes, 34 takes are with narrations, and 34 takes are without narrations. The length of each participant’s making each dish twice with and without narrations is about 35 minutes, ranging from 24 min 27 sec (Sushi roll) to 55 min 4 sec (Salad). The length includes time for the camera synchronization procedure.

During the recording of the cooking scenario, we discovered a flickering issue in some of the video data due to the incompatibility between the Aria Glass sampling rate and the power frequency in Tokyo. To overcome this issue, we attempted to shoot as much as possible in daylight and adjusted the fps when using artificial lighting. While processing the videos, we discovered some exo videos had decoding errors due to damaged frames. Each corrupted video contained one to three damaged frames for an unknown reason. To address this, we re-encoded these videos by replacing the damaged frames preventing decoding with the nearest good frame. Note that some videos still contain damaged frames as long as those frames did not influence decoding. In addition, the original MP4 files recorded by GoPro contain 4 streams: video, audio, data0, and data1, but the re-encoded videos only contain a video stream and an audio stream.

\paragraph{Health} 
We recruited 17 Japanese participants living around Tokyo, Japan, through a temporary employment agency. The gender and age of the participants were balanced to collect diverse behavior patterns. We recorded videos of the 17 participants performing two tasks: COVID-19 rapid antigen testing using three test kits and performing CPR on a mannequin. We conducted all recordings in the same meeting room on campus 
over two days. For the COVID test, an instruction manual of each test kit was provided to the participants before and during the recording. Also, we did not show the participants or record any COVID test results for privacy protection. For CPR, the participants took an introductory lifesaving course provided by the Tokyo Fire Department before recording. Besides, a one-page summary of the CPR procedure was provided before the data collection and shown during the video recording. This is so that the participants can perform CPR smoothly and the procedure is consistent among the participants. We collected 73 takes from 17 participants. All of the CPR procedures (17 takes) were recorded without narration. For the COVID test, we recorded the videos of the 17 participants using the three test kits (51 takes) without narrations. The video length of each participant’s performing CPR is 9 min 48 sec on average, including the camera synchronization procedure. Similarly, the video length of each participant’s using the three COVID test kits is 29 min 22 sec on average. Additionally, we recorded extra takes of 5 participants out of the 17 using a test kit with narrations (5 takes in total).

\paragraph{Soccer} 
We gathered videos of 14 Japanese participants, each performing three fundamental soccer drills: dribbling and juggling for two minutes each and penalty kicks ten times. Of the 14 participants, 13 are soccer players from a university football club.
We recruited them through the staff of the club. The remaining one participant is not from the club but is an expert with over ten years of soccer experience. All the participants are male, and their age ranges from 18 to 30s. We recorded the videos on an outdoor soccer field at a local university %
over four days, with three to four participants participating each day. For juggling, we instructed the participants to include various movements such as juggling with thigh, inside and outside of feet, and alternating feet. For penalty kicks, we instructed them to shoot to the right side of the goal 5 times and to the left side five times. During penalty kicks, a helper aids the participant in retrieving the ball. This helper stands within the goal area and might be recorded by some cameras. We collected 42 takes from 14 participants. All the takes were recorded without narrations.

Our university's institutional review board
reviewed and approved our study protocol. We explained the objective and the range of use of the videos through documents and took consent from each participant before the recording. In particular, we took the consent not to blur their faces to keep the naturalness of the videos.

%% file: sec/appendices/A_people.tex
\clearpage
\section{~~Participants}\label{sec:appendix-people}

We provide self-declared information on ethnic groups by the participants. Sharing this information was optional for all research subjects. Ethnicity is reported based on location specific categories as defined by the relevant partner lab. No such information was gathered from research subjects participating in our collections in California, New York, and Pittsburgh, Pennsylvania.

\paragraph{Atlanta, Georgia, USA }100\% of participants that reside in Fulton County, Georgia self-reported their ethnic group membership as follows:

\begin{tabular}{lc}
\toprule
\textbf{Ethnicity} & {\textbf{Number of participants}} \\
\midrule
Asian   & 23
\\
White   & 8
\\
Hispanic/Latino & 3
\\
\bottomrule
\end{tabular}

\paragraph{Bloomington, Indiana, USA } 100\% of participants that reside in Monroe County, Indiana self-reported their ethnic group membership as follows:

\begin{tabular}{lc}
\toprule
\textbf{Ethnicity} & {\textbf{Number of participants}} \\
\midrule
Asian   & 22
\\
Black   & 1
\\
Middle Eastern & 1 
\\
White & 18
\\
Prefer not to say & 1
\\
\bottomrule
\end{tabular}

\paragraph{Minneapolis, Minnesota, USA} 100\% of participants that reside in Hennepin County, Minnesota self-reported their ethnic group membership as follows:

\begin{tabular}{lc}
\toprule
\textbf{Ethnicity} & {\textbf{Number of participants}} \\
\midrule
White   & 41
\\
Hispanic/Latino   & 4
\\
Asian & 8 
\\
Black & 1
\\
\bottomrule
\end{tabular}

\paragraph{Tokyo, Japan} 100\% of participants that reside in Tokyo self-reported their ethnic group membership as follows:

\begin{tabular}{lc}
\toprule
\textbf{Ethnicity} & {\textbf{Number of participants}} \\
\midrule
Asian (Japanese) & 45
\\
\bottomrule
\end{tabular}

\paragraph{Hyderabad, India} 100\% of participants that reside in Hyderabad self-reported their ethnic group membership as follows:

\begin{tabular}{lc}
\toprule
\textbf{Ethnicity} & {\textbf{Number of participants}} \\
\midrule
Asian (Indian) & 95 \\
\bottomrule
\end{tabular}

\paragraph{Chapel Hill, North Carolina, USA} 100\% of participants that reside in Orange County, North Carolina self-reported their ethnic group membership as follows:

\begin{tabular}{lc}
\toprule
\textbf{Ethnicity} & {\textbf{Number of participants}} \\
\midrule
White & 20
\\
Indian & 1
\\
Asian & 13
\\
African American & 9 
\\
Hispanic/Latino & 3
\\
Prefer not to say & 3
\\
\bottomrule
\end{tabular}

\paragraph{Vancouver, British Columbia, Canada} 100\% of participants that reside in Vancouver self-reported their ethnic group membership as follows. Please note that research subjects in this case opted not to use any assigned category and independently described their identity.  

\begin{tabular}{lc}
\toprule
\textbf{Ethnicity} & {\textbf{Number of participants}} \\
\midrule
African/Nigerian & 4
\\
Asian & 9
\\
White/Caucasian & 10
\\
Chinese & 26 
\\
European & 1
\\
Iranian/Persian & 14
\\
Italian & 1 
\\
Jamaican & 2
\\
Kazakh & 1 
\\
Kyrgyz & 2
\\
Middle Eastern & 1
\\
Mixed & 3
\\
South Asian & 2
\\
\bottomrule
\end{tabular}

\paragraph{Philadelphia, Pennsylvania, USA} 100\% of participants that reside in Philadelphia Country, Pennsylvania self-reported their ethnic group membership as follows:

\begin{tabular}{lc}
\toprule
\textbf{Ethnicity} & {\textbf{Number of participants}} \\
\midrule
White/Caucasian & 10
\\
Asian & 30
\\
African American & 3
\\
Hispanic/Latino & 4 
\\
Prefer not to say & 43
\\
\bottomrule
\end{tabular}

\paragraph{Singapore} 100\% of participants that reside in Singapore self-reported their ethnic group membership as follows:

\begin{tabular}{lc}
\toprule
\textbf{Ethnicity} & {\textbf{Number of participants}} \\
\midrule
Chinese & 65
\\
Indian & 3
\\
Singaporean & 2
\\
Indian/Chinese & 2 
\\
Prefer not to say & 17
\\
\bottomrule
\end{tabular}

\paragraph{Bogota, Colombia} 100\% of participants that reside in Singapore self-reported their ethnic group membership as follows:

\begin{tabular}{lc}
\toprule
\textbf{Ethnicity} & {\textbf{Number of participants}} \\
\midrule
Black/&\\
Afro-descendant/&\\
Afro-Colombian & 7
\\
Mixed & 104
\\
Palenquero & 1
\\
Raizal & 1 
\\
White/Caucasian & 38
\\
Prefer not to say & 23
\\
\bottomrule
\end{tabular}

Participant surveys were separated into two: a pre-task questionnaire and a post-task questionnaire.
The pre-task questionnaire aims to capture the participant's perceived skill level whereas the post-task questionnaire captures the participant's reflection on how well the task went.
The list of questions for both questionnaires can be found in Table~\ref{tab:questionnaire_qs} with questions/answers designed for consistency and ease of filling in, as participants would be filling these out before/after each recording.
This involved using multiple choice and Yes/No answers with open text fields being utilized sparingly.

\begin{table*}[t]
    \centering
    \resizebox{\linewidth}{!}{
    \begin{tabular}{lll}
    \toprule
    & Question & Answer Type \\ \midrule
    \multirow{10}{*}{\rotatebox[origin=c]{90}{\textbf{Pre-Task}}} & Recording Location    & multiple choice \\
    & How many times do you estimate you have done this task? & multiple choice \\
    & How often do you carry out this task? & multiple choice\\
    & How many years have you been doing this task? & multiple choice\\
    & Have you taught this activity to others before? & Yes/No \\
    & Have you recorded a video of yourself carrying out or explaining this task before? & Yes/No \\
    & Have you watched videos of others doing this task before? & Yes/No \\
    & Do you have any qualifications/professional training that are related to the task? & Yes/No \\
    & How long does it typically take you to complete this task?$^*$ & text \\
    & How long would you typically spend in one practice session of this task?$^\dagger$ & text \\ \midrule
    \multirow{8}{*}{\rotatebox[origin=c]{90}{\textbf{Post-Task}}} & Self Reported Quality & multiple choice \\ 
    & Completed Route?$^\ddag$ & Yes/No \\
    & What mistakes/errors did you make during this task? & text \\
    & Any issues with the familiarity of the tools/location? & text \\
    & Did it take longer/shorter than your initial expectation and why? & text \\
    & How did you find wearing the camera? & multiple choice \\
    & How easy was the setup for recording? & multiple choice \\
    & Any other comments to take on board? & text \\ \bottomrule
    \end{tabular}
    }
    \caption{Questions for the pre-task and post-task questionnaires. *: Only applicable for non-dance/non-music scenarios. $\dagger$: Only applicable for Dance/Music scenarios. $\ddag$: Bouldering scenario only.}
    \label{tab:questionnaire_qs}
\end{table*}

%% file: sec/appendices/A_language.tex
\clearpage
\section{Language Descriptions}\label{sec:appendix-language}

\begin{table*}[t]
\begin{center}\footnotesize
\hspace*{-0.25in}
\begin{tabularx}{1.1\textwidth}{|l|s|s|X|}
\hline
\textbf{Domain} & \begin{tabular}[c]{@{}l@{}}\NEW{\textbf{Atomic Action}} \\ \NEW{\textbf{Description}}\end{tabular} & \NEW{\textbf{Narrate and Act}} & \textbf{Example commentary} \\
\hline
\NEW{Cooking} & \NEW{C turns on heat on the gas burner.} & \NEW{So I'm going to start out by boiling some water.} & \NEW{Here the preparer is checking the pasta for done-ness. It's important to do this and not rely on what a package says. Use a package that gives you cooking time as a guideline and start to check your pasta, you know, a few minutes before the maximum amount of time given for cooking that specific pasta.} \\ \hline
Health & \NEW{C inserts the nasal swab in the buffer test tube on the covid test kit pack with his right hand.} & \NEW{Open the newly picked up tube, place the swab in the tube, stirring the swab in the tube.} & So this individual has done a great job of making sure that her nasal passageways have adequate time in contact with her nasal swab. Something that might make it a little bit easier for her is if she could tilt her head back just a bit so that she wouldn't have to strain quite so much to get that access. Additionally, she did a great job making sure that the nasal swab was about an inch into her nose. \\ \hline
Bike repair & \NEW{C holds the bike wheel with her left hand.} & \NEW{And then I will locate the location of the valve cap and pull the tube out of the wheel.} & It's a great method to always double check or do a pre-check before beginning work on a bicycle to make sure the issue that you are working to fix is the only issue that is occurring. If not, you could find a secondary issue or something else that may be greater than the one you are currently working on. \\ \hline
\NEW{Music} & \NEW{C puts the bow on the violin with his right hand.} & \NEW{So regardless of how tricky left hand passage work is you want to always keep your bow completely independent.} & \NEW{This is a really great use of the bow and decision to play in this middle third of the bow. This is exactly where they should be playing. And we can hear that the note envelope is very consistent and that it's very controlled and that it also allows the rhythm to be stable...} \\ \hline
\NEW{Basketball} & \NEW{C runs towards the hoop with the basketball.} & \NEW{Now I'm going to do a reverse layup, stepping right, left, going up with the right hand.} & \NEW{As the ball goes through the basket, she catches the ball and does an excellent job of keeping the ball high, never allowing the ball to drop down to her waist area, but keeping the ball high in her upper chest, neck area throughout the drill...} \\ \hline
Bouldering & \NEW{C places both hands on a red hand hold.} & \NEW{So I know that a lot of these holds, I'm going to need my weight leaning to the left to utilize} & Once the climber recovered from the foot cut, the climber pasted the right foot on this foot jib and then did a toe match. So brought this foot in and then dropped the right foot down and to the right to again counterbalance so that the climber can then move their left hand out left. But at this point the climber is just a little too gassed to be able to make this move, which is unfortunate. \\ \hline
\NEW{Soccer} & \NEW{C kicks the ball to the right with his right foot.} & & \NEW{Angle approach, start position is good, maybe slightly squarer than 45, but again because the intended outcome from previous actions is into the left, by being a little squarer is going to help him be able to rotate his hips to move to the left, but on a slight angle is good and help him with his technical action.} \\ \hline
\NEW{Dance} & \NEW{C moves her right leg forward while swinging both hands.} & \NEW{Ring and wing, one, two, one, two, three} & \NEW{She is doing these steps in place when she's traveling forward. At this point, she really could be further forward all the way, still on the screen, but towards the edge of the screen, if she was to take bigger steps. And she could take bigger steps if she bends her knees and lowers her center of gravity and then extends her leg outward...} \\ \hline
\end{tabularx}
\end{center}
\caption{Example excerpts from all three language types. 
 Experts are charged with critiquing the performance of the participants, pointing out strengths and weaknesses and explaining how the participant's approach influences the quality of their skill demonstration.  Narrate and act focuses more on what the camera wearer is doing and, sometimes, briefly why.  Atomic action descriptions are about the specific actions seen.}\label{tab:example-commentary}
\end{table*}

\begin{figure*}[t]
    \centering
    \includegraphics[width=.85\textwidth]{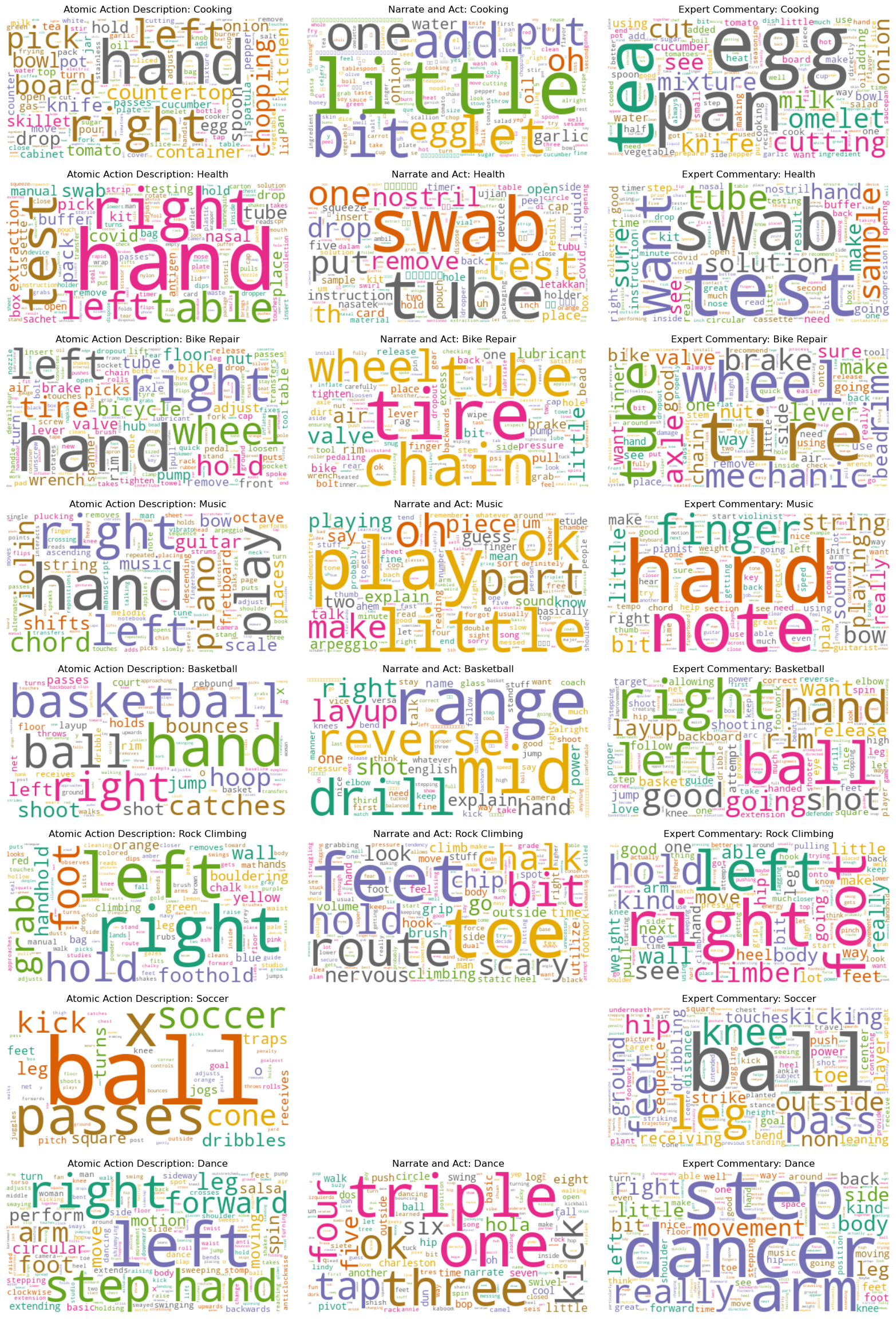}
    \caption{\NEW{Word clouds for each scenario and annotation type. The vocabulary for atomic action descriptions typically focuses on the person's hands and how they complete the actions (e.g. using left/right/hand) whereas narrate and act describe the high level goals/objects. The expert commentary has the largest variety of words, including specialist words for each scenario such as swab/solution for health and axle/valve for bike repair.}}
   \label{fig:appendix:annotations:wordcloud:fig}
\end{figure*}

As introduced in the main paper, Ego-Exo4D provides three forms of parallel text corpora for the video: expert commentary, narrate and act, and atomic action descriptions. 
\NEW{Table~\ref{tab:example-commentary} shows examples from different scenarios highlighting their distinctions in style and point of view.}  
 \NEW{Figure~\ref{fig:appendix:annotations:wordcloud:fig} shows word clouds per scenario and annotation type highlighting the differences in vocabulary and word frequency.}

\begin{figure*}[t]
    \centering
    \includegraphics[width=\textwidth]{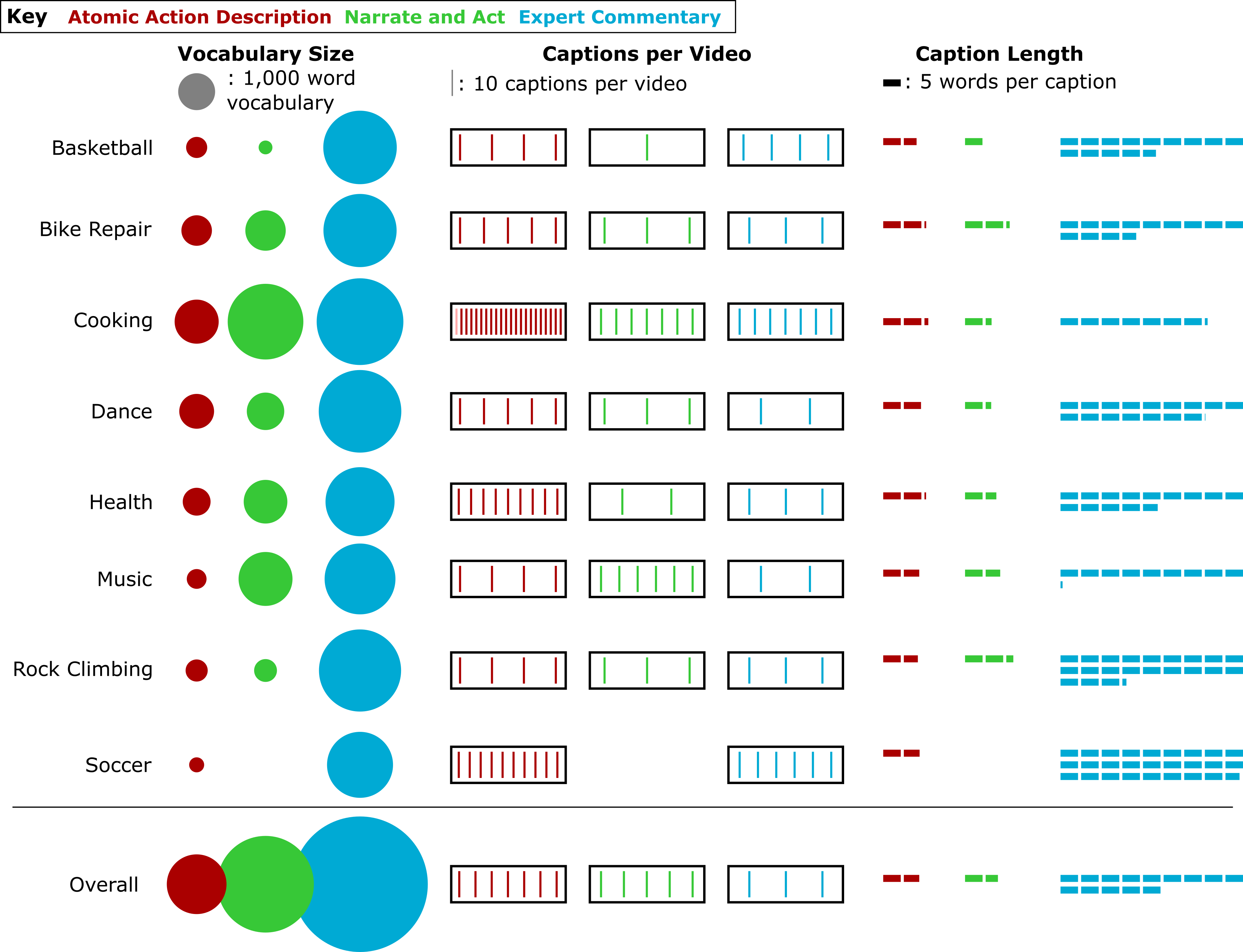}
    \caption{Comparisons between the vocabulary size (left) number of captions per video (center) and length of caption (right) for the atomic action description, narrate and act, and expert commentaries. Statistics are shown both per scenario and over the entire dataset.  We see that the expert commentary tends to use a much larger vocabulary and more lengthy statements, since commentators are giving more elaborate statements of advice and explanation.  The temporal density of the atomic action descriptions is greater than the other two forms, since the annotators are pausing to describe every single action of the camera wearer.  Narrate-and-act comments use a vocabulary size in between the other two, reflecting the more free-form speech (compared to the written atomic actions) is used.  Trends are mostly similar across scenarios, with the most noticeable differences being the temporal density; it is particularly high for both cooking and soccer.  In the former, there are many procedural steps, whereas in the latter there are many instances of the drill being executed.}
    \label{fig:appendix:annotations:statistics:fig}
\end{figure*}

\NEW{In Figure~\ref{fig:appendix:annotations:statistics:fig} we further emphasize the characteristics of each text corpus across three axes: total vocabulary size, average number of captions per video, and caption length. 
 See caption for details.}

\subsection{Expert Commentary Tool}\label{sec:appendix-commentary}

\begin{figure*}[tp]
\includegraphics[width=\linewidth]{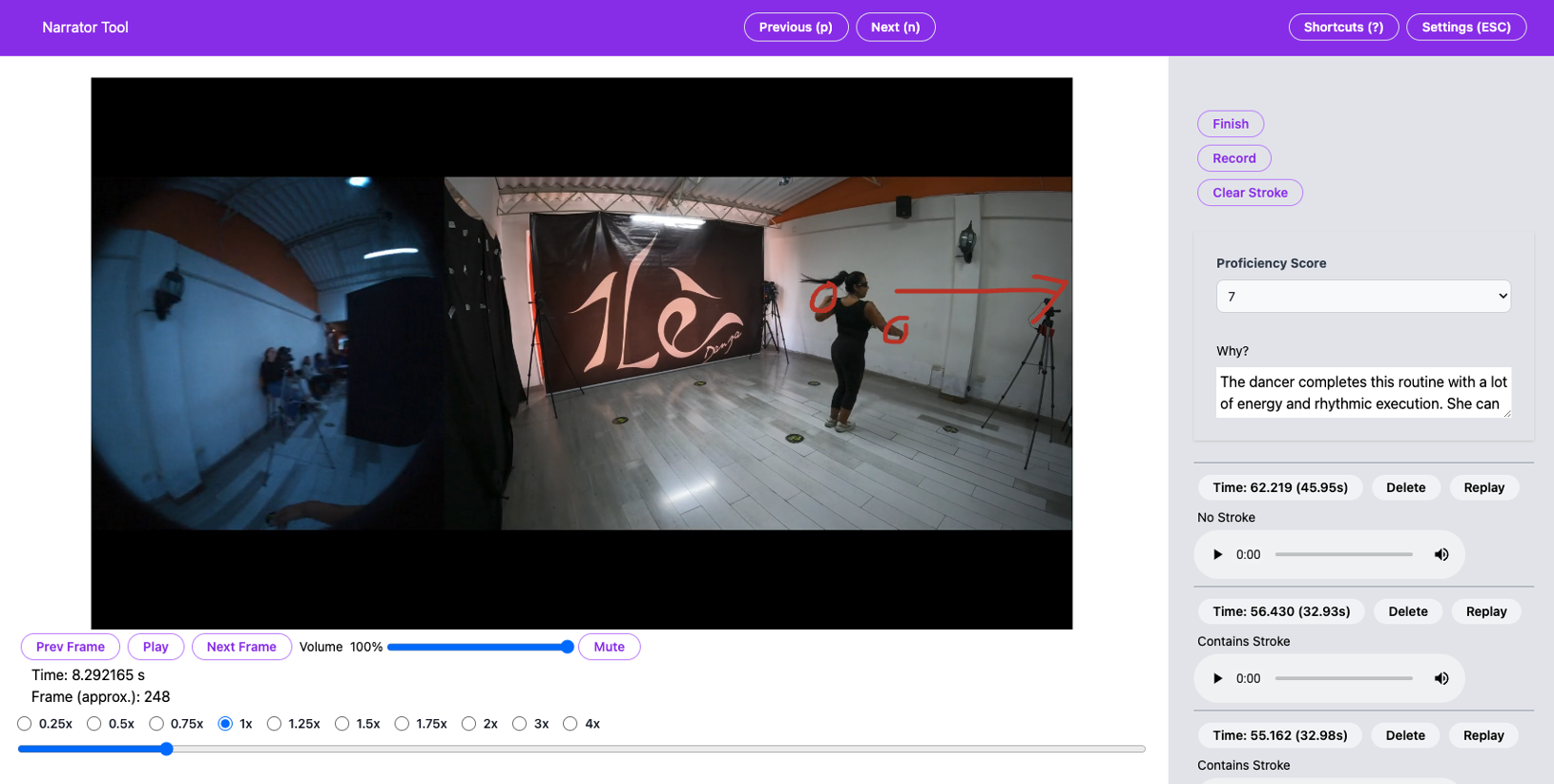}
\caption{Our expert commentary web tool called \emph{Narrator} provides an easy-to-use platform for experts. Experts can stream video, record audio commentaries, and provide proficiency ratings and justifications. The tool also supports drawing on the video feed (see red arrow and circles on the right frame), allowing for manual spatial grounding during commentary.
}
\label{fig:expert_narrator}
\end{figure*}

To collect expert commentary, we developed a web-based tool, which is open sourced as part of the Ego-Exo4D dataset and benchmark suite.   See Figure~\ref{fig:expert_narrator}.
Known as the Narrator, this application supports video playback for Ego-Exo4D skills demonstrations, records time-stamped verbal commentary, and allows exporting and viewing commented videos.  As a web-based platform, the Narrator can be simply accessed through a browser, with minimal set-up and less restrictive system requirements compared to tools requiring local installation.
These attributes made it efficient
to onboard and manage our geographically distributed experts.
We acknowledge the EPIC Narrator~\citep{Damen2018EPICKITCHENS} as the open-sourced inspiration and source code for this initiative.

\subsection{Atomic Action Description Statistics}\label{sec:appendix-shortnarrations}

See Table~\ref{tab:atomic_stats} for atomic action descriptions summary statistics.

\begin{table*}[]
\footnotesize
\centering
\begin{tabular}{ccccccc}
\toprule
Category & 1x Coverage & 2x Coverage & \# of Descriptions & Descriptions Per Minute & Unique Nouns & Unique Verbs\\[0pt]
\hline
Basketball & 778 & 116 & 50299 & 53.330 (+- 26.049) & 201 & 134 \\
Bike Repair & 202 & 160 & 31317 & 24.891 (+- 9.555) & 642 & 393 \\
Cooking & 360 & 266 & 189225 & 27.745 (+- 12.843) & 1744 & 823 \\
Dance & 307 & 417 & 43663 & 30.852 (+- 13.915) & 504 & 468 \\
Health & 299 & 97 & 43769 & 24.304 (+- 11.234) & 619 & 384 \\
Music & 85 & 75 & 10695 & 4.278 (+- 8.969) & 255 & 163 \\
Rock Climbing & 1270 & 103 & 32246 & 32.350 (+- 11.974) & 301 & 224 \\
Soccer & 225 & 53 & 31253 & 38.467 (+- 23.957) & 229 & 125 \\

\hline
All & 3526 & 1287 & 432467 & 31.293 (+- 20.209) & 2924 & 1481 \\
\bottomrule
\end{tabular}
\caption{\NEW{Atomic action descriptions per-domain statistics.} }
\label{tab:atomic_stats}
\end{table*}

%% file: sec/appendices/A_benchmarks.tex
\clearpage
\section{Benchmarks: annotations and baselines}\label{sec:appendix-benchmarks}

We employed a unified dataset split for EgoExo4D applicable for all benchmarks. Splits were defined at the ``take" level, i.e. each take was allocated to either the training, validation, or testing set. Any derivatives from individual takes (e.g. segment clips) simply inherit the split assignment from the original take.

Additionally, the splits were stratified according to activity types and proficiency scores. This was done to ensure a balanced distribution of data across the training, validation, and testing sets, with proportional representation of activities and proficiency levels.  To prevent data leakage, participants and all their associated artifacts were assigned exclusively to one of the splits. This constraint was particularly important because several participants contributed multiple takes, and it was crucial to avoid sharing their data between the splits. In the end we divided a total of 5045 takes into 3082 training, 842 validation and 1121 testing takes.

\subsection{Relation}
\label{sec:appendix:relation}
\input{sec/appendices/annotation/correspondence-annotation}

\subsubsection{Ego-exo correspondence}
\label{sec:appendix:correspondence}

\input{sec/appendices/benchmarks-appendix/correspondence-baseline}

\subsubsection{Ego-exo translation}
\label{sec:appendix:translation}

\input{sec/appendices/benchmarks-appendix/translation-baseline}

\subsection{Fine-grained keystep recognition}
\label{sec:appendix:keystep_recognition}
\input{sec/appendices/benchmarks-appendix/keysteps-baseline}

\subsection{Energy-efficient multimodal keystep recognition}
\label{sec:appendix:energy_efficient}
\input{sec/appendices/benchmarks-appendix/multimodal-baseline}

\subsection{Procedure Understanding}
\label{sec:appendix:proc_understanding}

\input{sec/appendices/benchmarks-appendix/taskgraph-baseline}

\subsection{Proficiency estimation}
\label{sec:appendix:proficiency_estimation}
\input{sec/appendices/annotation/proficiency-annotation}

\input{sec/appendices/benchmarks-appendix/proficiency-baseline}

\subsection{Ego Pose}
\label{sec:appendix:ego_pose}
\input{sec/appendices/benchmarks-appendix/bodypose-baseline}

\input{sec/appendices/benchmarks-appendix/handpose-baseline}

%% file: sec/appendices/annotation/correspondence-annotation.tex
\paragraph{Annotations}

\begin{figure*}[tp]
\includegraphics[width=\linewidth]{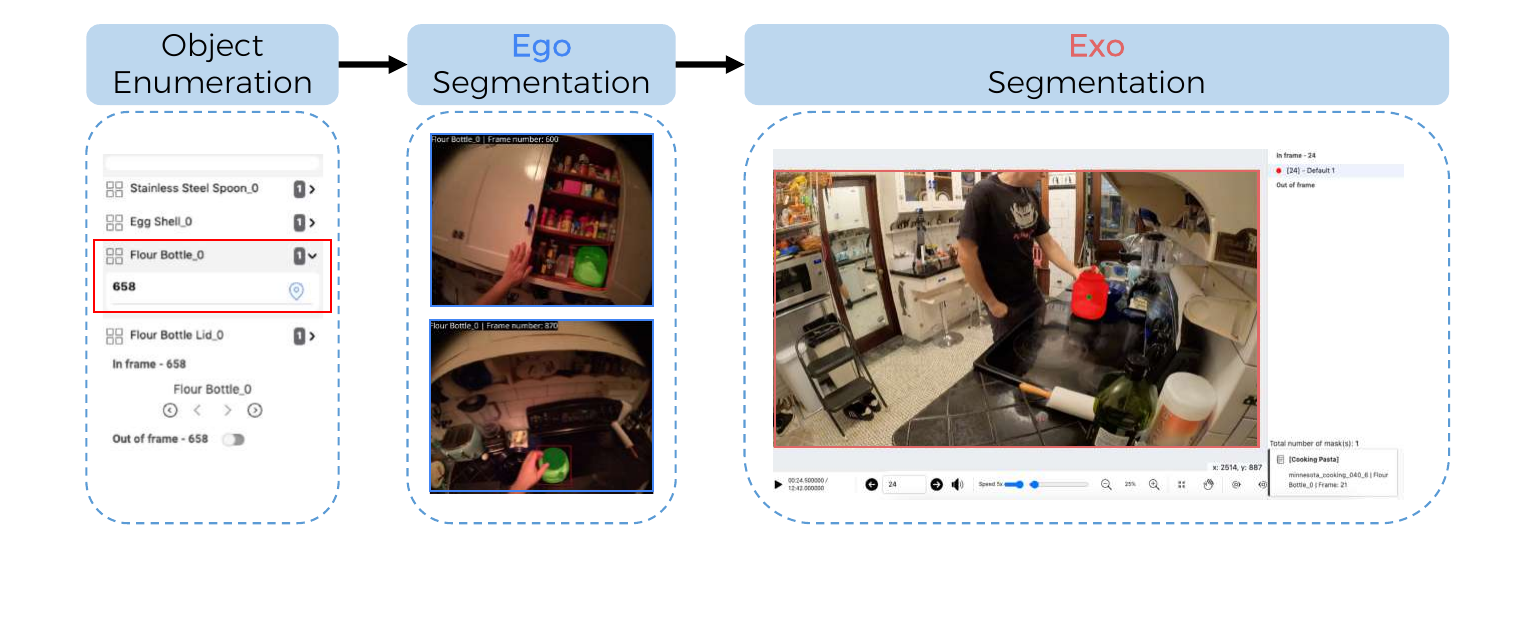}
\caption{
\EM{\textbf{Multi-stage annotation process for Ego-Exo Relation annotations.} After enumerating all active objects in the egocentric video, an object is selected and annotated with segmentation masks in all frames of the egocentric video. Then, annotators are given the exo video as well as the textual descriptions and sample egocentric segmentation masks for the object of the interest, and mark segmentation masks for the specified object of interest in all the frames where it is visible.}
}
\label{fig:relation_stage_12_ui}
\end{figure*}

We used a multi-stage annotation process for annotating paired ego-exo videos: 

\begin{itemize}
    \item {\em Stage 0: Object Enumeration}. %
    Annotator marks each \EM{object that is active at some point of the egocentric video}
    with a bounding box in a frame where it is clearly visible and provides a free-form textual description. 

    \item {\em Stage 1: Egocentric video annotation}. Annotator watches the egocentric video and is also shown (a) text and (b) a bounding box for one of the objects annotated in the previous stage. Annotator then marks a segmentation mask for that object in all the video frames where the object is visible. Segment Anything \citep{Kirillov_2023_ICCV} is leveraged to generate segmentation masks efficiently using only point clicks. 

    \item {\em Stage 2: Exocentric video annotation}. As shown in Figure~\ref{fig:relation_stage_12_ui}, the annotator watches a temporally synchronized exocentric video
    and is also provided with the (a) text and (b) several ego segmentation masks of this object. Annotator then marks a segmentation mask for this object in all the exo video frames, whenever the object is visible. \\
\end{itemize}

\begin{table}[tp]
\centering
\resizebox{\linewidth}{!}{%
\begin{tabular}{lcccc}
\toprule
 Scenario &  \# Takes &  \# Objects &  \# Ego Masks &  \# Exo Masks \\
\midrule
 basketball &        394 &          602 &           21820 &           31165 \\
       bike &        210 &          714 &           53886 &           71763 \\
    cooking &        478 &         3481 &          549507 &          888384 \\
     health &        127 &          570 &           77596 &           86585 \\
      music &        112 &          153 &           33624 &            5599 \\
     soccer &         12 &           22 &            2411 &            2475 \\
\midrule
      Total &       1335 &         5566 &          741965 &         1091135 \\
\bottomrule
\end{tabular}
}
\caption{{\bf Relation annotation statistics.} We show statistics for each scenario including the number of takes, total number of objects annotated and the number of egocentric and exocentric segmentation masks. %
}
\label{tab:relation_annotation_stats}
\end{table}

\noindent \EM{{\em What are the objects of interest?}
We focus on objects that are \emph{active} at some point during the execution of the activity. These objects are not only interesting because they are essential to the activity, but they are also challenging to track, since they are moving/changing state}. In particular,
our annotation guidelines requested annotators to  list (a) objects that the camera-wearer interacts with through their body or tools; 
(b) \EM{other} objects that are
\EM{relevant} to the activity (e.g., supporting surfaces like kitchen top); and (c) \EM{body parts} (hands and legs). 
\EM{Note that every time an object changes visual state (adopting the Point-of-No-Return definition from~\citep{grauman2022ego4d}), it is marked as a new object (e.g., annotators list \emph{tomato} and \emph{sliced tomato} as two distinct object instances)}.\\

\noindent {\em Which objects to annotate with masks?}
For scenarios that involve few objects (Music, Basketball and Soccer), 
we annotated all object instances. 
Instead, for Cooking, Health and Bike Repair we \EM{sampled object instances based on their frequency of occurrence and their size}, \EM{due to time and budget constraints}. In particular, we binned each \EM{object annotated in the Object Enumeration stage}
into bins based on their frequency of occurrence across the dataset (high, low) 
and object size (small, large). 
 We then uniformly sampled object instances from these bins while accounting for annotation time and budget\efi{Does this mean some extra constraint that modifies uniform random sampling?} and proceeded with segmentation mask annotations.
We ignored all objects with area $< 150$ pixels. For Cooking, specifically, we also filtered out a few objects such as spices, \EM{mixtures} and liquids, as they tend to be too small to match in the exo view. Finally, we skipped \EM{exocentric mask annotations} for objects that were visible in \EM{fewer than 10 frames of the egocentric video.}\\
   
\noindent {\em What frame rate to annotate at?} %
We annotated segmentation masks at 1 frame per second, except for videos from the \emph{Music scenario} which we annotated at 0.1 fps due to extremely long video durations. \\

\KGCR{In total, our annotation process yielded segmentation masks for 5,566 objects in 1,335 ego-exo video-pairs. Approximately 4M million frames were annotated resulting in a total of 742K ego and 1.1M exo  paired segmentation masks. Apart from this we also annotated 367K ego only segmentation masks. Collectively this results in a total of 2.2M segmentation masks. Table~\ref{tab:relation_annotation_stats} shows a detailed breakdown per scenario for the paired masks.}

%% file: sec/appendices/benchmarks-appendix/correspondence-baseline.tex
\paragraph{Correspondence Baseline Implementation Details}

\noindent \textit{Spatial baseline model.}  To adapt the architecture of SegSwap~\citep{shen2022learning} for our correspondence problem, we additionally condition the model on the segmentation mask of the object of interest by feeding the query mask as a third input to the model. In particular, we first pass the egocentric frame, the exocentric frame and the query mask (as a binary mask) through a visual backbone network. We then flatten the resulting features into three sequences and pass them through the cross-image transformer with alternating self-attention and cross-attention layers. We first use the query mask features to attend to the features in the query view which are then used to cross-attend over features from the target view. This allows the model to reason over features from both views conditioned on the input mask. The resulting sequences for both views are ``unflattened'' and passed through a decoder
to predict object segmentation masks in both views. We also pass the target view features through a classification head to classify if the query object is visible in the target view.

We train the model to perform mask prediction using a point-wise binary cross-entropy loss and a dice loss over the predicted and ground truth masks. We use only pairs of frames where the object of interest is visible on both views and apply the losses on predicted masks in both the views. %
During inference, we only consider the mask predicted in the target view and discard the predicted mask in the query view. %
We train the head performing Visibility classification using a binary cross-entropy loss on all the frames of the sequence. \\

\begin{figure}[t]
  \centering
  \begin{minipage}[b]{\linewidth}
    \centering
    \includegraphics[width=1.15\textwidth]{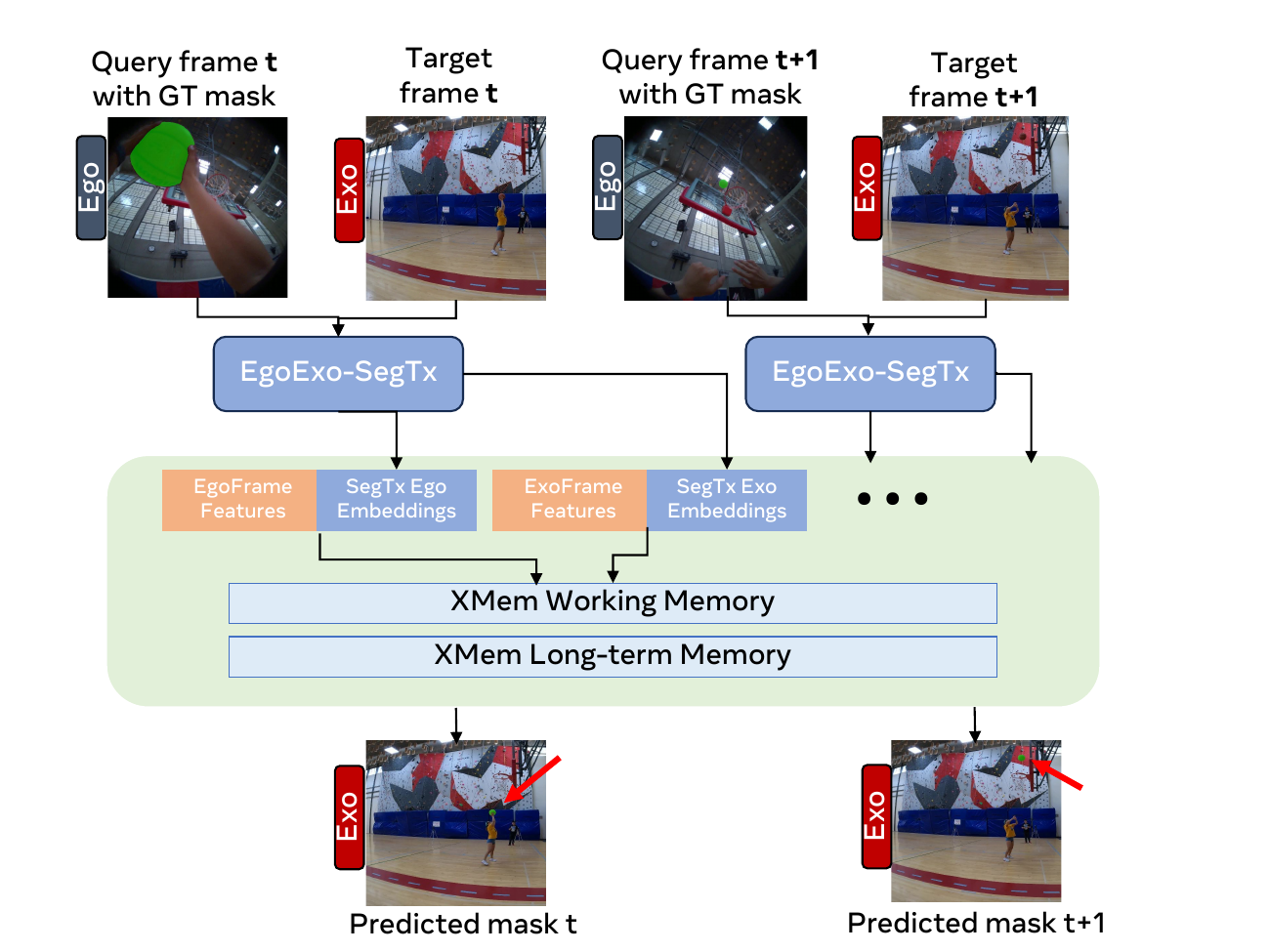}
    \caption{Overview of our spatio-temporal XView-XMem baseline model for the correspondence task. %
    }
    \label{fig:spatiotemporal_baseline}
  \end{minipage}
\end{figure}

\noindent\textit{Spatio-temporal baseline model.}
To encourage the model to learn associations of the objects between egocentric and exocentric views, we train XView-XMem to track the object in a sequence of interleaved frames of egocentric and exocentric views, i.e., each egocentric frame is followed by an exocentric frame and vice versa, as shown in Figure~\ref{fig:spatiotemporal_baseline}.

To mitigate track drift (within and across views), we also explore feeding the %
XSegTx embeddings to the XMem working memory. Since these embeddings are trained to guide the mask decoder at each frame independently, they capture rich information about the object of interest. 
The extracted image features from the ResNet in XMem are fused with the encoded embeddings from multiple layers of SA (self-attention) and CA (cross-attention) layers of XSegTx.
They are then projected into keys and stored in memory for tracking. 

For our spatial baseline model, we downsample the images to 480x480 resolution for all the views while using padding to keep the original aspect ratio of the images. For the image backbone we use the same ResNet50~\citep{he2016deep} checkpoint as SegSwap and freeze its weights during training. Our cross-image transformer architecture also follows~\citep{shen2022learning}. We use a batch size of $32$ and Adam~\citep{kingma2014adam} as our optimizer with a learning rate of $0.0002$ which decays to $0.0001$ after 50,000 iterations. We run all our experiments on a single Nvidia RTX A6000 GPU for 200,000 iterations. %

For our spatio-temporal baseline model, we use the same visual backbone (ResNet50~\citep{he2016deep}) and architecture as XMem~\citep{cheng2022XMem}. Our only modification is in the information that gets inserted in the working memory at each frame.  We first extract features from both ResNet and XSegTx for both both query and target frames. The corresponding features are then concatenated and projected to the original feature dimension through simple 2D convolution. We train on sequences of 8 interleaved ego and exo frames.
The model is trained using AdamW as our optimizer with a learning rate of $0.00001$ for 50,000 iterations and weight decay 0.05.  The batch size is 8 clip pairs. We initialize our model with the original pretrained XMem, and keep both the ResNet backbone as well as our finetuned XSegTx models frozen. 
Note that we do not apply any data augmentations.

\paragraph{Data}
We use \NEWRESULT{1028} takes from the Ego-Exo dataset to train and evaluate models for this benchmark. 
In particular, we use the common split shared across benchmarks, with \arxivb{838} takes for training, \arxivb{201} takes for validation and \arxivb{295} takes for testing.
We extract pairs of images between egocentric and exocentric views which have corresponding object masks annotated for training. This gives us a total of about \NEWRESULT{193k} pairs for training.

\paragraph{Results}

\begin{figure}[ht]
    \centering
    \includegraphics[width=\linewidth]{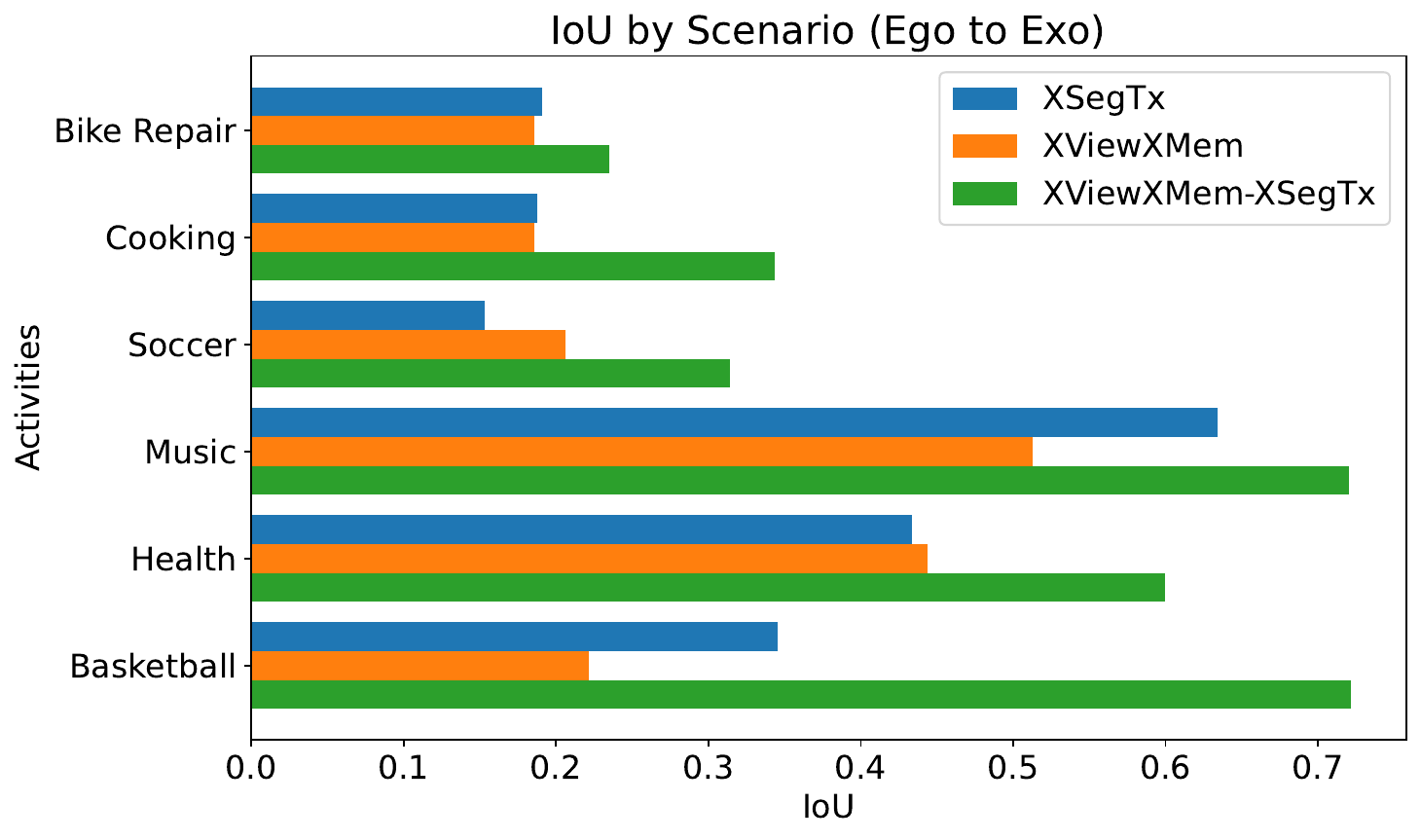} 
    \caption{Performance of both baselines per activity scenario.}
    \label{fig:iou_act}
\end{figure}

We break down our results across different activities in Fig.~\ref{fig:iou_act}. We note that some activities are generally easier to model (e.g., basketball, soccer) because of limited variation in object shape and appearance whereas some activities (e.g., cooking and bike repair) are much harder to model due to larger diversity in appearance, shape and size of the objects across views. We also explicitly evaluate our baselines on their ability to predict masks for very small objects. To do so, we split our validation set based on the predicted object size in proportion to pixels in the image. We see that, all our baselines struggle on very small objects and perform increasingly well on larger object sizes.

\begin{figure}[ht]
    \centering
    \includegraphics[width=\linewidth]{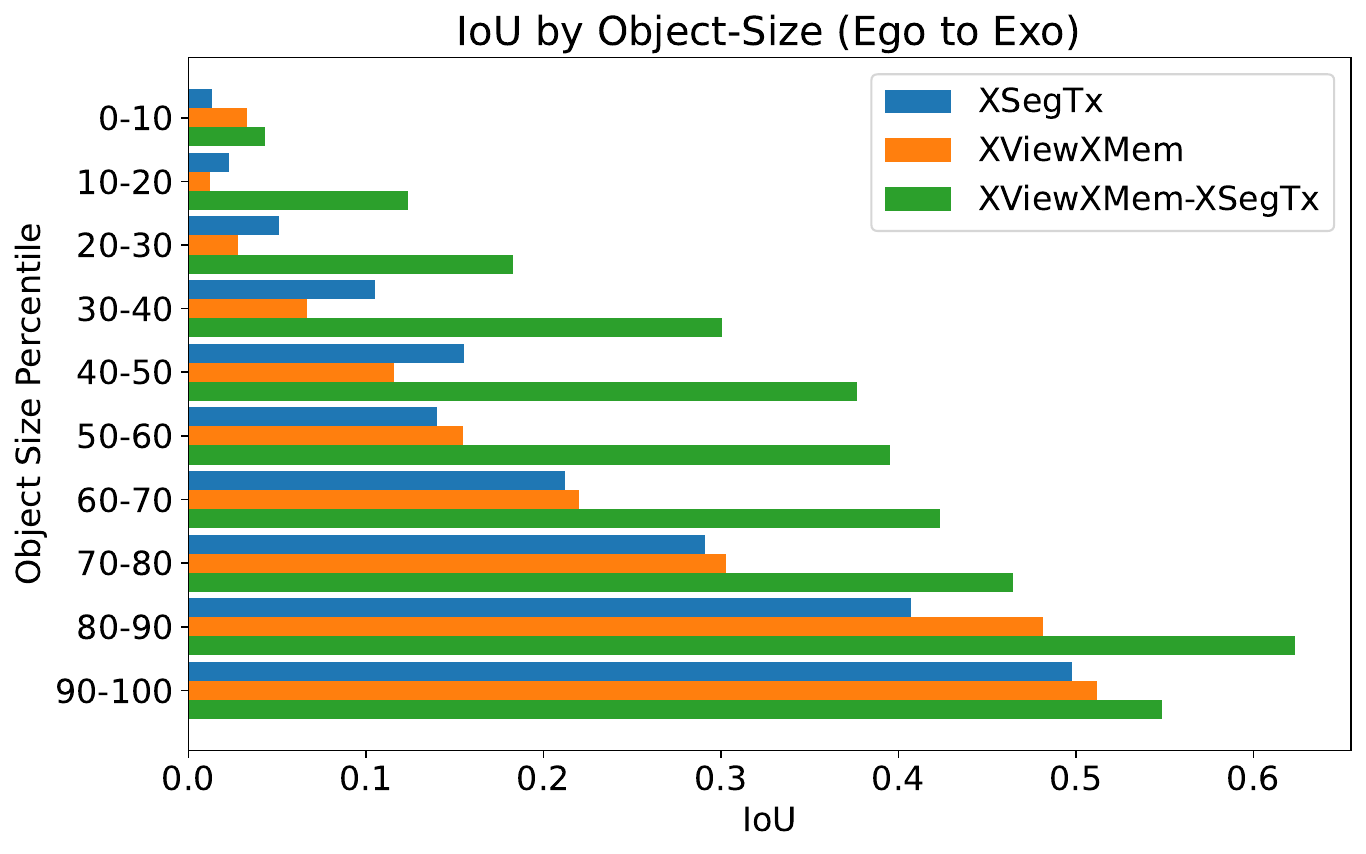} 
    \caption{Correspondence evaluated across different object sizes in the target (exo) view. The object sizes range from $7e^{-6}\%$ to $11\%$ pixels in the target view.}
    \label{fig:iou_objsize}
\end{figure}

%% file: sec/appendices/benchmarks-appendix/translation-baseline.tex
{
In Table~\ref{tab:translation:res:scenario-breakdown} we provide a break down of ego-exo translation results across different scenarios{, using GNT-mask for track prediction and DiT-pix for clip generation}. We can observe similar trends for the two subtasks: the methods achieve better results in basketball and soccer scenarios than in  bike and cook scenarios, which is reasonable as the objects in bike and cook scenarios are more complex and diverse.
}

\begin{table}
\small
  \centering
  \begin{tabular}{lcc}
\toprule
 Scenario & IoU (\%) $\uparrow$& LPIPS $\downarrow$ \\
\hline
Basketball& 14.5 & 0.41\\
Soccer & 17.0 & 0.51\\
Music&4.5& {0.30} \\
Health &12.9& 0.40\\
Bike &6.4& 0.52 \\
Cook &9.5& 0.47 \\
\hline
  \end{tabular}
  \caption{{Breakdown of ego-exo translation results per scenario for the subtasks of ego track prediction (IoU) and ego clip generation (LPIPS).}}
  \label{tab:translation:res:scenario-breakdown}  
\end{table}

%% file: sec/appendices/benchmarks-appendix/keysteps-baseline.tex
\paragraph{Data annotation details}

\begin{figure*}[tp]
\includegraphics[width=\linewidth]{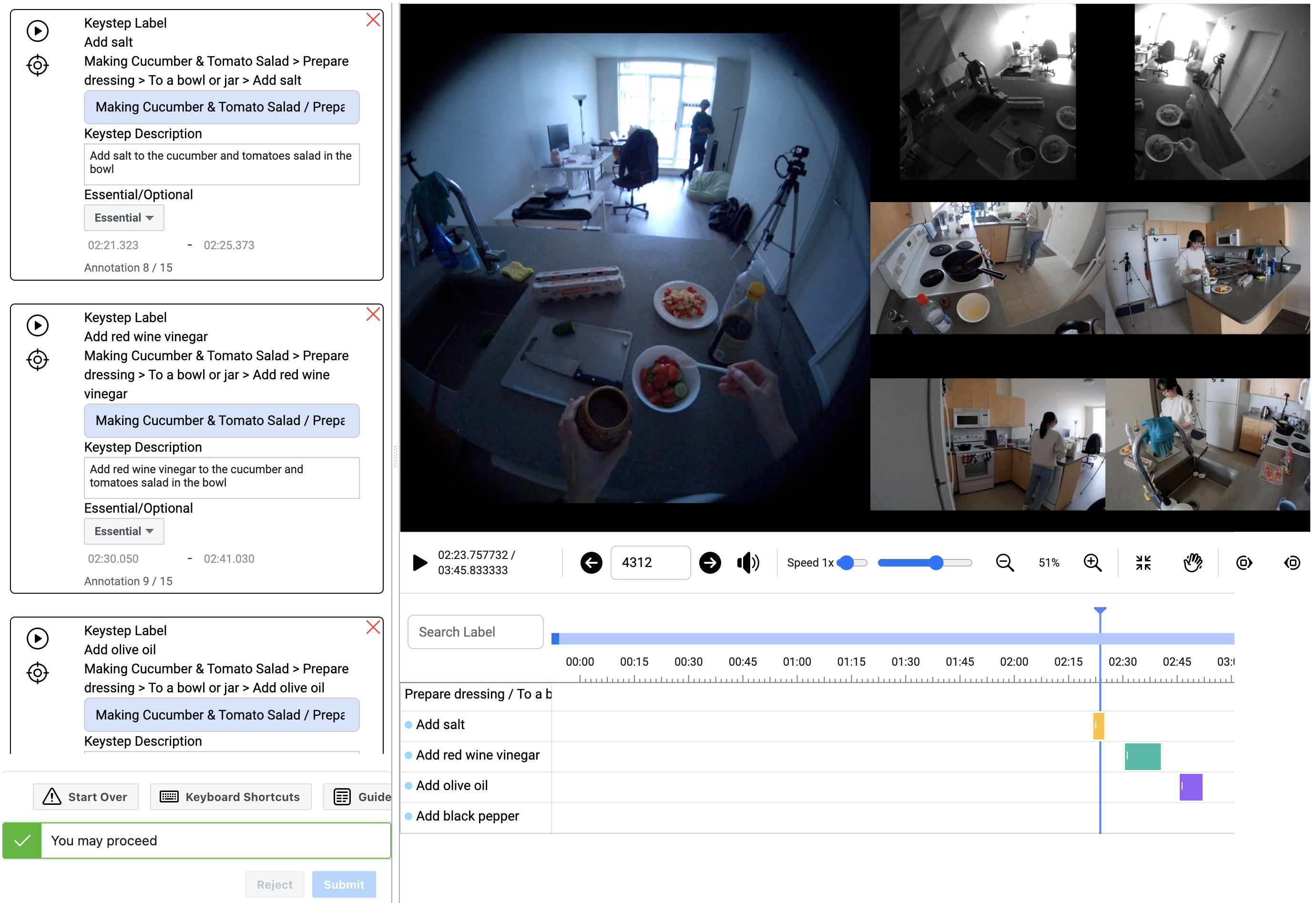}
\caption{
\textbf{The keystep annotation tool} shows a composite view of the time-synchronized ego-exo videos and the keystep time segment annotations. Each annotation consists of the start and end timestamps, a category label, a natural language description, and an essential/optional flag.
}
\label{fig:keystep_annotation_ui}
\end{figure*}

\begin{figure}[tp]
\includegraphics[width=\linewidth]{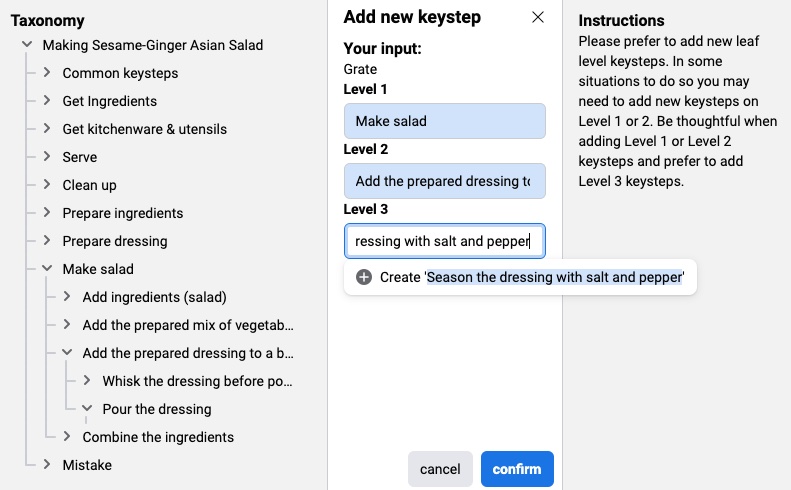}
\caption{
\textbf{Adding new keysteps to a taxonomy.} Annotators utilize a specialized widget to introduce new keysteps at any level within the existing taxonomy hierarchy. 
}
\label{fig:keystep_addition_widget}
\end{figure}

\TNIJCV{We collect manual annotations for keysteps (actions that contribute  towards the completion of a procedural task) and build a keystep taxonomy in parallel.}
Figure~\ref{fig:keystep_annotation_ui} shows the annotation user interface. We provide annotators a composite view of time-synchronized ego and exo videos. Each keystep annotation contains the start and end timestamps, a category label, a natural language description, and a flag indicating whether the keystep is essential or optional for task completion. Annotators interact with a search widget which displays keystep labels with their complete path within a hierarchical tree, e.g., Making cucumber \& tomato salad $>$ Prepare dressing $>$ To a bowl or jar $>$ Add salt. 

As the activities performed by the camera wearers are unscripted, it is not possible to establish a comprehensive keystep taxonomy prior to annotation. To address this challenge, we designed an iterative, data-driven process for taxonomy development. We first initialize the taxonomy using various resources including recipes and instruction articles from the Web. This initial taxonomy captures keysteps that are generally expected in the activities, but it is assumed to be incomplete for the specific variations the camera wearers performed in the recordings. Subsequently, in each iteration, annotators receive the current taxonomy and are instructed to add new keysteps when they encounter actions not represented in it (see Figure~\ref{fig:keystep_addition_widget}). Any newly added keysteps are kept valid only for the duration of each annotation session and are not visible in other sessions. After a batch of videos have been annotated, we review the newly added keysteps to ensure their validity and update the taxonomy before repeating the process. We finalized the taxonomy after the third iteration, after which we re-annotated the entire set of videos with the final taxonomy for consistency.

\begin{table*}[tp]
\centering
\small
\resizebox{\linewidth}{!}{
\begin{tabular}{r|rr|rr|rr|rr}
\multirow{2}{*}{Scenario} & 
    \multicolumn{2}{c|}{Takes} & 
    \multicolumn{2}{c|}{Ego Keystep Segments} & 
    \multicolumn{2}{c|}{Ego + Exo Keystep Segments} & 
    \multicolumn{2}{c}{Taxonomy} \\
 & 
    Count & Duration (total / avg$^\dagger$) & 
    Count (total / avg$^\dagger$) & Duration (total / avg$^\ddagger$) & 
    Count (total / avg$^\dagger$) & Duration (total / avg$^\ddagger$) & 
    Activity & Keystep \\ \midrule
Cooking     &  464 & 65.47h / 8.47m & 19,034 / 41.02 & 58.08h / 10.99s &  99,854 / 215.20 & 307.71h / 10.99s & 11 & 527 \\
Bike repair &   293 & 13.51h / 2.77m &  2,573 /  \:\:8.78 & 11.82h / 16.54s &  12,865 /  \:\:43.91 &  59.12h / 16.54s & 4 & 82 \\
Health      &   331 & 18.72h / 3.39m &  5,995 / 18.11 & 17.03h / 10.23s &  30,723 /  \:\:92.82 &  86.99h / 10.23s & 2 & 58 \\ \hline
Total       & 1,088 & 97.71h / 5.39m & 27,602 / 25.37 & 86.94h / 11.34s & 143,442 / 131.84 & 453.82h / 11.34s & 17 & 664
\end{tabular}
}
\caption{
\textbf{Keystep annotation statistics.} We report the statistics by grouping our 17 activities into three scenarios: cooking (11), bike repair (4), and health (2).  Statistics are listed for takes$^\dagger$ and keystep segments$^\ddagger$.
}
\label{tab:keystep_descriptive_stats}
\end{table*}

\paragraph{Dataset splits} 
The keysteps in our dataset exhibit a very long-tailed distribution. To address this challenge, we set a cutoff threshold at 20 samples per keystep, limiting our analysis to 278 unique keysteps. For simplicity, we consider only the leaf node keysteps in the hierarchy. Exploring the hierarchical structure including parent nodes is a promising direction but we leave this as future work. In all, the dataset for keystep recognition comprises 130,979 segments, with an average duration of 11.34 seconds each. Specifically, the training set contains 74,342 segments, of which 14,326 are from the ego view and the rest from the exo view. The validation set consists of 23,636 segments, including 4,517 ego-view segments, and the test set has 33,001 segments with 6,373 in the ego view.

\paragraph{Implementation details} We use clips of size $8\times224\times224$, with frames sampled at a rate of $1/32$ for all baselines except for EgoVLPv2 (where we adhere to its pretraining scheme and sample 4 frames). The patch size is $16\times16$. For training, we resize the shorter side of the frame to a random value within the range of [256, 320], followed by randomly sampling a $224\times224$ region from the resized video. For evaluation, we sample a single temporal clip in the middle of the video, scale down the shorter spatial side of the video to 224 pixels and select 3 spatial crops (top-left, center, bottom-right) from the temporal clip to cover a larger spatial extent within the clip. The final prediction is derived by averaging the scores obtained for these 3 crops. We train our model for a total of 100 epochs on 4 NVIDIA V100 GPUs with a batch size of 32. The model checkpoint yielding the best performance on the validation set is selected and evaluated on the test set. 

\paragraph{Results breakdown by keystep} 

\begin{figure*}[t]
    \centering
    \includegraphics[width=\linewidth]{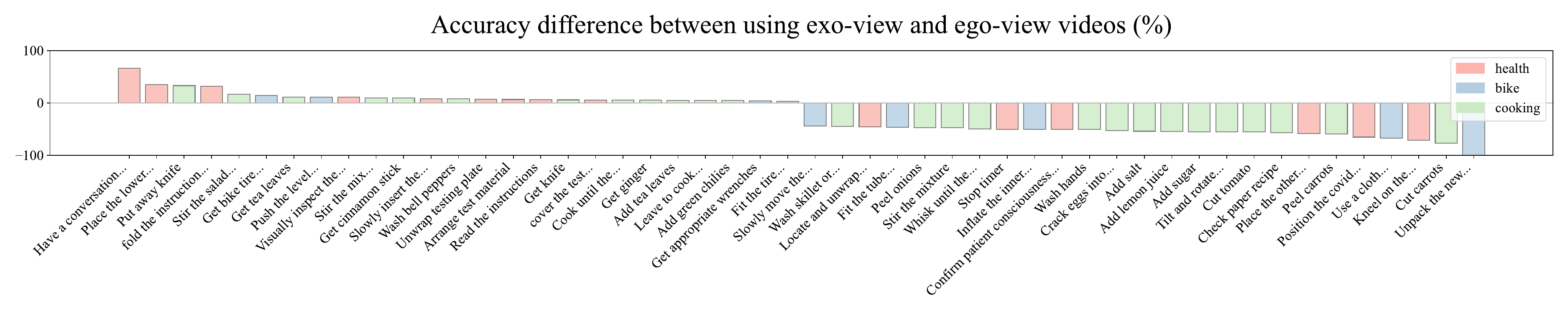} 
    \caption{Keystep recognition evaluated per keystep label on a validation split, comparing training with only ego-view videos versus exo-view videos. The accuracy delta (exo-ego) is displayed, where a positive value indicates better exo and a negative value indicates better ego performance.}
    \label{fig:baseline_keystep_detailed}
\end{figure*}

We present a more detailed analysis of per-step performance in Figure \ref{fig:baseline_keystep_detailed}, comparing training with ego-view videos and exo-view videos. We can observe that exo views show performance advantages over ego views in several steps, with the keystep `have a conversation asking different questions' benefiting the most from exo. Conversely, ego views are more effective in steps involving manipulation of small objects, like `cut carrots' and `unpack the new tube'. This observation can be linked to the positioning of exo cameras, which are often placed further away from the subject, enabling them to capture a broader view, though possibly missing finer details of the activity. We hope these findings provide insight for future research on the effective use of exo-view videos during training.

%% file: sec/appendices/benchmarks-appendix/multimodal-baseline.tex
\paragraph{Energy profiler} We adapt off-the-shelf profiler software built for PyTorch to compute the total multiply-accumulate operations (MACs) and memory transfer (MB) required to estimate total energy in Eqn.~\ref{eq:energy}. The quantities are time-normalized --- total energy consumption is expressed as power (mW). We describe each component of the profiler below.

\begin{itemize}
\item \textbf{Compute operations (MACs)} We use the native PyTorch FLOP counter to get the total FLOP count in the forward pass. We convert this to MACs (approximately 2 FLOPs = 1 MAC).
\item \textbf{Memory transfer (bytes)} We consider GPUs as our processing device, and use the PyTorch memory profiler to get the list of all operations executed in the forward pass (\texttt{model.forward()} call) and their associated GPU memory usage. The total memory is the sum of the individual operation memory costs.
\item \textbf{Sensor capture} For each modality, we measure the time for which it is active as the number of observations sampled containing the modality. We require that the sensors capture at least 1 second worth of samples (roughly 100 samples) as energy consumption is ill-defined for an \emph{instantaneous capture}. 
\end{itemize}

\paragraph{Energy tiers}
As mentioned in the main paper, there is a natural trade-off between efficiency and better performance. Thus, we evaluate models in two tiers by setting a budget for the power consumption in each tier, namely 20 mW for the \emph{high-efficiency} tier and 2.8W for the \emph{high-performance} tier. We select 
the \emph{high-efficiency} budget
based on the energy consumption of current single-modality, efficient architectures (e.g., X3D-XS~\citep{feichtenhofer2020x3d}) with an eye to the future where multi-modal models operate within it. 
For the \emph{high-performance} tier, we set the budget to a value that permits the use of powerful transformer-based action recognition models like LaViLa~\citep{zhao2023learning}. Once a model runs out of budget, in our setup it uses its latest prediction for all future steps. 

\paragraph{Complete baseline details}

\begin{itemize}
\item \textbf{X3D-XS~\citep{feichtenhofer2020x3d}.} This is a vision-only model comprising the X3D-XS feature encoder, which progressively expands the feature size and representational capacity of its layers, and later contracts them for achieving better performance-efficiency trade-off. This is the most lightweight model in our family of keystep predictors.
The encoder has a depth factor of 2.2, and takes 4 RGB frames of size $160 \times 160$ sampled at 15 fps, as inputs.

\item \textbf{LaViLa~\citep{zhao2023learning}.} This is another vision-only model where the visual feature encoder is trained through CLIP-style video-language pre-training.
To improve the feature quality over vanilla CLIP-style pre-training, this method augments the number of video-text pairs by leveraging pre-trained large language models (LLM) to generate textual descriptions of un-annotated videos and rephrase existing narrations.
In particular, we use the frozen TimeSformer~\citep{bertasius2021space}-Base (TSF-B) 
visual encoder pre-trained on the Ego4D dataset. To generate the feature for a target frame, the encoder samples 12 RGB frames of size $224 \times 224$ at 30 fps from a time window centered around the target frame and pads the samples with the boundary frames on both ends to create a 16-frame clip.

\item \textbf{Light-ASDNet~\citep{liao2023light}.} This is an audio-only model that
represents audio as spectrograms and efficiently encodes them by splitting 2D convolutions into 1D convolutions along the spectrogram temporal dimension~\citep{liao2023light}. In our setup, the spectrograms are Kaldi~\citep{Povey2011TheKS}-compliant, and consist of 196 temporal windows and 160 Mel-frequency bins, respectively.

\item \textbf{Audio-Visual Late Fusion (AV-LF).} This is an audio-visual model that does late fusion of visual features (encoded with X3D-XS or LaViLa) and audio features from Light-ASDNet by using linear layers.
\end{itemize}

\paragraph{Experimental setup}

We instantiate the task by considering keystep prediction episodes where the multimodal samples arrive in a streaming fashion. 
As mentioned above, we use vision and audio \SM{as our task modalities}, where vision comprises RGB frames that are streaming at 30 frames per second (fps), and the audio modality is made up of time-aligned single-channel chunks that are 0.4 seconds long and sampled at 16 kHz. However, the setup can be extended to include IMU, and potentially other sensors as well. 
We evaluate all models at the rate of 5 fps on a total of 211 test episodes of variable length, where the shortest episode is $\sim$15 seconds, and the longest episode is $\sim$34 minutes. 
We filter out episodes where all steps belong to the background class.

\paragraph{Implementation details} 
We train all keystep prediction models for 150 epochs using the cross entropy loss. We use the AdamW~\citep{loshchilov2018decoupled} optimizer with an initial learning rate of $10^{-4}$ and a weight decay of $10^{-5}$. We set the batch size to 512 for vision-only models, and 384 for audio-only and audio-visual models.

\begin{figure*}[ht]
\captionsetup[subfloat]{}
\centering
\begin{subfigure}[t]{0.45\textwidth}
\includegraphics[width=\textwidth]{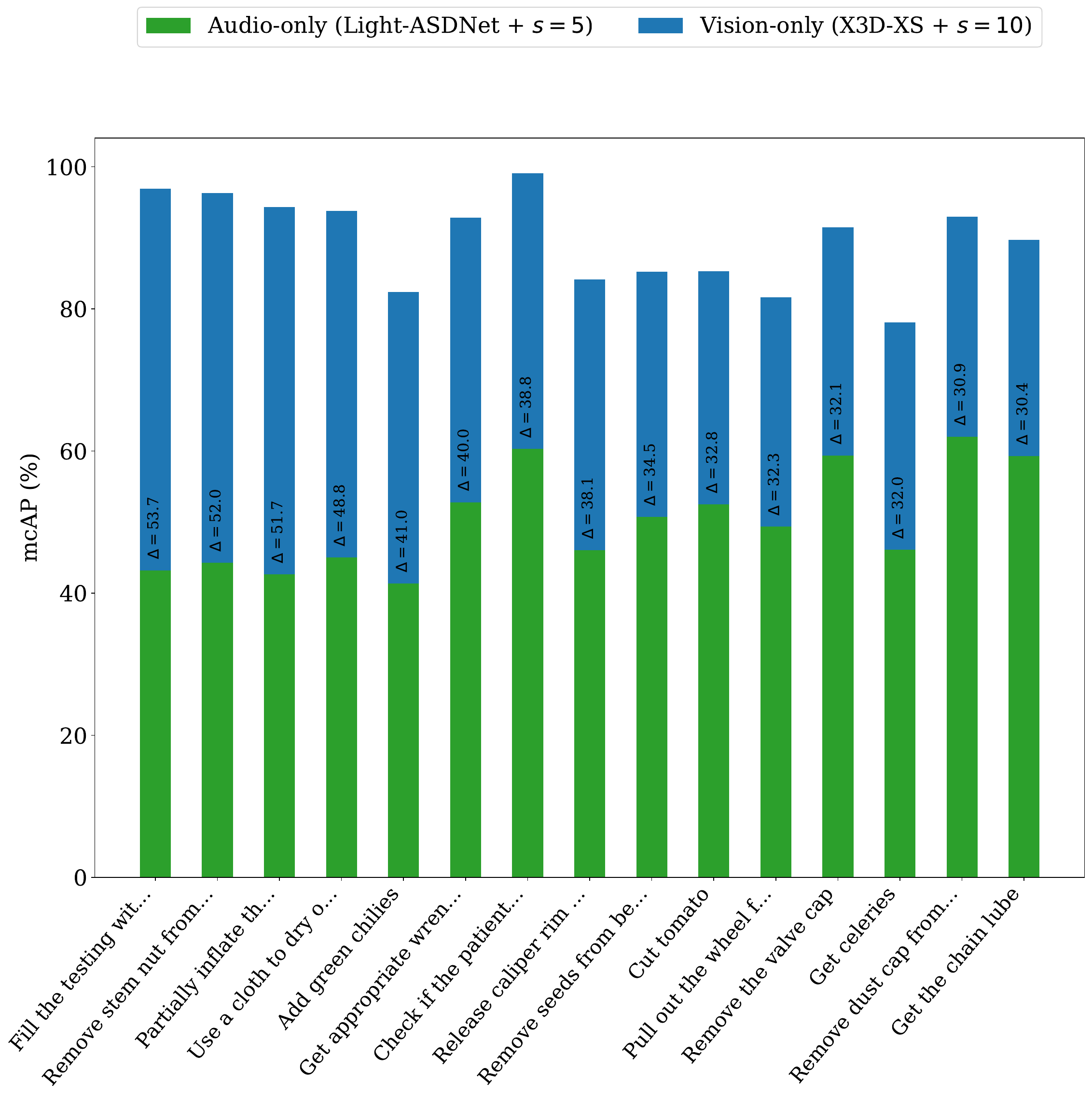}
  \caption{Vision improves over audio}
  \label{subfig:delPerf_x3d}
\end{subfigure}%
~
\begin{subfigure}[t]{0.45\textwidth}
\includegraphics[width=\textwidth]{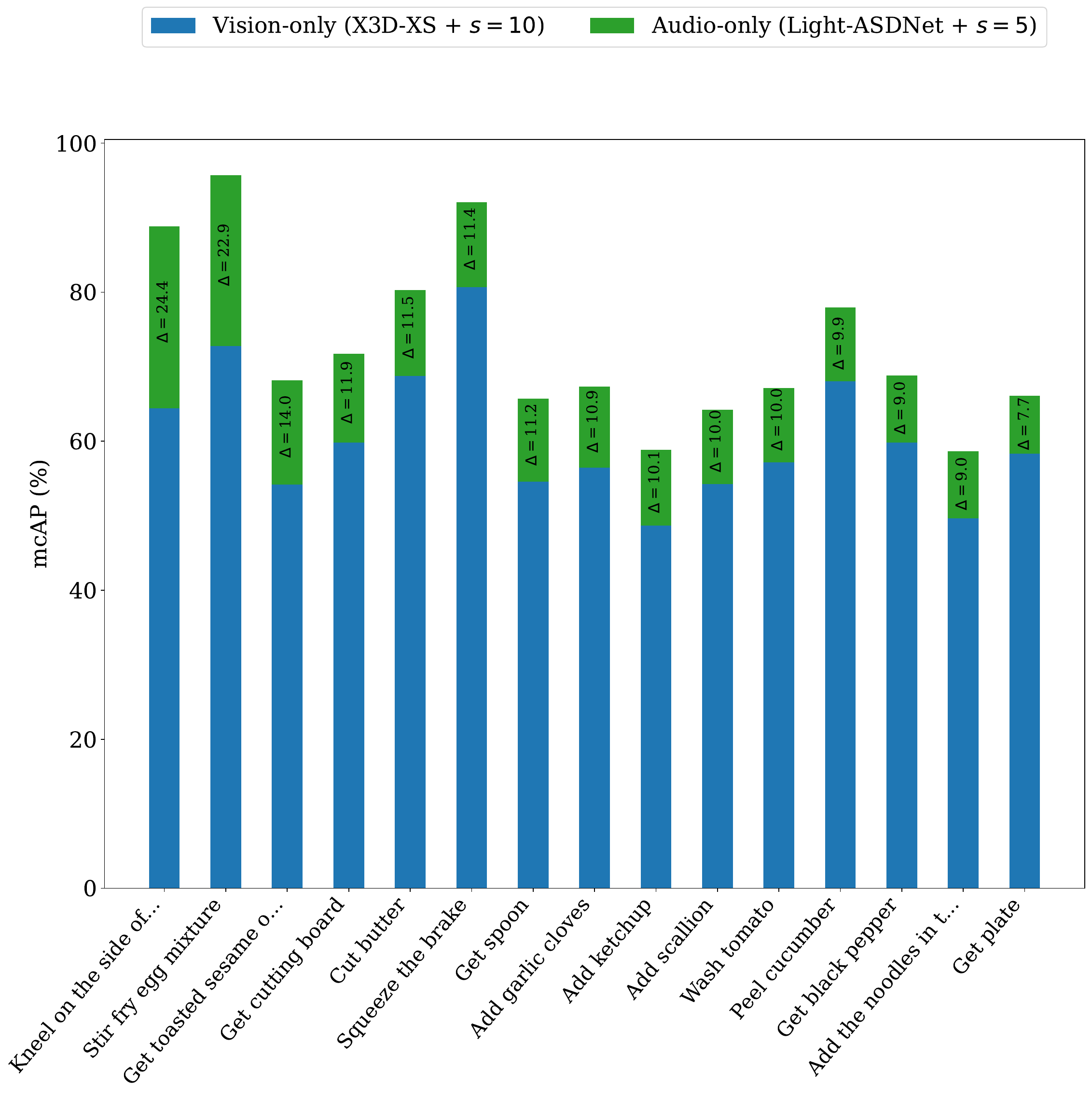}
  \caption{Audio improves over vision}
  \label{subfig:delPerf_lavila}
\end{subfigure}

\small
\caption{Improvement (left) or degradation (right) in keystep recognition performance per keystep label, when comparing the most efficient vision-only (X3D-XS~\citep{feichtenhofer2020x3d} + $s=10$) and audio-only (Light-ASDNet~\citep{liao2023light} + $s=5$) models from the \emph{high-efficiency} tier. The plots show the 15 keysteps where improvement or degradation are largest. $\Delta$ reports the amount of improvement/degradation.}
\label{figure:del_perf} 
\end{figure*}

\paragraph{Performance breakdown by modality}
In Fig.~\ref{figure:del_perf}, we present a detailed analysis of the keystep labels where the best vision-only model from the \emph{high-efficiency} tier yields the maximum improvement or decline in performance compared to its audio-only counterpart. We observe that the vision-only model produces a large improvement over the audio-only model usually in steps where the activity does not produce distinctive sounds (e.g., 
\emph{add green chillies},
\emph{get celeries},
etc). %
On the other hand, using audio alone helps the most when the activities involve sounds that are strongly indicative of the nature of the task (e.g., 
\emph{stir fry egg mixture},
\emph{cut butter},
etc). %

Finally, we envision that future work on this task will explore more sophisticated {\em learned} policies, potentially trained using reinforcement learning, in order to adaptively decide when to sample which modality instead of using fixed heuristics. Another promising direction is to investigate efficient transformer-based recognition backbones~\citep{xu2021long, zhao2022real} that can improve recognition performance without significantly affecting the model efficiency.

%% file: sec/appendices/benchmarks-appendix/taskgraph-baseline.tex
\paragraph{Complete baseline details}

\noindent
\textit{Graph-based baselines.}
Graph-based baselines are composed of a keystep assignment and a procedural reasoning component (see Figure~\ref{fig:baselines}(a)). 

Keystep assignment (A1) is applied to obtain a pseudo-labeling of the provided video segments when the supervision is at the procedure-level (i.e., segments are unlabeled and only keystep names are provided). 
This is achieved by means of an EgoVLPv2 model~\citep{kevin2022egovlp} pre-trained on ego-exo videos and narrations.
Video segments and keystep names are projected to the shared video-language space using EgoVLPv2.
We hence assigned each video segment to the closest keystep in the representation space according to the cosine distance.
In the problem formulation with instance-level supervision (i.e., when keystep labels are available for all segments during both training and testing), we use ground truth labels instead of those obtained from keystep assignment.
Additionally, we provide a baseline where the keystep assignment step is replaced with label predictions from the Keystep Recognition task (A2). Note that in the training set, segments with a confidence score below 20\% have been discarded.

The procedural reasoning component (B) creates for each procedure a transition graph based on keystep co-occurrences.
In the graph, each node represents a keystep category, while directed edges represent the probability of transitioning from one node to another one. 
An edge $A \to B$ is assigned the following weight based on statistics collected from the training videos:
\[ P(B|A) = \frac{\text{\# times keystep } B \text{ follows keystep } A}{\text{\# occurrences of keystep } A} \]
At test time, the graph is used to perform procedure understanding and answer the keystep-level questions. Specifically, given current segment $s_i$: 1) keystep $y_{prev}$ is predicted as the previous keystep with confidence score equal to the transition probability $P(y_i|y_{prev})$, where $y_i$ is the inferred or ground truth keystep label for segment $s_i$; 2) segment $s_i$ is predicted as optional based on the empirical probability $\frac{\text{\# training videos containing } y_i}{\text{\# training videos}}$; 3) segment $s_i$ is predicted as a procedural mistake with a score equal to the sum of the transition probabilities to $y_i$ from keysteps $y^{prev}$ that are missing from the keystep history, i.e., $\sum_{y^{prev}} [y^{prev} \notin S_{i-1}] \cdot  P(y_i|y_{prev})$, where $[\cdot]$ is the indicator function; 4) keystep $y$ is predicted as a possible missing keystep with probability $[y_i \notin Y_{:i-1}] \cdot P(y | y_i)$; 5) keystep $y$ is predicted as a future keystep with probability $P(y_i | y)$. \\

\noindent
\textit{End-to-end baseline.}
This baseline aims to provide an end-to-end approach to perform procedure understanding directly from the input clip. 
The baseline predicts previous keysteps, optional keysteps, and next keysteps by feeding video segment features extracted with EgoVLPv2~\citep{pramanick2023egovlpv2} to three dedicated MLPs.
Figure~\ref{fig:baselines}(b) illustrates the architecture of the baseline.
At training time, MLPs are supervised from the pseudo-labels obtained by graph-based baselines using Mean Squared Error (MSE) score to align the predicted probability distributions to the supervising ones. Missing keysteps and procedural mistakes are predicted from the outputs of the MLP components as in graph-based baselines.

%% file: sec/appendices/annotation/proficiency-annotation.tex
\paragraph{Annotations for demonstrator proficiency estimation} We derive annotations for this task from participant surveys (see Section~\ref{sec:appendix-people}) and expert commentary (see Section~\ref{sec:appendix-commentary}). Participant surveys contain responses to questions about prior experiences in the task such as ``How many years have you been doing this task?", and ``Do you have any qualifications/professional training related to the task?" (see Table~\ref{tab:questionnaire_qs} for the complete list). On the other hand, expert commentary is performed by task-specific experts and includes 1 to 10 proficiency scores for each video from the participant (see Section~\ref{sec:appendix-commentary}). After consulting with experts hired for each scenario, we designed scenario-specific conversion functions that use the surveys and expert commentaries to produce an estimate of a participant's proficiency score (see Table~\ref{tab:demonstrator_proficiency_skills}). For example, in basketball and soccer, we use the years of experience to determine skill level since we found this to be an accurate indicator of skill based on analyzing the videos. On the other hand, to determine skill level in bouldering, we use the highest difficulty level of the route solved by the participant. 

\begin{table*}[]
\small
\centering
\begin{tabular}{@{}cP{2.5cm}P{2.5cm}P{3.5cm}P{2.5cm}@{}}
\toprule
Scenario   & Novice            & Early Expert                  & Intermediate Expert         & Late Expert \\ \midrule
Basketball & $\valrng{X}{0}{1}$   & $\valrng{X}{1}{3}$ & $\valrng{X}{3}{10}$ & $X \ge 10$\\
Soccer     & $\valrng{X}{0}{1}$   & $\valrng{X}{1}{3}$ & $\valrng{X}{3}{10}$ & $X \ge 10$\\
Dancing    & $\valrng{X}{0}{3}$   & $\valrng{X}{3}{5}$ & $(\valrng{X}{5}{10})\lor((X \ge 10)\land\neg P)$  & $(X \ge 10) \land T$           \\
Bouldering & $H \le \textrm{V3}$ & $H == \textrm{V4}$          & $H == \textrm{V5}$          & $H \ge \textrm{V6}$ \\
Music (violin) & $(\valrng{X}{0}{3}) \lor (\valrng{N}{0}{500})$ & $(\valrng{X}{3}{5}) \lor (\valrng{N}{500}{1000})$ & $(\valrng{X}{5}{10}) \lor (\valrng{N}{1000}{10000})$ &  $(X \ge 10) \lor (N \ge 10000)$ \\
Music (guitar) & $(\valrng{X}{0}{1}) \lor (\valrng{N}{0}{500})$ & $(\valrng{X}{1}{3}) \lor (\valrng{N}{500}{1000})$ & $(\valrng{X}{3}{10}) \lor (\valrng{N}{1000}{10000})$ &  $(X \ge 10) \lor (N \ge 10000)$ \\
Music (piano)  & $(\valrng{X}{0}{1}) \lor (\valrng{N}{0}{500})$ & $(\valrng{X}{1}{5}) \lor (\valrng{N}{500}{1000})$ & $(\valrng{X}{5}{10}) \lor (\valrng{N}{1000}{10000})$ &  $(X \ge 10) \lor (N \ge 10000)$ \\
Cooking    &     $P < 3.5$              &          $\valrng{P}{3.5}{5}$                &     $\valrng{P}{5}{8}$                     &   $P \ge 8$       \\
\bottomrule
\end{tabular}
\caption{\textbf{Annotations for demonstrator proficiency estimation.} We designed scenario-specific conversion functions that take in participant surveys and expert commentary assessments to estimate proficiency of participants (i.e., novice, early expert, intermediate expert, and late expert). Legend: $X$ = years of experience performing the task, $T$ = professional training in the task, $H$ = highest difficulty level solved by participant in bouldering, $N$ = estimated number of times performing the task, $P$ = average proficiency rating from expert commentary.}
\label{tab:demonstrator_proficiency_skills}
\end{table*}

\paragraph{Annotations for demonstration proficiency estimation} Table \ref{tab:demonstration_proficiency_annotations} shows examples of expert comments and corresponding annotation tags derived from them indicating whether the comment suggests a good execution and/or tips for improvements. 

\begin{table*}[]
\small
\centering
\begin{tabular}{@{}cP{10cm}P{1.5cm}P{1.5cm}@{}}
\toprule
Scenario   & Expert comment                                                                                                                                                                                                    & Good execution & Tips to improve \\ \midrule
Basketball & Nice release. I like the follow through here. You'd like to see the guide hand maybe up a little bit higher on the release of that shot. Maybe to give it better ball control when you're letting go of the shot. & Yes            & Yes                  \\
Basketball & Great footwork, left foot take off, lifting of the right knee and extending that body up. Love how he's looking up, checking out the backboard, shooting hand behind the basketball. Nice job. &    Yes         &    No                \\
Basketball & He's also really far away from his body and the more he can keep his arm up by his ear, it will give him the most opportunity to make the basket without the defense interrupting. &    No          &        Yes           \\
Bike repair & It's a great method to always double check or do a pre-check before beginning work on a bicycle to make sure the issue that you are working to fix is the only issue that is occurring. If not, you could find a secondary issue or something else that may be greater than the one you are currently working on. &    Yes         &        No           \\
Bike repair & As you can see she clearly slipped on loosening the nut which essentially creates damage to the surface of the nut itself and can round out the nut. &    No          &        Yes           \\
Bouldering & The climber was efficiently able to position herself with one hand on each hold at the start and had, once her hands were positioned, she matched her feet on the hold and efficiently moved to the next hold. &    Yes         &        No            \\
Bouldering & And since she popped out and is swinging out, she can't really keep the tension through her one arm because she's so locked off. So it caused her to kind of just fall off the wall and lose all tension throughout all of her body. &    No          &        Yes           \\
Cooking & You can see there, she's not able to stir properly. She has to push it around, which means that the lime is not gonna be very evenly distributed among the pieces of tomato and cucumber. &    No          &        Yes           \\
Cooking & Using a grinder for fresh pepper is an excellent way to get a lot of flavor. The fresh grind of pepper as opposed to buying already ground pepper really expels the oils and everything in those peppercorns and allows the flavor to be as big as it can possibly be. &    Yes         &        No            \\
\bottomrule
\end{tabular}
\caption{\textbf{Annotations for demonstration proficiency estimation.} We annotated expert comments about a participant's task execution with tags indicating whether each comment describes a good execution or suggests tips for improving skills. Note that the same comment might describe one aspect of the task as being good while suggesting improvements in another aspect (e.g., see row 1).}
\label{tab:demonstration_proficiency_annotations}
\end{table*}

\begin{figure}
    \centering
    \includegraphics[width=0.5\textwidth,trim={0.5cm 0.0cm 10.5cm 0},clip]{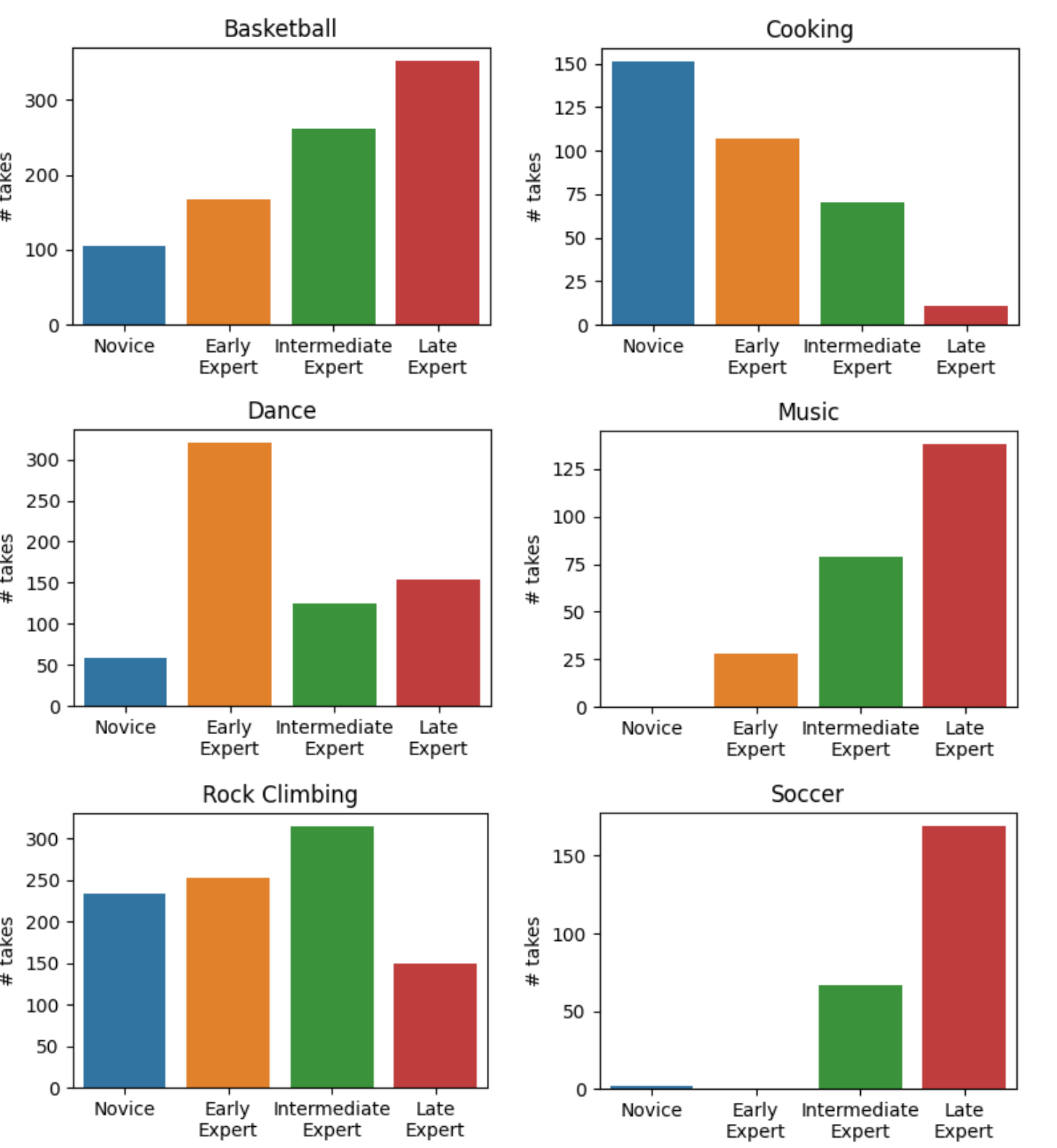}
    \caption{Distribution of demonstrator proficiency scores per scenario.}
    \label{fig:appendix-demonstrator-scores}
\end{figure}

%% file: sec/appendices/benchmarks-appendix/proficiency-baseline.tex
\paragraph{Baseline implementation details}

\textbf{Demonstration proficiency estimation:} We use the TimeSFormer~\citep{bertasius2021space} architecture for the baseline. TimeSFormer is a video transformer designed for video action recognition/classification that introduces a novel decoupled spatiotemporal attention mechanism. We resize all videos to 448 pixels along the smallest dimension and use a clip size of 16 frames with a frame rate of 16 FPS. The models are trained to classify individual clips using the cross-entropy loss on 8 Quadro RTX 6000 GPUs for 15 epochs. \vspace{-0.05in}\\

\noindent \textbf{Demonstration proficiency estimation:} We adopt ActionFormer~\citep{zhang2022actionformer}, a video action localization model for our experiments. Unlike traditional action localization that defines time windows as outputs, we instead perform timestamp regression since our annotations contain only a single point in time for each good execution or tip for improvement. We accordingly adapt ActionFormer's post-processing strategy and evaluation metrics. In our task, the predicted timestamps correspond to frames retained after non-maximum suppression (NMS). We remove the regression head of ActionFormer and infer the predicted timestamps from the indices of frames retained after NMS. We also modify the NMS module of ActionFormer to rely on the $\mathcal{L}_1$-distance between predicted timestamps instead of the tIoU between segments used in \citep{zhang2022actionformer}. During training, we keep the classification loss from~\citep{zhang2022actionformer} and replace the regression loss with the loss function defined in~\citep{kwak2020detecting}. We train our models with Omnivore features~\citep{omnivore} extracted with a clip size of 32 frames and a stride of 16 frames from all the videos. 

\paragraph{Scenario-specific results on demonstrator proficiency estimation}
We show scenario-specific results in Table~\ref{tab:demonstrator_proficiency_results_breakdown}. TimeSFormer achieves good performance with the egocentric view in cooking since a close-up view of the objects of interest and hand poses is essential to assessing skill in these scenarios. On the other hand, the model performs better with the exocentric view in bouldering on the test data since the overall body pose is a useful indicator of proficiency. Unfortunately, it fails to improve over the majority-class baseline is most scenarios on the test splits except basketball, highlighting a distribution shift between the val and test splits. 

\begin{table*}[t]
\small
\centering
\setlength{\tabcolsep}{3pt} %
\begin{tabular}{@{}ccccc|cccc@{}}
\toprule
                & \multicolumn{4}{c}{Val} & \multicolumn{4}{c}{Test}       \\
Scenario        & Majority-class & Ego            & Exos           & Ego + Exos & Majority-class & Ego & Exos & Ego + Exos \\ 
\midrule
Basketball      & 35.66          & \textbf{56.64} & \textbf{56.64} & 51.75      & 46.71          & \textbf{79.64} & 71.86          & 76.65         \\ 
Cooking         & 50.94          & \textbf{56.60} & 45.28          & 49.06      & \textbf{39.54} & 19.77          & 33.72          & 32.56         \\
Dancing         & \textbf{43.31} & 42.52          & 31.50          & 32.28      & \textbf{46.62} & 43.24          & 40.55          & 42.57         \\
Music           & 44.44          & \textbf{69.44} & 50.00          & 58.33      & \textbf{73.24} & 56.34          & 45.07          & 50.70         \\
Bouldering      & 0.00           & \textbf{27.67} & 18.24          & 17.61      & 19.57          & 39.13          & \textbf{43.04} & \textbf{43.04}\\
Soccer          & \textbf{74.42} & 65.11          & 37.21          & 41.86      & \textbf{72.73} & 65.15          & 31.82          & 30.30         \\ 
\bottomrule
\end{tabular}
\caption{\textbf{Breakdown of results for demonstrator proficiency estimation across scenarios.} Top-1 accuracies per scenario for the TimeSFormer model.}
\label{tab:demonstrator_proficiency_results_breakdown}
\end{table*}

%% file: sec/appendices/benchmarks-appendix/bodypose-baseline.tex
\paragraph{CVPR 2024 challenges}
We organized two Ego Pose challenges as part of the EgoVis workshop at CVPR 2024, aiming to encourage researchers to explore and utilize the dataset. We had eight participants in total, divided evenly between the Body Pose challenge and the Hand Pose challenge.

\paragraph{Body Pose}
There were a total of four submissions to the hand pose challenge, with three outperforming the best baseline method, as shown in Table \ref{tab:body_pose_challenge}. The best participant outperformed the baseline by 3.19 cm in MPJPE. 

\begin{table}[]
\begin{tabular}{l|l|l}
\hline
                  & MPJPE & MPJVE \\ \hline
Location-based (baseline) & 18.51 & 0.64    \\ 
\hline
Levelwise att ViT  & 18.09 & 0.62     \\ 
UCB Ego            & 17.19 & 0.57     \\ 
Multi-Scale Model Fusion   & 15.32 & 0.55    \\ 
\end{tabular}
\caption{Leaderboard of Ego-Exo4D Body Pose challenge at EgoVis CVPR 2024.}
\label{tab:body_pose_challenge}
\end{table}

\noindent\textbf{First place: Multi-Scale model fusion} \textit{(Baoqi Pei, Yifei Huang, Guo Chen, Jilan Xu, Yicheng Liu, Yuping He, Kanghua Pan, Tong Lu, Limin Wang, Yali Wang, Yu Qiao)}

Instead of implementing a new baseline model, this approach leverages the existing location-based baseline and focused on improving performance by fusing predictions from multiple models with varying numbers of transformer layers (ranging from 1 to 6). This approach allowed them to address variations in data distributions and stabilize predictions. By combining the outputs from the different versions of the baseline, they achieved an improved MPJPE of 15.32. 

\noindent\textbf{Second place: UCB Ego} \textit{(Brent Yi, Vickie Ye, Georgios Pavlakos, Lea Müller, Maya Zheng, Yi Ma, Jitendra Malik, Angjoo Kanazawa)}

The submission by UCB EGO to the Ego-Exo4D Body Pose Challenge centers around their development of a conditional human motion diffusion model that operates using 30Hz SLAM pose data to directly sample SMPL human body parameters. The model was trained exclusively on the AMASS dataset, which poses a limitation of domain transfer to the Ego-Exo4D dataset. To address this limitation, the team introduced an AdapterNet to estimate floor height and route procedural tasks to the baseline model while focusing the diffusion model on physical activities. This hybrid approach allowed the team to achieve a MPJPE of 17.19 cm.

\noindent\textbf{Third place: Levelwise attention ViT} \textit{(Congsheng Xu, Jinfan Liu, Yifan Liu, Shuwen Wu)} 

The authors proposed a model that combines two Vision Transformer (ViT) structures to leverage both coarse-grained and fine-grained information for 3D human pose estimation. They used a baseline ViT model with 8 attention heads and 3 layers for coarse-grained estimation, and a Huge-ViT model with 16 attention heads and 32 layers for fine-grained estimation. They combine the outputs from both ViTs through a weighting strategy. By assigning a higher weight to the fine-grained results, the method balances local detail and global information. This weighting approach resulted in a MPJPE of 18.09 cm.

\begin{table}[!t]
\begin{tabular}{l|l|l}
\hline
                  & MPJPE & PA-MPJPE \\ \hline
POTTER (baseline) & 28.94 & 11.07    \\ 
\hline
PCIE EgoHandPose  & 25.51 & 8.49     \\ 
Hand3D            & 30.52 & 9.30     \\ 
POTTER-ensemble  & 28.68 & 10.24    \\ \hline
\end{tabular}
\caption{Leaderboard of Ego-Exo4D Hand Pose challenge at EgoVis CVPR 2024.}
\label{tab:hand_pose_challenge}
\end{table}

\paragraph{Hand Pose}

There are in total 4 submissions to the hand pose challenge, with 3 outperformed the baseline method POTTER~\citep{zheng2023potter} as shown in Table~\ref{tab:hand_pose_challenge}. The best participant outperformed the baseline model by 11.85\% in  MPJPE, and by 23.31\% in PA-MPJPE.

\noindent\textbf{First place: PCIE EgoHandPose}~\citep{chen2024pcie_egohandpose} 

The authors propose the Hand Pose Vision Transformer (HP-ViT). The HP-ViT comprises a ViT backbone and transformer head to estimate joint position in 3D, utilizing MPJPE and RLE loss function. To be more specific, the model employed regression loss based on RLE~\citep{li2021human} to minimize the gap between output and input distributions. The experiments show that the model with ViT-Huge~\citep{dosovitskiy2020image} as backbone achieves the best performance. The ensembling model with different setting also contributes to decrease the overall error.

\noindent\textbf{Second place: Hand3D}~\citep{pavlakos2024reconstructing}  

The authors apply recently introduced HaMeR out-of-the-box on the images of the EgoExo4D Challenge, and observe very strong performance. HaMeR is a feed-forward model that takes as input a single image of a hand and hand side (left or right), and estimates a 3D reconstruction of the hand in the form of the MANO parametric hand model~\citep{MANO:SIGGRAPHASIA:2017}. HaMeR adopts a fully transformerized architecture design using a ViT-H~\citep{dosovitskiy2020image} backbone, followed by a transformer head. The authors further show the limit of the method, particularly when the occlusions/truncations are very extreme (only a few visible fingers), the wrist location or hand orientation is ambiguous, and finally, if the original hand bounding box crop is not very accurate.

\noindent\textbf{Third place: POTTER-ensemble} \textit{(Baoqi Pei, Yifei Huang, Guo Chen, Jilan Xu, Yicheng Liu, Yuping He, Kanghua Pan, Tong Lu, Limin Wang, Yali Wang, Yu Qiao)}

The authors handle the problems through a multi-scale model fusion method. To be specific, the authors enhance the baseline method POTTER~\citep{zheng2023potter} by (1) including multiple upsampling dimensions, specifically 128, 256, and 512, (2) integrating a dynamic pooling operation before the 3D convolution operation of the model, and (3) combining/ensembling these models to improve generalization capabilities.

%% file: sec/contributions.tex
\clearpage
\section{Detailed contribution statement}

 This project is the result of a large collaboration between many institutions over the last two years. Initial authors represent the leadership team of the project. Kristen Grauman initiated the project, served as the technical lead, initiated the recognition and proficiency benchmarks and expert commentary, and coordinated their working groups. Andrew Westbury served as the program manager and operations lead for all aspects of the project. Lorenzo Torresani led development of the capture domains, initiated the relation and ego-pose benchmarks, and coordinated their working groups.  Kris Kitani led development of the multi-camera rig and supported the Ego-Exo4D engineering team on all aspects of the data annotation and organization.  Jitendra Malik served as a scientific advisor.  Authors with stars ($^\ast$) were key drivers of implementation, collection, and/or annotation development throughout the project.  Authors with daggers ($^\dagger$) are faculty and senior researcher PIs for the project.

\paragraph{Camera rig}

Rawal Khirodkar proposed the hardware and software specifications for the ego-exo camera rig and helped design the capture protocol. Sean Crane investigated various hardware setups leading to the final rig configuration and helped draft the capture guidelines including recommended gear. Devansh Kukreja developed the sync and take separation algorithms, experimented with different equipment options (e.g., camera, timecode boxes, mount options), and designed the interface to transfer data; he also managed the ingestion pipeline, the collaboration with Aria, the integration of their code for EgoExo, and usage of the .vrs files.

\paragraph{Aria}

Jing Dong and Vijay Baiyya were responsible for obtaining camera poses, calibration, pointclouds and eye gaze using Aria MPS, created the 3D/4D visualizations for the paper and supplementary material, and acted as main contact points from the Aria team throughout the program; with Jing leading the algorithm development and verification, and Vijay leading the Aria MPS workflow and infrastructure development. Jakob Engel acted as technical and scientific advisor, and led the team that built the Aria Localization and Point Cloud algorithms. Kiran Somasundram helped design the capture setup and time-synchronization. Xiaqing Pan helped to align the Aria engineering team to support the EgoExo4D project. Mingfei Yan, Prince Gupta, and Sach Lakhavani acted as product managers of Aria and organizational leads for the successful use of Aria in the program. Kelly Forbes helped setting up agreements and working through the legal requirements of using Aria devices for recording the EgoExo4D dataset across the globe. Richard Newcombe initiated the Aria/Ego4D collaboration and acted as a scientific advisor throughout the program. Furthermore, we want to acknowledge the contribution of the entire Project Aria team as listed in~\citep{aria_2023}, including Carl Ren and Sean Diener leading the Aria software and hardware engineering organization, and Renzo De Nardi as technical lead for the Aria device.

\paragraph{Data collection}

\textbf{Los Andes University} 
Pablo Arbeláez -  lead coordinator for data collection and collaborator on the overall project design; Maria Escobar - data collection for all phases, design of the collection setup and workflow, data inspection, ingestion, encoding, and metadata generation; Cristhian Forigua -  data collection for all phases, participant recruitment, consent forms design, data inspection, communication with recording sites; Cristina González - data collection for phase 1, design of the collection setup and workflow, IRB management; Angela Castillo - data collection for phase 2, manual data inspection, and data analysis.

\textbf{Georgia Tech}
James M. Rehg - lead coordinator for data collection and protocol design, and overall project manager; Bikram Boote - lead coordinator for data collection, including recruiting and ingestion; Fiona Ryan - contributed to data collection; Audrey Southerland - lead coordinator for IRB development, contributed to recruiting.

\textbf{National University Singapore}
Mike Zheng Shou - lead coordinator for data collection and protocol design, and overall project manager; Joya Chen - contributed to protocol design, camera setup design, data collection for all phases; Jia-Wei Liu - contributed to protocol design, camera setup design, data collection for all phases; Xinzhu Fu - contributed to data collection for all phases; Chenan Song - contributed to data collection for all phases.

\textbf{Meta}
Andrew Westbury was the lead for data collection at our site, selecting scenarios, organizing capture sessions, recruiting participants, organizing and transferring data, and obtaining required approvals. In California, Hao Tang and Kevin Liang also supported all these functions, focused on bike repair. In New York, Devansh Kukreja and Alex Dinh lead collection for cooking scenarios. Miguel Martin also supported California-based collections and organized our local camera rig. Chefs Eton Chan and Dominic Ainza supported all culinary collections with technical guidance, recruitment, and coordination. Dimitri Elston coordinated and was the technical lead on bike collections. Adrian Salas supported pilot bouldering collections in California. Across all Ego-Exo4D collections, Devansh Kukreja continuously communicated and refined the recording procedure with universities, and problem-solved local recording issues.   

\textbf{University of North Carolina at Chapel Hill }
Gedas Bertasius - lead coordinator for data collection; Md Mohaiminul Islam - the main contributor to data collection and metadata processing across all scenarios; Wei Shan - contributed to data collection and metadata processing for the music and soccer scenarios; Jeff Zhuo - contributed to data collection and metadata processing for the soccer scenarios; Oluwatumininu Oguntola - contributed to participant recruiting and data collection for the music scenario. 

\textbf{Carnegie Mellon University}
Rawal Khirodkar developed the automatic 3D body keypoints extraction pipeline and collected a subset of the soccer, bike mechanic, and cooking sequences for the CMU portion of the dataset. Sean Crane was in charge of the data collection, IRB documents, capturing data, working with participants and processing the data for CMU. Abrham Gebreselasie ran the actionformer-based baseline for demonstration proficiency benchmark. Eugene Byrne served as the engineering lead for the initial design and implementation of the dataset, camera rig, processing pipeline and keystep annotations while at Meta. Subsequently at CMU, he assisted in the recognition benchmarks, implemented Ego-Exo transfer ~\cite{li2021ego} (1 of the 3 baseline methods for keystep recognition) and the initial implementation of keystep action detection ~\cite{actionformer}, and assisted in annotation/data quality generally.

\textbf{Simon Fraser University}
Sanjay Haresh was the lead coordinator for data collection, including recruiting, data ingestion, and data analysis. Yongsen Mao also contributed to the data collection pipeline, recruiting, data ingestion, metadata annotations, and statistics computations. Manolis Savva advised on data collection, protocol design, and overall project management. We acknowledge the assistance of Hanxiao Jiang and Armin Kavian with recruitment and data collection.

\NEW{
\textbf{University of Pennsylvania}
Edward Zhang led data collection efforts at UPenn and played a key role in subject recruitment, subject information collection, and on-site data recording. Jinxu Zhang led data management, information logging, and data transfer.  Shan Su is the overall project lead, focusing on determining good camera configurations based on 3D reconstruction feasibility and fixing issues in data post-processing of time synchronization and take separation. }

\NEW {
\textbf{University of Tokyo}
Yoichi Sato served as the primary coordinator for data collection, while Ryosuke Furuta was responsible for data collection across all three scenarios, participant recruitment, and IRB submission. Zecheng Yu and Masatoshi Tateno provided support for data collection and were responsible for managing and transferring data. Takuma Yagi helped with the IRB submission process. }

\NEW{
\textbf{Indiana University}
David Crandall oversaw the overall effort at Indiana University, including protocol design and data collection. Weslie Khoo led the IRB protocol design and compliance, arranged logistics such as ordering equipment, and designed and oversaw the cooking scenario data collection. Yuchen Wang and Ziwei Zhao co-led the participant recruitment and data collection for all three scenarios. Ziwei Zhao led data preparation and transfer. We also acknowledge Manasi Swaminathan who assisted with data collection and video synchronization.
}

\NEW{
\textbf{IIIT-Hyderabad}
Avijit Dasgupta was the lead on the ground for data collection in Hyderabad helping in organizing capture sessions, data collection, and managing and transferring data. Siddhant Bansal helped in the early stages with IRB application, consent forms, and pilot studies. C. V. Jawahar was the lead coordinator for data collection helped in selecting the scenarios, and recruiting the participants.}

\NEW {

\textbf{University of Minnesota}
Hyun Soo Park oversaw the overall effort at the University of Minnesota, Twin Cities, including protocol design and data collection. Zachary Chavis led the IRB protocol design and compliance, arranged logistics such as ordering equipment, participant recruitment, and designed and oversaw all scenarios of data collection. Anush Kumar assisted data collection for all scenarios.}

\paragraph{Language annotations}

\NEW{Kevin J Liang co-developed the expert commentary guidelines, interviewed and onboarded experts, helped test and suggest features for the narrator tool, and contributed to program management; he also developed the atomic action descriptions guidelines, helped coordinate annotations, and contributed to paper writing.}
\NEW{Michael Wray contributed to the definition of the expert commentary guidelines; provided feedback for experts; co-developed the narrate-and-act guidelines and the pre-task/post-task questionnaires; and contributed to paper writing.}
\NEW{Kristen Grauman proposed the expert commentary idea, co-developed the guidelines, interviewed and provided feedback to experts, and contributed to paper writing.}
\NEW{Andrew Westbury implemented expert commentary, recruiting, mobilizing and managing our experts and workplan.}
Miguel Martin contributed to the atomic action descriptions annotation guidelines and produced the annotation files and associated tutorial code, and for expert commentary, he authored the initial version of the Narrator tool, transcribed the commentaries, and produced the annotation files and associated tutorial code.

\NEW{Changan Chen contributed to the development of the  narrator tool; provided feedback for experts; and contributed to paper writing.}
Siddhant Bansal contributed to the design of narrate-and-act, user questionnaires, object dictionaries, expert commentary cooking. 
Dima Damen proposed narrate-and-act data collection and user questionnaires and contributed to their design, and also contributed to expert commentary for cooking scenarios.
\NEW{Tiffany Davis provided significant program management support throughout expert commentary.}
\NEW{Devansh Kukreja built the render flow to generate video collages for annotations.}

\NEW{Domain resource people from our consortium were Dima Damen and Michael Wray [cooking], Kristen Grauman and Changan Chen [soccer], Gedas Bertasius [basketball], Kristen Grauman and Jianbo Shi [music], Andrew Westbury [health and bike repair], Kevin Liang [dance], Pablo Arbelaez and Maria Escobar [bouldering] }

\NEW{ Ego-Exo4D's panel of expert commentators is: 
\textbf{Soccer}: John Bello, Phillip O'Kennedy, Lee Bakewell, Radcliffe McDougald, Thomas Harris
\textbf{Music}: James Peterson, Trevor Minton, Andrea LaPlante, Ethan Fallis, Alex Rogers, Jacqueline Burd
\textbf{Health}: Jasmine Higa, Angela Liszewski, Kristin Blanset, Melissa Robinson, Sonya Johnson
\textbf{Dance}: Rolanda Williams, Deanna Martinez, Enya-Kalia Jordan, Rachel Repinz, Yauri Dalencour, Kathryn Hightower
\textbf{Cooking}: Mark Manigault, Mary Drennen, Tiffany Davis, Reginald Howell, Rosanne Field, Donnie Murphy, Kiet Duong, Laura de Vera, Keegan Taylor
\textbf{Bouldering}: Daniel Ramos, Mike Kimmel, Roy Quanstrom, Christopher Deal, Carmen Acuna, Kelsey Hanson
\textbf{Bike repair}: Cesar Pineda, Walker Wilkson, Frank Trotter, Cordell Bushey, Dimitri Elston, Sam Arsenault, Aaron Hill
\textbf{Basketball}: Elizabeth Blose, Raven Benton, Joseph McCarron, Cornelius Gilleyen, Cecil Brown. Aaron Jones
}

\paragraph{Benchmarks}

\NEW{
\textbf{Ego-exo correspondence}
{Manolis Savva} co-led the correspondence benchmark and contributed to the task definition, the annotation guidelines, the baseline design, and paper writing.
{Effrosyni Mavroudi} co-led the correspondence benchmark and contributed to the task definition, the annotation guidelines, the baseline design, and paper writing.
Lorenzo Torresani contributed to the task formulation.
{Sanjay Haresh} developed the spatial baselines and contributed to data analysis, experimental results, and paper writing.
{Yongsen Mao} developed the spatiotemporal baselines and contributed to data analysis, experimental results, and paper writing.
{Suyog Jain} formulated the annotation pipeline, developed annotation tools and contributed to annotation guidelines and paper writing.
{Santhosh Ramakrishnan} contributed to the annotation guidelines and the formulation of the annotation pipeline.
{Xitong Yang} contributed to the annotation guidelines and the task definition.
We would like to acknowledge Hanxiao Jiang for helpful discussions and preliminary ideas on baseline implementation. Devansh Kukreja built the render flow to generate frame-aligned videos of each camera for each take as model input.}

\NEW{
\textbf{Ego-exo translation}  
Lorenzo Torresani co-led the translation benchmark, developed the task formulation, and advised the baseline development. Judy Hoffman co-led the translation benchmark and advised the baseline development. Feng Cheng led the baseline development and implemented the pix2pix and DiT models for track prediction and clip generation. Mi Luo implemented the GNT baseline and the evaluation pipeline for ego track prediction. Ziwei Zhao contributed the pix2pix baseline for multi-frame input and led the evaluation for ego clip generation. Huiyu Wang advised the baseline development and contributed to the task definition, the baseline design, and the metric selection and analysis.}

\NEW{
\textbf{Fine-grained keystep recognition}  
Tushar Nagarajan co-led the keystep recognition benchmark, co-developed the task formulation, and advised the baseline development.
{David Crandall co-led the keystep recognition benchmark and contributed to the task formulation.}
{}
Yale Song co-led the keystep recognition benchmark and led the keystep annotation effort, including design of annotation guidelines, taxonomy development and coordination of annotation workflows; he also contributed to the task formulation, advised baseline design, and facilitated the delivery of EgoVLPv2 pretrained backbone. Triantafyllos Afouras contributed to the taxonomy definition, managed the labeling effort, and developed software for post-processing the annotations.
Zihui Xue led the baseline development effort, implemented the TimeSFormer, EgoVLP, VI Encoder and Viewpoint distillation baselines, and performed analysis of results.
Eugene Byrne contributed to the taxonomy development and to the Ego-Exo Transfer baseline implementation and analysis.
Avijit Dasgupta contributed to the annotation and taxonomy development, and to the early stage of baseline design.
Miguel Martin contributed to the annotation and the taxonomy development.
Shraman Pramanick contributed the EgoVLPv2 pretrained backbone.
Yifei Huang contributed to the early stages of task definition and baseline design.
Devansh Kukreja built the render flow to generate frame-aligned videos of each camera for each take as model input, and 
to produce video collages for annotations. Kristen Grauman contributed to the task formulation.}

\NEW{
\textbf{Energy-efficient multimodal keystep recognition}  
Tushar Nagarajan led the energy-efficient multimodal benchmark, co-developed the task formulation, and advised the baseline development.
Sagnik Majumder led the baseline development effort and contributed all baseline implementations and analysis of results for the benchmark.
Merey Ramazanova developed the energy profiler used to evaluate all baselines and contributed to the experimental analysis. Mitesh Kumar Singh helped design the energy formula.
Miao Liu and Shengxin Cindy Zha initiated this benchmark and developed an early version of the task formulation.}

\NEW{
\textbf{Procedure understanding}
{Antonino Furnari} led the procedure understanding benchmark, and contributed to the task definition, the annotation guidelines, the baseline design, and paper writing.
{Giovanni Maria Farinella} contributed to the task definition, the annotation guidelines, the baseline design, and paper writing.
{Luigi Seminara} contributed to the annotation guidelines, the baseline design, and paper writing; he also developed tools for data annotation, and the baselines for the benchmark.
{Francesco Ragusa} contributed to the annotation guidelines, the baseline design, and paper writing; he also developed tools for data annotation, and the baselines for the benchmark.
{Kumar Ashutosh} contributed to the annotation guidelines, baseline design, and development of data annotation tools.
{Michael Wray} contributed to the task definition, the annotation guidelines, the baseline design, and paper writing.
{Siddhant Bansal} contributed to the task definition, the annotation guidelines, the baseline design, and paper writing.
{Gene Byrne} contributed to the task definition, the annotation guidelines and the baseline design.
{Tushar Nagarajan} contributed to the task definition, and the baseline design.}

\NEW{
\textbf{Ego-exo proficiency estimation}
Santhosh Kumar Ramakrishnan co-led the proficiency estimation benchmark, co-developed the task formulation, and advised the baseline development. Gedas Bertasius co-led the proficiency estimation benchmark, co-developed the task formulation, and advised the baseline development. Arjun Somayazulu developed the demonstrator proficiency estimation baselines. Abrham Gebreselasie developed the demonstration proficiency estimation baselines. Maria Escobar contributed to the task definition and the baseline design. Eugene Byrne contributed to the task definition and the baseline design. Miguel Martin developed an interface for obtaining proficiency estimation scores from the recruited experts. Suyog Jain contributed to an annotation pipeline for demonstration proficiency estimation. Devansh Kukreja built the render flow to generate frame-aligned videos of each camera for each take as model input. Kristen Grauman contributed to the task formulation. }

\NEW{
\textbf{Ego pose}
Kris Kitani co-led the ego-pose benchmark, and provided directional guidance on the task definition, the annotation methodology, and the baseline development. Jianbo Shi co-led the ego-pose benchmark, and provided directional guidance on automatic 3D hand pose generation and development of ego hand pose baseline methods. Maria Escobar led the ego-pose body baseline development, implemented the IMU-based baseline and contributed to experiment analysis. Cristhian Forigua developed the static pose baseline and contributed to the implementation of the IMU-based baseline. Fu-Jen Chu developed the multi-view annotation UI, the hand pose annotation guidelines, and the data preprocessing code for ego-pose annotation; he also trained and evaluated the HandOccNet baseline. Rawal Khirodkar developed the multi-view triangulation and 3D body keypoint estimation pipeline. Zhengyi Luo contributed to the Kinpoly baseline and to the coordinate transform for Aria head poses. Shan Su led the ego-pose hands baseline development, and contributed to automatic 3D hand pose generation, task definition, and annotation development; she also evaluated the baseline using METRO. Suyog Jain developed the annotation pipeline to scale the annotation collection, worked on training the annotators and managed the overall annotation process. Miguel Martin contributed to the automatic ground truth generation pipeline and provided high-level coordination of the body and hands automatic ground truth generation. Jinxu Zhang developed automatic ground truth generation for 3D hand pose; he also trained and evaluated baseline model POTTER. Yiming Huang trained and evaluated the baseline model THOR-net. Zhifan Zhu developed the METRO hand pose baseline method. Jing Huang led the automatic ground truth generation effort, refined body pose annotation guidelines, coordinated ego-pose body baseline development, and ran the EgoEgo body pose baseline. 
}

%% file: main.bbl

\begin{thebibliography}{222}
\ifx \bisbn   \undefined \def \bisbn  #1{ISBN #1}\fi
\ifx \binits  \undefined \def \binits#1{#1}\fi
\ifx \bauthor  \undefined \def \bauthor#1{#1}\fi
\ifx \batitle  \undefined \def \batitle#1{#1}\fi
\ifx \bjtitle  \undefined \def \bjtitle#1{#1}\fi
\ifx \bvolume  \undefined \def \bvolume#1{\textbf{#1}}\fi
\ifx \byear  \undefined \def \byear#1{#1}\fi
\ifx \bissue  \undefined \def \bissue#1{#1}\fi
\ifx \bfpage  \undefined \def \bfpage#1{#1}\fi
\ifx \blpage  \undefined \def \blpage #1{#1}\fi
\ifx \burl  \undefined \def \burl#1{\textsf{#1}}\fi
\ifx \doiurl  \undefined \def \doiurl#1{\url{https://doi.org/#1}}\fi
\ifx \betal  \undefined \def \betal{\textit{et al.}}\fi
\ifx \binstitute  \undefined \def \binstitute#1{#1}\fi
\ifx \binstitutionaled  \undefined \def \binstitutionaled#1{#1}\fi
\ifx \bctitle  \undefined \def \bctitle#1{#1}\fi
\ifx \beditor  \undefined \def \beditor#1{#1}\fi
\ifx \bpublisher  \undefined \def \bpublisher#1{#1}\fi
\ifx \bbtitle  \undefined \def \bbtitle#1{#1}\fi
\ifx \bedition  \undefined \def \bedition#1{#1}\fi
\ifx \bseriesno  \undefined \def \bseriesno#1{#1}\fi
\ifx \blocation  \undefined \def \blocation#1{#1}\fi
\ifx \bsertitle  \undefined \def \bsertitle#1{#1}\fi
\ifx \bsnm \undefined \def \bsnm#1{#1}\fi
\ifx \bsuffix \undefined \def \bsuffix#1{#1}\fi
\ifx \bparticle \undefined \def \bparticle#1{#1}\fi
\ifx \barticle \undefined \def \barticle#1{#1}\fi
\bibcommenthead
\ifx \bconfdate \undefined \def \bconfdate #1{#1}\fi
\ifx \botherref \undefined \def \botherref #1{#1}\fi
\ifx \url \undefined \def \url#1{\textsf{#1}}\fi
\ifx \bchapter \undefined \def \bchapter#1{#1}\fi
\ifx \bbook \undefined \def \bbook#1{#1}\fi
\ifx \bcomment \undefined \def \bcomment#1{#1}\fi
\ifx \oauthor \undefined \def \oauthor#1{#1}\fi
\ifx \citeauthoryear \undefined \def \citeauthoryear#1{#1}\fi
\ifx \endbibitem  \undefined \def \endbibitem {}\fi
\ifx \bconflocation  \undefined \def \bconflocation#1{#1}\fi
\ifx \arxivurl  \undefined \def \arxivurl#1{\textsf{#1}}\fi
\csname PreBibitemsHook\endcsname

\bibitem[\protect\citeauthoryear{Flavell et~al.}{1981}]{flavell}
\begin{botherref}
\oauthor{\bsnm{Flavell}, \binits{J.H.}},
\oauthor{\bsnm{Flavell}, \binits{E.R.}},
\oauthor{\bsnm{Green}, \binits{F.L.}},
\oauthor{\bsnm{Wilcox}, \binits{S.A.}}:
The development of three spatial perspective-taking rules.
Child Development
(1981)
\end{botherref}
\endbibitem

\bibitem[\protect\citeauthoryear{Newcombe}{1989}]{newcombe}
\begin{botherref}
\oauthor{\bsnm{Newcombe}, \binits{N.}}:
The development of spatial perspective taking.
Advances in child development and behavior
(1989)
\end{botherref}
\endbibitem

\bibitem[\protect\citeauthoryear{Sigurdsson et~al.}{2018}]{sigurdsson2018charades}
\begin{botherref}
\oauthor{\bsnm{Sigurdsson}, \binits{G.A.}},
\oauthor{\bsnm{Gupta}, \binits{A.}},
\oauthor{\bsnm{Schmid}, \binits{C.}},
\oauthor{\bsnm{Farhadi}, \binits{A.}},
\oauthor{\bsnm{Alahari}, \binits{K.}}:
Charades-ego: A large-scale dataset of paired third and first person videos.
arXiv preprint arXiv:1804.09626
(2018)
\end{botherref}
\endbibitem

\bibitem[\protect\citeauthoryear{Sener et~al.}{2022}]{sener2022assembly101}
\begin{bchapter}
\bauthor{\bsnm{Sener}, \binits{F.}},
\bauthor{\bsnm{Chatterjee}, \binits{D.}},
\bauthor{\bsnm{Shelepov}, \binits{D.}},
\bauthor{\bsnm{He}, \binits{K.}},
\bauthor{\bsnm{Singhania}, \binits{D.}},
\bauthor{\bsnm{Wang}, \binits{R.}},
\bauthor{\bsnm{Yao}, \binits{A.}}:
\bctitle{Assembly101: A large-scale multi-view video dataset for understanding procedural activities}.
In: \bbtitle{Proceedings of the IEEE/CVF Conference on Computer Vision and Pattern Recognition},
pp. \bfpage{21096}--\blpage{21106}
(\byear{2022})
\end{bchapter}
\endbibitem

\bibitem[\protect\citeauthoryear{Kwon et~al.}{2021}]{Kwon_2021_ICCV}
\begin{bchapter}
\bauthor{\bsnm{Kwon}, \binits{T.}},
\bauthor{\bsnm{Tekin}, \binits{B.}},
\bauthor{\bsnm{St\"uhmer}, \binits{J.}},
\bauthor{\bsnm{Bogo}, \binits{F.}},
\bauthor{\bsnm{Pollefeys}, \binits{M.}}:
\bctitle{H2o: Two hands manipulating objects for first person interaction recognition}.
In: \bbtitle{Proceedings of the IEEE/CVF International Conference on Computer Vision (ICCV)},
pp. \bfpage{10138}--\blpage{10148}
(\byear{2021})
\end{bchapter}
\endbibitem

\bibitem[\protect\citeauthoryear{la~Torre et~al.}{2009}]{cmu-kitchens}
\begin{bchapter}
\bauthor{\bsnm{Torre}, \binits{F.D.}},
\bauthor{\bsnm{Hodgins}, \binits{J.}},
\bauthor{\bsnm{Montano}, \binits{J.}},
\bauthor{\bsnm{Valcarcel}, \binits{S.}},
\bauthor{\bsnm{Forcada}, \binits{R.}},
\bauthor{\bsnm{Macey}, \binits{J.}}:
\bctitle{Guide to the carnegie mellon university multimodal activity (cmu-mmac) database}.
In: \bbtitle{Tech. Report CMU-RI-TR-08-22, Robotics Institute, Carnegie Mellon University}
(\byear{2009})
\end{bchapter}
\endbibitem

\bibitem[\protect\citeauthoryear{Rai et~al.}{2021}]{homage}
\begin{bchapter}
\bauthor{\bsnm{Rai}, \binits{N.}},
\bauthor{\bsnm{Chen}, \binits{H.}},
\bauthor{\bsnm{Ji}, \binits{J.}},
\bauthor{\bsnm{Desai}, \binits{R.}},
\bauthor{\bsnm{Kozuka}, \binits{K.}},
\bauthor{\bsnm{Ishizaka}, \binits{S.}},
\bauthor{\bsnm{Adeli}, \binits{E.}},
\bauthor{\bsnm{Niebles}, \binits{J.C.}}:
\bctitle{Home action genome: Contrastive compositional action understanding}.
In: \bbtitle{CVPR}
(\byear{2021})
\end{bchapter}
\endbibitem

\bibitem[\protect\citeauthoryear{Damen et~al.}{2021}]{Damen2020RESCALING}
\begin{botherref}
\oauthor{\bsnm{Damen}, \binits{D.}},
\oauthor{\bsnm{Doughty}, \binits{H.}},
\oauthor{\bsnm{Farinella}, \binits{G.M.}},
\oauthor{},
\oauthor{\bsnm{Furnari}, \binits{A.}},
\oauthor{\bsnm{Ma}, \binits{J.}},
\oauthor{\bsnm{Kazakos}, \binits{E.}},
\oauthor{\bsnm{Moltisanti}, \binits{D.}},
\oauthor{\bsnm{Munro}, \binits{J.}},
\oauthor{\bsnm{Perrett}, \binits{T.}},
\oauthor{\bsnm{Price}, \binits{W.}},
\oauthor{\bsnm{Wray}, \binits{M.}}:
Rescaling egocentric vision.
IJCV
(2021)
\end{botherref}
\endbibitem

\bibitem[\protect\citeauthoryear{Grauman et~al.}{2022}]{grauman2022ego4d}
\begin{bchapter}
\bauthor{\bsnm{Grauman}, \binits{K.}},
\bauthor{\bsnm{Westbury}, \binits{A.}},
\bauthor{\bsnm{Byrne}, \binits{E.}},
\bauthor{\bsnm{Chavis}, \binits{Z.}},
\bauthor{\bsnm{Furnari}, \binits{A.}},
\bauthor{\bsnm{Girdhar}, \binits{R.}},
\bauthor{\bsnm{Hamburger}, \binits{J.}},
\bauthor{\bsnm{Jiang}, \binits{H.}},
\bauthor{\bsnm{Liu}, \binits{M.}},
\bauthor{\bsnm{Liu}, \binits{X.}},
\bauthor{\bsnm{Martin}, \binits{M.}},
\bauthor{\bsnm{Nagarajan}, \binits{T.}},
\bauthor{\bsnm{Radosavovic}, \binits{I.}},
\bauthor{\bsnm{Ramakrishnan}, \binits{S.K.}},
\bauthor{\bsnm{Ryan}, \binits{F.}},
\bauthor{\bsnm{Sharma}, \binits{J.}},
\bauthor{\bsnm{Wray}, \binits{M.}},
\bauthor{\bsnm{Xu}, \binits{M.}},
\bauthor{\bsnm{Xu}, \binits{E.Z.}},
\bauthor{\bsnm{Zhao}, \binits{C.}},
\bauthor{\bsnm{Bansal}, \binits{S.}},
\bauthor{\bsnm{Batra}, \binits{D.}},
\bauthor{\bsnm{Cartillier}, \binits{V.}},
\bauthor{\bsnm{Crane}, \binits{S.}},
\bauthor{\bsnm{Do}, \binits{T.}},
\bauthor{\bsnm{Doulaty}, \binits{M.}},
\bauthor{\bsnm{Erapalli}, \binits{A.}},
\bauthor{\bsnm{Feichtenhofer}, \binits{C.}},
\bauthor{\bsnm{Fragomeni}, \binits{A.}},
\bauthor{\bsnm{Fu}, \binits{Q.}},
\bauthor{\bsnm{Gebreselasie}, \binits{A.}},
\bauthor{\bsnm{Gonz\'alez}, \binits{C.}},
\bauthor{\bsnm{Hillis}, \binits{J.}},
\bauthor{\bsnm{Huang}, \binits{X.}},
\bauthor{\bsnm{Huang}, \binits{Y.}},
\bauthor{\bsnm{Jia}, \binits{W.}},
\bauthor{\bsnm{Khoo}, \binits{W.}},
\bauthor{\bsnm{Kol\'a\v{r}}, \binits{J.}},
\bauthor{\bsnm{Kottur}, \binits{S.}},
\bauthor{\bsnm{Kumar}, \binits{A.}},
\bauthor{\bsnm{Landini}, \binits{F.}},
\bauthor{\bsnm{Li}, \binits{C.}},
\bauthor{\bsnm{Li}, \binits{Y.}},
\bauthor{\bsnm{Li}, \binits{Z.}},
\bauthor{\bsnm{Mangalam}, \binits{K.}},
\bauthor{\bsnm{Modhugu}, \binits{R.}},
\bauthor{\bsnm{Munro}, \binits{J.}},
\bauthor{\bsnm{Murrell}, \binits{T.}},
\bauthor{\bsnm{Nishiyasu}, \binits{T.}},
\bauthor{\bsnm{Price}, \binits{W.}},
\bauthor{\bsnm{Ruiz}, \binits{P.}},
\bauthor{\bsnm{Ramazanova}, \binits{M.}},
\bauthor{\bsnm{Sari}, \binits{L.}},
\bauthor{\bsnm{Somasundaram}, \binits{K.}},
\bauthor{\bsnm{Southerland}, \binits{A.}},
\bauthor{\bsnm{Sugano}, \binits{Y.}},
\bauthor{\bsnm{Tao}, \binits{R.}},
\bauthor{\bsnm{Vo}, \binits{M.}},
\bauthor{\bsnm{Wang}, \binits{Y.}},
\bauthor{\bsnm{Wu}, \binits{X.}},
\bauthor{\bsnm{Yagi}, \binits{T.}},
\bauthor{\bsnm{Zhao}, \binits{Z.}},
\bauthor{\bsnm{Zhu}, \binits{Y.}},
\bauthor{\bsnm{Arbel\'aez}, \binits{P.}},
\bauthor{\bsnm{Crandall}, \binits{D.}},
\bauthor{\bsnm{Damen}, \binits{D.}},
\bauthor{\bsnm{Farinella}, \binits{G.M.}},
\bauthor{\bsnm{Fuegen}, \binits{C.}},
\bauthor{\bsnm{Ghanem}, \binits{B.}},
\bauthor{\bsnm{Ithapu}, \binits{V.K.}},
\bauthor{\bsnm{Jawahar}, \binits{C.V.}},
\bauthor{\bsnm{Joo}, \binits{H.}},
\bauthor{\bsnm{Kitani}, \binits{K.}},
\bauthor{\bsnm{Li}, \binits{H.}},
\bauthor{\bsnm{Newcombe}, \binits{R.}},
\bauthor{\bsnm{Oliva}, \binits{A.}},
\bauthor{\bsnm{Park}, \binits{H.S.}},
\bauthor{\bsnm{Rehg}, \binits{J.M.}},
\bauthor{\bsnm{Sato}, \binits{Y.}},
\bauthor{\bsnm{Shi}, \binits{J.}},
\bauthor{\bsnm{Shou}, \binits{M.Z.}},
\bauthor{\bsnm{Torralba}, \binits{A.}},
\bauthor{\bsnm{Torresani}, \binits{L.}},
\bauthor{\bsnm{Yan}, \binits{M.}},
\bauthor{\bsnm{Malik}, \binits{J.}}:
\bctitle{Ego4{D}: Around the world in 3,000 hours of egocentric video}.
In: \bbtitle{CVPR}
(\byear{2022})
\end{bchapter}
\endbibitem

\bibitem[\protect\citeauthoryear{Kay et~al.}{2017}]{kinetics}
\begin{botherref}
\oauthor{\bsnm{Kay}, \binits{W.}},
\oauthor{\bsnm{Carreira}, \binits{J.}},
\oauthor{\bsnm{Simonyan}, \binits{K.}},
\oauthor{\bsnm{Zhang}, \binits{B.}},
\oauthor{\bsnm{Hillier}, \binits{C.}},
\oauthor{\bsnm{Vijayanarasimhan}, \binits{S.}},
\oauthor{\bsnm{Viola}, \binits{F.}},
\oauthor{\bsnm{Green}, \binits{T.}},
\oauthor{\bsnm{Back}, \binits{T.}},
\oauthor{\bsnm{Natsev}, \binits{P.}}, et al.:
The kinetics human action video dataset.
arXiv preprint arXiv:1705.06950
(2017)
\end{botherref}
\endbibitem

\bibitem[\protect\citeauthoryear{Gu et~al.}{2018}]{ava}
\begin{bchapter}
\bauthor{\bsnm{Gu}, \binits{C.}},
\bauthor{\bsnm{Sun}, \binits{C.}},
\bauthor{\bsnm{Ross}, \binits{D.A.}},
\bauthor{\bsnm{Vondrick}, \binits{C.}},
\bauthor{\bsnm{Pantofaru}, \binits{C.}},
\bauthor{\bsnm{Li}, \binits{Y.}},
\bauthor{\bsnm{Vijayanarasimhan}, \binits{S.}},
\bauthor{\bsnm{Toderici}, \binits{G.}},
\bauthor{\bsnm{Ricco}, \binits{S.}},
\bauthor{\bsnm{Sukthankar}, \binits{R.}},
\bauthor{\bsnm{Schmid}, \binits{C.}},
\bauthor{\bsnm{Malik}, \binits{J.}}:
\bctitle{Ava: A video dataset of spatio-temporally localized atomic visual actions}.
In: \bbtitle{CVPR}
(\byear{2018})
\end{bchapter}
\endbibitem

\bibitem[\protect\citeauthoryear{Monfort et~al.}{2019}]{moments}
\begin{botherref}
\oauthor{\bsnm{Monfort}, \binits{M.}},
\oauthor{\bsnm{Andonian}, \binits{A.}},
\oauthor{\bsnm{Zhou}, \binits{B.}},
\oauthor{\bsnm{Ramakrishnan}, \binits{K.}},
\oauthor{\bsnm{Bargal}, \binits{S.A.}},
\oauthor{\bsnm{Yan}, \binits{T.}},
\oauthor{\bsnm{Brown}, \binits{L.}},
\oauthor{\bsnm{Fan}, \binits{Q.}},
\oauthor{\bsnm{Gutfreund}, \binits{D.}},
\oauthor{\bsnm{Vondrick}, \binits{C.}},
\oauthor{\bsnm{Oliva}, \binits{A.}}:
Moments in time dataset: one million videos for event understanding.
PAMI
(2019)
\end{botherref}
\endbibitem

\bibitem[\protect\citeauthoryear{Soomro et~al.}{2012}]{ucf}
\begin{bchapter}
\bauthor{\bsnm{Soomro}, \binits{K.}},
\bauthor{\bsnm{Zamir}, \binits{A.R.}},
\bauthor{\bsnm{Shah}, \binits{M.}}:
\bctitle{Ucf101: A dataset of 101 human action classes from videos in the wild}.
In: \bbtitle{CRCV-TR-12-01}
(\byear{2012})
\end{bchapter}
\endbibitem

\bibitem[\protect\citeauthoryear{Miech et~al.}{2019}]{miech19howto100m}
\begin{bchapter}
\bauthor{\bsnm{Miech}, \binits{A.}},
\bauthor{\bsnm{Zhukov}, \binits{D.}},
\bauthor{\bsnm{Alayrac}, \binits{J.-B.}},
\bauthor{\bsnm{Tapaswi}, \binits{M.}},
\bauthor{\bsnm{Laptev}, \binits{I.}},
\bauthor{\bsnm{Sivic}, \binits{J.}}:
\bctitle{How{T}o100{M}: {L}earning a {T}ext-{V}ideo {E}mbedding by {W}atching {H}undred {M}illion {N}arrated {V}ideo {C}lips}.
In: \bbtitle{ICCV}
(\byear{2019})
\end{bchapter}
\endbibitem

\bibitem[\protect\citeauthoryear{Tang et~al.}{2020}]{tang2020comprehensive}
\begin{botherref}
\oauthor{\bsnm{Tang}, \binits{Y.}},
\oauthor{\bsnm{Lu}, \binits{J.}},
\oauthor{\bsnm{Zhou}, \binits{J.}}:
Comprehensive instructional video analysis: The coin dataset and performance evaluation.
IEEE transactions on pattern analysis and machine intelligence
(2020)
\end{botherref}
\endbibitem

\bibitem[\protect\citeauthoryear{Zhukov et~al.}{2019}]{zhukov2019cross}
\begin{bchapter}
\bauthor{\bsnm{Zhukov}, \binits{D.}},
\bauthor{\bsnm{Alayrac}, \binits{J.-B.}},
\bauthor{\bsnm{Cinbis}, \binits{R.G.}},
\bauthor{\bsnm{Fouhey}, \binits{D.}},
\bauthor{\bsnm{Laptev}, \binits{I.}},
\bauthor{\bsnm{Sivic}, \binits{J.}}:
\bctitle{Cross-task weakly supervised learning from instructional videos}.
In: \bbtitle{Proceedings of the IEEE/CVF Conference on Computer Vision and Pattern Recognition}
(\byear{2019})
\end{bchapter}
\endbibitem

\bibitem[\protect\citeauthoryear{Zhou et~al.}{2018}]{youcook2}
\begin{bchapter}
\bauthor{\bsnm{Zhou}, \binits{L.}},
\bauthor{\bsnm{Louis}, \binits{N.}},
\bauthor{\bsnm{Corso}, \binits{J.}}:
\bctitle{Weakly-supervised video object grounding from text by loss weighting and object interaction}.
In: \bbtitle{BMVC}
(\byear{2018})
\end{bchapter}
\endbibitem

\bibitem[\protect\citeauthoryear{Engel et~al.}{2023}]{aria_2023}
\begin{botherref}
\oauthor{\bsnm{Engel}, \binits{J.}},
\oauthor{\bsnm{Somasundaram}, \binits{K.}},
\oauthor{\bsnm{Goesele}, \binits{M.}},
\oauthor{\bsnm{Sun}, \binits{A.}},
\oauthor{\bsnm{Gamino}, \binits{A.}},
\oauthor{\bsnm{Turner}, \binits{A.}},
\oauthor{\bsnm{Talattof}, \binits{A.}},
\oauthor{\bsnm{Yuan}, \binits{A.}},
\oauthor{\bsnm{Souti}, \binits{B.}},
\oauthor{\bsnm{Meredith}, \binits{B.}},
\oauthor{\bsnm{Peng}, \binits{C.}},
\oauthor{\bsnm{Sweeney}, \binits{C.}},
\oauthor{\bsnm{Wilson}, \binits{C.}},
\oauthor{\bsnm{Barnes}, \binits{D.}},
\oauthor{\bsnm{DeTone}, \binits{D.}},
\oauthor{\bsnm{Caruso}, \binits{D.}},
\oauthor{\bsnm{Valleroy}, \binits{D.}},
\oauthor{\bsnm{Ginjupalli}, \binits{D.}},
\oauthor{\bsnm{Frost}, \binits{D.}},
\oauthor{\bsnm{Miller}, \binits{E.}},
\oauthor{\bsnm{Mueggler}, \binits{E.}},
\oauthor{\bsnm{Oleinik}, \binits{E.}},
\oauthor{\bsnm{Zhang}, \binits{F.}},
\oauthor{\bsnm{Somasundaram}, \binits{G.}},
\oauthor{\bsnm{Solaira}, \binits{G.}},
\oauthor{\bsnm{Lanaras}, \binits{H.}},
\oauthor{\bsnm{Howard-Jenkins}, \binits{H.}},
\oauthor{\bsnm{Tang}, \binits{H.}},
\oauthor{\bsnm{Kim}, \binits{H.J.}},
\oauthor{\bsnm{Rivera}, \binits{J.}},
\oauthor{\bsnm{Luo}, \binits{J.}},
\oauthor{\bsnm{Dong}, \binits{J.}},
\oauthor{\bsnm{Straub}, \binits{J.}},
\oauthor{\bsnm{Bailey}, \binits{K.}},
\oauthor{\bsnm{Eckenhoff}, \binits{K.}},
\oauthor{\bsnm{Ma}, \binits{L.}},
\oauthor{\bsnm{Pesqueira}, \binits{L.}},
\oauthor{\bsnm{Schwesinger}, \binits{M.}},
\oauthor{\bsnm{Monge}, \binits{M.}},
\oauthor{\bsnm{Yang}, \binits{N.}},
\oauthor{\bsnm{Charron}, \binits{N.}},
\oauthor{\bsnm{Raina}, \binits{N.}},
\oauthor{\bsnm{Parkhi}, \binits{O.}},
\oauthor{\bsnm{Borschowa}, \binits{P.}},
\oauthor{\bsnm{Moulon}, \binits{P.}},
\oauthor{\bsnm{Gupta}, \binits{P.}},
\oauthor{\bsnm{Mur-Artal}, \binits{R.}},
\oauthor{\bsnm{Pennington}, \binits{R.}},
\oauthor{\bsnm{Kulkarni}, \binits{S.}},
\oauthor{\bsnm{Miglani}, \binits{S.}},
\oauthor{\bsnm{Gondi}, \binits{S.}},
\oauthor{\bsnm{Solanki}, \binits{S.}},
\oauthor{\bsnm{Diener}, \binits{S.}},
\oauthor{\bsnm{Cheng}, \binits{S.}},
\oauthor{\bsnm{Green}, \binits{S.}},
\oauthor{\bsnm{Saarinen}, \binits{S.}},
\oauthor{\bsnm{Patra}, \binits{S.}},
\oauthor{\bsnm{Mourikis}, \binits{T.}},
\oauthor{\bsnm{Whelan}, \binits{T.}},
\oauthor{\bsnm{Singh}, \binits{T.}},
\oauthor{\bsnm{Balntas}, \binits{V.}},
\oauthor{\bsnm{Baiyya}, \binits{V.}},
\oauthor{\bsnm{Dreewes}, \binits{W.}},
\oauthor{\bsnm{Pan}, \binits{X.}},
\oauthor{\bsnm{Lou}, \binits{Y.}},
\oauthor{\bsnm{Zhao}, \binits{Y.}},
\oauthor{\bsnm{Mansour}, \binits{Y.}},
\oauthor{\bsnm{Zou}, \binits{Y.}},
\oauthor{\bsnm{Lv}, \binits{Z.}},
\oauthor{\bsnm{Wang}, \binits{Z.}},
\oauthor{\bsnm{Yan}, \binits{M.}},
\oauthor{\bsnm{Ren}, \binits{C.}},
\oauthor{\bsnm{Nardi}, \binits{R.D.}},
\oauthor{\bsnm{Newcombe}, \binits{R.}}:
{P}roject {A}ria: A New Tool for Egocentric Multi-Modal {AI} Research
(2023)
\end{botherref}
\endbibitem

\bibitem[\protect\citeauthoryear{Li et~al.}{2018}]{li2018eye}
\begin{bchapter}
\bauthor{\bsnm{Li}, \binits{Y.}},
\bauthor{\bsnm{Liu}, \binits{M.}},
\bauthor{\bsnm{Rehg}, \binits{J.M.}}:
\bctitle{In the eye of beholder: Joint learning of gaze and actions in first person video}.
In: \bbtitle{ECCV}
(\byear{2018})
\end{bchapter}
\endbibitem

\bibitem[\protect\citeauthoryear{Ragusa et~al.}{2021}]{ragusa2021meccano}
\begin{bchapter}
\bauthor{\bsnm{Ragusa}, \binits{F.}},
\bauthor{\bsnm{Furnari}, \binits{A.}},
\bauthor{\bsnm{Livatino}, \binits{S.}},
\bauthor{\bsnm{Farinella}, \binits{G.M.}}:
\bctitle{The meccano dataset: Understanding human-object interactions from egocentric videos in an industrial-like domain}.
In: \bbtitle{WACV}
(\byear{2021})
\end{bchapter}
\endbibitem

\bibitem[\protect\citeauthoryear{Damen et~al.}{2022}]{epic-kitchens-100}
\begin{barticle}
\bauthor{\bsnm{Damen}, \binits{D.}},
\bauthor{\bsnm{Doughty}, \binits{H.}},
\bauthor{\bsnm{Farinella}, \binits{G.M.}},
\bauthor{\bsnm{Furnari}, \binits{A.}},
\bauthor{\bsnm{Kazakos}, \binits{E.}},
\bauthor{\bsnm{Ma}, \binits{J.}},
\bauthor{\bsnm{Moltisanti}, \binits{D.}},
\bauthor{\bsnm{Munro}, \binits{J.}},
\bauthor{\bsnm{Perrett}, \binits{T.}},
\bauthor{\bsnm{Price}, \binits{W.}}, \betal:
\batitle{Rescaling egocentric vision: collection, pipeline and challenges for epic-kitchens-100}.
\bjtitle{International Journal of Computer Vision}
\bvolume{130}(\bissue{1}),
\bfpage{33}--\blpage{55}
(\byear{2022})
\end{barticle}
\endbibitem

\bibitem[\protect\citeauthoryear{Wang et~al.}{2023}]{wang2023holoassist}
\begin{bchapter}
\bauthor{\bsnm{Wang}, \binits{X.}},
\bauthor{\bsnm{Kwon}, \binits{T.}},
\bauthor{\bsnm{Rad}, \binits{M.}},
\bauthor{\bsnm{Pan}, \binits{B.}},
\bauthor{\bsnm{Chakraborty}, \binits{I.}},
\bauthor{\bsnm{Andrist}, \binits{S.}},
\bauthor{\bsnm{Bohus}, \binits{D.}},
\bauthor{\bsnm{Feniello}, \binits{A.}},
\bauthor{\bsnm{Tekin}, \binits{B.}},
\bauthor{\bsnm{Frujeri}, \binits{F.V.}}, \betal:
\bctitle{Holoassist: an egocentric human interaction dataset for interactive ai assistants in the real world}.
In: \bbtitle{Proceedings of the IEEE/CVF International Conference on Computer Vision},
pp. \bfpage{20270}--\blpage{20281}
(\byear{2023})
\end{bchapter}
\endbibitem

\bibitem[\protect\citeauthoryear{Weinland et~al.}{2006}]{ixmas}
\begin{botherref}
\oauthor{\bsnm{Weinland}, \binits{D.}},
\oauthor{\bsnm{Ronfard}, \binits{R.}},
\oauthor{\bsnm{Boyer}, \binits{E.}}:
Free viewpoint action recognition using motion history volumes.
Computer Vision and Image Understanding (CVIU)
(2006)
\end{botherref}
\endbibitem

\bibitem[\protect\citeauthoryear{Corona et~al.}{2021}]{meva}
\begin{bchapter}
\bauthor{\bsnm{Corona}, \binits{K.}},
\bauthor{\bsnm{Osterdahl}, \binits{K.}},
\bauthor{\bsnm{Collins}, \binits{R.}},
\bauthor{\bsnm{Hoogs}, \binits{A.}}:
\bctitle{Meva: A large-scale multiview, multimodal video dataset for activity detection}.
In: \bbtitle{WACV}
(\byear{2021})
\end{bchapter}
\endbibitem

\bibitem[\protect\citeauthoryear{Sigurdsson et~al.}{2018}]{sigurdsson2018actor}
\begin{bchapter}
\bauthor{\bsnm{Sigurdsson}, \binits{G.A.}},
\bauthor{\bsnm{Gupta}, \binits{A.}},
\bauthor{\bsnm{Schmid}, \binits{C.}},
\bauthor{\bsnm{Farhadi}, \binits{A.}},
\bauthor{\bsnm{Alahari}, \binits{K.}}:
\bctitle{Actor and observer: Joint modeling of first and third-person videos}.
In: \bbtitle{CVPR}
(\byear{2018})
\end{bchapter}
\endbibitem

\bibitem[\protect\citeauthoryear{Jia et~al.}{2020}]{jia2020lemma}
\begin{bchapter}
\bauthor{\bsnm{Jia}, \binits{B.}},
\bauthor{\bsnm{Chen}, \binits{Y.}},
\bauthor{\bsnm{Huang}, \binits{S.}},
\bauthor{\bsnm{Zhu}, \binits{Y.}},
\bauthor{\bsnm{Zhu}, \binits{S.-c.}}:
\bctitle{Lemma: A multi-view dataset for le arning m ulti-agent m ulti-task a ctivities}.
In: \bbtitle{European Conference on Computer Vision},
pp. \bfpage{767}--\blpage{786}
(\byear{2020}).
\bcomment{Springer}
\end{bchapter}
\endbibitem

\bibitem[\protect\citeauthoryear{Ohkawa et~al.}{2023}]{ohkawa2023assemblyhands}
\begin{bchapter}
\bauthor{\bsnm{Ohkawa}, \binits{T.}},
\bauthor{\bsnm{He}, \binits{K.}},
\bauthor{\bsnm{Sener}, \binits{F.}},
\bauthor{\bsnm{Hodan}, \binits{T.}},
\bauthor{\bsnm{Tran}, \binits{L.}},
\bauthor{\bsnm{Keskin}, \binits{C.}}:
\bctitle{Assemblyhands: Towards egocentric activity understanding via 3d hand pose estimation}.
In: \bbtitle{Proceedings of the IEEE/CVF Conference on Computer Vision and Pattern Recognition},
pp. \bfpage{12999}--\blpage{13008}
(\byear{2023})
\end{bchapter}
\endbibitem

\bibitem[\protect\citeauthoryear{Huang et~al.}{2024}]{huang2024egoexolearn}
\begin{bchapter}
\bauthor{\bsnm{Huang}, \binits{Y.}},
\bauthor{\bsnm{Chen}, \binits{G.}},
\bauthor{\bsnm{Xu}, \binits{J.}},
\bauthor{\bsnm{Zhang}, \binits{M.}},
\bauthor{\bsnm{Yang}, \binits{L.}},
\bauthor{\bsnm{Pei}, \binits{B.}},
\bauthor{\bsnm{Zhang}, \binits{H.}},
\bauthor{\bsnm{Dong}, \binits{L.}},
\bauthor{\bsnm{Wang}, \binits{Y.}},
\bauthor{\bsnm{Wang}, \binits{L.}}, \betal:
\bctitle{Egoexolearn: A dataset for bridging asynchronous ego-and exo-centric view of procedural activities in real world}.
In: \bbtitle{Proceedings of the IEEE/CVF Conference on Computer Vision and Pattern Recognition},
pp. \bfpage{22072}--\blpage{22086}
(\byear{2024})
\end{bchapter}
\endbibitem

\bibitem[\protect\citeauthoryear{Damen et~al.}{2018}]{Damen2018EPICKITCHENS}
\begin{bchapter}
\bauthor{\bsnm{Damen}, \binits{D.}},
\bauthor{\bsnm{Doughty}, \binits{H.}},
\bauthor{\bsnm{Farinella}, \binits{G.M.}},
\bauthor{\bsnm{Fidler}, \binits{S.}},
\bauthor{\bsnm{Furnari}, \binits{A.}},
\bauthor{\bsnm{Kazakos}, \binits{E.}},
\bauthor{\bsnm{Moltisanti}, \binits{D.}},
\bauthor{\bsnm{Munro}, \binits{J.}},
\bauthor{\bsnm{Perrett}, \binits{T.}},
\bauthor{\bsnm{Price}, \binits{W.}},
\bauthor{\bsnm{Wray}, \binits{M.}}:
\bctitle{Scaling egocentric vision: The epic-kitchens dataset}.
In: \bbtitle{European Conference on Computer Vision (ECCV)}
(\byear{2018})
\end{bchapter}
\endbibitem

\bibitem[\protect\citeauthoryear{Tschernezki et~al.}{2023}]{EPICFields2023}
\begin{bchapter}
\bauthor{\bsnm{Tschernezki}, \binits{V.}},
\bauthor{\bsnm{Darkhalil}, \binits{A.}},
\bauthor{\bsnm{Zhu}, \binits{Z.}},
\bauthor{\bsnm{Fouhey}, \binits{D.}},
\bauthor{\bsnm{Larina}, \binits{I.}},
\bauthor{\bsnm{Larlus}, \binits{D.}},
\bauthor{\bsnm{Damen}, \binits{D.}},
\bauthor{\bsnm{Vedaldi}, \binits{A.}}:
\bctitle{{EPIC Fields}: {M}arrying {3D} {G}eometry and {V}ideo {U}nderstanding}.
In: \bbtitle{Proceedings of the Neural Information Processing Systems (NeurIPS)}
(\byear{2023})
\end{bchapter}
\endbibitem

\bibitem[\protect\citeauthoryear{Lee et~al.}{2012}]{lee-cvpr2012}
\begin{bchapter}
\bauthor{\bsnm{Lee}, \binits{Y.J.}},
\bauthor{\bsnm{Ghosh}, \binits{J.}},
\bauthor{\bsnm{Grauman}, \binits{K.}}:
\bctitle{Discovering important people and objects for egocentric video summarization}.
In: \bbtitle{CVPR}
(\byear{2012})
\end{bchapter}
\endbibitem

\bibitem[\protect\citeauthoryear{Pirsiavash and Ramanan}{2012}]{pirsiavash2012detecting}
\begin{bchapter}
\bauthor{\bsnm{Pirsiavash}, \binits{H.}},
\bauthor{\bsnm{Ramanan}, \binits{D.}}:
\bctitle{Detecting activities of daily living in first-person camera views}.
In: \bbtitle{CVPR}
(\byear{2012})
\end{bchapter}
\endbibitem

\bibitem[\protect\citeauthoryear{Singh et~al.}{2016}]{krishnacam}
\begin{bchapter}
\bauthor{\bsnm{Singh}, \binits{K.K.}},
\bauthor{\bsnm{Fatahalian}, \binits{K.}},
\bauthor{\bsnm{Efros}, \binits{A.A.}}:
\bctitle{Krishnacam: Using a longitudinal, single-person, egocentric dataset for scene understanding tasks}.
In: \bbtitle{WACV}
(\byear{2016})
\end{bchapter}
\endbibitem

\bibitem[\protect\citeauthoryear{Wong et~al.}{2022}]{wong2022assistq}
\begin{bchapter}
\bauthor{\bsnm{Wong}, \binits{B.}},
\bauthor{\bsnm{Chen}, \binits{J.}},
\bauthor{\bsnm{Wu}, \binits{Y.}},
\bauthor{\bsnm{Lei}, \binits{S.W.}},
\bauthor{\bsnm{Mao}, \binits{D.}},
\bauthor{\bsnm{Gao}, \binits{D.}},
\bauthor{\bsnm{Shou}, \binits{M.Z.}}:
\bctitle{Assistq: Affordance-centric question-driven task completion for egocentric assistant}.
In: \bbtitle{European Conference on Computer Vision}
(\byear{2022})
\end{bchapter}
\endbibitem

\bibitem[\protect\citeauthoryear{Bansal et~al.}{2022}]{EgoProceLECCV2022}
\begin{bchapter}
\bauthor{\bsnm{Bansal}, \binits{S.}},
\bauthor{\bsnm{Arora}, \binits{C.}},
\bauthor{\bsnm{Jawahar}, \binits{C.V.}}:
\bctitle{My view is the best view: Procedure learning from egocentric videos}.
In: \bbtitle{European Conference on Computer Vision (ECCV)}
(\byear{2022})
\end{bchapter}
\endbibitem

\bibitem[\protect\citeauthoryear{Chang et~al.}{2017}]{matterport3d}
\begin{bchapter}
\bauthor{\bsnm{Chang}, \binits{A.}},
\bauthor{\bsnm{Dai}, \binits{A.}},
\bauthor{\bsnm{Funkhouser}, \binits{T.}},
\bauthor{\bsnm{Nie{\ss}ner}, \binits{M.}},
\bauthor{\bsnm{Savva}, \binits{M.}},
\bauthor{\bsnm{Song}, \binits{S.}},
\bauthor{\bsnm{Zeng}, \binits{A.}},
\bauthor{\bsnm{Zhang}, \binits{Y.}}:
\bctitle{Matterport3d: Learning from rgb-d data in indoor environments}.
In: \bbtitle{Proceedings of the International Conference on 3D Vision (3DV)}
(\byear{2017}).
\bcomment{MatterPort3D dataset license available at: \url{http://kaldir.vc.in.tum.de/matterport/MP_TOS.pdf}.}
\end{bchapter}
\endbibitem

\bibitem[\protect\citeauthoryear{Xia et~al.}{2018}]{gibson}
\begin{bchapter}
\bauthor{\bsnm{Xia}, \binits{F.}},
\bauthor{\bsnm{R.~Zamir}, \binits{A.}},
\bauthor{\bsnm{He}, \binits{Z.-Y.}},
\bauthor{\bsnm{Sax}, \binits{A.}},
\bauthor{\bsnm{Malik}, \binits{J.}},
\bauthor{\bsnm{Savarese}, \binits{S.}}:
\bctitle{Gibson {E}nv: real-world perception for embodied agents}.
In: \bbtitle{CVPR}
(\byear{2018}).
\bcomment{IEEE. Gibson license is available at \url{http://svl.stanford.edu/gibson2/assets/GDS_agreement.pdf}.}
\end{bchapter}
\endbibitem

\bibitem[\protect\citeauthoryear{Straub et~al.}{2019}]{replica19arxiv}
\begin{botherref}
\oauthor{\bsnm{Straub}, \binits{J.}},
\oauthor{\bsnm{Whelan}, \binits{T.}},
\oauthor{\bsnm{Ma}, \binits{L.}},
\oauthor{\bsnm{Chen}, \binits{Y.}},
\oauthor{\bsnm{Wijmans}, \binits{E.}},
\oauthor{\bsnm{Green}, \binits{S.}},
\oauthor{\bsnm{Engel}, \binits{J.J.}},
\oauthor{\bsnm{Mur-Artal}, \binits{R.}},
\oauthor{\bsnm{Ren}, \binits{C.}},
\oauthor{\bsnm{Verma}, \binits{S.}},
\oauthor{\bsnm{Clarkson}, \binits{A.}},
\oauthor{\bsnm{Yan}, \binits{M.}},
\oauthor{\bsnm{Budge}, \binits{B.}},
\oauthor{\bsnm{Yan}, \binits{Y.}},
\oauthor{\bsnm{Pan}, \binits{X.}},
\oauthor{\bsnm{Yon}, \binits{J.}},
\oauthor{\bsnm{Zou}, \binits{Y.}},
\oauthor{\bsnm{Leon}, \binits{K.}},
\oauthor{\bsnm{Carter}, \binits{N.}},
\oauthor{\bsnm{Briales}, \binits{J.}},
\oauthor{\bsnm{Gillingham}, \binits{T.}},
\oauthor{\bsnm{Mueggler}, \binits{E.}},
\oauthor{\bsnm{Pesqueira}, \binits{L.}},
\oauthor{\bsnm{Savva}, \binits{M.}},
\oauthor{\bsnm{Batra}, \binits{D.}},
\oauthor{\bsnm{Strasdat}, \binits{H.M.}},
\oauthor{\bsnm{Nardi}, \binits{R.D.}},
\oauthor{\bsnm{Goesele}, \binits{M.}},
\oauthor{\bsnm{Lovegrove}, \binits{S.}},
\oauthor{\bsnm{Newcombe}, \binits{R.}}:
The {R}eplica dataset: A digital replica of indoor spaces.
arXiv preprint arXiv:1906.05797
(2019)
\end{botherref}
\endbibitem

\bibitem[\protect\citeauthoryear{Ramakrishnan et~al.}{2021}]{ramakrishnan2021habitat}
\begin{bchapter}
\bauthor{\bsnm{Ramakrishnan}, \binits{S.K.}},
\bauthor{\bsnm{Gokaslan}, \binits{A.}},
\bauthor{\bsnm{Wijmans}, \binits{E.}},
\bauthor{\bsnm{Maksymets}, \binits{O.}},
\bauthor{\bsnm{Clegg}, \binits{A.}},
\bauthor{\bsnm{Turner}, \binits{J.M.}},
\bauthor{\bsnm{Undersander}, \binits{E.}},
\bauthor{\bsnm{Galuba}, \binits{W.}},
\bauthor{\bsnm{Westbury}, \binits{A.}},
\bauthor{\bsnm{Chang}, \binits{A.X.}},
\bauthor{\bsnm{Savva}, \binits{M.}},
\bauthor{\bsnm{Zhao}, \binits{Y.}},
\bauthor{\bsnm{Batra}, \binits{D.}}:
\bctitle{Habitat-matterport 3d dataset ({HM}3d): 1000 large-scale 3d environments for embodied {AI}}.
In: \bbtitle{Thirty-fifth Conference on Neural Information Processing Systems Datasets and Benchmarks Track}
(\byear{2021}).
\burl{https://arxiv.org/abs/2109.08238}
\end{bchapter}
\endbibitem

\bibitem[\protect\citeauthoryear{Xiao et~al.}{2013}]{sun3d}
\begin{bchapter}
\bauthor{\bsnm{Xiao}, \binits{J.}},
\bauthor{\bsnm{Owens}, \binits{A.}},
\bauthor{\bsnm{Torralba}, \binits{A.}}:
\bctitle{Sun3d: A database of big spaces reconstructed using sfm and object labels}.
In: \bbtitle{ICCV}
(\byear{2013})
\end{bchapter}
\endbibitem

\bibitem[\protect\citeauthoryear{Reizenstein et~al.}{2021}]{co3d}
\begin{bchapter}
\bauthor{\bsnm{Reizenstein}, \binits{J.}},
\bauthor{\bsnm{Shapovalov}, \binits{R.}},
\bauthor{\bsnm{Henzler}, \binits{P.}},
\bauthor{\bsnm{Sbordone}, \binits{L.}},
\bauthor{\bsnm{Labatut}, \binits{P.}},
\bauthor{\bsnm{Novotny}, \binits{D.}}:
\bctitle{Common objects in 3d: Large-scale learning and evaluation of real-life 3d category reconstruction}.
In: \bbtitle{ICCV}
(\byear{2021})
\end{bchapter}
\endbibitem

\bibitem[\protect\citeauthoryear{Wu et~al.}{2015}]{shapenet}
\begin{bchapter}
\bauthor{\bsnm{Wu}, \binits{Z.}},
\bauthor{\bsnm{Song}, \binits{S.}},
\bauthor{\bsnm{Khosla}, \binits{A.}},
\bauthor{\bsnm{Yu}, \binits{F.}},
\bauthor{\bsnm{Zhang}, \binits{L.}},
\bauthor{\bsnm{Tang}, \binits{X.}},
\bauthor{\bsnm{Xiao}, \binits{J.}}:
\bctitle{3d shapenets: A deep representation for volumetric shapes}.
In: \bbtitle{Computer Vision and Pattern Recognition, IEEE Conference On}
(\byear{2015})
\end{bchapter}
\endbibitem

\bibitem[\protect\citeauthoryear{Zhang et~al.}{2022}]{egobody}
\begin{bchapter}
\bauthor{\bsnm{Zhang}, \binits{S.}},
\bauthor{\bsnm{Ma}, \binits{Q.}},
\bauthor{\bsnm{Zhang}, \binits{Y.}},
\bauthor{\bsnm{Qian}, \binits{Z.}},
\bauthor{\bsnm{Kwon}, \binits{T.}},
\bauthor{\bsnm{Pollefeys}, \binits{M.}},
\bauthor{\bsnm{Bogo}, \binits{F.}},
\bauthor{\bsnm{Tang}, \binits{S.}}:
\bctitle{Egobody: Human body shape and motion of interacting people from head-mounted devices}.
In: \bbtitle{ECCV}
(\byear{2022})
\end{bchapter}
\endbibitem

\bibitem[\protect\citeauthoryear{Li et~al.}{2023}]{li2023ego}
\begin{bchapter}
\bauthor{\bsnm{Li}, \binits{J.}},
\bauthor{\bsnm{Liu}, \binits{K.}},
\bauthor{\bsnm{Wu}, \binits{J.}}:
\bctitle{Ego-body pose estimation via ego-head pose estimation}.
In: \bbtitle{Proceedings of the IEEE/CVF Conference on Computer Vision and Pattern Recognition},
pp. \bfpage{17142}--\blpage{17151}
(\byear{2023})
\end{bchapter}
\endbibitem

\bibitem[\protect\citeauthoryear{Joo et~al.}{2017}]{Joo_2017_TPAMI}
\begin{botherref}
\oauthor{\bsnm{Joo}, \binits{H.}},
\oauthor{\bsnm{Simon}, \binits{T.}},
\oauthor{\bsnm{Li}, \binits{X.}},
\oauthor{\bsnm{Liu}, \binits{H.}},
\oauthor{\bsnm{Tan}, \binits{L.}},
\oauthor{\bsnm{Gui}, \binits{L.}},
\oauthor{\bsnm{Banerjee}, \binits{S.}},
\oauthor{\bsnm{Godisart}, \binits{T.S.}},
\oauthor{\bsnm{Nabbe}, \binits{B.}},
\oauthor{\bsnm{Matthews}, \binits{I.}},
\oauthor{\bsnm{Kanade}, \binits{T.}},
\oauthor{\bsnm{Nobuhara}, \binits{S.}},
\oauthor{\bsnm{Sheikh}, \binits{Y.}}:
Panoptic studio: A massively multiview system for social interaction capture.
IEEE Transactions on Pattern Analysis and Machine Intelligence
(2017)
\end{botherref}
\endbibitem

\bibitem[\protect\citeauthoryear{Khirodkar et~al.}{2023}]{egohumans}
\begin{bchapter}
\bauthor{\bsnm{Khirodkar}, \binits{R.}},
\bauthor{\bsnm{Bansal}, \binits{A.}},
\bauthor{\bsnm{Ma}, \binits{L.}},
\bauthor{\bsnm{Newcombe}, \binits{R.}},
\bauthor{\bsnm{Vo}, \binits{M.}},
\bauthor{\bsnm{Kitani}, \binits{K.}}:
\bctitle{Egohumans: An egocentric 3d multi-human benchmark}.
In: \bbtitle{ICCV}
(\byear{2023})
\end{bchapter}
\endbibitem

\bibitem[\protect\citeauthoryear{Guzov et~al.}{2021}]{hps}
\begin{bchapter}
\bauthor{\bsnm{Guzov}, \binits{V.}},
\bauthor{\bsnm{Mir}, \binits{A.}},
\bauthor{\bsnm{Sattler}, \binits{T.}},
\bauthor{\bsnm{Pons-Moll}, \binits{G.}}:
\bctitle{Human poseitioning system (hps): 3d human pose estimation and self-localization in large scenes from body-mounted sensors}.
In: \bbtitle{CVPR}
(\byear{2021})
\end{bchapter}
\endbibitem

\bibitem[\protect\citeauthoryear{Pirsiavash et~al.}{2014}]{assessing-eccv2014}
\begin{bchapter}
\bauthor{\bsnm{Pirsiavash}, \binits{H.}},
\bauthor{\bsnm{Vondrick}, \binits{C.}},
\bauthor{\bsnm{Torralba}, \binits{A.}}:
\bctitle{Assessing the quality of actions}.
In: \bbtitle{ECCV}
(\byear{2014})
\end{bchapter}
\endbibitem

\bibitem[\protect\citeauthoryear{Bertasius et~al.}{2017}]{baller-iccv2017}
\begin{bchapter}
\bauthor{\bsnm{Bertasius}, \binits{G.}},
\bauthor{\bsnm{Park}, \binits{H.S.}},
\bauthor{\bsnm{Yu}, \binits{S.}},
\bauthor{\bsnm{Shi}, \binits{J.}}:
\bctitle{Am i a baller? basketball performance assessment from first-person videos}.
In: \bbtitle{ICCV}
(\byear{2017})
\end{bchapter}
\endbibitem

\bibitem[\protect\citeauthoryear{Parmar and Morris}{2019}]{action-quality-2019}
\begin{bchapter}
\bauthor{\bsnm{Parmar}, \binits{P.}},
\bauthor{\bsnm{Morris}, \binits{B.}}:
\bctitle{Action quality assessment across multiple actions}.
In: \bbtitle{WACV}
(\byear{2019})
\end{bchapter}
\endbibitem

\bibitem[\protect\citeauthoryear{Doughty et~al.}{2018}]{skill-determination}
\begin{bchapter}
\bauthor{\bsnm{Doughty}, \binits{H.}},
\bauthor{\bsnm{Damen}, \binits{D.}},
\bauthor{\bsnm{Mayol-Cuevas}, \binits{W.}}:
\bctitle{Who's better? who's best? pairwise deep ranking for skill determination}.
In: \bbtitle{CVPR}
(\byear{2018})
\end{bchapter}
\endbibitem

\bibitem[\protect\citeauthoryear{Doughty et~al.}{2019}]{Doughty_2019_CVPR}
\begin{botherref}
\oauthor{\bsnm{Doughty}, \binits{H.}},
\oauthor{\bsnm{Mayol-Cuevas}, \binits{W.}},
\oauthor{\bsnm{Damen}, \binits{D.}}:
{T}he {P}ros and {C}ons: {R}ank-aware {T}emporal {A}ttention for {S}kill {D}etermination in {L}ong {V}ideos
(2019)
\end{botherref}
\endbibitem

\bibitem[\protect\citeauthoryear{Zhang et~al.}{2023}]{logo}
\begin{bchapter}
\bauthor{\bsnm{Zhang}, \binits{S.}},
\bauthor{\bsnm{Dai}, \binits{W.}},
\bauthor{\bsnm{Wang}, \binits{S.}},
\bauthor{\bsnm{Shen}, \binits{X.}},
\bauthor{\bsnm{Lu}, \binits{J.}},
\bauthor{\bsnm{Zhou}, \binits{J.}},
\bauthor{\bsnm{Tang}, \binits{Y.}}:
\bctitle{Logo: A long-form video dataset for group action quality assessment}.
In: \bbtitle{CVPR}
(\byear{2023})
\end{bchapter}
\endbibitem

\bibitem[\protect\citeauthoryear{Ben-Shabat et~al.}{2020}]{ikea-assembly}
\begin{botherref}
\oauthor{\bsnm{Ben-Shabat}, \binits{Y.}},
\oauthor{\bsnm{Yu}, \binits{X.}},
\oauthor{\bsnm{Saleh}, \binits{F.}},
\oauthor{\bsnm{Campbell}, \binits{D.}},
\oauthor{\bsnm{Rodriguez-Opazo}, \binits{C.}},
\oauthor{\bsnm{Li}, \binits{H.}},
\oauthor{\bsnm{Gould}, \binits{S.}}:
The ikea asm dataset: Understanding people assembling furniture through actions, objects and pose
(2020)
\end{botherref}
\endbibitem

\bibitem[\protect\citeauthoryear{Elhamifar and Huynh}{2020}]{elhamifar2020self}
\begin{bchapter}
\bauthor{\bsnm{Elhamifar}, \binits{E.}},
\bauthor{\bsnm{Huynh}, \binits{D.}}:
\bctitle{Self-supervised multi-task procedure learning from instructional videos}.
In: \bbtitle{European Conference on Computer Vision},
pp. \bfpage{557}--\blpage{573}
(\byear{2020}).
\bcomment{Springer}
\end{bchapter}
\endbibitem

\bibitem[\protect\citeauthoryear{Xu et~al.}{2021}]{videoclip}
\begin{botherref}
\oauthor{\bsnm{Xu}, \binits{H.}},
\oauthor{\bsnm{Ghosh}, \binits{G.}},
\oauthor{\bsnm{Huang}, \binits{P.-Y.}},
\oauthor{\bsnm{Okhonko}, \binits{D.}},
\oauthor{\bsnm{Aghajanyan}, \binits{A.}},
\oauthor{\bsnm{Metze}, \binits{F.}},
\oauthor{\bsnm{Zettlemoyer}, \binits{L.}},
\oauthor{\bsnm{Feichtenhofer}, \binits{C.}}:
Videoclip: Contrastive pre-training for zero-shot video-text understanding.
arXiv preprint arXiv:2109.14084
(2021)
\end{botherref}
\endbibitem

\bibitem[\protect\citeauthoryear{Miech et~al.}{2020}]{mil-nce}
\begin{bchapter}
\bauthor{\bsnm{Miech}, \binits{A.}},
\bauthor{\bsnm{Alayrac}, \binits{J.-B.}},
\bauthor{\bsnm{Smaira}, \binits{L.}},
\bauthor{\bsnm{Laptev}, \binits{I.}},
\bauthor{\bsnm{Sivic}, \binits{J.}},
\bauthor{\bsnm{Zisserman}, \binits{A.}}:
\bctitle{End-to-end learning of visual representations from uncurated instructional videos}.
In: \bbtitle{Proceedings of the IEEE/CVF Conference on Computer Vision and Pattern Recognition},
pp. \bfpage{9879}--\blpage{9889}
(\byear{2020})
\end{bchapter}
\endbibitem

\bibitem[\protect\citeauthoryear{Dvornik et~al.}{2022}]{dvornik2022flow}
\begin{bchapter}
\bauthor{\bsnm{Dvornik}, \binits{N.}},
\bauthor{\bsnm{Hadji}, \binits{I.}},
\bauthor{\bsnm{Pham}, \binits{H.}},
\bauthor{\bsnm{Bhatt}, \binits{D.}},
\bauthor{\bsnm{Martinez}, \binits{B.}},
\bauthor{\bsnm{Fazly}, \binits{A.}},
\bauthor{\bsnm{Jepson}, \binits{A.D.}}:
\bctitle{Flow graph to video grounding for weakly-supervised multi-step localization}.
In: \bbtitle{ECCV},
pp. \bfpage{319}--\blpage{335}
(\byear{2022}).
\bcomment{Springer}
\end{bchapter}
\endbibitem

\bibitem[\protect\citeauthoryear{Lin et~al.}{2022}]{video-distant}
\begin{bchapter}
\bauthor{\bsnm{Lin}, \binits{X.}},
\bauthor{\bsnm{Petroni}, \binits{F.}},
\bauthor{\bsnm{Bertasius}, \binits{G.}},
\bauthor{\bsnm{Rohrbach}, \binits{M.}},
\bauthor{\bsnm{Chang}, \binits{S.-F.}},
\bauthor{\bsnm{Torresani}, \binits{L.}}:
\bctitle{Learning to recognize procedural activities with distant supervision}.
In: \bbtitle{Proceedings of the IEEE/CVF Conference on Computer Vision and Pattern Recognition},
pp. \bfpage{13853}--\blpage{13863}
(\byear{2022})
\end{bchapter}
\endbibitem

\bibitem[\protect\citeauthoryear{Chang et~al.}{2020}]{procedure-learning-fei-fei-li}
\begin{bchapter}
\bauthor{\bsnm{Chang}, \binits{C.-Y.}},
\bauthor{\bsnm{Huang}, \binits{D.-A.}},
\bauthor{\bsnm{Xu}, \binits{D.}},
\bauthor{\bsnm{Adeli}, \binits{E.}},
\bauthor{\bsnm{Fei-Fei}, \binits{L.}},
\bauthor{\bsnm{Niebles}, \binits{J.C.}}:
\bctitle{Procedure planning in instructional videos}.
In: \bbtitle{Computer Vision--ECCV 2020: 16th European Conference, Glasgow, UK, August 23--28, 2020, Proceedings, Part XI},
pp. \bfpage{334}--\blpage{350}
(\byear{2020}).
\bcomment{Springer}
\end{bchapter}
\endbibitem

\bibitem[\protect\citeauthoryear{Bi et~al.}{2021}]{procedure2}
\begin{bchapter}
\bauthor{\bsnm{Bi}, \binits{J.}},
\bauthor{\bsnm{Luo}, \binits{J.}},
\bauthor{\bsnm{Xu}, \binits{C.}}:
\bctitle{Procedure planning in instructional videos via contextual modeling and model-based policy learning}.
In: \bbtitle{Proceedings of the IEEE/CVF International Conference on Computer Vision},
pp. \bfpage{15611}--\blpage{15620}
(\byear{2021})
\end{bchapter}
\endbibitem

\bibitem[\protect\citeauthoryear{Zhao et~al.}{2022}]{p3iv}
\begin{bchapter}
\bauthor{\bsnm{Zhao}, \binits{H.}},
\bauthor{\bsnm{Hadji}, \binits{I.}},
\bauthor{\bsnm{Dvornik}, \binits{N.}},
\bauthor{\bsnm{Derpanis}, \binits{K.G.}},
\bauthor{\bsnm{Wildes}, \binits{R.P.}},
\bauthor{\bsnm{Jepson}, \binits{A.D.}}:
\bctitle{P3iv: Probabilistic procedure planning from instructional videos with weak supervision}.
In: \bbtitle{Proceedings of the IEEE/CVF Conference on Computer Vision and Pattern Recognition},
pp. \bfpage{2938}--\blpage{2948}
(\byear{2022})
\end{bchapter}
\endbibitem

\bibitem[\protect\citeauthoryear{Wang et~al.}{2023}]{pdpp}
\begin{botherref}
\oauthor{\bsnm{Wang}, \binits{H.}},
\oauthor{\bsnm{Wu}, \binits{Y.}},
\oauthor{\bsnm{Guo}, \binits{S.}},
\oauthor{\bsnm{Wang}, \binits{L.}}:
Pdpp: Projected diffusion for procedure planning in instructional videos.
arXiv preprint arXiv:2303.14676
(2023)
\end{botherref}
\endbibitem

\bibitem[\protect\citeauthoryear{Zhong et~al.}{2023}]{procedure3}
\begin{botherref}
\oauthor{\bsnm{Zhong}, \binits{Y.}},
\oauthor{\bsnm{Yu}, \binits{L.}},
\oauthor{\bsnm{Bai}, \binits{Y.}},
\oauthor{\bsnm{Li}, \binits{S.}},
\oauthor{\bsnm{Yan}, \binits{X.}},
\oauthor{\bsnm{Li}, \binits{Y.}}:
Learning procedure-aware video representation from instructional videos and their narrations.
arXiv preprint arXiv:2303.17839
(2023)
\end{botherref}
\endbibitem

\bibitem[\protect\citeauthoryear{Shvetsova et~al.}{2022}]{shvetsova2022everything}
\begin{bchapter}
\bauthor{\bsnm{Shvetsova}, \binits{N.}},
\bauthor{\bsnm{Chen}, \binits{B.}},
\bauthor{\bsnm{Rouditchenko}, \binits{A.}},
\bauthor{\bsnm{Thomas}, \binits{S.}},
\bauthor{\bsnm{Kingsbury}, \binits{B.}},
\bauthor{\bsnm{Feris}, \binits{R.S.}},
\bauthor{\bsnm{Harwath}, \binits{D.}},
\bauthor{\bsnm{Glass}, \binits{J.}},
\bauthor{\bsnm{Kuehne}, \binits{H.}}:
\bctitle{Everything at once-multi-modal fusion transformer for video retrieval}.
In: \bbtitle{Proceedings of the IEEE/CVF Conference on Computer Vision and Pattern Recognition},
pp. \bfpage{20020}--\blpage{20029}
(\byear{2022})
\end{bchapter}
\endbibitem

\bibitem[\protect\citeauthoryear{Ko et~al.}{2022}]{ko2022video}
\begin{bchapter}
\bauthor{\bsnm{Ko}, \binits{D.}},
\bauthor{\bsnm{Choi}, \binits{J.}},
\bauthor{\bsnm{Ko}, \binits{J.}},
\bauthor{\bsnm{Noh}, \binits{S.}},
\bauthor{\bsnm{On}, \binits{K.-W.}},
\bauthor{\bsnm{Kim}, \binits{E.-S.}},
\bauthor{\bsnm{Kim}, \binits{H.J.}}:
\bctitle{Video-text representation learning via differentiable weak temporal alignment}.
In: \bbtitle{Proceedings of the IEEE/CVF Conference on Computer Vision and Pattern Recognition},
pp. \bfpage{5016}--\blpage{5025}
(\byear{2022})
\end{bchapter}
\endbibitem

\bibitem[\protect\citeauthoryear{Cao et~al.}{2022}]{cao2022locvtp}
\begin{bchapter}
\bauthor{\bsnm{Cao}, \binits{M.}},
\bauthor{\bsnm{Yang}, \binits{T.}},
\bauthor{\bsnm{Weng}, \binits{J.}},
\bauthor{\bsnm{Zhang}, \binits{C.}},
\bauthor{\bsnm{Wang}, \binits{J.}},
\bauthor{\bsnm{Zou}, \binits{Y.}}:
\bctitle{Locvtp: Video-text pre-training for temporal localization}.
In: \bbtitle{European Conference on Computer Vision}
(\byear{2022})
\end{bchapter}
\endbibitem

\bibitem[\protect\citeauthoryear{Narasimhan et~al.}{2023}]{task-structure}
\begin{botherref}
\oauthor{\bsnm{Narasimhan}, \binits{M.}},
\oauthor{\bsnm{Yu}, \binits{L.}},
\oauthor{\bsnm{Bell}, \binits{S.}},
\oauthor{\bsnm{Zhang}, \binits{N.}},
\oauthor{\bsnm{Darrell}, \binits{T.}}:
Learning and verification of task structure in instructional videos.
arXiv preprint arXiv:2303.13519
(2023)
\end{botherref}
\endbibitem

\bibitem[\protect\citeauthoryear{Zhou et~al.}{2018}]{zhou2018towards}
\begin{bchapter}
\bauthor{\bsnm{Zhou}, \binits{L.}},
\bauthor{\bsnm{Xu}, \binits{C.}},
\bauthor{\bsnm{Corso}, \binits{J.J.}}:
\bctitle{Towards automatic learning of procedures from web instructional videos}.
In: \bbtitle{AAAI}
(\byear{2018})
\end{bchapter}
\endbibitem

\bibitem[\protect\citeauthoryear{Alayrac et~al.}{2016}]{alayrac}
\begin{bchapter}
\bauthor{\bsnm{Alayrac}, \binits{J.-B.}},
\bauthor{\bsnm{Bojanowski}, \binits{P.}},
\bauthor{\bsnm{Agrawal}, \binits{N.}},
\bauthor{\bsnm{Sivic}, \binits{J.}},
\bauthor{\bsnm{Laptev}, \binits{I.}},
\bauthor{\bsnm{Lacoste-Julien}, \binits{S.}}:
\bctitle{Unsupervised learning from narrated instruction videos}.
In: \bbtitle{CVPR}
(\byear{2016})
\end{bchapter}
\endbibitem

\bibitem[\protect\citeauthoryear{Ashutosh et~al.}{2023}]{ashutosh-neurips2023}
\begin{bchapter}
\bauthor{\bsnm{Ashutosh}, \binits{K.}},
\bauthor{\bsnm{Ramakrishnan}, \binits{S.K.}},
\bauthor{\bsnm{Afouras}, \binits{T.}},
\bauthor{\bsnm{Grauman}, \binits{K.}}:
\bctitle{Video-mined task graphs for keystep recognition in instructional videos}.
In: \bbtitle{NeurIPS}
(\byear{2023})
\end{bchapter}
\endbibitem

\bibitem[\protect\citeauthoryear{Zhou et~al.}{2023}]{zhou2023paprika}
\begin{botherref}
\oauthor{\bsnm{Zhou}, \binits{H.}},
\oauthor{\bsnm{Martin-Martin}, \binits{R.}},
\oauthor{\bsnm{Kapadia}, \binits{M.}},
\oauthor{\bsnm{Savarese}, \binits{S.}},
\oauthor{\bsnm{Niebles}, \binits{J.C.}}:
Procedure-aware pretraining for instructional video understanding.
Proceedings of the IEEE/CVF Conference on Computer Vision and Pattern Recognition
(2023)
\end{botherref}
\endbibitem

\bibitem[\protect\citeauthoryear{Ardeshir and Borji}{2016}]{ardeshir2016ego2top}
\begin{bchapter}
\bauthor{\bsnm{Ardeshir}, \binits{S.}},
\bauthor{\bsnm{Borji}, \binits{A.}}:
\bctitle{Ego2top: Matching viewers in egocentric and top-view videos}.
In: \bbtitle{ECCV}
(\byear{2016})
\end{bchapter}
\endbibitem

\bibitem[\protect\citeauthoryear{Ardeshir and Borji}{2018}]{ardeshir2018egocentric}
\begin{botherref}
\oauthor{\bsnm{Ardeshir}, \binits{S.}},
\oauthor{\bsnm{Borji}, \binits{A.}}:
Egocentric meets top-view.
IEEE transactions on pattern analysis and machine intelligence
\textbf{41}(6)
(2018)
\end{botherref}
\endbibitem

\bibitem[\protect\citeauthoryear{Fan et~al.}{2017}]{fan2017identifying}
\begin{bchapter}
\bauthor{\bsnm{Fan}, \binits{C.}},
\bauthor{\bsnm{Lee}, \binits{J.}},
\bauthor{\bsnm{Xu}, \binits{M.}},
\bauthor{\bsnm{Kumar~Singh}, \binits{K.}},
\bauthor{\bsnm{Jae~Lee}, \binits{Y.}},
\bauthor{\bsnm{Crandall}, \binits{D.J.}},
\bauthor{\bsnm{Ryoo}, \binits{M.S.}}:
\bctitle{Identifying first-person camera wearers in third-person videos}.
In: \bbtitle{CVPR}
(\byear{2017})
\end{bchapter}
\endbibitem

\bibitem[\protect\citeauthoryear{Xu et~al.}{2018}]{xu2018joint}
\begin{bchapter}
\bauthor{\bsnm{Xu}, \binits{M.}},
\bauthor{\bsnm{Fan}, \binits{C.}},
\bauthor{\bsnm{Wang}, \binits{Y.}},
\bauthor{\bsnm{Ryoo}, \binits{M.S.}},
\bauthor{\bsnm{Crandall}, \binits{D.J.}}:
\bctitle{Joint person segmentation and identification in synchronized first-and third-person videos}.
In: \bbtitle{ECCV}
(\byear{2018})
\end{bchapter}
\endbibitem

\bibitem[\protect\citeauthoryear{Wen et~al.}{2021}]{Wen_2021_ICCV}
\begin{bchapter}
\bauthor{\bsnm{Wen}, \binits{Y.}},
\bauthor{\bsnm{Singh}, \binits{K.K.}},
\bauthor{\bsnm{Anderson}, \binits{M.}},
\bauthor{\bsnm{Jan}, \binits{W.-P.}},
\bauthor{\bsnm{Lee}, \binits{Y.J.}}:
\bctitle{Seeing the unseen: Predicting the first-person camera wearer's location and pose in third-person scenes}.
In: \bbtitle{Proceedings of the IEEE/CVF International Conference on Computer Vision (ICCV) Workshops},
pp. \bfpage{3446}--\blpage{3455}
(\byear{2021})
\end{bchapter}
\endbibitem

\bibitem[\protect\citeauthoryear{Ardeshir and Borji}{2018}]{ardeshir2018exocentric}
\begin{botherref}
\oauthor{\bsnm{Ardeshir}, \binits{S.}},
\oauthor{\bsnm{Borji}, \binits{A.}}:
An exocentric look at egocentric actions and vice versa.
Computer Vision and Image Understanding
\textbf{171}
(2018)
\end{botherref}
\endbibitem

\bibitem[\protect\citeauthoryear{Sermanet et~al.}{2018}]{Sermanet2017TCN}
\begin{botherref}
\oauthor{\bsnm{Sermanet}, \binits{P.}},
\oauthor{\bsnm{Lynch}, \binits{C.}},
\oauthor{\bsnm{Chebotar}, \binits{Y.}},
\oauthor{\bsnm{Hsu}, \binits{J.}},
\oauthor{\bsnm{Jang}, \binits{E.}},
\oauthor{\bsnm{Schaal}, \binits{S.}},
\oauthor{\bsnm{Levine}, \binits{S.}}:
Time-contrastive networks: Self-supervised learning from video.
Proceedings of International Conference in Robotics and Automation (ICRA)
(2018)
\end{botherref}
\endbibitem

\bibitem[\protect\citeauthoryear{Yu et~al.}{2019}]{yu2019see}
\begin{bchapter}
\bauthor{\bsnm{Yu}, \binits{H.}},
\bauthor{\bsnm{Cai}, \binits{M.}},
\bauthor{\bsnm{Liu}, \binits{Y.}},
\bauthor{\bsnm{Lu}, \binits{F.}}:
\bctitle{What i see is what you see: Joint attention learning for first and third person video co-analysis}.
In: \bbtitle{ACM MM}
(\byear{2019})
\end{bchapter}
\endbibitem

\bibitem[\protect\citeauthoryear{Yu et~al.}{2020}]{yu2020first}
\begin{botherref}
\oauthor{\bsnm{Yu}, \binits{H.}},
\oauthor{\bsnm{Cai}, \binits{M.}},
\oauthor{\bsnm{Liu}, \binits{Y.}},
\oauthor{\bsnm{Lu}, \binits{F.}}:
First-and third-person video co-analysis by learning spatial-temporal joint attention.
IEEE Transactions on Pattern Analysis and Machine Intelligence
(2020)
\end{botherref}
\endbibitem

\bibitem[\protect\citeauthoryear{Xue and Grauman}{2023}]{sherry-neurips2023}
\begin{bchapter}
\bauthor{\bsnm{Xue}, \binits{Z.}},
\bauthor{\bsnm{Grauman}, \binits{K.}}:
\bctitle{Learning fine-grained view-invariant representations from unpaired ego-exo videos via temporal alignment}.
In: \bbtitle{NeurIPS}
(\byear{2023})
\end{bchapter}
\endbibitem

\bibitem[\protect\citeauthoryear{Li et~al.}{2021}]{li2021ego}
\begin{bchapter}
\bauthor{\bsnm{Li}, \binits{Y.}},
\bauthor{\bsnm{Nagarajan}, \binits{T.}},
\bauthor{\bsnm{Xiong}, \binits{B.}},
\bauthor{\bsnm{Grauman}, \binits{K.}}:
\bctitle{Ego-exo: Transferring visual representations from third-person to first-person videos}.
In: \bbtitle{Proceedings of the IEEE/CVF Conference on Computer Vision and Pattern Recognition},
pp. \bfpage{6943}--\blpage{6953}
(\byear{2021})
\end{bchapter}
\endbibitem

\bibitem[\protect\citeauthoryear{Regmi and Borji}{2018}]{Regmi_2018_CVPR}
\begin{bchapter}
\bauthor{\bsnm{Regmi}, \binits{K.}},
\bauthor{\bsnm{Borji}, \binits{A.}}:
\bctitle{Cross-view image synthesis using conditional gans}.
In: \bbtitle{The IEEE Conference on Computer Vision and Pattern Recognition (CVPR)}
(\byear{2018})
\end{bchapter}
\endbibitem

\bibitem[\protect\citeauthoryear{Regmi and Borji}{2019}]{REGMI2019}
\begin{barticle}
\bauthor{\bsnm{Regmi}, \binits{K.}},
\bauthor{\bsnm{Borji}, \binits{A.}}:
\batitle{Cross-view image synthesis using geometry-guided conditional gans}.
\bjtitle{Computer Vision and Image Understanding}
(\byear{2019})
\doiurl{10.1016/j.cviu.2019.07.008}
\end{barticle}
\endbibitem

\bibitem[\protect\citeauthoryear{Tang et~al.}{2019}]{tang2019selectiongan}
\begin{bchapter}
\bauthor{\bsnm{Tang}, \binits{H.}},
\bauthor{\bsnm{Xu}, \binits{D.}},
\bauthor{\bsnm{Sebe}, \binits{N.}},
\bauthor{\bsnm{Wang}, \binits{Y.}},
\bauthor{\bsnm{Corso}, \binits{J.J.}},
\bauthor{\bsnm{Yan}, \binits{Y.}}:
\bctitle{Multi-channel attention selection gan with cascaded semantic guidance for cross-view image translation}.
In: \bbtitle{Proceedings of the IEEE/CVF Conference on Computer Vision and Pattern Recognition},
pp. \bfpage{2417}--\blpage{2426}
(\byear{2019})
\end{bchapter}
\endbibitem

\bibitem[\protect\citeauthoryear{Ren et~al.}{2021}]{ren2021crossmlp}
\begin{botherref}
\oauthor{\bsnm{Ren}, \binits{B.}},
\oauthor{\bsnm{Tang}, \binits{H.}},
\oauthor{\bsnm{Sebe}, \binits{N.}}:
Cascaded cross mlp-mixer gans for cross-view image translation.
arXiv preprint arXiv:2110.10183
(2021)
\end{botherref}
\endbibitem

\bibitem[\protect\citeauthoryear{Luo et~al.}{2024}]{exo2ego-eccv2024}
\begin{bchapter}
\bauthor{\bsnm{Luo}, \binits{M.}},
\bauthor{\bsnm{Xue}, \binits{Z.}},
\bauthor{\bsnm{Dimakis}, \binits{A.}},
\bauthor{\bsnm{Grauman}, \binits{K.}}:
\bctitle{Put myself in your shoes: Lifting the egocentric perspective from exocentric videos}.
In: \bbtitle{ECCV}
(\byear{2024})
\end{bchapter}
\endbibitem

\bibitem[\protect\citeauthoryear{Cheng et~al.}{2024}]{4diff}
\begin{bchapter}
\bauthor{\bsnm{Cheng}, \binits{F.}},
\bauthor{\bsnm{Luo}, \binits{M.}},
\bauthor{\bsnm{Wang}, \binits{H.}},
\bauthor{\bsnm{Dimakis}, \binits{A.}},
\bauthor{\bsnm{Torresani}, \binits{L.}},
\bauthor{\bsnm{Bertasius}, \binits{G.}},
\bauthor{\bsnm{Grauman}, \binits{K.}}:
\bctitle{4{D}{I}{F}{F}: 3d-aware diffusion model for third-to-first viewpoint translation}.
In: \bbtitle{ECCV}
(\byear{2024})
\end{bchapter}
\endbibitem

\bibitem[\protect\citeauthoryear{Liu et~al.}{2021}]{liu2021infinite}
\begin{bchapter}
\bauthor{\bsnm{Liu}, \binits{A.}},
\bauthor{\bsnm{Tucker}, \binits{R.}},
\bauthor{\bsnm{Jampani}, \binits{V.}},
\bauthor{\bsnm{Makadia}, \binits{A.}},
\bauthor{\bsnm{Snavely}, \binits{N.}},
\bauthor{\bsnm{Kanazawa}, \binits{A.}}:
\bctitle{Infinite nature: Perpetual view generation of natural scenes from a single image}.
In: \bbtitle{Proceedings of the IEEE/CVF International Conference on Computer Vision},
pp. \bfpage{14458}--\blpage{14467}
(\byear{2021})
\end{bchapter}
\endbibitem

\bibitem[\protect\citeauthoryear{Ren and Wang}{2022}]{ren2022look}
\begin{bchapter}
\bauthor{\bsnm{Ren}, \binits{X.}},
\bauthor{\bsnm{Wang}, \binits{X.}}:
\bctitle{Look outside the room: Synthesizing a consistent long-term 3d scene video from a single image}.
In: \bbtitle{Proceedings of the IEEE/CVF Conference on Computer Vision and Pattern Recognition},
pp. \bfpage{3563}--\blpage{3573}
(\byear{2022})
\end{bchapter}
\endbibitem

\bibitem[\protect\citeauthoryear{Rombach et~al.}{2021}]{rombach2021geometry}
\begin{bchapter}
\bauthor{\bsnm{Rombach}, \binits{R.}},
\bauthor{\bsnm{Esser}, \binits{P.}},
\bauthor{\bsnm{Ommer}, \binits{B.}}:
\bctitle{Geometry-free view synthesis: Transformers and no 3d priors}.
In: \bbtitle{Proceedings of the IEEE/CVF International Conference on Computer Vision},
pp. \bfpage{14356}--\blpage{14366}
(\byear{2021})
\end{bchapter}
\endbibitem

\bibitem[\protect\citeauthoryear{Wiles et~al.}{2020}]{wiles2020synsin}
\begin{bchapter}
\bauthor{\bsnm{Wiles}, \binits{O.}},
\bauthor{\bsnm{Gkioxari}, \binits{G.}},
\bauthor{\bsnm{Szeliski}, \binits{R.}},
\bauthor{\bsnm{Johnson}, \binits{J.}}:
\bctitle{Synsin: End-to-end view synthesis from a single image}.
In: \bbtitle{Proceedings of the IEEE/CVF Conference on Computer Vision and Pattern Recognition},
pp. \bfpage{7467}--\blpage{7477}
(\byear{2020})
\end{bchapter}
\endbibitem

\bibitem[\protect\citeauthoryear{Watson et~al.}{2022}]{watson2022novel}
\begin{botherref}
\oauthor{\bsnm{Watson}, \binits{D.}},
\oauthor{\bsnm{Chan}, \binits{W.}},
\oauthor{\bsnm{Martin-Brualla}, \binits{R.}},
\oauthor{\bsnm{Ho}, \binits{J.}},
\oauthor{\bsnm{Tagliasacchi}, \binits{A.}},
\oauthor{\bsnm{Norouzi}, \binits{M.}}:
Novel view synthesis with diffusion models.
arXiv preprint arXiv:2210.04628
(2022)
\end{botherref}
\endbibitem

\bibitem[\protect\citeauthoryear{Tseng et~al.}{2023}]{tseng2023consistent}
\begin{botherref}
\oauthor{\bsnm{Tseng}, \binits{H.-Y.}},
\oauthor{\bsnm{Li}, \binits{Q.}},
\oauthor{\bsnm{Kim}, \binits{C.}},
\oauthor{\bsnm{Alsisan}, \binits{S.}},
\oauthor{\bsnm{Huang}, \binits{J.-B.}},
\oauthor{\bsnm{Kopf}, \binits{J.}}:
Consistent view synthesis with pose-guided diffusion models.
arXiv preprint arXiv:2303.17598
(2023)
\end{botherref}
\endbibitem

\bibitem[\protect\citeauthoryear{Chan et~al.}{2023}]{chan2023generative}
\begin{botherref}
\oauthor{\bsnm{Chan}, \binits{E.R.}},
\oauthor{\bsnm{Nagano}, \binits{K.}},
\oauthor{\bsnm{Chan}, \binits{M.A.}},
\oauthor{\bsnm{Bergman}, \binits{A.W.}},
\oauthor{\bsnm{Park}, \binits{J.J.}},
\oauthor{\bsnm{Levy}, \binits{A.}},
\oauthor{\bsnm{Aittala}, \binits{M.}},
\oauthor{\bsnm{De~Mello}, \binits{S.}},
\oauthor{\bsnm{Karras}, \binits{T.}},
\oauthor{\bsnm{Wetzstein}, \binits{G.}}:
Generative novel view synthesis with 3d-aware diffusion models.
arXiv preprint arXiv:2304.02602
(2023)
\end{botherref}
\endbibitem

\bibitem[\protect\citeauthoryear{Regmi and Shah}{2019}]{shah-aerial}
\begin{bchapter}
\bauthor{\bsnm{Regmi}, \binits{K.}},
\bauthor{\bsnm{Shah}, \binits{M.}}:
\bctitle{Bridging the domain gap for ground-to-aerial image matching}.
In: \bbtitle{ICCV}
(\byear{2019})
\end{bchapter}
\endbibitem

\bibitem[\protect\citeauthoryear{Lin et~al.}{2015}]{deepgeo}
\begin{bchapter}
\bauthor{\bsnm{Lin}, \binits{T.}},
\bauthor{\bsnm{Cui}, \binits{Y.}},
\bauthor{\bsnm{Belongie}, \binits{S.}},
\bauthor{\bsnm{Hays}, \binits{J.}}:
\bctitle{Learning deep representations for ground-to-aerial geolocalization}.
In: \bbtitle{CVPR}
(\byear{2015})
\end{bchapter}
\endbibitem

\bibitem[\protect\citeauthoryear{Kukelova et~al.}{2016}]{kukelova2016efficient}
\begin{bchapter}
\bauthor{\bsnm{Kukelova}, \binits{Z.}},
\bauthor{\bsnm{Heller}, \binits{J.}},
\bauthor{\bsnm{Fitzgibbon}, \binits{A.}}:
\bctitle{Efficient intersection of three quadrics and applications in computer vision}.
In: \bbtitle{Proceedings of the IEEE Conference on Computer Vision and Pattern Recognition},
pp. \bfpage{1799}--\blpage{1808}
(\byear{2016})
\end{bchapter}
\endbibitem

\bibitem[\protect\citeauthoryear{Lin et~al.}{2022}]{kevin2022egovlp}
\begin{botherref}
\oauthor{\bsnm{Lin}, \binits{K.Q.}},
\oauthor{\bsnm{Wang}, \binits{A.J.}},
\oauthor{\bsnm{Soldan}, \binits{M.}},
\oauthor{\bsnm{Wray}, \binits{M.}},
\oauthor{\bsnm{Yan}, \binits{R.}},
\oauthor{\bsnm{Xu}, \binits{E.Z.}},
\oauthor{\bsnm{Gao}, \binits{D.}},
\oauthor{\bsnm{Tu}, \binits{R.}},
\oauthor{\bsnm{Zhao}, \binits{W.}},
\oauthor{\bsnm{Kong}, \binits{W.}}, et al.:
Egocentric video-language pretraining.
NeurIPS
(2022)
\end{botherref}
\endbibitem

\bibitem[\protect\citeauthoryear{Pramanick et~al.}{2023}]{pramanick2023egovlpv2}
\begin{bchapter}
\bauthor{\bsnm{Pramanick}, \binits{S.}},
\bauthor{\bsnm{Song}, \binits{Y.}},
\bauthor{\bsnm{Nag}, \binits{S.}},
\bauthor{\bsnm{Lin}, \binits{K.Q.}},
\bauthor{\bsnm{Shah}, \binits{H.}},
\bauthor{\bsnm{Shou}, \binits{M.Z.}},
\bauthor{\bsnm{Chellappa}, \binits{R.}},
\bauthor{\bsnm{Zhang}, \binits{P.}}:
\bctitle{Egovlpv2: Egocentric video-language pre-training with fusion in the backbone}.
In: \bbtitle{Proceedings of the IEEE/CVF International Conference on Computer Vision},
pp. \bfpage{5285}--\blpage{5297}
(\byear{2023})
\end{bchapter}
\endbibitem

\bibitem[\protect\citeauthoryear{Pan et~al.}{2020}]{pan2020spatio}
\begin{bchapter}
\bauthor{\bsnm{Pan}, \binits{B.}},
\bauthor{\bsnm{Cai}, \binits{H.}},
\bauthor{\bsnm{Huang}, \binits{D.-A.}},
\bauthor{\bsnm{Lee}, \binits{K.-H.}},
\bauthor{\bsnm{Gaidon}, \binits{A.}},
\bauthor{\bsnm{Adeli}, \binits{E.}},
\bauthor{\bsnm{Niebles}, \binits{J.C.}}:
\bctitle{Spatio-temporal graph for video captioning with knowledge distillation}.
In: \bbtitle{CVPR}
(\byear{2020})
\end{bchapter}
\endbibitem

\bibitem[\protect\citeauthoryear{Iashin and Rahtu}{2020}]{iashin2020multi}
\begin{bchapter}
\bauthor{\bsnm{Iashin}, \binits{V.}},
\bauthor{\bsnm{Rahtu}, \binits{E.}}:
\bctitle{Multi-modal dense video captioning}.
In: \bbtitle{Proceedings of the IEEE/CVF Conference on Computer Vision and Pattern Recognition Workshops},
pp. \bfpage{958}--\blpage{959}
(\byear{2020})
\end{bchapter}
\endbibitem

\bibitem[\protect\citeauthoryear{Zhao et~al.}{2023}]{zhao2023learning}
\begin{bchapter}
\bauthor{\bsnm{Zhao}, \binits{Y.}},
\bauthor{\bsnm{Misra}, \binits{I.}},
\bauthor{\bsnm{Kr{\"a}henb{\"u}hl}, \binits{P.}},
\bauthor{\bsnm{Girdhar}, \binits{R.}}:
\bctitle{Learning video representations from large language models}.
In: \bbtitle{Proceedings of the IEEE/CVF Conference on Computer Vision and Pattern Recognition},
pp. \bfpage{6586}--\blpage{6597}
(\byear{2023})
\end{bchapter}
\endbibitem

\bibitem[\protect\citeauthoryear{Radford et~al.}{2023}]{radford2023robust}
\begin{bchapter}
\bauthor{\bsnm{Radford}, \binits{A.}},
\bauthor{\bsnm{Kim}, \binits{J.W.}},
\bauthor{\bsnm{Xu}, \binits{T.}},
\bauthor{\bsnm{Brockman}, \binits{G.}},
\bauthor{\bsnm{McLeavey}, \binits{C.}},
\bauthor{\bsnm{Sutskever}, \binits{I.}}:
\bctitle{Robust speech recognition via large-scale weak supervision}.
In: \bbtitle{International Conference on Machine Learning},
pp. \bfpage{28492}--\blpage{28518}
(\byear{2023}).
\bcomment{PMLR}
\end{bchapter}
\endbibitem

\bibitem[\protect\citeauthoryear{Ashutosh et~al.}{2023}]{hiervl}
\begin{bchapter}
\bauthor{\bsnm{Ashutosh}, \binits{K.}},
\bauthor{\bsnm{Girdhar}, \binits{R.}},
\bauthor{\bsnm{Torresani}, \binits{L.}},
\bauthor{\bsnm{Grauman}, \binits{K.}}:
\bctitle{Hiervl: Learning hierarchical video-language embeddings}.
In: \bbtitle{Proceedings of the IEEE/CVF Conference on Computer Vision and Pattern Recognition}
(\byear{2023})
\end{bchapter}
\endbibitem

\bibitem[\protect\citeauthoryear{Jiang et~al.}{2021}]{jiang2021cotr}
\begin{bchapter}
\bauthor{\bsnm{Jiang}, \binits{W.}},
\bauthor{\bsnm{Trulls}, \binits{E.}},
\bauthor{\bsnm{Hosang}, \binits{J.}},
\bauthor{\bsnm{Tagliasacchi}, \binits{A.}},
\bauthor{\bsnm{Yi}, \binits{K.M.}}:
\bctitle{Cotr: Correspondence transformer for matching across images}.
In: \bbtitle{Proceedings of the IEEE/CVF International Conference on Computer Vision},
pp. \bfpage{6207}--\blpage{6217}
(\byear{2021})
\end{bchapter}
\endbibitem

\bibitem[\protect\citeauthoryear{Vicente et~al.}{2011}]{vicente2011object}
\begin{bchapter}
\bauthor{\bsnm{Vicente}, \binits{S.}},
\bauthor{\bsnm{Rother}, \binits{C.}},
\bauthor{\bsnm{Kolmogorov}, \binits{V.}}:
\bctitle{Object cosegmentation}.
In: \bbtitle{CVPR 2011},
pp. \bfpage{2217}--\blpage{2224}
(\byear{2011}).
\bcomment{IEEE}
\end{bchapter}
\endbibitem

\bibitem[\protect\citeauthoryear{Tang et~al.}{2023}]{tang2023egotracks}
\begin{botherref}
\oauthor{\bsnm{Tang}, \binits{H.}},
\oauthor{\bsnm{Liang}, \binits{K.}},
\oauthor{\bsnm{Grauman}, \binits{K.}},
\oauthor{\bsnm{Feiszli}, \binits{M.}},
\oauthor{\bsnm{Wang}, \binits{W.}}:
Egotracks: A long-term egocentric visual object tracking dataset.
Advances in Neural Information Processing Systems
(2023)
\end{botherref}
\endbibitem

\bibitem[\protect\citeauthoryear{Perazzi et~al.}{2016}]{perazzi2016benchmark}
\begin{bchapter}
\bauthor{\bsnm{Perazzi}, \binits{F.}},
\bauthor{\bsnm{Pont-Tuset}, \binits{J.}},
\bauthor{\bsnm{McWilliams}, \binits{B.}},
\bauthor{\bsnm{Van~Gool}, \binits{L.}},
\bauthor{\bsnm{Gross}, \binits{M.}},
\bauthor{\bsnm{Sorkine-Hornung}, \binits{A.}}:
\bctitle{A benchmark dataset and evaluation methodology for video object segmentation}.
In: \bbtitle{Proceedings of the IEEE Conference on Computer Vision and Pattern Recognition},
pp. \bfpage{724}--\blpage{732}
(\byear{2016})
\end{bchapter}
\endbibitem

\bibitem[\protect\citeauthoryear{Brodersen et~al.}{2010}]{brodersen2010Balanced}
\begin{bchapter}
\bauthor{\bsnm{Brodersen}, \binits{K.H.}},
\bauthor{\bsnm{Ong}, \binits{C.S.}},
\bauthor{\bsnm{Stephan}, \binits{K.E.}},
\bauthor{\bsnm{Buhmann}, \binits{J.M.}}:
\bctitle{The balanced accuracy and its posterior distribution}.
In: \bbtitle{2010 20th International Conference on Pattern Recognition},
pp. \bfpage{3121}--\blpage{3124}
(\byear{2010}).
\bcomment{IEEE}
\end{bchapter}
\endbibitem

\bibitem[\protect\citeauthoryear{Shen et~al.}{2022}]{shen2022learning}
\begin{bchapter}
\bauthor{\bsnm{Shen}, \binits{X.}},
\bauthor{\bsnm{Efros}, \binits{A.A.}},
\bauthor{\bsnm{Joulin}, \binits{A.}},
\bauthor{\bsnm{Aubry}, \binits{M.}}:
\bctitle{Learning co-segmentation by segment swapping for retrieval and discovery}.
In: \bbtitle{Proceedings of the IEEE/CVF Conference on Computer Vision and Pattern Recognition},
pp. \bfpage{5082}--\blpage{5092}
(\byear{2022})
\end{bchapter}
\endbibitem

\bibitem[\protect\citeauthoryear{Cheng and Schwing}{2022}]{cheng2022XMem}
\begin{bchapter}
\bauthor{\bsnm{Cheng}, \binits{H.K.}},
\bauthor{\bsnm{Schwing}, \binits{A.G.}}:
\bctitle{Xmem: Long-term video object segmentation with an atkinson-shiffrin memory model}.
In: \bbtitle{European Conference on Computer Vision},
pp. \bfpage{640}--\blpage{658}
(\byear{2022}).
\bcomment{Springer}
\end{bchapter}
\endbibitem

\bibitem[\protect\citeauthoryear{Lu et~al.}{2020}]{Lu_2020_CVPR}
\begin{bchapter}
\bauthor{\bsnm{Lu}, \binits{X.}},
\bauthor{\bsnm{Li}, \binits{Z.}},
\bauthor{\bsnm{Cui}, \binits{Z.}},
\bauthor{\bsnm{Oswald}, \binits{M.R.}},
\bauthor{\bsnm{Pollefeys}, \binits{M.}},
\bauthor{\bsnm{Qin}, \binits{R.}}:
\bctitle{Geometry-aware satellite-to-ground image synthesis for urban areas}.
In: \bbtitle{Proceedings of the IEEE/CVF Conference on Computer Vision and Pattern Recognition (CVPR)}
(\byear{2020})
\end{bchapter}
\endbibitem

\bibitem[\protect\citeauthoryear{Liu et~al.}{2020}]{liu2020pgan}
\begin{bchapter}
\bauthor{\bsnm{Liu}, \binits{G.}},
\bauthor{\bsnm{Tang}, \binits{H.}},
\bauthor{\bsnm{Latapie}, \binits{H.}},
\bauthor{\bsnm{Yan}, \binits{Y.}}:
\bctitle{Exocentric to egocentric image generation via parallel generative adversarial network}.
In: \bbtitle{ICASSP 2020-2020 IEEE International Conference on Acoustics, Speech and Signal Processing (ICASSP)},
pp. \bfpage{1843}--\blpage{1847}
(\byear{2020}).
\bcomment{IEEE}
\end{bchapter}
\endbibitem

\bibitem[\protect\citeauthoryear{Luo et~al.}{2024}]{romy-exo2ego}
\begin{botherref}
\oauthor{\bsnm{Luo}, \binits{M.}},
\oauthor{\bsnm{Xue}, \binits{Z.}},
\oauthor{\bsnm{Dimakis}, \binits{A.}},
\oauthor{\bsnm{Grauman}, \binits{K.}}:
Put myself in your shoes: Lifting the egocentric perspective from exocentric videos.
arXiv:2403.06351
(2024)
\end{botherref}
\endbibitem

\bibitem[\protect\citeauthoryear{Liu et~al.}{2021}]{liu2021stagan}
\begin{bchapter}
\bauthor{\bsnm{Liu}, \binits{G.}},
\bauthor{\bsnm{Tang}, \binits{H.}},
\bauthor{\bsnm{Latapie}, \binits{H.M.}},
\bauthor{\bsnm{Corso}, \binits{J.J.}},
\bauthor{\bsnm{Yan}, \binits{Y.}}:
\bctitle{Cross-view exocentric to egocentric video synthesis}.
In: \bbtitle{Proceedings of the 29th ACM International Conference on Multimedia},
pp. \bfpage{974}--\blpage{982}
(\byear{2021})
\end{bchapter}
\endbibitem

\bibitem[\protect\citeauthoryear{Hore and Ziou}{2010}]{hore2010SSIM}
\begin{bchapter}
\bauthor{\bsnm{Hore}, \binits{A.}},
\bauthor{\bsnm{Ziou}, \binits{D.}}:
\bctitle{Image quality metrics: Psnr vs. ssim}.
In: \bbtitle{2010 20th International Conference on Pattern Recognition},
pp. \bfpage{2366}--\blpage{2369}
(\byear{2010}).
\bcomment{IEEE}
\end{bchapter}
\endbibitem

\bibitem[\protect\citeauthoryear{Ding et~al.}{2020}]{ding2020dists}
\begin{barticle}
\bauthor{\bsnm{Ding}, \binits{K.}},
\bauthor{\bsnm{Ma}, \binits{K.}},
\bauthor{\bsnm{Wang}, \binits{S.}},
\bauthor{\bsnm{Simoncelli}, \binits{E.P.}}:
\batitle{Image quality assessment: Unifying structure and texture similarity}.
\bjtitle{IEEE transactions on pattern analysis and machine intelligence}
\bvolume{44}(\bissue{5}),
\bfpage{2567}--\blpage{2581}
(\byear{2020})
\end{barticle}
\endbibitem

\bibitem[\protect\citeauthoryear{Zhang et~al.}{2018}]{zhang2018lpips}
\begin{bchapter}
\bauthor{\bsnm{Zhang}, \binits{R.}},
\bauthor{\bsnm{Isola}, \binits{P.}},
\bauthor{\bsnm{Efros}, \binits{A.A.}},
\bauthor{\bsnm{Shechtman}, \binits{E.}},
\bauthor{\bsnm{Wang}, \binits{O.}}:
\bctitle{The unreasonable effectiveness of deep features as a perceptual metric}.
In: \bbtitle{CVPR}
(\byear{2018})
\end{bchapter}
\endbibitem

\bibitem[\protect\citeauthoryear{Radford et~al.}{2021}]{radford2021clip}
\begin{bchapter}
\bauthor{\bsnm{Radford}, \binits{A.}},
\bauthor{\bsnm{Kim}, \binits{J.W.}},
\bauthor{\bsnm{Hallacy}, \binits{C.}},
\bauthor{\bsnm{Ramesh}, \binits{A.}},
\bauthor{\bsnm{Goh}, \binits{G.}},
\bauthor{\bsnm{Agarwal}, \binits{S.}},
\bauthor{\bsnm{Sastry}, \binits{G.}},
\bauthor{\bsnm{Askell}, \binits{A.}},
\bauthor{\bsnm{Mishkin}, \binits{P.}},
\bauthor{\bsnm{Clark}, \binits{J.}}, \betal:
\bctitle{Learning transferable visual models from natural language supervision}.
In: \bbtitle{International Conference on Machine Learning},
pp. \bfpage{8748}--\blpage{8763}
(\byear{2021}).
\bcomment{PMLR}
\end{bchapter}
\endbibitem

\bibitem[\protect\citeauthoryear{Isola et~al.}{2017}]{pix2pix2017}
\begin{botherref}
\oauthor{\bsnm{Isola}, \binits{P.}},
\oauthor{\bsnm{Zhu}, \binits{J.-Y.}},
\oauthor{\bsnm{Zhou}, \binits{T.}},
\oauthor{\bsnm{Efros}, \binits{A.A.}}:
Image-to-image translation with conditional adversarial networks.
CVPR
(2017)
\end{botherref}
\endbibitem

\bibitem[\protect\citeauthoryear{Varma et~al.}{2023}]{gnt2023}
\begin{bchapter}
\bauthor{\bsnm{Varma}, \binits{M.}},
\bauthor{\bsnm{Wang}, \binits{P.}},
\bauthor{\bsnm{Chen}, \binits{X.}},
\bauthor{\bsnm{Chen}, \binits{T.}},
\bauthor{\bsnm{Venugopalan}, \binits{S.}},
\bauthor{\bsnm{Wang}, \binits{Z.}}:
\bctitle{Is attention all that ne{RF} needs?}
In: \bbtitle{The Eleventh International Conference on Learning Representations}
(\byear{2023}).
\burl{https://openreview.net/forum?id=xE-LtsE-xx}
\end{bchapter}
\endbibitem

\bibitem[\protect\citeauthoryear{Peebles and Xie}{2022}]{Peebles2022DiT}
\begin{botherref}
\oauthor{\bsnm{Peebles}, \binits{W.}},
\oauthor{\bsnm{Xie}, \binits{S.}}:
Scalable diffusion models with transformers.
arXiv preprint arXiv:2212.09748
(2022)
\end{botherref}
\endbibitem

\bibitem[\protect\citeauthoryear{Reed et~al.}{2014}]{reed2014bootce}
\begin{botherref}
\oauthor{\bsnm{Reed}, \binits{S.}},
\oauthor{\bsnm{Lee}, \binits{H.}},
\oauthor{\bsnm{Anguelov}, \binits{D.}},
\oauthor{\bsnm{Szegedy}, \binits{C.}},
\oauthor{\bsnm{Erhan}, \binits{D.}},
\oauthor{\bsnm{Rabinovich}, \binits{A.}}:
Training deep neural networks on noisy labels with bootstrapping.
arXiv preprint arXiv:1412.6596
(2014)
\end{botherref}
\endbibitem

\bibitem[\protect\citeauthoryear{Sudre et~al.}{2017}]{sudre2017dice}
\begin{bchapter}
\bauthor{\bsnm{Sudre}, \binits{C.H.}},
\bauthor{\bsnm{Li}, \binits{W.}},
\bauthor{\bsnm{Vercauteren}, \binits{T.}},
\bauthor{\bsnm{Ourselin}, \binits{S.}},
\bauthor{\bsnm{Jorge~Cardoso}, \binits{M.}}:
\bctitle{Generalised dice overlap as a deep learning loss function for highly unbalanced segmentations}.
In: \bbtitle{Deep Learning in Medical Image Analysis and Multimodal Learning for Clinical Decision Support: Third International Workshop, DLMIA 2017, and 7th International Workshop, ML-CDS 2017, Held in Conjunction with MICCAI 2017, Qu{\'e}bec City, QC, Canada, September 14, Proceedings 3},
pp. \bfpage{240}--\blpage{248}
(\byear{2017}).
\bcomment{Springer}
\end{bchapter}
\endbibitem

\bibitem[\protect\citeauthoryear{Perez et~al.}{2018}]{perez2018adaln}
\begin{bchapter}
\bauthor{\bsnm{Perez}, \binits{E.}},
\bauthor{\bsnm{Strub}, \binits{F.}},
\bauthor{\bsnm{De~Vries}, \binits{H.}},
\bauthor{\bsnm{Dumoulin}, \binits{V.}},
\bauthor{\bsnm{Courville}, \binits{A.}}:
\bctitle{Film: Visual reasoning with a general conditioning layer}.
In: \bbtitle{Proceedings of the AAAI Conference on Artificial Intelligence},
vol. \bseriesno{32}
(\byear{2018})
\end{bchapter}
\endbibitem

\bibitem[\protect\citeauthoryear{Bertasius et~al.}{2021}]{bertasius2021space}
\begin{bchapter}
\bauthor{\bsnm{Bertasius}, \binits{G.}},
\bauthor{\bsnm{Wang}, \binits{H.}},
\bauthor{\bsnm{Torresani}, \binits{L.}}:
\bctitle{Is space-time attention all you need for video understanding?}
In: \bbtitle{ICML}
(\byear{2021})
\end{bchapter}
\endbibitem

\bibitem[\protect\citeauthoryear{Goyal et~al.}{2017}]{goyal2017something}
\begin{bchapter}
\bauthor{\bsnm{Goyal}, \binits{R.}},
\bauthor{\bsnm{Ebrahimi~Kahou}, \binits{S.}},
\bauthor{\bsnm{Michalski}, \binits{V.}},
\bauthor{\bsnm{Materzynska}, \binits{J.}},
\bauthor{\bsnm{Westphal}, \binits{S.}},
\bauthor{\bsnm{Kim}, \binits{H.}},
\bauthor{\bsnm{Haenel}, \binits{V.}},
\bauthor{\bsnm{Fruend}, \binits{I.}},
\bauthor{\bsnm{Yianilos}, \binits{P.}},
\bauthor{\bsnm{Mueller-Freitag}, \binits{M.}}, \betal:
\bctitle{The" something something" video database for learning and evaluating visual common sense}.
In: \bbtitle{Proceedings of the IEEE International Conference on Computer Vision},
pp. \bfpage{5842}--\blpage{5850}
(\byear{2017})
\end{bchapter}
\endbibitem

\bibitem[\protect\citeauthoryear{Song et~al.}{2023}]{goalstep}
\begin{bchapter}
\bauthor{\bsnm{Song}, \binits{Y.}},
\bauthor{\bsnm{Byrne}, \binits{E.}},
\bauthor{\bsnm{Nagarajan}, \binits{T.}},
\bauthor{\bsnm{Wang}, \binits{H.}},
\bauthor{\bsnm{Martin}, \binits{M.}},
\bauthor{\bsnm{Torresani}, \binits{L.}}:
\bctitle{Ego4d goal-step: Toward hierarchical understanding of procedural activities}.
In: \bbtitle{NeurIPS}
(\byear{2023})
\end{bchapter}
\endbibitem

\bibitem[\protect\citeauthoryear{Mavroudi et~al.}{2022}]{htstep}
\begin{bchapter}
\bauthor{\bsnm{Mavroudi}, \binits{E.}},
\bauthor{\bsnm{Afouras}, \binits{T.}},
\bauthor{\bsnm{Torresani}, \binits{L.}}:
\bctitle{Learning to ground instructional articles in videos through narrations}.
(\byear{2022})
\end{bchapter}
\endbibitem

\bibitem[\protect\citeauthoryear{Oord et~al.}{2018}]{oord2018representation}
\begin{botherref}
\oauthor{\bsnm{Oord}, \binits{A.v.d.}},
\oauthor{\bsnm{Li}, \binits{Y.}},
\oauthor{\bsnm{Vinyals}, \binits{O.}}:
Representation learning with contrastive predictive coding.
arXiv preprint arXiv:1807.03748
(2018)
\end{botherref}
\endbibitem

\bibitem[\protect\citeauthoryear{Sermanet et~al.}{2017}]{sermanet2017time}
\begin{bchapter}
\bauthor{\bsnm{Sermanet}, \binits{P.}},
\bauthor{\bsnm{Lynch}, \binits{C.}},
\bauthor{\bsnm{Hsu}, \binits{J.}},
\bauthor{\bsnm{Levine}, \binits{S.}}:
\bctitle{Time-contrastive networks: Self-supervised learning from multi-view observation}.
In: \bbtitle{2017 IEEE Conference on Computer Vision and Pattern Recognition Workshops (CVPRW)},
pp. \bfpage{486}--\blpage{487}
(\byear{2017}).
\bcomment{IEEE}
\end{bchapter}
\endbibitem

\bibitem[\protect\citeauthoryear{Hinton et~al.}{2015}]{hinton2015distilling}
\begin{botherref}
\oauthor{\bsnm{Hinton}, \binits{G.}},
\oauthor{\bsnm{Vinyals}, \binits{O.}},
\oauthor{\bsnm{Dean}, \binits{J.}}:
Distilling the knowledge in a neural network.
arXiv preprint arXiv:1503.02531
(2015)
\end{botherref}
\endbibitem

\bibitem[\protect\citeauthoryear{Tong et~al.}{2022}]{videomae}
\begin{bchapter}
\bauthor{\bsnm{Tong}, \binits{Z.}},
\bauthor{\bsnm{Song}, \binits{Y.}},
\bauthor{\bsnm{Wang}, \binits{J.}},
\bauthor{\bsnm{Wang}, \binits{L.}}:
\bctitle{Videomae: Masked autoencoders are data-efficient learners for self-supervised video pre-training}.
(\byear{2022}).
\burl{https://arxiv.org/abs/2203.12602}
\end{bchapter}
\endbibitem

\bibitem[\protect\citeauthoryear{Feichtenhofer}{2020}]{feichtenhofer2020x3d}
\begin{bchapter}
\bauthor{\bsnm{Feichtenhofer}, \binits{C.}}:
\bctitle{X3d: Expanding architectures for efficient video recognition}.
In: \bbtitle{Proceedings of the IEEE/CVF Conference on Computer Vision and Pattern Recognition},
pp. \bfpage{203}--\blpage{213}
(\byear{2020})
\end{bchapter}
\endbibitem

\bibitem[\protect\citeauthoryear{Vasu et~al.}{2023}]{vasu2023mobileone}
\begin{bchapter}
\bauthor{\bsnm{Vasu}, \binits{P.K.A.}},
\bauthor{\bsnm{Gabriel}, \binits{J.}},
\bauthor{\bsnm{Zhu}, \binits{J.}},
\bauthor{\bsnm{Tuzel}, \binits{O.}},
\bauthor{\bsnm{Ranjan}, \binits{A.}}:
\bctitle{Mobileone: An improved one millisecond mobile backbone}.
In: \bbtitle{Proceedings of the IEEE/CVF Conference on Computer Vision and Pattern Recognition},
pp. \bfpage{7907}--\blpage{7917}
(\byear{2023})
\end{bchapter}
\endbibitem

\bibitem[\protect\citeauthoryear{Howard et~al.}{2017}]{howard2017mobilenets}
\begin{botherref}
\oauthor{\bsnm{Howard}, \binits{A.G.}},
\oauthor{\bsnm{Zhu}, \binits{M.}},
\oauthor{\bsnm{Chen}, \binits{B.}},
\oauthor{\bsnm{Kalenichenko}, \binits{D.}},
\oauthor{\bsnm{Wang}, \binits{W.}},
\oauthor{\bsnm{Weyand}, \binits{T.}},
\oauthor{\bsnm{Andreetto}, \binits{M.}},
\oauthor{\bsnm{Adam}, \binits{H.}}:
Mobilenets: Efficient convolutional neural networks for mobile vision applications.
arXiv preprint arXiv:1704.04861
(2017)
\end{botherref}
\endbibitem

\bibitem[\protect\citeauthoryear{Zhang et~al.}{2018}]{zhang2018shufflenet}
\begin{bchapter}
\bauthor{\bsnm{Zhang}, \binits{X.}},
\bauthor{\bsnm{Zhou}, \binits{X.}},
\bauthor{\bsnm{Lin}, \binits{M.}},
\bauthor{\bsnm{Sun}, \binits{J.}}:
\bctitle{Shufflenet: An extremely efficient convolutional neural network for mobile devices}.
In: \bbtitle{Proceedings of the IEEE Conference on Computer Vision and Pattern Recognition},
pp. \bfpage{6848}--\blpage{6856}
(\byear{2018})
\end{bchapter}
\endbibitem

\bibitem[\protect\citeauthoryear{Tan et~al.}{2019}]{tan2019mnasnet}
\begin{bchapter}
\bauthor{\bsnm{Tan}, \binits{M.}},
\bauthor{\bsnm{Chen}, \binits{B.}},
\bauthor{\bsnm{Pang}, \binits{R.}},
\bauthor{\bsnm{Vasudevan}, \binits{V.}},
\bauthor{\bsnm{Sandler}, \binits{M.}},
\bauthor{\bsnm{Howard}, \binits{A.}},
\bauthor{\bsnm{Le}, \binits{Q.V.}}:
\bctitle{Mnasnet: Platform-aware neural architecture search for mobile}.
In: \bbtitle{Proceedings of the IEEE/CVF Conference on Computer Vision and Pattern Recognition},
pp. \bfpage{2820}--\blpage{2828}
(\byear{2019})
\end{bchapter}
\endbibitem

\bibitem[\protect\citeauthoryear{Mehta and Rastegari}{2021}]{mehta2021mobilevit}
\begin{botherref}
\oauthor{\bsnm{Mehta}, \binits{S.}},
\oauthor{\bsnm{Rastegari}, \binits{M.}}:
Mobilevit: light-weight, general-purpose, and mobile-friendly vision transformer.
arXiv preprint arXiv:2110.02178
(2021)
\end{botherref}
\endbibitem

\bibitem[\protect\citeauthoryear{Gao et~al.}{2020}]{gao2020listen}
\begin{bchapter}
\bauthor{\bsnm{Gao}, \binits{R.}},
\bauthor{\bsnm{Oh}, \binits{T.-H.}},
\bauthor{\bsnm{Grauman}, \binits{K.}},
\bauthor{\bsnm{Torresani}, \binits{L.}}:
\bctitle{Listen to look: Action recognition by previewing audio}.
In: \bbtitle{CVPR}
(\byear{2020})
\end{bchapter}
\endbibitem

\bibitem[\protect\citeauthoryear{Korbar et~al.}{2019}]{korbar2019scsampler}
\begin{bchapter}
\bauthor{\bsnm{Korbar}, \binits{B.}},
\bauthor{\bsnm{Tran}, \binits{D.}},
\bauthor{\bsnm{Torresani}, \binits{L.}}:
\bctitle{Scsampler: Sampling salient clips from video for efficient action recognition}.
In: \bbtitle{Proceedings of the IEEE/CVF International Conference on Computer Vision}
(\byear{2019})
\end{bchapter}
\endbibitem

\bibitem[\protect\citeauthoryear{Ghodrati et~al.}{2021}]{ghodrati2021frameexit}
\begin{bchapter}
\bauthor{\bsnm{Ghodrati}, \binits{A.}},
\bauthor{\bsnm{Bejnordi}, \binits{B.E.}},
\bauthor{\bsnm{Habibian}, \binits{A.}}:
\bctitle{Frameexit: Conditional early exiting for efficient video recognition}.
In: \bbtitle{Proceedings of the IEEE/CVF Conference on Computer Vision and Pattern Recognition}
(\byear{2021})
\end{bchapter}
\endbibitem

\bibitem[\protect\citeauthoryear{Meng et~al.}{2020}]{meng2020ar}
\begin{bchapter}
\bauthor{\bsnm{Meng}, \binits{Y.}},
\bauthor{\bsnm{Lin}, \binits{C.-C.}},
\bauthor{\bsnm{Panda}, \binits{R.}},
\bauthor{\bsnm{Sattigeri}, \binits{P.}},
\bauthor{\bsnm{Karlinsky}, \binits{L.}},
\bauthor{\bsnm{Oliva}, \binits{A.}},
\bauthor{\bsnm{Saenko}, \binits{K.}},
\bauthor{\bsnm{Feris}, \binits{R.}}:
\bctitle{Ar-net: Adaptive frame resolution for efficient action recognition}.
In: \bbtitle{ECCV 2020}
(\byear{2020})
\end{bchapter}
\endbibitem

\bibitem[\protect\citeauthoryear{Tan et~al.}{2023}]{tan2023egodistill}
\begin{botherref}
\oauthor{\bsnm{Tan}, \binits{S.}},
\oauthor{\bsnm{Nagarajan}, \binits{T.}},
\oauthor{\bsnm{Grauman}, \binits{K.}}:
Egodistill: Egocentric head motion distillation for efficient video understanding.
NeurIPS
(2023)
\end{botherref}
\endbibitem

\bibitem[\protect\citeauthoryear{Iandola et~al.}{2016}]{iandola2016squeezenet}
\begin{botherref}
\oauthor{\bsnm{Iandola}, \binits{F.N.}},
\oauthor{\bsnm{Han}, \binits{S.}},
\oauthor{\bsnm{Moskewicz}, \binits{M.W.}},
\oauthor{\bsnm{Ashraf}, \binits{K.}},
\oauthor{\bsnm{Dally}, \binits{W.J.}},
\oauthor{\bsnm{Keutzer}, \binits{K.}}:
Squeezenet: Alexnet-level accuracy with 50x fewer parameters and< 0.5 mb model size.
arXiv preprint arXiv:1602.07360
(2016)
\end{botherref}
\endbibitem

\bibitem[\protect\citeauthoryear{Esser et~al.}{2019}]{esser2019learned}
\begin{botherref}
\oauthor{\bsnm{Esser}, \binits{S.K.}},
\oauthor{\bsnm{McKinstry}, \binits{J.L.}},
\oauthor{\bsnm{Bablani}, \binits{D.}},
\oauthor{\bsnm{Appuswamy}, \binits{R.}},
\oauthor{\bsnm{Modha}, \binits{D.S.}}:
Learned step size quantization.
arXiv preprint arXiv:1902.08153
(2019)
\end{botherref}
\endbibitem

\bibitem[\protect\citeauthoryear{Polino et~al.}{2018}]{polino2018model}
\begin{botherref}
\oauthor{\bsnm{Polino}, \binits{A.}},
\oauthor{\bsnm{Pascanu}, \binits{R.}},
\oauthor{\bsnm{Alistarh}, \binits{D.}}:
Model compression via distillation and quantization.
arXiv preprint arXiv:1802.05668
(2018)
\end{botherref}
\endbibitem

\bibitem[\protect\citeauthoryear{Zhu and Gupta}{2017}]{zhu2017prune}
\begin{botherref}
\oauthor{\bsnm{Zhu}, \binits{M.}},
\oauthor{\bsnm{Gupta}, \binits{S.}}:
To prune, or not to prune: exploring the efficacy of pruning for model compression.
arXiv preprint arXiv:1710.01878
(2017)
\end{botherref}
\endbibitem

\bibitem[\protect\citeauthoryear{Wu et~al.}{2018}]{blockdrop}
\begin{bchapter}
\bauthor{\bsnm{Wu}, \binits{Z.}},
\bauthor{\bsnm{Nagarajan}, \binits{T.}},
\bauthor{\bsnm{Kumar}, \binits{A.}},
\bauthor{\bsnm{Rennie}, \binits{S.}},
\bauthor{\bsnm{Davis}, \binits{L.S.}},
\bauthor{\bsnm{Grauman}, \binits{K.}},
\bauthor{\bsnm{Feris}, \binits{R.}}:
\bctitle{Blockdrop: Dynamic inference paths in residual networks}.
In: \bbtitle{CVPR}
(\byear{2018})
\end{bchapter}
\endbibitem

\bibitem[\protect\citeauthoryear{Abrash}{2021}]{abrash2021creating}
\begin{bchapter}
\bauthor{\bsnm{Abrash}, \binits{M.}}:
\bctitle{Creating the future: Augmented reality, the next human-machine interface}.
In: \bbtitle{2021 IEEE International Electron Devices Meeting (IEDM)}
(\byear{2021})
\end{bchapter}
\endbibitem

\bibitem[\protect\citeauthoryear{Chen et~al.}{2019}]{chen2019eyeriss}
\begin{botherref}
\oauthor{\bsnm{Chen}, \binits{Y.-H.}},
\oauthor{\bsnm{Yang}, \binits{T.-J.}},
\oauthor{\bsnm{Emer}, \binits{J.}},
\oauthor{\bsnm{Sze}, \binits{V.}}:
Eyeriss v2: A flexible accelerator for emerging deep neural networks on mobile devices.
IEEE Journal on Emerging and Selected Topics in Circuits and Systems
(2019)
\end{botherref}
\endbibitem

\bibitem[\protect\citeauthoryear{Yang et~al.}{2022}]{yang2022three}
\begin{botherref}
\oauthor{\bsnm{Yang}, \binits{L.}},
\oauthor{\bsnm{Radway}, \binits{R.M.}},
\oauthor{\bsnm{Chen}, \binits{Y.-H.}},
\oauthor{\bsnm{Wu}, \binits{T.F.}},
\oauthor{\bsnm{Liu}, \binits{H.}},
\oauthor{\bsnm{Ansari}, \binits{E.}},
\oauthor{\bsnm{Chandra}, \binits{V.}},
\oauthor{\bsnm{Mitra}, \binits{S.}},
\oauthor{\bsnm{Beign{\'e}}, \binits{E.}}:
Three-dimensional stacked neural network accelerator architectures for ar/vr applications.
IEEE Micro
(2022)
\end{botherref}
\endbibitem

\bibitem[\protect\citeauthoryear{Sze et~al.}{2020}]{sze2020evaluate}
\begin{botherref}
\oauthor{\bsnm{Sze}, \binits{V.}},
\oauthor{\bsnm{Chen}, \binits{Y.-H.}},
\oauthor{\bsnm{Yang}, \binits{T.-J.}},
\oauthor{\bsnm{Emer}, \binits{J.S.}}:
How to evaluate deep neural network processors: Tops/w (alone) considered harmful.
IEEE Solid-State Circuits Magazine
(2020)
\end{botherref}
\endbibitem

\bibitem[\protect\citeauthoryear{Desislavov et~al.}{2023}]{desislavov2023trends}
\begin{barticle}
\bauthor{\bsnm{Desislavov}, \binits{R.}},
\bauthor{\bsnm{Mart{\'\i}nez-Plumed}, \binits{F.}},
\bauthor{\bsnm{Hern{\'a}ndez-Orallo}, \binits{J.}}:
\batitle{Trends in ai inference energy consumption: Beyond the performance-vs-parameter laws of deep learning}.
\bjtitle{Sustainable Computing: Informatics and Systems}
\bvolume{38},
\bfpage{100857}
(\byear{2023})
\end{barticle}
\endbibitem

\bibitem[\protect\citeauthoryear{Horowitz}{2014}]{horowitz20141}
\begin{bchapter}
\bauthor{\bsnm{Horowitz}, \binits{M.}}:
\bctitle{1.1 computing's energy problem (and what we can do about it)}.
In: \bbtitle{2014 IEEE International Solid-state Circuits Conference Digest of Technical Papers (ISSCC)}
(\byear{2014})
\end{bchapter}
\endbibitem

\bibitem[\protect\citeauthoryear{Liu et~al.}{2020}]{liu20204}
\begin{bchapter}
\bauthor{\bsnm{Liu}, \binits{C.}},
\bauthor{\bsnm{Bainbridge}, \binits{L.}},
\bauthor{\bsnm{Berkovich}, \binits{A.}},
\bauthor{\bsnm{Chen}, \binits{S.}},
\bauthor{\bsnm{Gao}, \binits{W.}},
\bauthor{\bsnm{Tsai}, \binits{T.-H.}},
\bauthor{\bsnm{Mori}, \binits{K.}},
\bauthor{\bsnm{Ikeno}, \binits{R.}},
\bauthor{\bsnm{Uno}, \binits{M.}},
\bauthor{\bsnm{Isozaki}, \binits{T.}}, \betal:
\bctitle{A 4.6 $\mu$m, 512$\times$ 512, ultra-low power stacked digital pixel sensor with triple quantization and 127db dynamic range}.
In: \bbtitle{2020 IEEE International Electron Devices Meeting (IEDM)}
(\byear{2020})
\end{bchapter}
\endbibitem

\bibitem[\protect\citeauthoryear{De~Geest et~al.}{2016}]{de2016online}
\begin{bchapter}
\bauthor{\bsnm{De~Geest}, \binits{R.}},
\bauthor{\bsnm{Gavves}, \binits{E.}},
\bauthor{\bsnm{Ghodrati}, \binits{A.}},
\bauthor{\bsnm{Li}, \binits{Z.}},
\bauthor{\bsnm{Snoek}, \binits{C.}},
\bauthor{\bsnm{Tuytelaars}, \binits{T.}}:
\bctitle{Online action detection}.
In: \bbtitle{Computer Vision--ECCV 2016: 14th European Conference, Amsterdam, The Netherlands, October 11-14, 2016, Proceedings, Part V 14},
pp. \bfpage{269}--\blpage{284}
(\byear{2016}).
\bcomment{Springer}
\end{bchapter}
\endbibitem

\bibitem[\protect\citeauthoryear{Liao et~al.}{2023}]{liao2023light}
\begin{bchapter}
\bauthor{\bsnm{Liao}, \binits{J.}},
\bauthor{\bsnm{Duan}, \binits{H.}},
\bauthor{\bsnm{Feng}, \binits{K.}},
\bauthor{\bsnm{Zhao}, \binits{W.}},
\bauthor{\bsnm{Yang}, \binits{Y.}},
\bauthor{\bsnm{Chen}, \binits{L.}}:
\bctitle{A light weight model for active speaker detection}.
In: \bbtitle{Proceedings of the IEEE/CVF Conference on Computer Vision and Pattern Recognition},
pp. \bfpage{22932}--\blpage{22941}
(\byear{2023})
\end{bchapter}
\endbibitem

\bibitem[\protect\citeauthoryear{Chang et~al.}{2020}]{chang2020procedure}
\begin{bchapter}
\bauthor{\bsnm{Chang}, \binits{C.-Y.}},
\bauthor{\bsnm{Huang}, \binits{D.-A.}},
\bauthor{\bsnm{Xu}, \binits{D.}},
\bauthor{\bsnm{Adeli}, \binits{E.}},
\bauthor{\bsnm{Fei-Fei}, \binits{L.}},
\bauthor{\bsnm{Niebles}, \binits{J.C.}}:
\bctitle{Procedure planning in instructional videos}.
In: \bbtitle{European Conference on Computer Vision},
pp. \bfpage{334}--\blpage{350}
(\byear{2020}).
\bcomment{Springer}
\end{bchapter}
\endbibitem

\bibitem[\protect\citeauthoryear{Bi et~al.}{2021}]{bi2021procedure}
\begin{bchapter}
\bauthor{\bsnm{Bi}, \binits{J.}},
\bauthor{\bsnm{Luo}, \binits{J.}},
\bauthor{\bsnm{Xu}, \binits{C.}}:
\bctitle{Procedure planning in instructional videos via contextual modeling and model-based policy learning}.
In: \bbtitle{Proceedings of the IEEE/CVF International Conference on Computer Vision},
pp. \bfpage{15611}--\blpage{15620}
(\byear{2021})
\end{bchapter}
\endbibitem

\bibitem[\protect\citeauthoryear{Zhou et~al.}{2023}]{zhou2023procedure}
\begin{bchapter}
\bauthor{\bsnm{Zhou}, \binits{H.}},
\bauthor{\bsnm{Mart{\'\i}n-Mart{\'\i}n}, \binits{R.}},
\bauthor{\bsnm{Kapadia}, \binits{M.}},
\bauthor{\bsnm{Savarese}, \binits{S.}},
\bauthor{\bsnm{Niebles}, \binits{J.C.}}:
\bctitle{Procedure-aware pretraining for instructional video understanding}.
In: \bbtitle{CVPR},
pp. \bfpage{10727}--\blpage{10738}
(\byear{2023})
\end{bchapter}
\endbibitem

\bibitem[\protect\citeauthoryear{Seminara et~al.}{2024}]{seminara2024differentiable}
\begin{botherref}
\oauthor{\bsnm{Seminara}, \binits{L.}},
\oauthor{\bsnm{Farinella}, \binits{G.M.}},
\oauthor{\bsnm{Furnari}, \binits{A.}}:
Differentiable Task Graph Learning: Procedural Activity Representation and Online Mistake Detection from Egocentric Videos
(2024)
\end{botherref}
\endbibitem

\bibitem[\protect\citeauthoryear{Jang et~al.}{2023}]{jang2023multimodal}
\begin{botherref}
\oauthor{\bsnm{Jang}, \binits{Y.}},
\oauthor{\bsnm{Sohn}, \binits{S.}},
\oauthor{\bsnm{Logeswaran}, \binits{L.}},
\oauthor{\bsnm{Luo}, \binits{T.}},
\oauthor{\bsnm{Lee}, \binits{M.}},
\oauthor{\bsnm{Lee}, \binits{H.}}:
Multimodal subtask graph generation from instructional videos.
arXiv preprint arXiv:2302.08672
(2023)
\end{botherref}
\endbibitem

\bibitem[\protect\citeauthoryear{Xu et~al.}{2020}]{xu2020benchmark}
\begin{botherref}
\oauthor{\bsnm{Xu}, \binits{F.F.}},
\oauthor{\bsnm{Ji}, \binits{L.}},
\oauthor{\bsnm{Shi}, \binits{B.}},
\oauthor{\bsnm{Du}, \binits{J.}},
\oauthor{\bsnm{Neubig}, \binits{G.}},
\oauthor{\bsnm{Bisk}, \binits{Y.}},
\oauthor{\bsnm{Duan}, \binits{N.}}:
A benchmark for structured procedural knowledge extraction from cooking videos.
arXiv preprint arXiv:2005.00706
(2020)
\end{botherref}
\endbibitem

\bibitem[\protect\citeauthoryear{Soran et~al.}{2015}]{soran2015generating}
\begin{bchapter}
\bauthor{\bsnm{Soran}, \binits{B.}},
\bauthor{\bsnm{Farhadi}, \binits{A.}},
\bauthor{\bsnm{Shapiro}, \binits{L.}}:
\bctitle{Generating notifications for missing actions: Don't forget to turn the lights off!}
In: \bbtitle{ICCV},
pp. \bfpage{4669}--\blpage{4677}
(\byear{2015})
\end{bchapter}
\endbibitem

\bibitem[\protect\citeauthoryear{Ding et~al.}{2023}]{ding2023every}
\begin{botherref}
\oauthor{\bsnm{Ding}, \binits{G.}},
\oauthor{\bsnm{Sener}, \binits{F.}},
\oauthor{\bsnm{Ma}, \binits{S.}},
\oauthor{\bsnm{Yao}, \binits{A.}}:
Every mistake counts in assembly.
arXiv preprint arXiv:2307.16453
(2023)
\end{botherref}
\endbibitem

\bibitem[\protect\citeauthoryear{Parmar and Morris}{2017}]{parmar2017learning}
\begin{botherref}
\oauthor{\bsnm{Parmar}, \binits{P.}},
\oauthor{\bsnm{Morris}, \binits{B.T.}}:
Learning To Score Olympic Events
(2017)
\end{botherref}
\endbibitem

\bibitem[\protect\citeauthoryear{Parmar and Tran~Morris}{2019}]{mtlaqa}
\begin{bchapter}
\bauthor{\bsnm{Parmar}, \binits{P.}},
\bauthor{\bsnm{Tran~Morris}, \binits{B.}}:
\bctitle{What and how well you performed? a multitask learning approach to action quality assessment}.
In: \bbtitle{Proceedings of the IEEE Conference on Computer Vision and Pattern Recognition},
pp. \bfpage{304}--\blpage{313}
(\byear{2019})
\end{bchapter}
\endbibitem

\bibitem[\protect\citeauthoryear{Ismail~Fawaz et~al.}{2018}]{10.1007/978-3-030-00937-3_25}
\begin{bchapter}
\bauthor{\bsnm{Ismail~Fawaz}, \binits{H.}},
\bauthor{\bsnm{Forestier}, \binits{G.}},
\bauthor{\bsnm{Weber}, \binits{J.}},
\bauthor{\bsnm{Idoumghar}, \binits{L.}},
\bauthor{\bsnm{Muller}, \binits{P.-A.}}:
\bctitle{Evaluating surgical skills from kinematic data using convolutional neural networks}.
In: \bbtitle{Medical Image Computing and Computer Assisted Intervention -- MICCAI 2018},
pp. \bfpage{214}--\blpage{221}
(\byear{2018})
\end{bchapter}
\endbibitem

\bibitem[\protect\citeauthoryear{Liu et~al.}{2021}]{Liu_2021_CVPR}
\begin{bchapter}
\bauthor{\bsnm{Liu}, \binits{D.}},
\bauthor{\bsnm{Li}, \binits{Q.}},
\bauthor{\bsnm{Jiang}, \binits{T.}},
\bauthor{\bsnm{Wang}, \binits{Y.}},
\bauthor{\bsnm{Miao}, \binits{R.}},
\bauthor{\bsnm{Shan}, \binits{F.}},
\bauthor{\bsnm{Li}, \binits{Z.}}:
\bctitle{Towards unified surgical skill assessment}.
In: \bbtitle{Proceedings of the IEEE/CVF Conference on Computer Vision and Pattern Recognition (CVPR)},
pp. \bfpage{9522}--\blpage{9531}
(\byear{2021})
\end{bchapter}
\endbibitem

\bibitem[\protect\citeauthoryear{Zhang and Li}{2013}]{Zhang_2013_CVPR}
\begin{bchapter}
\bauthor{\bsnm{Zhang}, \binits{Q.}},
\bauthor{\bsnm{Li}, \binits{B.}}:
\bctitle{Relative hidden markov models for evaluating motion skill}.
In: \bbtitle{Proceedings of the IEEE Conference on Computer Vision and Pattern Recognition (CVPR)}
(\byear{2013})
\end{bchapter}
\endbibitem

\bibitem[\protect\citeauthoryear{Zia et~al.}{2017}]{DBLP:journals/corr/ZiaSBSE17}
\begin{botherref}
\oauthor{\bsnm{Zia}, \binits{A.}},
\oauthor{\bsnm{Sharma}, \binits{Y.}},
\oauthor{\bsnm{Bettadapura}, \binits{V.}},
\oauthor{\bsnm{Sarin}, \binits{E.L.}},
\oauthor{\bsnm{Essa}, \binits{I.A.}}:
Video and accelerometer-based motion analysis for automated surgical skills assessment.
CoRR
\textbf{abs/1702.07772}
(2017)
{\href{https://arxiv.org/abs/1702.07772}{{1702.07772}}}
\end{botherref}
\endbibitem

\bibitem[\protect\citeauthoryear{Yu et~al.}{2021}]{Yu_2021_ICCV}
\begin{bchapter}
\bauthor{\bsnm{Yu}, \binits{X.}},
\bauthor{\bsnm{Rao}, \binits{Y.}},
\bauthor{\bsnm{Zhao}, \binits{W.}},
\bauthor{\bsnm{Lu}, \binits{J.}},
\bauthor{\bsnm{Zhou}, \binits{J.}}:
\bctitle{Group-aware contrastive regression for action quality assessment}.
In: \bbtitle{Proceedings of the IEEE/CVF International Conference on Computer Vision (ICCV)},
pp. \bfpage{7919}--\blpage{7928}
(\byear{2021})
\end{bchapter}
\endbibitem

\bibitem[\protect\citeauthoryear{Zhang et~al.}{2022}]{zhang2022actionformer}
\begin{bchapter}
\bauthor{\bsnm{Zhang}, \binits{C.-L.}},
\bauthor{\bsnm{Wu}, \binits{J.}},
\bauthor{\bsnm{Li}, \binits{Y.}}:
\bctitle{Actionformer: Localizing moments of actions with transformers}.
In: \bbtitle{European Conference on Computer Vision}.
\bsertitle{LNCS},
vol. \bseriesno{13664},
pp. \bfpage{492}--\blpage{510}
(\byear{2022})
\end{bchapter}
\endbibitem

\bibitem[\protect\citeauthoryear{Girdhar et~al.}{2022}]{omnivore}
\begin{bchapter}
\bauthor{\bsnm{Girdhar}, \binits{R.}},
\bauthor{\bsnm{Singh}, \binits{M.}},
\bauthor{\bsnm{Ravi}, \binits{N.}},
\bauthor{\bsnm{Maaten}, \binits{L.}},
\bauthor{\bsnm{Joulin}, \binits{A.}},
\bauthor{\bsnm{Misra}, \binits{I.}}:
\bctitle{Omnivore: A single model for many visual modalities}.
In: \bbtitle{Proceedings of the IEEE/CVF Conference on Computer Vision and Pattern Recognition},
pp. \bfpage{16102}--\blpage{16112}
(\byear{2022})
\end{bchapter}
\endbibitem

\bibitem[\protect\citeauthoryear{Lin et~al.}{2014}]{lin2014microsoft}
\begin{bchapter}
\bauthor{\bsnm{Lin}, \binits{T.-Y.}},
\bauthor{\bsnm{Maire}, \binits{M.}},
\bauthor{\bsnm{Belongie}, \binits{S.}},
\bauthor{\bsnm{Hays}, \binits{J.}},
\bauthor{\bsnm{Perona}, \binits{P.}},
\bauthor{\bsnm{Ramanan}, \binits{D.}},
\bauthor{\bsnm{Doll{\'a}r}, \binits{P.}},
\bauthor{\bsnm{Zitnick}, \binits{C.L.}}:
\bctitle{Microsoft {C}{O}{C}{O}: Common objects in context}.
In: \bbtitle{ECCV}
(\byear{2014})
\end{bchapter}
\endbibitem

\bibitem[\protect\citeauthoryear{Jiang and Grauman}{2017}]{jiang2017seeing}
\begin{bchapter}
\bauthor{\bsnm{Jiang}, \binits{H.}},
\bauthor{\bsnm{Grauman}, \binits{K.}}:
\bctitle{Seeing invisible poses: Estimating 3d body pose from egocentric video}.
In: \bbtitle{CVPR}
(\byear{2017})
\end{bchapter}
\endbibitem

\bibitem[\protect\citeauthoryear{Yuan and Kitani}{2018}]{Yuan_2018_ECCV}
\begin{bchapter}
\bauthor{\bsnm{Yuan}, \binits{Y.}},
\bauthor{\bsnm{Kitani}, \binits{K.}}:
\bctitle{3d ego-pose estimation via imitation learning}.
In: \bbtitle{Proceedings of the European Conference on Computer Vision (ECCV)}
(\byear{2018})
\end{bchapter}
\endbibitem

\bibitem[\protect\citeauthoryear{Yuan and Kitani}{2019}]{Yuan_2019_ICCV}
\begin{bchapter}
\bauthor{\bsnm{Yuan}, \binits{Y.}},
\bauthor{\bsnm{Kitani}, \binits{K.}}:
\bctitle{Ego-pose estimation and forecasting as real-time pd control}.
In: \bbtitle{Proceedings of the IEEE/CVF International Conference on Computer Vision (ICCV)}
(\byear{2019})
\end{bchapter}
\endbibitem

\bibitem[\protect\citeauthoryear{Luo et~al.}{2021}]{Luo2021DynamicsRegulatedKP}
\begin{bchapter}
\bauthor{\bsnm{Luo}, \binits{Z.}},
\bauthor{\bsnm{Hachiuma}, \binits{R.}},
\bauthor{\bsnm{Yuan}, \binits{Y.}},
\bauthor{\bsnm{Kitani}, \binits{K.}}:
\bctitle{Dynamics-regulated kinematic policy for egocentric pose estimation}.
In: \bbtitle{Advances in Neural Information Processing Systems}
(\byear{2021})
\end{bchapter}
\endbibitem

\bibitem[\protect\citeauthoryear{Rhodin et~al.}{2016}]{rhodin2016egocap}
\begin{barticle}
\bauthor{\bsnm{Rhodin}, \binits{H.}},
\bauthor{\bsnm{Richardt}, \binits{C.}},
\bauthor{\bsnm{Casas}, \binits{D.}},
\bauthor{\bsnm{Insafutdinov}, \binits{E.}},
\bauthor{\bsnm{Shafiei}, \binits{M.}},
\bauthor{\bsnm{Seidel}, \binits{H.-P.}},
\bauthor{\bsnm{Schiele}, \binits{B.}},
\bauthor{\bsnm{Theobalt}, \binits{C.}}:
\batitle{Egocap: egocentric marker-less motion capture with two fisheye cameras}.
\bjtitle{ACM Transactions on Graphics (TOG)}
\bvolume{35}(\bissue{6}),
\bfpage{1}--\blpage{11}
(\byear{2016})
\end{barticle}
\endbibitem

\bibitem[\protect\citeauthoryear{Tome et~al.}{2019}]{Tome_2019_ICCV}
\begin{bchapter}
\bauthor{\bsnm{Tome}, \binits{D.}},
\bauthor{\bsnm{Peluse}, \binits{P.}},
\bauthor{\bsnm{Agapito}, \binits{L.}},
\bauthor{\bsnm{Badino}, \binits{H.}}:
\bctitle{xr-egopose: Egocentric 3d human pose from an hmd camera}.
In: \bbtitle{Proceedings of the IEEE/CVF International Conference on Computer Vision (ICCV)}
(\byear{2019})
\end{bchapter}
\endbibitem

\bibitem[\protect\citeauthoryear{Xu et~al.}{2019}]{xu2019mo}
\begin{barticle}
\bauthor{\bsnm{Xu}, \binits{W.}},
\bauthor{\bsnm{Chatterjee}, \binits{A.}},
\bauthor{\bsnm{Zollhoefer}, \binits{M.}},
\bauthor{\bsnm{Rhodin}, \binits{H.}},
\bauthor{\bsnm{Fua}, \binits{P.}},
\bauthor{\bsnm{Seidel}, \binits{H.-P.}},
\bauthor{\bsnm{Theobalt}, \binits{C.}}:
\batitle{Mo 2 cap 2: Real-time mobile 3d motion capture with a cap-mounted fisheye camera}.
\bjtitle{IEEE transactions on visualization and computer graphics}
\bvolume{25}(\bissue{5}),
\bfpage{2093}--\blpage{2101}
(\byear{2019})
\end{barticle}
\endbibitem

\bibitem[\protect\citeauthoryear{Ahuja et~al.}{2019}]{ahuja2019mecap}
\begin{bchapter}
\bauthor{\bsnm{Ahuja}, \binits{K.}},
\bauthor{\bsnm{Harrison}, \binits{C.}},
\bauthor{\bsnm{Goel}, \binits{M.}},
\bauthor{\bsnm{Xiao}, \binits{R.}}:
\bctitle{Mecap: Whole-body digitization for low-cost vr/ar headsets}.
In: \bbtitle{Proceedings of the 32nd Annual ACM Symposium on User Interface Software and Technology},
pp. \bfpage{453}--\blpage{462}
(\byear{2019})
\end{bchapter}
\endbibitem

\bibitem[\protect\citeauthoryear{Hwang et~al.}{2020}]{hwang2020monoeye}
\begin{bchapter}
\bauthor{\bsnm{Hwang}, \binits{D.-H.}},
\bauthor{\bsnm{Aso}, \binits{K.}},
\bauthor{\bsnm{Yuan}, \binits{Y.}},
\bauthor{\bsnm{Kitani}, \binits{K.}},
\bauthor{\bsnm{Koike}, \binits{H.}}:
\bctitle{Monoeye: Multimodal human motion capture system using a single ultra-wide fisheye camera}.
In: \bbtitle{Proceedings of the 33rd Annual ACM Symposium on User Interface Software and Technology},
pp. \bfpage{98}--\blpage{111}
(\byear{2020})
\end{bchapter}
\endbibitem

\bibitem[\protect\citeauthoryear{Simon et~al.}{2017}]{simon2017hand}
\begin{bchapter}
\bauthor{\bsnm{Simon}, \binits{T.}},
\bauthor{\bsnm{Joo}, \binits{H.}},
\bauthor{\bsnm{Matthews}, \binits{I.}},
\bauthor{\bsnm{Sheikh}, \binits{Y.}}:
\bctitle{Hand keypoint detection in single images using multiview bootstrapping}.
In: \bbtitle{Proceedings of the IEEE Conference on Computer Vision and Pattern Recognition},
pp. \bfpage{1145}--\blpage{1153}
(\byear{2017})
\end{bchapter}
\endbibitem

\bibitem[\protect\citeauthoryear{Moon et~al.}{2020}]{moon2020interhand2}
\begin{bchapter}
\bauthor{\bsnm{Moon}, \binits{G.}},
\bauthor{\bsnm{Yu}, \binits{S.-I.}},
\bauthor{\bsnm{Wen}, \binits{H.}},
\bauthor{\bsnm{Shiratori}, \binits{T.}},
\bauthor{\bsnm{Lee}, \binits{K.M.}}:
\bctitle{Interhand2. 6m: A dataset and baseline for 3d interacting hand pose estimation from a single rgb image}.
In: \bbtitle{Computer Vision--ECCV 2020: 16th European Conference, Glasgow, UK, August 23--28, 2020, Proceedings, Part XX 16},
pp. \bfpage{548}--\blpage{564}
(\byear{2020}).
\bcomment{Springer}
\end{bchapter}
\endbibitem

\bibitem[\protect\citeauthoryear{Hampali et~al.}{2020}]{hampali2020honnotate}
\begin{bchapter}
\bauthor{\bsnm{Hampali}, \binits{S.}},
\bauthor{\bsnm{Rad}, \binits{M.}},
\bauthor{\bsnm{Oberweger}, \binits{M.}},
\bauthor{\bsnm{Lepetit}, \binits{V.}}:
\bctitle{Honnotate: A method for 3d annotation of hand and object poses}.
In: \bbtitle{Proceedings of the IEEE/CVF Conference on Computer Vision and Pattern Recognition},
pp. \bfpage{3196}--\blpage{3206}
(\byear{2020})
\end{bchapter}
\endbibitem

\bibitem[\protect\citeauthoryear{MMPoseContributors}{2020}]{mmpose2020}
\begin{botherref}
\oauthor{\bsnm{MMPoseContributors}}:
OpenMMLab Pose Estimation Toolbox and Benchmark.
\url{https://github.com/open-mmlab/mmpose}
(2020)
\end{botherref}
\endbibitem

\bibitem[\protect\citeauthoryear{Teed and Deng}{2021}]{teed2021droid}
\begin{botherref}
\oauthor{\bsnm{Teed}, \binits{Z.}},
\oauthor{\bsnm{Deng}, \binits{J.}}:
{DROID-SLAM: Deep Visual SLAM for Monocular, Stereo, and RGB-D Cameras}.
Advances in neural information processing systems
(2021)
\end{botherref}
\endbibitem

\bibitem[\protect\citeauthoryear{Ho et~al.}{2020}]{ho2020denoising}
\begin{barticle}
\bauthor{\bsnm{Ho}, \binits{J.}},
\bauthor{\bsnm{Jain}, \binits{A.}},
\bauthor{\bsnm{Abbeel}, \binits{P.}}:
\batitle{Denoising diffusion probabilistic models}.
\bjtitle{Advances in Neural Information Processing Systems}
\bvolume{33},
\bfpage{6840}--\blpage{6851}
(\byear{2020})
\end{barticle}
\endbibitem

\bibitem[\protect\citeauthoryear{Mahmood et~al.}{2019}]{AMASS:2019}
\begin{bchapter}
\bauthor{\bsnm{Mahmood}, \binits{N.}},
\bauthor{\bsnm{Ghorbani}, \binits{N.}},
\bauthor{\bsnm{F.~Troje}, \binits{N.}},
\bauthor{\bsnm{Pons-Moll}, \binits{G.}},
\bauthor{\bsnm{Black}, \binits{M.J.}}:
\bctitle{Amass: Archive of motion capture as surface shapes}.
In: \bbtitle{The IEEE International Conference on Computer Vision (ICCV)}
(\byear{2019}).
\burl{https://amass.is.tue.mpg.de}
\end{bchapter}
\endbibitem

\bibitem[\protect\citeauthoryear{Castillo et~al.}{2023}]{castillo2023bodiffusion}
\begin{botherref}
\oauthor{\bsnm{Castillo}, \binits{A.}},
\oauthor{\bsnm{Escobar}, \binits{M.}},
\oauthor{\bsnm{Jeanneret}, \binits{G.}},
\oauthor{\bsnm{Pumarola}, \binits{A.}},
\oauthor{\bsnm{Arbeláez}, \binits{P.}},
\oauthor{\bsnm{Thabet}, \binits{A.}},
\oauthor{\bsnm{Sanakoyeu}, \binits{A.}}:
Bodiffusion: Diffusing sparse observations for full-body human motion synthesis.
CV4Metaverse workshop, International Conference on Computer Vision
(2023)
\end{botherref}
\endbibitem

\bibitem[\protect\citeauthoryear{Jiang et~al.}{2022}]{jiang2022avatarposer}
\begin{bchapter}
\bauthor{\bsnm{Jiang}, \binits{J.}},
\bauthor{\bsnm{Streli}, \binits{P.}},
\bauthor{\bsnm{Qiu}, \binits{H.}},
\bauthor{\bsnm{Fender}, \binits{A.}},
\bauthor{\bsnm{Laich}, \binits{L.}},
\bauthor{\bsnm{Snape}, \binits{P.}},
\bauthor{\bsnm{Holz}, \binits{C.}}:
\bctitle{Avatarposer: Articulated full-body pose tracking from sparse motion sensing}.
In: \bbtitle{European Conference on Computer Vision},
pp. \bfpage{443}--\blpage{460}
(\byear{2022}).
\bcomment{Springer}
\end{bchapter}
\endbibitem

\bibitem[\protect\citeauthoryear{Aboukhadra et~al.}{2023}]{aboukhadra2023thor}
\begin{bchapter}
\bauthor{\bsnm{Aboukhadra}, \binits{A.T.}},
\bauthor{\bsnm{Malik}, \binits{J.}},
\bauthor{\bsnm{Elhayek}, \binits{A.}},
\bauthor{\bsnm{Robertini}, \binits{N.}},
\bauthor{\bsnm{Stricker}, \binits{D.}}:
\bctitle{Thor-net: End-to-end graformer-based realistic two hands and object reconstruction with self-supervision}.
In: \bbtitle{Proceedings of the IEEE/CVF Winter Conference on Applications of Computer Vision},
pp. \bfpage{1001}--\blpage{1010}
(\byear{2023})
\end{bchapter}
\endbibitem

\bibitem[\protect\citeauthoryear{Zhao et~al.}{2022}]{zhao2022graformer}
\begin{bchapter}
\bauthor{\bsnm{Zhao}, \binits{W.}},
\bauthor{\bsnm{Wang}, \binits{W.}},
\bauthor{\bsnm{Tian}, \binits{Y.}}:
\bctitle{Graformer: Graph-oriented transformer for 3d pose estimation}.
In: \bbtitle{Proceedings of the IEEE/CVF Conference on Computer Vision and Pattern Recognition},
pp. \bfpage{20438}--\blpage{20447}
(\byear{2022})
\end{bchapter}
\endbibitem

\bibitem[\protect\citeauthoryear{Park et~al.}{2022}]{park2022handoccnet}
\begin{bchapter}
\bauthor{\bsnm{Park}, \binits{J.}},
\bauthor{\bsnm{Oh}, \binits{Y.}},
\bauthor{\bsnm{Moon}, \binits{G.}},
\bauthor{\bsnm{Choi}, \binits{H.}},
\bauthor{\bsnm{Lee}, \binits{K.M.}}:
\bctitle{Handoccnet: Occlusion-robust 3d hand mesh estimation network}.
In: \bbtitle{Proceedings of the IEEE/CVF Conference on Computer Vision and Pattern Recognition},
pp. \bfpage{1496}--\blpage{1505}
(\byear{2022})
\end{bchapter}
\endbibitem

\bibitem[\protect\citeauthoryear{He et~al.}{2016}]{he2016deep}
\begin{bchapter}
\bauthor{\bsnm{He}, \binits{K.}},
\bauthor{\bsnm{Zhang}, \binits{X.}},
\bauthor{\bsnm{Ren}, \binits{S.}},
\bauthor{\bsnm{Sun}, \binits{J.}}:
\bctitle{Deep residual learning for image recognition}.
In: \bbtitle{CVPR}
(\byear{2016})
\end{bchapter}
\endbibitem

\bibitem[\protect\citeauthoryear{Lin et~al.}{2017}]{lin2017feature}
\begin{bchapter}
\bauthor{\bsnm{Lin}, \binits{T.-Y.}},
\bauthor{\bsnm{Doll{\'a}r}, \binits{P.}},
\bauthor{\bsnm{Girshick}, \binits{R.}},
\bauthor{\bsnm{He}, \binits{K.}},
\bauthor{\bsnm{Hariharan}, \binits{B.}},
\bauthor{\bsnm{Belongie}, \binits{S.}}:
\bctitle{Feature pyramid networks for object detection}.
In: \bbtitle{Proceedings of the IEEE Conference on Computer Vision and Pattern Recognition},
pp. \bfpage{2117}--\blpage{2125}
(\byear{2017})
\end{bchapter}
\endbibitem

\bibitem[\protect\citeauthoryear{Romero et~al.}{2017}]{MANO:SIGGRAPHASIA:2017}
\begin{botherref}
\oauthor{\bsnm{Romero}, \binits{J.}},
\oauthor{\bsnm{Tzionas}, \binits{D.}},
\oauthor{\bsnm{Black}, \binits{M.J.}}:
Embodied hands: Modeling and capturing hands and bodies together.
ACM Transactions on Graphics, (Proc. SIGGRAPH Asia)
\textbf{36}(6)
(2017)
\end{botherref}
\endbibitem

\bibitem[\protect\citeauthoryear{Zheng et~al.}{2023}]{zheng2023potter}
\begin{bchapter}
\bauthor{\bsnm{Zheng}, \binits{C.}},
\bauthor{\bsnm{Liu}, \binits{X.}},
\bauthor{\bsnm{Qi}, \binits{G.-J.}},
\bauthor{\bsnm{Chen}, \binits{C.}}:
\bctitle{Potter: Pooling attention transformer for efficient human mesh recovery}.
In: \bbtitle{Proceedings of the IEEE/CVF Conference on Computer Vision and Pattern Recognition},
pp. \bfpage{1611}--\blpage{1620}
(\byear{2023})
\end{bchapter}
\endbibitem

\bibitem[\protect\citeauthoryear{Li et~al.}{2021}]{li2021hybrik}
\begin{bchapter}
\bauthor{\bsnm{Li}, \binits{J.}},
\bauthor{\bsnm{Xu}, \binits{C.}},
\bauthor{\bsnm{Chen}, \binits{Z.}},
\bauthor{\bsnm{Bian}, \binits{S.}},
\bauthor{\bsnm{Yang}, \binits{L.}},
\bauthor{\bsnm{Lu}, \binits{C.}}:
\bctitle{Hybrik: A hybrid analytical-neural inverse kinematics solution for 3d human pose and shape estimation}.
In: \bbtitle{Proceedings of the IEEE/CVF Conference on Computer Vision and Pattern Recognition},
pp. \bfpage{3383}--\blpage{3393}
(\byear{2021})
\end{bchapter}
\endbibitem

\bibitem[\protect\citeauthoryear{Lin et~al.}{2021}]{lin2021end}
\begin{bchapter}
\bauthor{\bsnm{Lin}, \binits{K.}},
\bauthor{\bsnm{Wang}, \binits{L.}},
\bauthor{\bsnm{Liu}, \binits{Z.}}:
\bctitle{End-to-end human pose and mesh reconstruction with transformers}.
In: \bbtitle{Proceedings of the IEEE/CVF Conference on Computer Vision and Pattern Recognition},
pp. \bfpage{1954}--\blpage{1963}
(\byear{2021})
\end{bchapter}
\endbibitem

\bibitem[\protect\citeauthoryear{Kolotouros et~al.}{2019}]{kolotouros2019learning}
\begin{bchapter}
\bauthor{\bsnm{Kolotouros}, \binits{N.}},
\bauthor{\bsnm{Pavlakos}, \binits{G.}},
\bauthor{\bsnm{Black}, \binits{M.J.}},
\bauthor{\bsnm{Daniilidis}, \binits{K.}}:
\bctitle{Learning to reconstruct 3d human pose and shape via model-fitting in the loop}.
In: \bbtitle{Proceedings of the IEEE/CVF International Conference on Computer Vision},
pp. \bfpage{2252}--\blpage{2261}
(\byear{2019})
\end{bchapter}
\endbibitem

\bibitem[\protect\citeauthoryear{Tendulkar et~al.}{2023}]{Tendulkar_2023_CVPR}
\begin{bchapter}
\bauthor{\bsnm{Tendulkar}, \binits{P.}},
\bauthor{\bsnm{Sur{\'\i}s}, \binits{D.}},
\bauthor{\bsnm{Vondrick}, \binits{C.}}:
\bctitle{Flex: Full-body grasping without full-body grasps}.
In: \bbtitle{Proceedings of the IEEE/CVF Conference on Computer Vision and Pattern Recognition (CVPR)}
(\byear{2023})
\end{bchapter}
\endbibitem

\bibitem[\protect\citeauthoryear{Taheri et~al.}{2020}]{GRAB2020}
\begin{bchapter}
\bauthor{\bsnm{Taheri}, \binits{O.}},
\bauthor{\bsnm{Ghorbani}, \binits{N.}},
\bauthor{\bsnm{Black}, \binits{M.J.}},
\bauthor{\bsnm{Tzionas}, \binits{D.}}:
\bctitle{{GRAB}: A dataset of whole-body human grasping of objects}.
In: \bbtitle{European Conference on Computer Vision (ECCV)}
(\byear{2020})
\end{bchapter}
\endbibitem

\bibitem[\protect\citeauthoryear{Plizzari et~al.}{2024}]{plizzari2024outlook}
\begin{botherref}
\oauthor{\bsnm{Plizzari}, \binits{C.}},
\oauthor{\bsnm{Goletto}, \binits{G.}},
\oauthor{\bsnm{Furnari}, \binits{A.}},
\oauthor{\bsnm{Bansal}, \binits{S.}},
\oauthor{\bsnm{Ragusa}, \binits{F.}},
\oauthor{\bsnm{Farinella}, \binits{G.M.}},
\oauthor{\bsnm{Damen}, \binits{D.}},
\oauthor{\bsnm{Tommasi}, \binits{T.}}:
An outlook into the future of egocentric vision.
International Journal of Computer Vision
(2024)
\end{botherref}
\endbibitem

\bibitem[\protect\citeauthoryear{Ashutosh et~al.}{2024}]{ashutosh2024expertafexpertactionablefeedback}
\begin{botherref}
\oauthor{\bsnm{Ashutosh}, \binits{K.}},
\oauthor{\bsnm{Nagarajan}, \binits{T.}},
\oauthor{\bsnm{Pavlakos}, \binits{G.}},
\oauthor{\bsnm{Kitani}, \binits{K.}},
\oauthor{\bsnm{Grauman}, \binits{K.}}:
ExpertAF: Expert Actionable Feedback from Video
(2024)
\end{botherref}
\endbibitem

\bibitem[\protect\citeauthoryear{Kirillov et~al.}{2023}]{Kirillov_2023_ICCV}
\begin{bchapter}
\bauthor{\bsnm{Kirillov}, \binits{A.}},
\bauthor{\bsnm{Mintun}, \binits{E.}},
\bauthor{\bsnm{Ravi}, \binits{N.}},
\bauthor{\bsnm{Mao}, \binits{H.}},
\bauthor{\bsnm{Rolland}, \binits{C.}},
\bauthor{\bsnm{Gustafson}, \binits{L.}},
\bauthor{\bsnm{Xiao}, \binits{T.}},
\bauthor{\bsnm{Whitehead}, \binits{S.}},
\bauthor{\bsnm{Berg}, \binits{A.C.}},
\bauthor{\bsnm{Lo}, \binits{W.-Y.}},
\bauthor{\bsnm{Dollar}, \binits{P.}},
\bauthor{\bsnm{Girshick}, \binits{R.}}:
\bctitle{Segment anything}.
In: \bbtitle{Proceedings of the IEEE/CVF International Conference on Computer Vision (ICCV)},
pp. \bfpage{4015}--\blpage{4026}
(\byear{2023})
\end{bchapter}
\endbibitem

\bibitem[\protect\citeauthoryear{Kingma and Ba}{2014}]{kingma2014adam}
\begin{botherref}
\oauthor{\bsnm{Kingma}, \binits{D.P.}},
\oauthor{\bsnm{Ba}, \binits{J.}}:
Adam: A method for stochastic optimization.
arXiv preprint arXiv:1412.6980
(2014)
\end{botherref}
\endbibitem

\bibitem[\protect\citeauthoryear{Povey et~al.}{2011}]{Povey2011TheKS}
\begin{bchapter}
\bauthor{\bsnm{Povey}, \binits{D.}},
\bauthor{\bsnm{Ghoshal}, \binits{A.}},
\bauthor{\bsnm{Boulianne}, \binits{G.}},
\bauthor{\bsnm{Burget}, \binits{L.}},
\bauthor{\bsnm{Glembek}, \binits{O.}},
\bauthor{\bsnm{Goel}, \binits{N.K.}},
\bauthor{\bsnm{Hannemann}, \binits{M.}},
\bauthor{\bsnm{Motl{\'i}cek}, \binits{P.}},
\bauthor{\bsnm{Qian}, \binits{Y.}},
\bauthor{\bsnm{Schwarz}, \binits{P.}},
\bauthor{\bsnm{Silovsk{\'y}}, \binits{J.}},
\bauthor{\bsnm{Stemmer}, \binits{G.}},
\bauthor{\bsnm{Vesel{\'y}}, \binits{K.}}:
\bctitle{The kaldi speech recognition toolkit}.
(\byear{2011}).
\burl{https://api.semanticscholar.org/CorpusID:1774023}
\end{bchapter}
\endbibitem

\bibitem[\protect\citeauthoryear{Loshchilov and Hutter}{2019}]{loshchilov2018decoupled}
\begin{bchapter}
\bauthor{\bsnm{Loshchilov}, \binits{I.}},
\bauthor{\bsnm{Hutter}, \binits{F.}}:
\bctitle{Decoupled weight decay regularization}.
In: \bbtitle{International Conference on Learning Representations}
(\byear{2019}).
\burl{https://openreview.net/forum?id=Bkg6RiCqY7}
\end{bchapter}
\endbibitem

\bibitem[\protect\citeauthoryear{Xu et~al.}{2021}]{xu2021long}
\begin{barticle}
\bauthor{\bsnm{Xu}, \binits{M.}},
\bauthor{\bsnm{Xiong}, \binits{Y.}},
\bauthor{\bsnm{Chen}, \binits{H.}},
\bauthor{\bsnm{Li}, \binits{X.}},
\bauthor{\bsnm{Xia}, \binits{W.}},
\bauthor{\bsnm{Tu}, \binits{Z.}},
\bauthor{\bsnm{Soatto}, \binits{S.}}:
\batitle{Long short-term transformer for online action detection}.
\bjtitle{Advances in Neural Information Processing Systems}
\bvolume{34},
\bfpage{1086}--\blpage{1099}
(\byear{2021})
\end{barticle}
\endbibitem

\bibitem[\protect\citeauthoryear{Zhao and Kr{\"a}henb{\"u}hl}{2022}]{zhao2022real}
\begin{bchapter}
\bauthor{\bsnm{Zhao}, \binits{Y.}},
\bauthor{\bsnm{Kr{\"a}henb{\"u}hl}, \binits{P.}}:
\bctitle{Real-time online video detection with temporal smoothing transformers}.
In: \bbtitle{European Conference on Computer Vision}
(\byear{2022})
\end{bchapter}
\endbibitem

\bibitem[\protect\citeauthoryear{Kwak et~al.}{2020}]{kwak2020detecting}
\begin{bchapter}
\bauthor{\bsnm{Kwak}, \binits{I.}},
\bauthor{\bsnm{Guo}, \binits{J.-Z.}},
\bauthor{\bsnm{Hantman}, \binits{A.}},
\bauthor{\bsnm{Kriegman}, \binits{D.}},
\bauthor{\bsnm{Branson}, \binits{K.}}:
\bctitle{Detecting the starting frame of actions in video}.
In: \bbtitle{Proceedings of the IEEE/CVF Winter Conference on Applications of Computer Vision},
pp. \bfpage{489}--\blpage{497}
(\byear{2020})
\end{bchapter}
\endbibitem

\bibitem[\protect\citeauthoryear{Chen et~al.}{2024}]{chen2024pcie_egohandpose}
\begin{botherref}
\oauthor{\bsnm{Chen}, \binits{F.}},
\oauthor{\bsnm{Ding}, \binits{L.}},
\oauthor{\bsnm{Lertniphonphan}, \binits{K.}},
\oauthor{\bsnm{Li}, \binits{J.}},
\oauthor{\bsnm{Huang}, \binits{K.}},
\oauthor{\bsnm{Wang}, \binits{Z.}}:
Pcie\_egohandpose solution for egoexo4d hand pose challenge.
arXiv preprint arXiv:2406.12219
(2024)
\end{botherref}
\endbibitem

\bibitem[\protect\citeauthoryear{Li et~al.}{2021}]{li2021human}
\begin{bchapter}
\bauthor{\bsnm{Li}, \binits{J.}},
\bauthor{\bsnm{Bian}, \binits{S.}},
\bauthor{\bsnm{Zeng}, \binits{A.}},
\bauthor{\bsnm{Wang}, \binits{C.}},
\bauthor{\bsnm{Pang}, \binits{B.}},
\bauthor{\bsnm{Liu}, \binits{W.}},
\bauthor{\bsnm{Lu}, \binits{C.}}:
\bctitle{Human pose regression with residual log-likelihood estimation}.
In: \bbtitle{Proceedings of the IEEE/CVF International Conference on Computer Vision},
pp. \bfpage{11025}--\blpage{11034}
(\byear{2021})
\end{bchapter}
\endbibitem

\bibitem[\protect\citeauthoryear{Dosovitskiy}{2020}]{dosovitskiy2020image}
\begin{botherref}
\oauthor{\bsnm{Dosovitskiy}, \binits{A.}}:
An image is worth 16x16 words: Transformers for image recognition at scale.
arXiv preprint arXiv:2010.11929
(2020)
\end{botherref}
\endbibitem

\bibitem[\protect\citeauthoryear{Pavlakos et~al.}{2024}]{pavlakos2024reconstructing}
\begin{bchapter}
\bauthor{\bsnm{Pavlakos}, \binits{G.}},
\bauthor{\bsnm{Shan}, \binits{D.}},
\bauthor{\bsnm{Radosavovic}, \binits{I.}},
\bauthor{\bsnm{Kanazawa}, \binits{A.}},
\bauthor{\bsnm{Fouhey}, \binits{D.}},
\bauthor{\bsnm{Malik}, \binits{J.}}:
\bctitle{Reconstructing hands in 3d with transformers}.
In: \bbtitle{Proceedings of the IEEE/CVF Conference on Computer Vision and Pattern Recognition},
pp. \bfpage{9826}--\blpage{9836}
(\byear{2024})
\end{bchapter}
\endbibitem

\bibitem[\protect\citeauthoryear{Zhang et~al.}{2022}]{actionformer}
\begin{botherref}
\oauthor{\bsnm{Zhang}, \binits{C.}},
\oauthor{\bsnm{Wu}, \binits{J.}},
\oauthor{\bsnm{Li}, \binits{Y.}}:
Actionformer: Localizing moments of actions with transformers.
arXiv preprint arXiv:2202.07925
(2022)
\end{botherref}
\endbibitem

\end{thebibliography}
